\documentclass[10pt,twocolumn,letterpaper]{article}

\usepackage[pagenumbers]{cvpr} % To force page numbers, e.g. for an arXiv version

\usepackage[dvipsnames]{xcolor}
\usepackage{times}
\usepackage{epsfig}
\usepackage{graphicx}
\usepackage{amsmath}
\usepackage{amssymb}
\usepackage{booktabs}
\usepackage{float}
\usepackage{pifont}% http://ctan.org/pkg/pifont
\usepackage{multirow}
\usepackage[accsupp]{axessibility}  % Improves PDF readability for those with disabilities.

\usepackage{caption}
\usepackage{subcaption}
\usepackage{comment}

\usepackage[pagebackref,breaklinks,colorlinks]{hyperref}

\usepackage{dsfont}
\usepackage{etoolbox}

\newif\ifshowedits

\newcommand{\addeditor}[3]{%
  \definecolor{#1color}{rgb}{#3}
  \expandafter\newcommand\csname #1\endcsname[1]{%
  \ifshowedits
    {\color{#1color} ##1}%
  \else
    {##1}%
  \fi
  }%
  \expandafter\newcommand\csname #1rmk\endcsname[1]{%
  \ifshowedits
    {\color{#1color} {\bf [#2: ##1]}}
  \fi
  }%
  \expandafter\newcommand\csname #1rpl\endcsname[2]{%
  \ifshowedits
    {\color{#1color} ##1 \sout{##2}}
  \else
    {##1}
  \fi
  }%
}

\newcommand{\createtextvar}[1]{
  \expandafter\newcommand\csname #1\endcsname{%
  {\text{#1}}
}%
}
\newcommand{\textvars}[1]{\forcsvlist{\createtextvar}{#1}}

\newcommand{\moretextwithfigures}{
\renewcommand{\topfraction}{1}
\renewcommand{\dbltopfraction}{1}
\renewcommand{\bottomfraction}{1}
\renewcommand{\textfraction}{.0}
\renewcommand{\floatpagefraction}{1}
\renewcommand{\dblfloatpagefraction}{1}
}

\newcommand{\mycomment}[1]{}

\newcommand{\calL}{{\cal L}}

\newcommand{\calN}{{\cal N}}

\newcommand{\calS}{{\cal S}}

\addeditor{vincent}{VL}{0,0.6,0}
\addeditor{stefan}{SA}{0,0,1}
\addeditor{sinisa}{SS}{0.7.0.7,0}
\addeditor{ff}{FF}{1,0,0}

\showeditsfalse

\moretextwithfigures

\def\paperID{221} % *** Enter the Paper ID here
\def\confName{3DV\xspace}
\def\confYear{2024\xspace}

\title{HOC-Search: Efficient CAD Model and Pose Retrieval from RGB-D Scans}

\author{Stefan Ainetter\textsuperscript{(1)}, Sinisa Stekovic\textsuperscript{(1)}, Friedrich Fraundorfer\textsuperscript{(1)}, Vincent Lepetit\textsuperscript{(2)}
 \\
\textsuperscript{(1)}Institute for Computer Graphics and Vision, Graz University of Technology, Graz, Austria\\
\textsuperscript{(2)}LIGM, \'Ecole des Ponts, Univ Gustave Eiffel, CNRS, Marne-la-Vallée, France\\
{\tt\small \{stefan.ainetter, sinisa.stekovic, fraundorfer\}@icg.tugraz.at}, {\tt\small vincent.lepetit@enpc.fr}
}

\begin{document}
\maketitle

%%%%%%%%% ABSTRACT
\begin{abstract}
We present an automated and efficient approach for retrieving high-quality CAD models of objects and their poses in a scene captured by a moving RGB-D camera. We first investigate various objective functions to measure similarity between a candidate CAD object model and the available data, and the best objective function appears to be a "render-and-compare" method comparing depth and mask rendering. We thus introduce a fast-search method that approximates an exhaustive search based on this objective function for simultaneously retrieving the object category, a CAD model, and the pose of an object given an approximate 3D bounding box. This method involves a search tree that organizes the CAD models and object properties including object category and pose for fast retrieval and an algorithm inspired by Monte Carlo Tree Search, that efficiently searches this tree. We show that this method retrieves CAD models that fit the real objects very well, with a speed-up factor of 10x to \sinisa{120x} compared to  exhaustive search. 
\end{abstract}

%%%%%%%%% BODY TEXT

\begin{figure*}
\centering
\scalebox{0.78}{
\begin{tabular}{c}
\begin{tabular}{ccccc}
  \includegraphics[trim={11cm 3cm 7cm 2.cm},clip,width=0.22\linewidth]{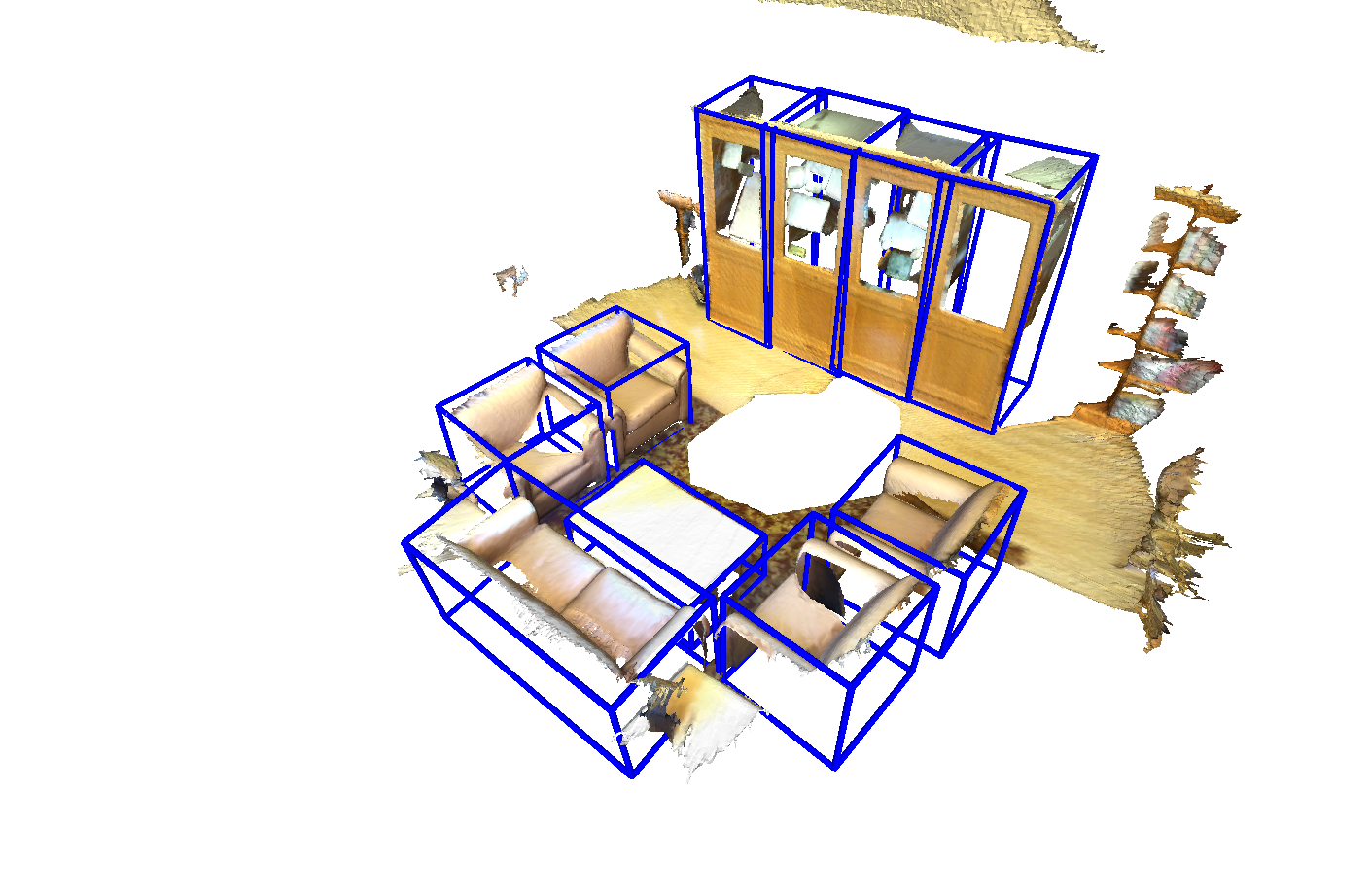} &
  \includegraphics[trim={11cm 3cm 7cm 2.cm},clip,width=0.22\linewidth]{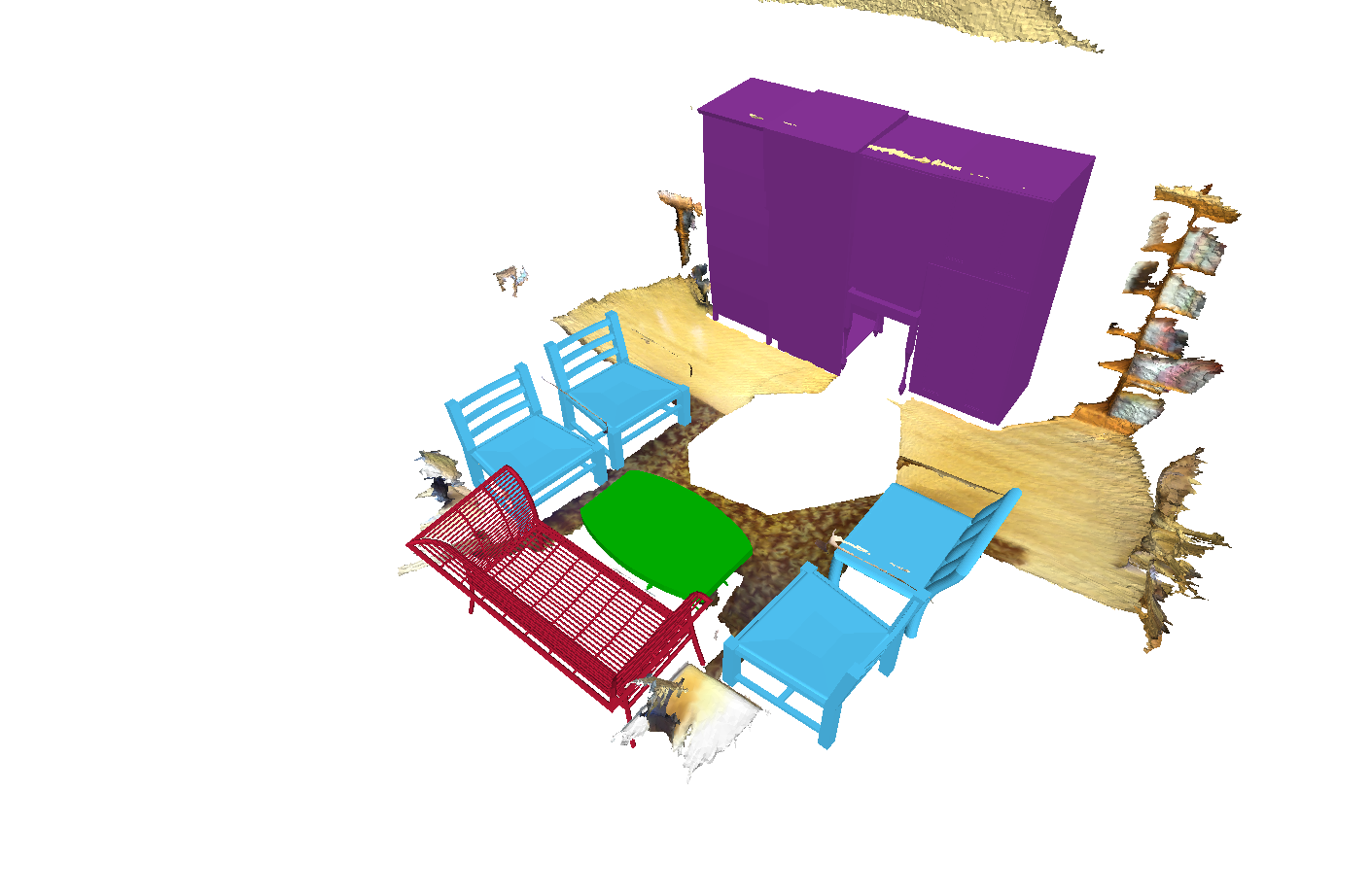} &
  \includegraphics[trim={11cm 3cm 7cm 2.cm},clip,width=0.22\linewidth]{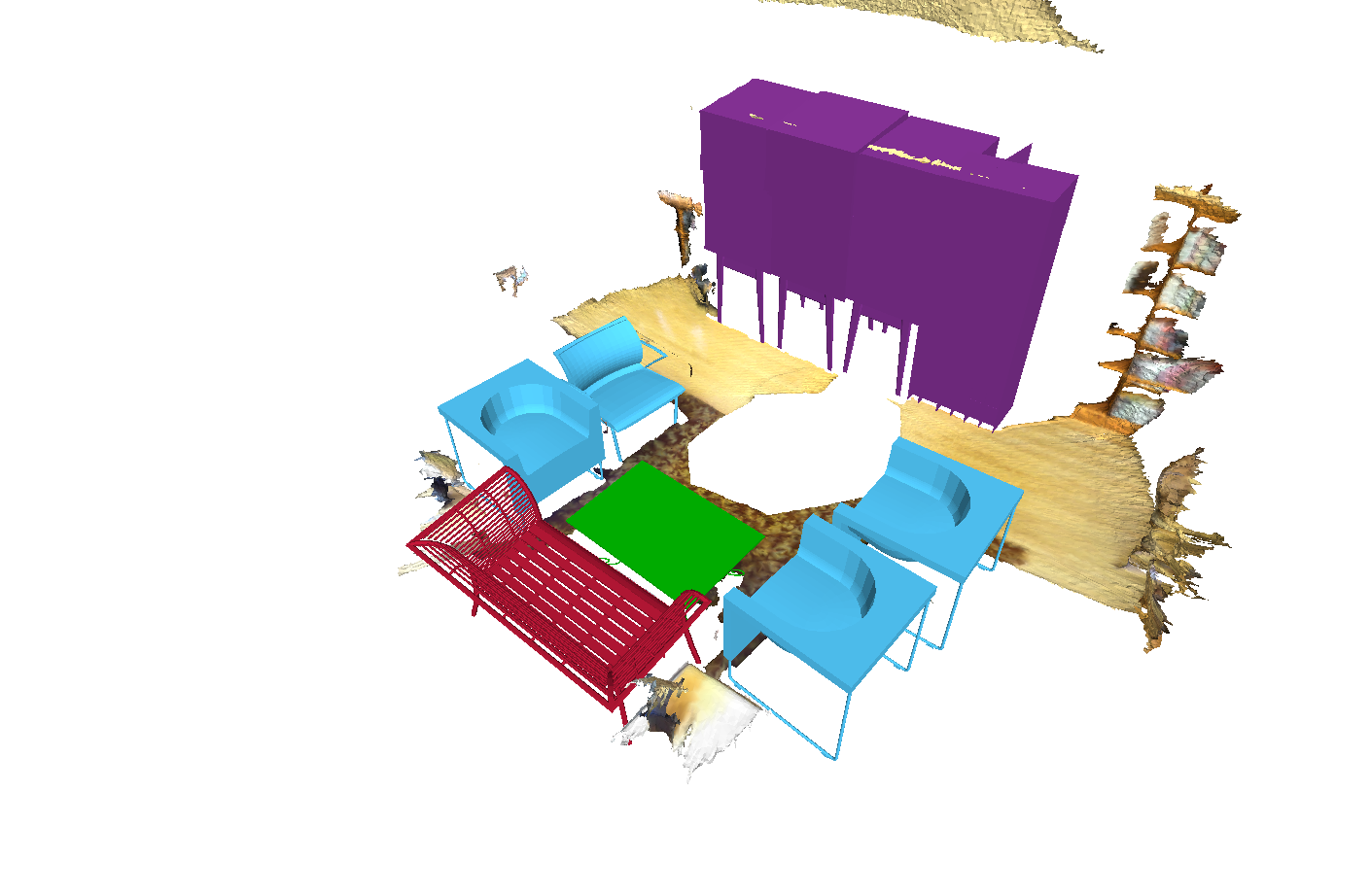} &
  \includegraphics[trim={11cm 3cm 7cm 2.cm},clip,width=0.22\linewidth]{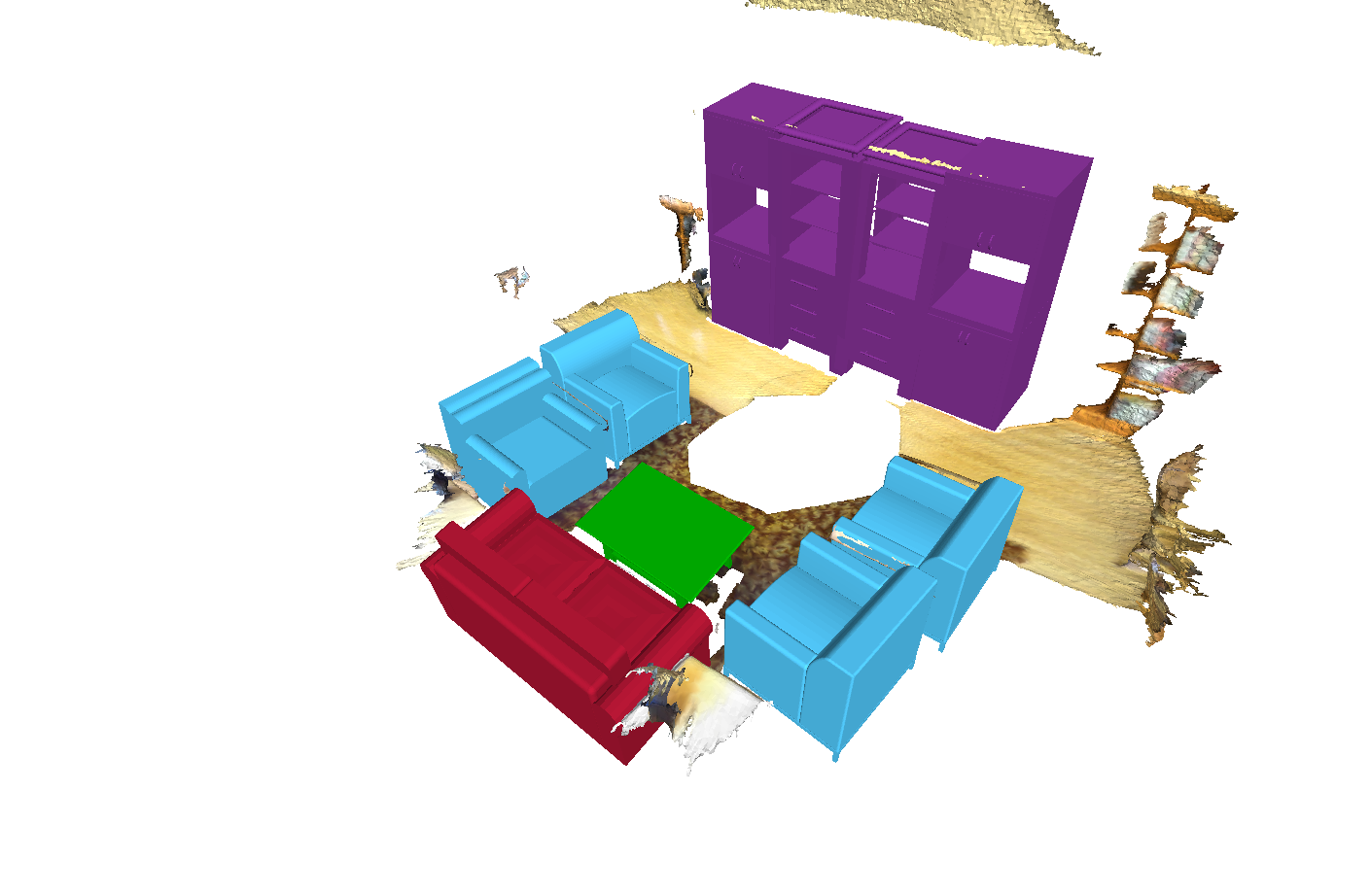} &
  \includegraphics[trim={11cm 3cm 7cm 2.cm},clip,width=0.22\linewidth]{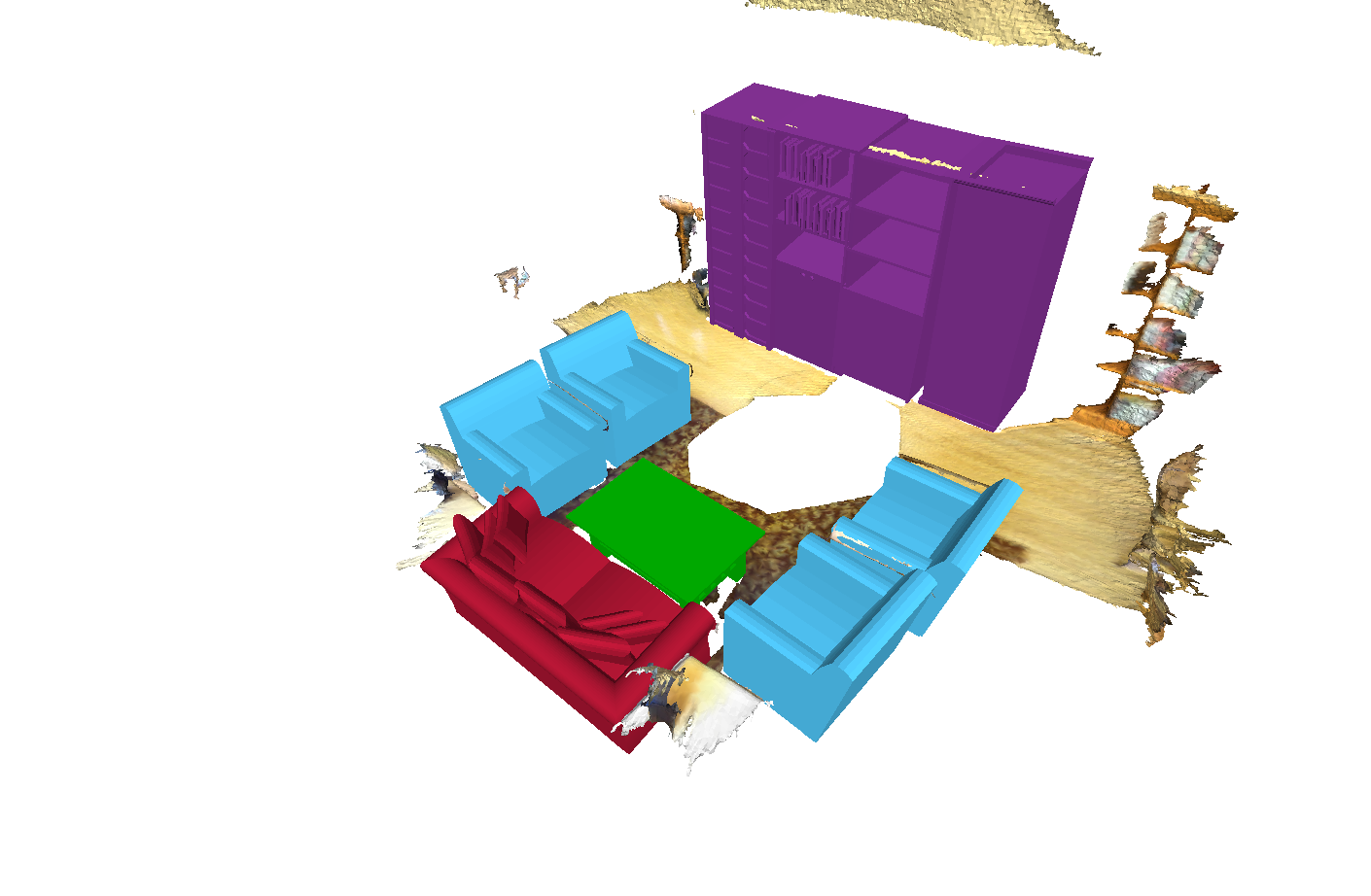}\\
  RGB-D Scan &  Embeddings distance & Chamfer distance  &Render-and-Compare & Render-and-Compare\\
            &   2.2 sec   &      1644 sec    &exhaustive search: 5096 sec & \textbf{our method}: 457 sec\\
\end{tabular}\\[2.5cm]
\includegraphics[width=1.2\linewidth]{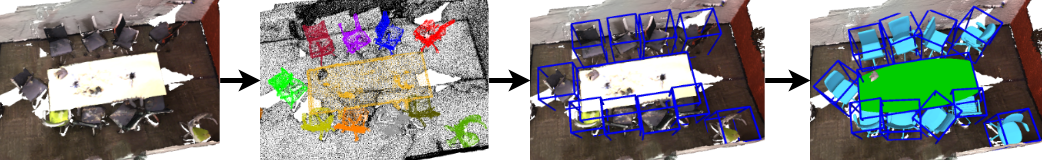}\\
\begin{tabular}{c@{\hspace{2.3cm}}c@{\hspace{1.3cm}}c@{\hspace{1cm}}c}
  \quad  RGB-D scan &   \quad Segmentation by SoftGroup &  \quad  Axis-aligned bounding boxes &  \quad  CAD models and poses \\ 
&&& \quad  retrieved by \textbf{our method}\\
\end{tabular}
\end{tabular}
}
\vspace{-0.25cm}
\caption{
\textbf{Top:} Using the distance between learned embeddings as an objective function for CAD model retrieval is extremely fast, but the accuracy is low. Using the chamfer distance returns slightly better CAD models, but is extremely slow. Using render-and-compare exhaustively on each possible CAD models returns models that fit the objects very well but is even slower. \textbf{Our method} also relies on render-and-compare, but is much faster and a very good approximation of the exhaustive search. 
\vincent{It produces much more accurate results compared to the faster embedding based search, while our computation times are still suitable for offline annotations.}
%It is still slower than search based on embeddings, but the results are much better and the computation times are suitable for offline annotations.
\textbf{Bottom:} We can use as input 3D bounding boxes obtained from an automated method, making our approach completely automatic. In practice, we use SoftGroup~\cite{Vu_2022_CVPR} to segment the objects and get axis-aligned bounding boxes. Our method can still recover accurate CAD models and poses even if the initial boxes are not accurate.}
\label{fig:teaser}
\end{figure*}

% \sinisarmk{I am not 100\% on this. But a reliable tweet from two days ago stated that chamfer is not a person, so it should not be capitalized XD Is this correct? (I am adapting this in the text) }
% \vincentrmk{that's interesting, i always saw Chamfer capitalized :) } \sinisarmk{Daniyar Turmukhambetov (You know the guy XD): Also, "chamfer distance", not "Chamfer distance". Chamfer is not a person.}

% \stefanrmk{In the last experiment, for CAD retrieval with unknown category label, we need 120x fewer render-and-compare iterations (which is roughly 120x speedup), should we integrate that into the Abstract?}\vincentrmk{it is done}

\section{Introduction}

Datasets play a crucial role in computer vision, serving as indispensable resources for training and evaluating algorithms. In the realm of 3D scene understanding in a multi-view setting, many  datasets have been introduced, such as SceneNN~\cite{hua2016scenenn}, ScanNet~\cite{dai2017scannet}, Matterport3D~\cite{Matterport3D}, and ARKitScenes~\cite{dehghan2021arkitscenes}. Unfortunately, except for ScanNet that benefits from annotations provided by the Scan2CAD dataset~\cite{avetisyan2019scan2cad}, these datasets do not have annotations for object shapes. This is because manual 3D annotations of object shapes can be highly challenging to create, as they require the joint estimation of a good 6D pose and retrieval of an appropriate CAD model for each object.

One alternative for training models is to use synthetic images, but generating photorealistic 3D virtual environments is a costly process that requires substantial resources for both creation and rendering~\cite{zheng2020structured3d,roberts2021hypersim}. Furthermore, it is essential to test models on real images, which cannot be substituted with synthetic images alone.

% , i.e., when both a point cloud and depth maps are available, which covers common scenarios

As shown in Figure~\ref{fig:teaser}, in this paper, we present an efficient and automated approach for retrieving high-quality CAD models of objects and their poses in a scene captured by a moving RGB-D camera. For each object, we start from a 3D bounding box and an associated object category label. These information may come with the dataset~(as in ScanNet), or we obtain it from the 3D semantic instance segmentation of the objects by SoftGroup~\cite{Vu_2022_CVPR}. In the latter case, the bounding boxes are only approximations: they are only axis-aligned---in other words we do not know the object's orientation---and they do not encompass the objects tightly. Our method still recovers accurate CAD models and poses. When using SoftGroup, our method thus becomes completely automated.

%We start from 3D bounding boxes that may come with the dataset (as in ScanNet), or 3D bounding boxes that we obtain from a segmentation of the objects by SoftGroup~\cite{Vu_2022_CVPR}. These bounding boxes are only axis-aligned and do not encompass tightly the objects, but our method can still recover accurate CAD models and poses, making our approach completely automatic.

% In a preliminary study, w

We first investigate possible objective functions between a potential CAD model and the available data, in order to select the best CAD model. We observe that matching the 3D points in a bounding box to a database of CAD models based on embeddings, as done in \cite{dahnert2019embedding,wei2022accurate}, is very fast but results in suboptimal CAD model retrieval. The best objective function we found relies on ``render-and-compare''~\cite{ainetter2023automatically} and compares depth and mask rendering of the CAD model with the captured depth data. Another advantage of this objective function is that it does not rely on learning and thus does not require registered 3D models for training.

Unfortunately, using this objective function for CAD model retrieval requires \emph{a priori} an exhaustive search over all possible models, which is very slow. To solve this, we introduce a fast-search method that approximates the exhaustive search. As shown in Figure~\ref{fig:teaser}, we retrieve CAD models from ShapeNet~\cite{chang2015shapenet} that fit the real objects very well, with a speed-up factor of 10x to 50x, \sinisa{and in some experiments 120x}, compared to the exhaustive search: One interesting property of our method is that we can trade of speed for fitness easily by simply increasing the number of iterations.

The closest work to ours is \cite{ainetter2023automatically}, which retrieves CAD models using an exhaustive search and render-and-compare as objective function. The Scan2CAD method from \cite{avetisyan2019scan2cad}~(not to be confused with the manual Scan2CAD annotations) does retrieve CAD models for ScanNet, but from a very small pool created for benchmarking the proposed method: The size of the pool is taken to be equal to the number of objects in the scene, i.e., in a range of 5 to 30 CAD models selected from the ShapeNet dataset. It is also designed to contain the ground truth models. \cite{avetisyan2019scan2cad} shows an experiment with more CAD models~(400), it is however only a qualitative result. \cite{wei2022accurate} uses a CAD model database consisting of 2827 CAD models for CAD model retrieval. By contrast, we consider the entire set of available CAD models from ShapeNet consisting of 32535 different models for the category classes present in ScanNet. For example, we consider the 8437 CAD Table models and the 6779 CAD Chair models in ShapeNet.

\vincent{

The method we propose has two key components: a data structure that organizes the CAD models for fast retrieval and an algorithm that efficiently searches this data structure given a 3D bounding box to find a CAD model that fits the input scene. Our data structure is a tree reminiscent of the Vocabulary Tree~\cite{nister2006scalable}, a popular option to build histograms of image patches for image retrieval. One fundamental difference with a vocabulary tree is that we introduce a novel type of node to integrate general search properties such as the pose and the category of objects. This will allow us to jointly retrieve the CAD model, pose and category.

It is possible to imagine different ways to use this tree for fast retrieval, and we consider several simple possibilities in the Experiments section. However, we propose a more sophisticated algorithm based on Monte Carlo Tree Search~(MCTS)~\cite{browne2012survey}. The advantage of basing our algorithm upon MCTS is that MCTS is an efficient and good approximation of the exhaustive search of the tree. We show this in our experiments by comparing our approach to several baselines. 
}

% The advantage of our approach is that objective evaluations are always performed directly for CAD models in leaf nodes instead of intermediate evaluations. % of clusters' centers.
% In turn, we can use these evaluations to iteratively discover promising nodes in the tree which are likely to minimize the objective function.

% However, while our tree structures jointly the CAD models and the object properties, it is not sufficient by itself to find the CAD model that optimizes our objective function.\vincentrmk{To show this, we compare our full method to several baselines that use this tree in straightforward ways. ??}
% We thus introduce an algorithm based on Monte Carlo Tree Search~(MCTS)~\cite{browne2012survey} that efficiently searches the tree of CAD models for a CAD model that optimizes the objective function. 

To further improve the pose of the retrieved CAD model, we extend our MCTS-based algorithm by adding a pose refinement step. In case of less accurate initial 3D object poses, such as those from the SoftGroup method~\cite{Vu_2022_CVPR}, it enables us to clearly outperform other methods on the challenging ScanNet dataset in terms of retrieval accuracy and computation time. 

To summarize, when used together with SoftGroup, our method retrieves fully automatically good CAD models and their poses for the objects in a scene captured with an RGB-D camera. It does not require training data and can therefore be applied easily to novel scenes.%, as shown in Figure~\ref{fig:teaser}.
\vincentrmk{depending on the results on the 2 scenes} \stefanrmk{ I added the results for the two scenes to the supp. material.}
In the remainder of the paper, we first discuss related work, describe our experiments to identify a good objective function to optimize, present our approach, and then our evaluations.

\section{Related Work}

In this section, we discuss related work on methods for CAD model retrieval and applications of Monte Carlo Tree Search~(MCTS) for  scene understanding. 

\subsection{CAD Model Retrieval}

Methods for CAD model retrieval aim to retrieve CAD models from a database of object models such as the ShapeNet dataset~\cite{chang2015shapenet} that best fit objects in the input scene. In some cases we also need to determine the object pose at the same time which makes this task even more challenging.

While it is possible to cast CAD model retrieval as a classification problem~\cite{aubry2014seeing,mottaghi2015coarse} such approaches do not scale to larger databases. Other line of works compare image descriptors of rendered CAD models and input images~\cite{aubry2015understanding,izadinia2017im2cad,massa2016deep}. Yet another possibility is to encode CAD models and input images into embedding vectors and calculate similarity in the embedding space~\cite{Grabner20183DPE,grabner2019location,kuo2020mask2cad}. One challenge with such approaches is that they need to bridge the domain gap between CAD models and images. Another challenge is that existing datasets such as Pix3D~\cite{sun2018pix3d} contain only samples where objects are fully visible. Therefore, methods trained on such datasets generalize poorly to real world scenarios that often include partial occlusions between objects in the scene. \cite{wei2022accurate} infers embeddings from an object scan, retrieves k-nearest CAD model embeddings, and performs re-ranking based on chamfer distance to the object scan. Such an approach counteracts the limited capability of embeddings to encode the precise shape of partially occluded objects. \cite{dahnert2019embedding} learns a joint embedding space between noisy and incomplete object scans and CAD models, and such methods are more resistant to noisy data. However, they use a small set of $100$ candidate objects, and rely on manual annotations to match CAD models with scans during training which are cumbersome to obtain.

It is also possible to learn alignments between input sets of CAD models and 3D scans~\cite{avetisyan2019scan2cad,avetisyan2019end}. These methods produce very good alignments when the object database contains only objects that are in the scene. Scan2CAD~\cite{avetisyan2019scan2cad} and \cite{avetisyan2019end} showed an experiment with a slightly larger database of objects, $400$ and $3000$ respectively, but only as a qualitative result. In addition these methods rely on manual annotations between CAD models and 3D scans.

A recent approach~\cite{ainetter2023automatically} relies on render-and-compare and geometric consistency between the 3D scan and the rendered CAD model. Therefore, this method does not rely on manual CAD model annotations. While it is able to recover accurate representations even for partially occluded objects in the scene, the method relies on exhaustive search which implies high time complexity. Instead, we propose a tree-based search method that notably increases efficiency of the search while maintaining high accuracy.

\subsection{Monte Carlo Tree Search for 3D Scene Understanding}

Monte Carlo Tree Search~(MCTS) is a popular algorithm for combinatorial optimization, especially for problems of high combinatorial complexity. It is the central component of AlphaGO~\cite{silver2016mastering}, an algorithm that achieved super human performance in the complex game of Go. \cite{hampali2021monte} used MCTS to identify arrangements of objects and structural elements in the scene, however without focusing on accuracy of retrieved CAD models. One possible search alternative is hill climbing~\cite{ZouGLH19}, however such greedy methods often lead to sub-optimal solutions. \cite{stekovic2021montefloor,MonteRoom} combined MCTS and gradient descent to jointly perform discrete and continuous optimization for 2D floor plan reconstruction from point clouds and room layout reconstruction from single images. 

In this work, we first design a tree structure to efficiently organize CAD models in a large database based on their similarity. Then, we introduce an MCTS-based approach that efficiently searches for the best CAD model in the dataset and, at the same time, optimizes the pose of the object. To the best of our knowledge, this is the first time MCTS is applied to a retrieval problem.

\textvars{dpt,Sil,CD,MSCD}

\section{What is the Best Objective Function for CAD Model Retrieval?}

The aim of this work is to provide a method for CAD model retrieval which does not suffer from one of the main drawbacks of learning-based methods, which is, the need of large amounts of manual annotations.

To achieve this, we first investigate objective functions which do not require manual CAD model annotations of scanned objects. A comparison of suitable objective functions serves as motivation for our novel algorithm,
which delivers accurate results and is computationally efficient.  Note that we provide here only a summary, more details are given in  the supplementary material. 

\subsection{Overview of Objective Functions}
%\paragraph{3D geometry-based function.}
\textbf{3D geometry-based function.} The chamfer distance is a commonly used metric to calculate distances of point sets, and is therefore widely used to calculate the similarity of 3D shapes. Considering two point clouds $P$ and $Q$, their chamfer distance can be defined as 
\begin{equation}
\CD = \frac{1}{|P|} \sum_{p \in P} \min_{q \in Q}  \|p - q\|_2  + \frac{1}{|Q|} \sum_{q \in Q} \min_{p \in P}  \|p - q\|_2\> , 
\label{eq:chamfer_dist}
\end{equation}
where $|P|$ and $|Q|$ are the number of points in $P$ and $Q$.%, respectively.

\begin{comment}
The authors of \cite{wei2022accurate} proposed the Modified Single-direction Chamfer Distance (MSCD) as objective function for CAD model retrieval which is defined as
%
\begin{equation}
\MSCD = \frac{1}{|P|} \sum_{p \in P} \min_{q \in Q}  \|p - q\|_2 \> .
\end{equation}

MSCD considers only the distance of points from a scanned object $P$ to the points from the CAD model $Q$, which increases robustness for incomplete point clouds by focusing on visible parts of the scanned object and ignoring unobserved/missing parts.
\end{comment}

%\paragraph{Render-and-compare.} 
\textbf{Render-and-compare.} The authors of~\cite{ainetter2023automatically} used an objective function which mainly focuses on comparing 2D observations of the scene before and after replacing the target object with a CAD model from the database.
This render-and-compare objective function can be defined as
\begin{equation}
\calL_{RC} = \calL_\dpt + \lambda_\Sil \calL_\Sil + \lambda_\CD \calL_\CD \> ,
\label{eq:rac}
\end{equation}
where $\calL_\dpt$ is a depth matching term defined as the L1-distance between depth maps before and after replacing the target object with a CAD model. $\calL_\Sil$ defines the Intersection-over-Union between silhouettes of real object and CAD model and $\calL_\CD$ defines the single-direction chamfer distance, with $\lambda_\Sil$ and $\lambda_\CD$ the corresponding weights.
%\stefan{In \cite{ainetter2023automatically}, the 2D silhouettes are obtained by rendering the 3D instance segmentation of the target objects, without using the RGB images. This way of generating the silhouettes still leads to an over-reliance on 3D data, and we will show in an experiment in Section~\ref{sec:Softgroup_experiment} that refining these rendered 2D silhouettes using the corresponding RGB images improves the quality of the CAD model retrieval for objects where the majority of 3D points are missing.}
By utilizing 2D observations to select and register the CAD models, this method is robust against incomplete point clouds. The retrieved models are well located in 3D, and also reproject well into the images of the RGB-D scan. %Additional information about our implementation of render-and-compare is provided in the supplementary material.

\textbf{Other objective functions.} We also consider the Modified Single-direction Chamfer Distance~(MSCD)~\cite{wei2022accurate}, and nearest neighbor search in embedding space using the encoder network from~\cite{cai2020learning} to extract feature embeddings. 

% Additional information and implementation details are provided in the supplementary material.

% for the described objective functions are provided in the supplementary material in Section~\ref{sec:supp_details_objective_function}.}

\begin{comment}

\paragraph{Nearest neighbor search in embedding space.}
Learned 3D shape descriptors can be used to efficiently encode information about geometric properties of objects in the embedding space.
Let $E(\cdot)$ denote an encoder network to extract a feature embedding for a given object. Using this encoder it is possible to calculate a feature embedding for each CAD model $x_{i}$ from the CAD model database $X = \{x_{i}, 1 \leq i \leq S\}$, where $S$ defines the number of CAD models. Nearest neighbor search in embedding space for a given target object $y$ can then be defined as 
%
\begin{equation}
    NN = \min_{x_{i} \in X}  \|E(y) - E(x_{i})\|_2 \> .
\end{equation}
%
For our experiments, we use the encoder network from~\cite{cai2020learning} which is based on PointNet~\cite{qi2017pointnet}. The network is trained on ShapeNet~\cite{chang2015shapenet} for the task of 3D shape reconstruction for point clouds. Note that several other learned feature embeddings can be effectively used for extracting discriminative features of shapes of objects, as shown in~\cite{wei2022accurate}.

\end{comment}

\subsection{Datasets and Evaluation Metrics}

We briefly describe the datasets and evaluation metrics that are used for all experiments throughout the paper.

\subsubsection{Datasets} 

For our experiments we use the validation set of ScanNet~\cite{dai2017scannet} which provides RGB-D scans of 312 real world indoor scenes. Additionally, we use Scan2CAD~\cite{avetisyan2019scan2cad} which provides CAD model annotations for 3184 objects in the ScanNet validation set. We use ShapeNet~\cite{chang2015shapenet} as CAD model database for retrieval, which consists of 32535 different models for the object categories present in ScanNet.

\subsubsection{Evaluation Metrics}
%We describe here the metrics we use to evaluate the accuracy and computation time for CAD model retrieval.
\paragraph{Chamfer distance.} We use the chamfer distance as defined in Equation~\eqref{eq:chamfer_dist} to quantify the agreement between point sets and to measure the similarity of 3D shapes. The number of sampled points for a specific  target object is dependent on the volume of the 3D box, to make sure that the point density of objects with different sizes is similar. Given an object with scaling $s_x$, $s_y$ and $s_z$ in $x$, $y$ and $z$ directions, the number of sampled points $N = m \cdot s_{x} \cdot s_{y} \cdot s_{z}$, where we take $m=25000$ as default value.
%\paragraph{Top-k Retrieval Accuracy.}

\textbf{Top-k Retrieval Accuracy.} The Top-k Retrieval Accuracy~(RA) can be defined as
\begin{equation}
    RA = \frac{1}{|Y|} \sum_{y \in Y} \sum_{x \in X(y)} 1(x = y_\text{CAD}) \> ,
\end{equation}
where $Y$ is the set of \stefan{target} objects, $X(y)$ the set of $k$ best retrieved CAD models for \stefan{target} object $y$, and $y_\text{CAD}$ the CAD model manually assigned to object $y$.

% \paragraph{Top-k Retrieval Accuracy.} The Top-k Retrieval Accuracy~(RA) can be defined as
% %
% \begin{equation}
%     RA = \frac{1}{T} \sum_{x \in X_{k}} \sum_{y_{t} \in Y} 1(x = y_{t}) \> ,
% \end{equation}
% %
% where $X_{k}$ is the set of $k$ best CAD models for a target object $y_{t}$, with $Y = \{y_{t}, 1 \leq t \leq T\}$ the set of all target objects of size $T$.

%\paragraph{Runtime analysis.}
\textbf{Runtime analysis.} For all experiments, we are using a single NVidia Titan V graphics card, and we report the mean runtime for CAD retrieval per scene. This indicates the order of magnitude of the computation time for the different experiments. For all experiments, our implementation runs on the CPU except for computing the objective function, which is done on the GPU.

\subsection{Quantitative Comparison}
To evaluate the performance of the previously described objective functions, we compare the accuracy of the retrieved CAD models for the objects in the ScanNet validation set using Scan2CAD as ground truth. We assume that the target objects are already detected and represented with a corresponding 3D box from the Scan2CAD annotations. However, we do not know the orientation of the objects inside the bounding box, hence we have to generate 
%four box proposals by rotating the box around the vertical axis by $[0^\circ,90^\circ,180^\circ,270^\circ]$.
four box proposals for each target object by rotating the box around the vertical axis by $[0^\circ,90^\circ,180^\circ,270^\circ]$.

Quantitative results using the described objective functions are shown in Table~\ref{tab:eval_obj_func}. Render-and-compare delivers the most accurate results but has the highest computation time. These results show the need for an efficient search algorithm to reduce the computational complexity of exhaustive search, while keeping the benefit of its accuracy.
We also provide qualitative results for this experiment in the supplementary material.

\begin{table}[t]
\centering
\scalebox{0.79}{
\begin{tabular}{@{}ccc@{}}
\toprule
Method      & \begin{tabular}[c]{@{}c@{}} Mean\\ Chamfer Distance\end{tabular} &  \begin{tabular}[c]{@{}c@{}} Mean Runtime \\ (seconds) \end{tabular} \\
\midrule
Nearest-Neighbor Embedding~\cite{cai2020learning}     & 12.64 $\cdot 10^{-3}$  & \textbf{2.24} \\
Exhaustive Chamfer Distance     & 10.03 $\cdot 10^{-3}$  & 1644 \\
Exhaustive MSCD~\cite{wei2022accurate}    & 7.24 $\cdot 10^{-3}$ & 1638 \\
Exhaustive Render-and-Compare~\cite{ainetter2023automatically} & \textbf{5.95 $\cdot 10^{-3}$}  & 5096 \\
\bottomrule
\end{tabular}
}
\vspace{-2mm}
\caption{Evaluation of different objective functions for CAD model retrieval. The mean chamfer distance is calculated between the retrieved CAD models and the ground truth CAD models from Scan2CAD~\cite{avetisyan2019scan2cad}. Exhaustive Render-and-Compare delivers the most accurate result while being the slowest method. The accuracy of the other functions is significantly lower.}
\label{tab:eval_obj_func}
\end{table}
\section{Method}

Our method for CAD model, category, and pose retrieval has two key components: A tree that organizes the CAD models but also the possible categories and poses for fast retrieval, and an algorithm that efficiently searches this tree given an input scene and an approximate 3D bounding box for the target object. 

As shown in Figure~\ref{fig:cadmodeltree}, we build the tree from a hierarchical clustering of the CAD models, where nodes correspond to the centroids of the clusters. Additionally, as shown in  Figure~\ref{fig:fulltree}, we introduce nodes that we call 'property nodes' to also search for  'properties' such as category and pose.

% The resulting tree structure combines efficient organization of the CAD models with the additional integration of general search properties. However, it is not trivial to efficiently explore this complex tree structure. A simple approach based on greedily selecting nodes that minimize the objective function is not easily possible due to the presence of meta nodes.} \sinisarmk{The last sentence is unclear for someone who does not know the method.}

%As shown in Figure~\ref{fig:cadmodeltree}, the tree is built from a hierarchical clustering of the CAD models, where nodes correspond to the centroids of the clusters. Then, we can search the tree to retrieve CAD models of objects in the scene: A simple approach is to greedily select nodes that minimize the objective function between an input set of 3D points and the centroid CAD model in the nodes. This CAD model is however in general far from being optimal. There is indeed only a small chance that such strategy retrieves the CAD model that minimizes our objective function.  

% \vincentrmk{Actually, we should provide this as a baseline!}\sinisarmk{You are right, but I do not think there is enough time to do that experiment. Stefan, would that be feasible?} \stefanrmk{Maybe we can try tomorrow if time is left, but I would not count on it as implementation and evaluation will take a bit of time.}
% \vincentrmk{it should be simple to implement (it is just going down the tree) and fast to run, but I understand time is limited. What about a rough evaluation on a subset of the dataset?}

\vincent{
As mentioned in the introduction, we propose an algorithm based on MCTS to efficiently search this tree. We compare our algorithm to several simpler options in Section~\ref{sec:experiments}  and show it performs significantly better. 
}

% The advantage of our approach is that objective evaluations are always performed directly for CAD models in leaf nodes instead of intermediate evaluations. % of clusters' centers.

% In turn, we can use these evaluations to iteratively discover promising nodes in the tree which are likely to minimize the objective function.

We call our tree structure 'HOC-Tree', for Hierarchical Object Clustering Tree, and our algorithm 'HOC-Search'. We describe them in more detail below.

\subsection{HOC-Tree: Hierarchical Object Clustering}
\label{sec:hoc_tree}

\newlength{\clusterfigwidth}
\setlength{\clusterfigwidth}{0.25\textwidth}
% \begin{figure*}
%     \centering
% %     \begin{tabular}{cc}
% %              % \includegraphics[width=0.3\linewidth]{figures/objectclusterTree.png}
% % % & \includegraphics[width=0.6\linewidth]{figures/rotationtree.png} \\
% %  (a) & (b)  
% %     \end{tabular}
%     % \includegraphics[width=0.8\linewidth]{figures/fulltree.png}

%     \scalebox{0.85}{
%     \begin{tabular}{c|c|c}
%         \includegraphics[width=\clusterfigwidth]{figures/clusters/cluster1.png} & \includegraphics[width=\clusterfigwidth]{figures/clusters/cluster2.png} & \includegraphics[width=\clusterfigwidth]{figures/clusters/cluster3.png} \\
%          Cluster level 1 & Cluster level 2 & Cluster level 3 \\
%          \includegraphics[width=\clusterfigwidth]{figures/clusters/cluster4.png} & \includegraphics[width=\clusterfigwidth]{figures/clusters/cluster5.png} & \includegraphics[width=\clusterfigwidth]{figures/clusters/cluster6.png} \\
%           Cluster level 4 & Cluster level 5 & Cluster level 6 \\
%     \end{tabular}}
%     \caption{Clusters' centers along a single path of our HOC-Tree for the Chair category. Cluster centers with a green outline are the parent nodes for the next level. The cluster centers become more and more similar to each other when depth increases, which makes searching our tree structure efficient.
%     }
%     \label{fig:cadmodeltree}
% \end{figure*}
\begin{figure}
    \centering
%     \begin{tabular}{cc}
%              % \includegraphics[width=0.3\linewidth]{figures/objectclusterTree.png}
% % & \includegraphics[width=0.6\linewidth]{figures/rotationtree.png} \\
%  (a) & (b)  
%     \end{tabular}
    % \includegraphics[width=0.8\linewidth]{figures/fulltree.png}

    \scalebox{0.8}{
    \begin{tabular}{c|c}
        \includegraphics[width=\clusterfigwidth]{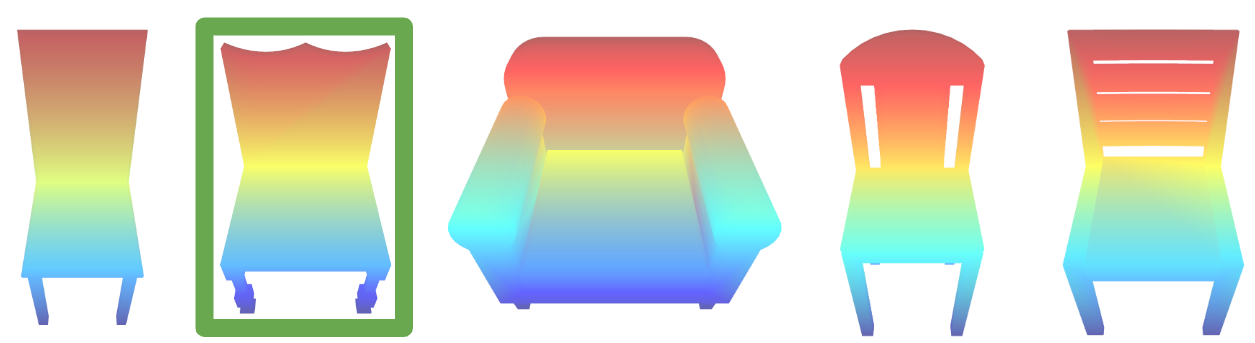} &
        \includegraphics[width=\clusterfigwidth]{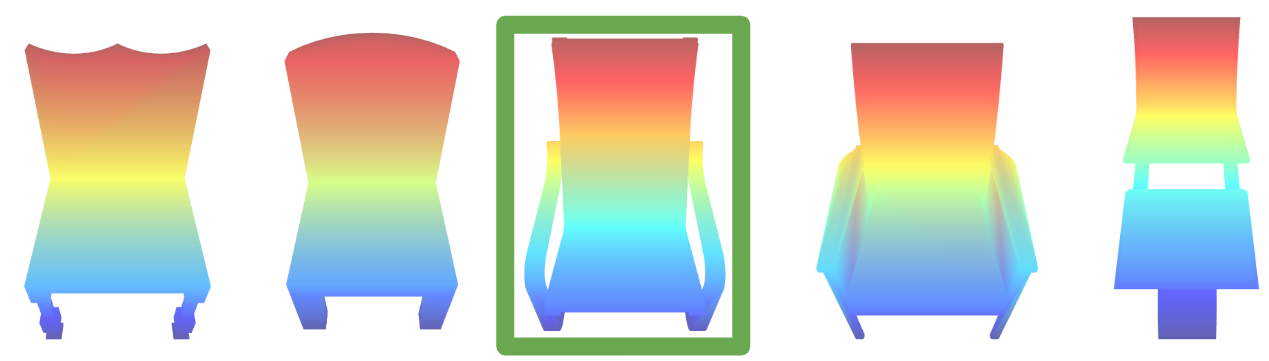} \\
        \vspace{-0.1cm} Cluster level 1 & Cluster level 2 \\
        % 
        % \includegraphics[width=\clusterfigwidth]{figures/clusters/cluster3.png} &
         % Cluster level 1 & Cluster level 2 & Cluster level 3 \\
         \includegraphics[width=\clusterfigwidth]{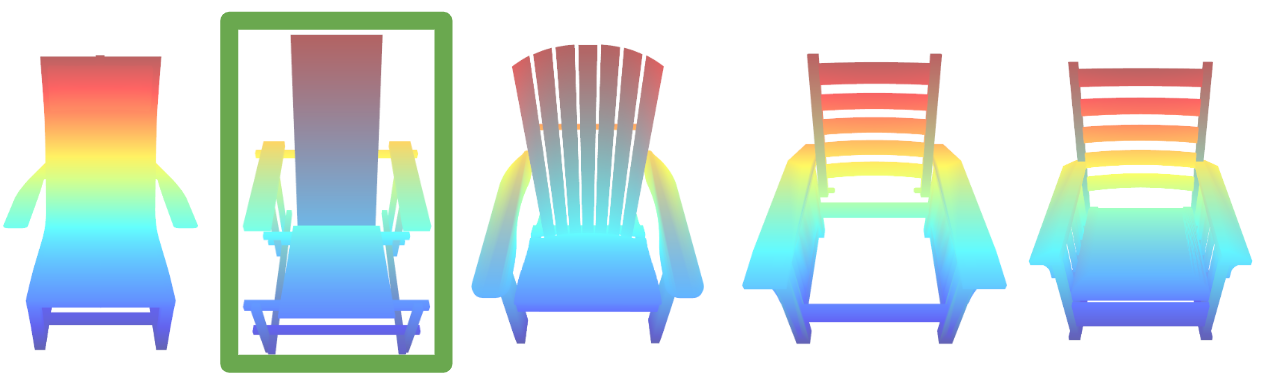} & 
         \includegraphics[width=\clusterfigwidth]{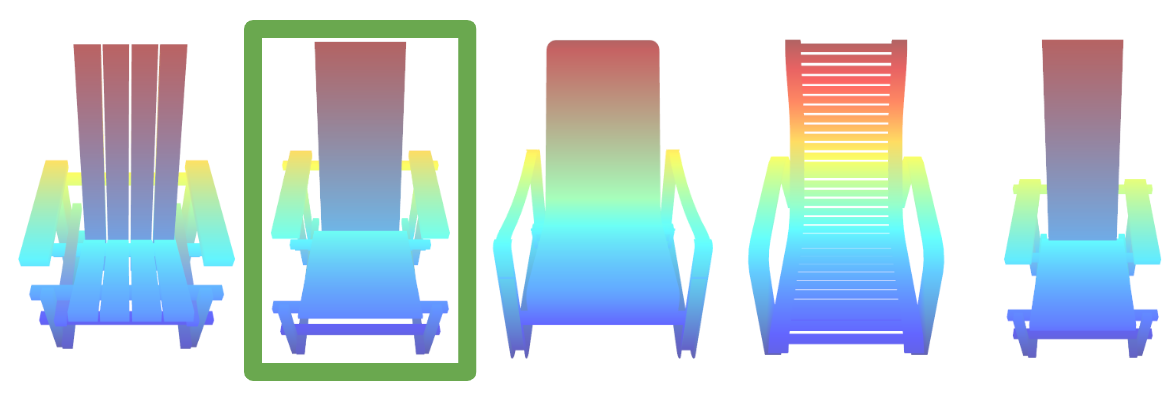} \\
         \vspace{-0.1cm} Cluster level 4 &  Cluster level 5 \\
    \end{tabular}}
    \vspace{-0.2cm}
    \caption{Clusters' centers along a single path of our HOC-Tree for the Chair category. Cluster centers with a green outline are the parent nodes for the next level. The cluster centers become more and more similar to each other when depth increases, which makes searching our tree structure efficient.
    }
    \label{fig:cadmodeltree}
\end{figure}

% \begin{figure}
%     \centering
%     \includegraphics[width=0.9\linewidth]{figures/fulltreewithoutzoomin.png}
%     \caption{Structure of our HOC-Tree for the Chair category. The first level of the HOC-Tree represents different rotations. We then copy the hierarchically clustered objects as a sub-tree of each rotation node. This allows our search method to jointly recover both the rotation and the CAD model of an object.}
%     \label{fig:fulltree}
% \end{figure}

\begin{figure}
    \centering
    %[trim={left bottom right top}
     \includegraphics[trim={0cm 0cm 0cm 0.2cm},clip,width=0.9\linewidth] {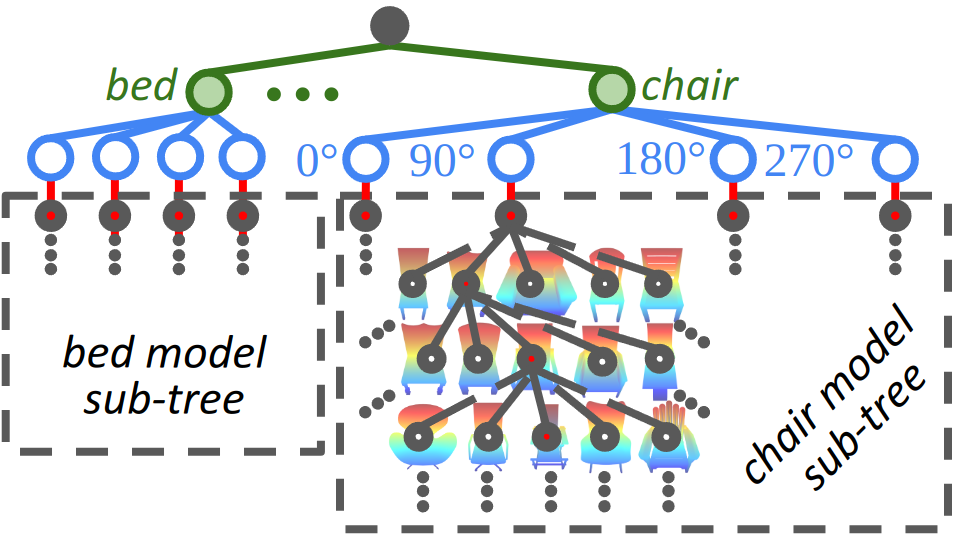}
    \vspace{-.25cm}
    \caption{\vincent{Structure of our HOC-Tree. The first levels of the tree are made of 'property nodes': We use category nodes to split the tree into different object categories, and we use pose nodes to represent different rotations. We then continue with the hierarchically clustered objects of the corresponding categories for each pose node. This allows our search method to jointly retrieve category, pose, and CAD model.}}
    \label{fig:fulltree}
\end{figure}

As we show in Figure~\ref{fig:cadmodeltree}, we want to organize the CAD models from a database into a tree structure in such a way that similarity between objects in sibling nodes increases with the depth of the tree. This property will be used by our search algorithm. To do so, we build the tree by hierarchically clustering the CAD models. For every CAD model, we first compute an embedding vector using the ShapeGF encoder~\cite{cai2020learning}, because such embeddings are correlated to the actual shape of the object. ShapeGF is an autoencoder trained on the ShapeNet dataset, so it does not require any manual annotations. Then, we perform $k$-means, with $k=5$, in the embedding space to assign CAD models to individual clusters. We repeat the procedure for remaining objects of each cluster until there is only one model left in the cluster. We can represent such hierarchical structure as a tree, where a node corresponds to a cluster and all children nodes are the sub-clusters of this cluster.

\vincent{
%\paragraph{Pose nodes.}
\textbf{Pose nodes.}
The resulting hierarchical tree structure efficiently organizes the CAD models, however, it is not yet sufficient for our goal: In practice, the objects' bounding boxes, whether obtained manually or predicted by SoftGroup for example, are not correctly oriented.  It is possible to refine the orientation once a CAD model is found by optimizing the objective function with gradient descent, however this converges correctly only with a good CAD model and from a good initial orientation. 

To solve this, we introduce the concept of pose nodes that are introduced into the HOC-Tree. We do this by first discretizing the orientation. In our application, the rotation of the object around the vertical axis is unknown, and we consider four possible rotations ($0^\circ, 90^\circ, 180^\circ, \text{and } 270^\circ$) in practice. 
We then attach the CAD model sub-tree to each of the pose nodes. This means that the first level of the tree corresponds to the rotation selection, and the lower levels correspond to the cluster selections. We pre-compute this tree for each category class in ScanNet, which we then use for retrieval at inference time. This will allow our HOC-Search algorithm to look for the CAD model and the discretized rotation simultaneously as described below. Note that HOC-Search also performs a refinement when we have a candidate CAD model to retrieve fine orientations. 
}

%\paragraph{Property nodes.}
\textbf{Property nodes.} In a more general setting, our tree structure can be adapted to contain more general properties which can be useful in practical scenarios. For example, by using category labels as property nodes, it is possible to combine the per-category trees to create a single large HOC-Tree, as we show in Figure~\ref{fig:fulltree}. In our experiments, we demonstrate that such tree structure can be used to perform category-agnostic retrieval.

\subsection{HOC-Search: MCTS-Based Model Retrieval}

Our HOC-Tree is similar to the concept of Vocabulary Trees~\cite{nister2006scalable}, a traditional method for image retrieval, except for the \stefan{property} nodes. While in the case of images, we can directly compare image descriptors of a target and a query image, here we want to optimize the render-and-compare objective function identified in Table~\ref{tab:eval_obj_func}. 
% More exactly, we use Monte Carlo Tree Search~(MCTS) and render-and-compare objective function when calculating the score.

Our HOC-Search algorithm for exploring the HOC-Tree while optimizing the objective function is inspired by MCTS. MCTS is a tree search algorithm that relies on Monte Carlo simulations, or random rollouts, to discover and focus on the most promising parts of the tree. The interested reader can refer to \cite{browne2012survey} for further details on MCTS. 
% In the following, we describe \sinisa{our iterative algorithm based on} MCTS for searching the CAD model tree. 
A single iteration of MCTS can be divided into three phases, which we describe below in the context of our algorithm.

\textvars{UCB,parent}

During the first phase, called \textit{selection}, we traverse the ``known'' part of the tree, until reaching a node not visited yet. At each level of the tree, we select the child node $\calN$ that maximizes the Upper Confidence Bound~(UCB) criterion, which balances exploitation and exploration:
\begin{equation}
    \UCB(\calN) = \calS(\calN) + \lambda \sqrt{\frac{\log n(\parent(\calN))}{n(\calN)}} \> ,
\end{equation}
where $\calS(\calN)$ is a score for node $\calN$, calculated later in the \emph{update} phase. The second term controls exploration. $n(\calN)$ is the total number of times that node $\calN$ has been visited during the previous iterations. $\parent(\calN)$ is the parent of node $\calN$. $\lambda$ is a scalar value that balances exploitation and exploration. In our case, $\lambda$ is linearly decreasing from $\lambda_{n} = 20.0$ to $\lambda_{N}=1.0$ with $ 1\leq n \leq N$, where $N$ is number of pre-defined HOC-Search iterations.

In case we run into a node that has not been visited yet (i.e.,  with $n(\calN) = 0$), we proceed to the \textit{simulation} phase, during which we randomly traverse the remaining tree branches until we reach a leaf node. The leaf node of our tree contains a CAD model, and we compute the score $s$ for this CAD model by taking the opposite of the render-and-compare objective function from Eq.~\eqref{eq:rac}: $s = - \calL_{RC}$.

We then proceed to the \textit{update} phase, in which we update the node scores of the nodes traversed during the current iteration: The new node score is taken as the maximum value between the old node score and the iteration score $s$. 

%In our implementation, the new node score is taken as the maximum value between the old node score and the iteration score $s$.

Then, we proceed to the next iteration and repeat the process. We halt after a pre-defined number of iterations and return the CAD model with the highest score. 

\textbf{Node locking.} We implement a node locking scheme  for MCTS \sinisa{to prevent re-visiting paths that were fully explored previously}. First, during the update phase the visited leaf node is locked \vincentrmk{what does locking mean exactly?}\sinisarmk{Setting a flag to true so that we know all descendants were visited before} and afterward the ancestor nodes are locked as well if all of their children are already locked. Therefore, we can avoid selecting locked nodes for which the sub-tree has already been fully explored to prevent evaluating the same solution more than once.

\textbf{CAD model retrieval with pose refinement.}
Accuracy of CAD model retrieval is highly dependent on the 3D pose estimate of the object. In practice, this 3D pose can be predicted using state-of-the-art methods for 3D object detection. However, such poses are often not perfectly aligned with the actual object and, consequently, this reduces the accuracy of CAD model retrieval. To counteract this, we integrate a pose refinement step to our MCTS-based search strategy, which enables us to perform simultaneous CAD model retrieval and pose refinement.

% One major problem for CAD model retrieval is that the selection of a suitable CAD model highly depends on the given initial 3D pose for the object. In practice, this 3D pose can be predicted using state-of-the-art methods for 3D object detection or 3D semantic instance segmentation. Although these methods recently made significant progress, predictions are still inaccurate which leads to initial poses which are most likely not optimal aligned with the corresponding target objects. Consequently, using these inaccurate 3D poses reduces the quality of the retrieved CAD models. To counteract this, we integrate a pose refinement step to our search strategy, which enables us to perform simultaneous CAD model retrieval and pose refinement. \\

During the score calculation step of MCTS, if the observed CAD model achieves a new high score, we perform a gradient-based optimization to iteratively adapt the 9-DOF pose $T$ (scale, rotation, and translation) of this current CAD model. More precisely, we optimize $T$ to minimize the objective function in Equation~\eqref{eq:rac} using the 
differentiable rendering pipeline of \cite{ravi2020pytorch3d} and the Adam optimizer~\cite{kingma2014adam}. Then, we use $T$ as the new initial pose for the upcoming search iterations. Note that for each of the \stefan{pose node} branches ($0^\circ$, $90^\circ$, $180^\circ$, $270^\circ$) the HOC-Tree has a different and independent initial pose, and therefore each one uses an independent refined pose $T_r$. We perform $150$ pose refinement steps each time a better model is found. This number of steps is in most cases not enough to fully align the current CAD model with the target object, but helps to correct significantly wrong poses. The partial refinement ensures that the refined pose is still general enough for other CAD models of upcoming HOC-Tree iterations. For the final CAD model with the highest score, we again perform pose refinement for another $800$ steps to obtain the final object pose.

\section{Evaluation and Experiments}
\label{sec:experiments}

Our evaluation is divided into three parts: First, we compare our HOC-Search to exhaustive search while optimizing the same objective function. This allows us to evaluate the impact of the approximation introduced by our method and the speedup it provides. 

\vincent{
Second, we compare our method to other possible approaches to show the advantage of our HOC-Tree structure in combination with our search algorithm.
}

% other competitive search strategies, namely nearest-neighbor search in embedding space + candidate re-ranking, and greedy search using the hierarchical cluster tree. %For these experiments, we use HOC-Search \textbf{without} pose refinement, to guarantee a fair comparison.}

Third, we show that our method can be efficiently used for CAD model retrieval in point clouds, using off-the-shelf SoftGroup~\cite{Vu_2022_CVPR} to extract initial 3D bounding boxes. 

% This last experiment shows the benefit of simultaneous CAD model retrieval and pose refinement. 

\vincent{For these first three experiments, we assume that the category labels of the target objects are given, which enables us to quantitatively compare our approach to other baselines. In an additional experiment, we provide qualitative results for CAD model and pose retrieval when the object category is not available \emph{a priori}.
}

% , by first generating a single HOC-Tree for all CAD models with the help of our proposed property nodes, and in order to highlight the potential of our tree structure and efficiency of our proposed search algorithm.

For all experiments, we use the ScanNet validation set~\cite{dai2017scannet} and the Scan2CAD manual annotations~\cite{avetisyan2019scan2cad} as ground truth for quantitative evaluation.

Due to the page limit, we show additional results and discuss limitations in the supplementary material.

% All experiments are conducted using the ScanNet validation set~\cite{dai2017scannet}, and we use the Scan2CAD manual annotations~\cite{avetisyan2019scan2cad} as ground truth to calculate the Chamfer distance metric. 

%We first compare our HOC-Search to exhaustive search while optimizing the same objective function. This allows us to evaluate the impact of the approximation introduced by our method and its speedup. 
%\stefan{Additionally, we compare our method to nearest-neighbor search in embedding space + candidate re-ranking using render-and-compare, to show the effectiveness of HOC-Search compared to this competitive baseline.}

%Second, we show that our method can be efficiently used for CAD model retrieval in point clouds, using the off-the-shelf SoftGroup method~\cite{Vu_2022_CVPR} to extract initial 3D poses for the target objects. Here, we show the benefit of simultaneous CAD model retrieval and pose refinement. 

\subsection{HOC-Search Efficiency Evaluation}
\label{sec:hoc_eff_eval}
%\stefan{
%To evaluate the efficiency of HOC-Search, we provide two experiments: First, we compare runtime and accuracy of our method with the exhaustive search proposed in~\cite{ainetter2023automatically}. This shows that our approach is an efficient approximation to the exhaustive search, and we measure in this experiment how close our retrieved CAD models are to the ones retrieved by the exhaustive search baseline in relation to our achieved speed-up factor. Second, we compare our approach to nearest neighbor search in embedding space with candidate re-ranking using render-and-compare. Both experiments are conducted using the ScanNet validation set~\cite{dai2017scannet} and the initial 3D bounding boxes from the Scan2CAD manual annotations~\cite{avetisyan2019scan2cad}. For both experiment, we use HOC-Search \textbf{without} pose refinement, to guarantee a fair comparison.}

To evaluate the efficiency of our MCTS-based search strategy, we compare runtime and accuracy of our method with the exhaustive search baseline from \cite{ainetter2023automatically} on the ScanNet validation set~\cite{dai2017scannet} and the initial poses from the Scan2CAD manual annotations~\cite{avetisyan2019scan2cad}. As our approach is an efficient approximation of the exhaustive search, we measure in this experiment how close our retrieved CAD models are to the ones retrieved by the exhaustive search baseline compared to our speed-up factor.
%\textbf{Comparison to exhaustive search baseline.}
Here, we use HOC-Search \textbf{without} pose refinement, to guarantee a fair comparison to the exhaustive search baseline.

The Top-k Retrieval Accuracy of our method compared to the exhaustive search baseline is shown in Table~\ref{tab:eval_topk_acc}. We emphasize that an average Top-1 Accuracy of 46.3\% for 800 iterations is significant: For example, to retrieve one Table CAD model, exhaustive search has to perform $8437 \times 4 = 33748$ iterations (the 4x factor is due to the unknown orientation of the CAD model inside a given 3D box). Therefore, randomly selecting 800 samples (without replacement) has a probability of $\frac{800}{33748} = 2.37\%$ to find the same model as the exhaustive search. Even when we are not able to retrieve exactly the same CAD model as the exhaustive search baseline, our retrieved models are of similar quality. 
\stefan{This is indicated by the chamfer distance to the ground truth models, which is comparable to $5.95 \cdot 10^{-3}$ for the exhaustive search (see Table~\ref{tab:eval_obj_func}).}
%This is indicated by the chamfer distance to the ground truth models from Scan2CAD that is comparable to the exhaustive search baseline. 
Visualizations of Top-5 retrieved CAD models, as well as per-category Top-k Accuracy results are provided in the supplementary material.

%The supplementary material provides visualizations of top-5 candidates retrieved by HOC-Search and exhaustive search and quantitative results for per-category Top-k Accuracy.

%In a large scale CAD model database~(more than 32k CAD models in total, e.g., 8437 Tables models and the 6779 Chairs models), it is highly likely that there exist multiple suitable CAD models. 

%\stefanrmk{I think we can skip Figure 4 (Visualization of top-3 candidates ) to make some space, and just refer to visualizations in the supp. material?}
%As shown in Figure~\ref{fig:vis_topk_acc}, often the top-3 candidates of HOC-Search and exhaustive search are very similar compared to the target object according to their 3D shape. Our method possibly retrieves a CAD model that is not the same model as from the exhaustive search but which fits the real object as well.

\begin{table}[]
\centering
\scalebox{0.83}{
\begin{tabular}{@{}ccccc@{}}
\toprule

\# iterations      & 
\begin{tabular}[c]{@{}c@{}} Mean Top-1 \\ Accuracy\end{tabular} &
\begin{tabular}[c]{@{}c@{}} Mean Top-1 \\ Chamfer Dist. \end{tabular} &
\begin{tabular}[c]{@{}c@{}} Mean Top-5 \\ Accuracy\end{tabular} &
\begin{tabular}[c]{@{}c@{}} Speedup\\ Factor\end{tabular}\\
\midrule
200     & 22.3\% & 7.33 $\cdot 10^{-3}$ & 47.8\% & 44.7\\
%300      & 29.4\% & 6.96 $\cdot 10^{-3}$ & 58.1\% & 29.8\\
400      & 34.3\% &6.91 $\cdot 10^{-3}$ & 64.7\%  & 22.3\\
600      & 41.7\% & 6.72 $\cdot 10^{-3}$ & 73.5\% & 14.9\\
800      & 46.3\% & 6.70 $\cdot 10^{-3}$ & 77.9\% & 11.1\\
%1000     & 50.2\% & 6.67 $\cdot 10^{-3}$ & 80.5\% & 8.9\\
\bottomrule
\end{tabular}
}
\vspace{-2mm}
\caption{CAD model retrieval efficiency and accuracy of our HOC-Search shown for different number of search iterations. We measure if the models retrieved by HOC-Search are within Top-k best candidates based on the exhaustive search. Our approach shows very high retrieval accuracy. As previously emphasized, a brute-force search with 800 iterations has a probability of 2.37\% of finding the same model. %The chamfer distance is relative to the Scan2CAD annotations and is comparable to $5.95 \cdot 10^{-3}$ for the slow exhaustive search (see Table~\ref{tab:eval_obj_func}). We refer to the supplementary material for results of per-category Top-k Accuracy.
}
\label{tab:eval_topk_acc}
\end{table}

\vincent{

\subsection{Comparison to Other Search Strategies}
To validate our HOC-Search method, we compare it to other possible search methods. 

\textbf{Nearest-neighbor baseline.} 
A first simple baseline for fast approximate search is to first identify $N$ good candidate CAD models by nearest neighbor~(NN) search, and re-rank them according to the render-and-compare objective function.\vincentrmk{does wei2022 really do that? }
\stefanrmk{wei2022 do NN-search in embedding space, and then re-rank the candidates, but for re-ranking they use the MSCD objective function; I removed the reference to wei2022 here} The idea is to exploit fast NN search to significantly decrease the number of evaluations of slow render-and-compare evaluations. For NN search, we use the ShapeGF embedding of the point cloud inside the input 3D box and the ShapeGF embeddings of the CAD models.

We compare Mean Top-1 Accuracy between this baseline and our HOC-Search in Table~\ref{tab:nn_search_re-rank} while varying $N$. For a fair comparison, we use the same number of evaluations of the render-and-compare objective function, which is the most costly operation by far and results in almost equal computation time for both approaches. The results indicate that our method is clearly superior to this baseline. 

% \vincentrmk{maybe we should remove this. it is dangerous as a reviewer could say we should retrain the embeddings}
% One of the reasons is that, due to the domain gap between synthetic and real data, target embeddings extracted by ShapeGF cannot be matched correctly to the CAD model embeddings. In contrast, our approach is not influenced by this domain gap, as the search is performed exclusively using the render-and-compare objective.

\textbf{Greedy search baseline.} We also consider another baseline that performs a greedy search of our HOC-Tree. Starting from the root, the search ranks the children nodes of the current node by computing the render-and-compare objective function for their centroid CAD models. It continues with the child node with the best  render-and-compare value and repeats this process until reaching a leaf. 

To deal with the property nodes, we simply evaluate all sub-levels of property nodes and in the end, return the result that minimizes the loss. The number of evaluations for the greedy search of the HOC-Tree depends on the size of the tree. For example, the hierarchical cluster tree for the Chair category consists of $d=7$ depth levels and each level consists of up to $k=5$ nodes. Because we have to independently search each property node branch (we have $4$ pose nodes in this experiment), the worst-case number of iterations is $4 \times d \times k$, or $140$ for the chair category.

On average, the Mean Top-1 Accuracy is $4.5\%$ which is significantly worse compared to $22.3\%$ achieved by our HOC-Search with comparable $200$ iterations. Greedy search often selects wrong nodes in early levels of the tree, and it cannot recover from such mistakes. In contrast, our MCTS-based strategy performs evaluations in leaf nodes of the tree and it is also able to revisit nodes in the upper part of the tree which leads to much better performance.

% Mean Top-1 Accuracy for the proposed search strategies compared to HOC-Search are given in Table~\ref{tab:nn_search_re-rank}. The results show that our method is a significantly better approximation of exhaustive search based on render-and-compare, and thus returns significantly better CAD models.
}

\begin{table}[]
\centering
\scalebox{0.87}{
%\begin{tabular}{cccccc}
%\toprule
%&\multicolumn{5}{c}{\# render-and-compare evaluations}\\
%\cmidrule{2-6}
% & 200  & 300  & 400  & 600  & 800  \\ 
%\midrule
% NN-Search + re-rank  & 14.7\% & 18.1\% & 20.8\% & 24.4\% & 27.6\% \\
%% \begin{tabular}[c]{@{}c@{}} NN-Search + \\ re-ranking \end{tabular}
%% & 14.7\% & 18.1\% & 20.8\% & 24.4\% & 27.6\% \\
%%\begin{tabular}[c]{@{}c@{}} HOC-Search \\ (ours) \end{tabular}    & %22.3\% & 29.4\% & 34.3\% & 41.7\% & 46.3\%
%HOC-Search (ours)         & 22.3\% & 29.4\% & 34.3\% & 41.7\% & 46.3\%
%\\

%\bottomrule
%\end{tabular}

\begin{tabular}{ccccc}
\toprule
&\multicolumn{4}{c}{\# render-and-compare evaluations}\\
\cmidrule{2-5}
 & 200   & 400  & 600  & 800  \\ 
\midrule
 NN-Search + re-ranking  & 14.7\% &  20.8\% & 24.4\% & 27.6\% \\
HOC-Search (ours)         & 22.3\% & 34.3\% & 41.7\% & 46.3\% \\
\bottomrule
\end{tabular}

}
\vspace{-2mm}
\caption{\label{tab:nn_search_re-rank} 
\stefan{Mean Top-1 Retrieval Accuracy for different search strategies in relation to exhaustive search results. For the same number of evaluations, our HOC-Search significantly outperforms the nearest-neighbor baseline.}
}
\end{table}

%\subsection{Experiment: CAD Model and Pose Retrieval from RGB-D Scans}
\subsection{Automatic CAD Model and Pose Retrieval %from RGB-D Scans
}

\label{sec:Softgroup_experiment}

\begin{table}[]
\centering
\scalebox{0.85}{
\begin{tabular}{@{}ccc@{}}
\toprule
\begin{tabular}[c]{@{}c@{}}Method\end{tabular}  & \begin{tabular}[c]{@{}c@{}} Mean \\ Chamfer Dist. \end{tabular} &  
%\begin{tabular}[c]{@{}c@{}} Runtime \\ (seconds) \end{tabular} &
\begin{tabular}[c]{@{}c@{}} Speedup\\ Factor\end{tabular}\\
\midrule
Exhaustive MSCD~\cite{wei2022accurate} & 16.75 $\cdot 10^{-3}$  %& 1792
& 4.5
\\
Exhaustive Render-and-Compare~\cite{ainetter2023automatically} & 14.59 $\cdot 10^{-3}$ %& 8199
& 1 \\
\midrule
HOC-Search (800 iterations)  \\ 
 w/o refinement (ours) & 15.45 $\cdot 10^{-3}$ % &  513 
 & \textbf{15.9}
\\
 w/ refinement (ours) & 13.72 $\cdot 10^{-3}$  %&  736 
 & 11.1
\\
\midrule
HOC-Search (1200 iterations) \\
 w/o refinement (ours) & 15.15 $\cdot 10^{-3}$ % & 749
 & 10.9
\\
w/ refinement (ours) & \textbf{13.66} $\cdot 10^{-3}$  %&  1023
& 8.0
\\
\bottomrule
\end{tabular}
}
\vspace{-2mm}
\caption{Results for CAD model retrieval on the ScanNet validation set when using SoftGroup predictions to obtain the initial object poses. Chamfer distance metric is relative to Scan2CAD annotations. Exhaustive MSCD does not cope well with the inaccurate initial poses. The accuracy of HOC-Search \textbf{without} refinement using 1200 iterations is close to the exhaustive search baseline. HOC-Search \textbf{with} refinement outperforms exhaustive search, while being significantly faster. \stefan{The speedup factor is relative to exhaustive render-and-compare.}} 
\label{tab:eval_softgroup}
\end{table}

\begin{figure}
\centering
\scalebox{.95}{
\begin{tabular}{cc}
  %[trim={left bottom right top}

  \includegraphics[trim={0cm 2cm 0cm 2.cm},clip,width=0.47\linewidth]{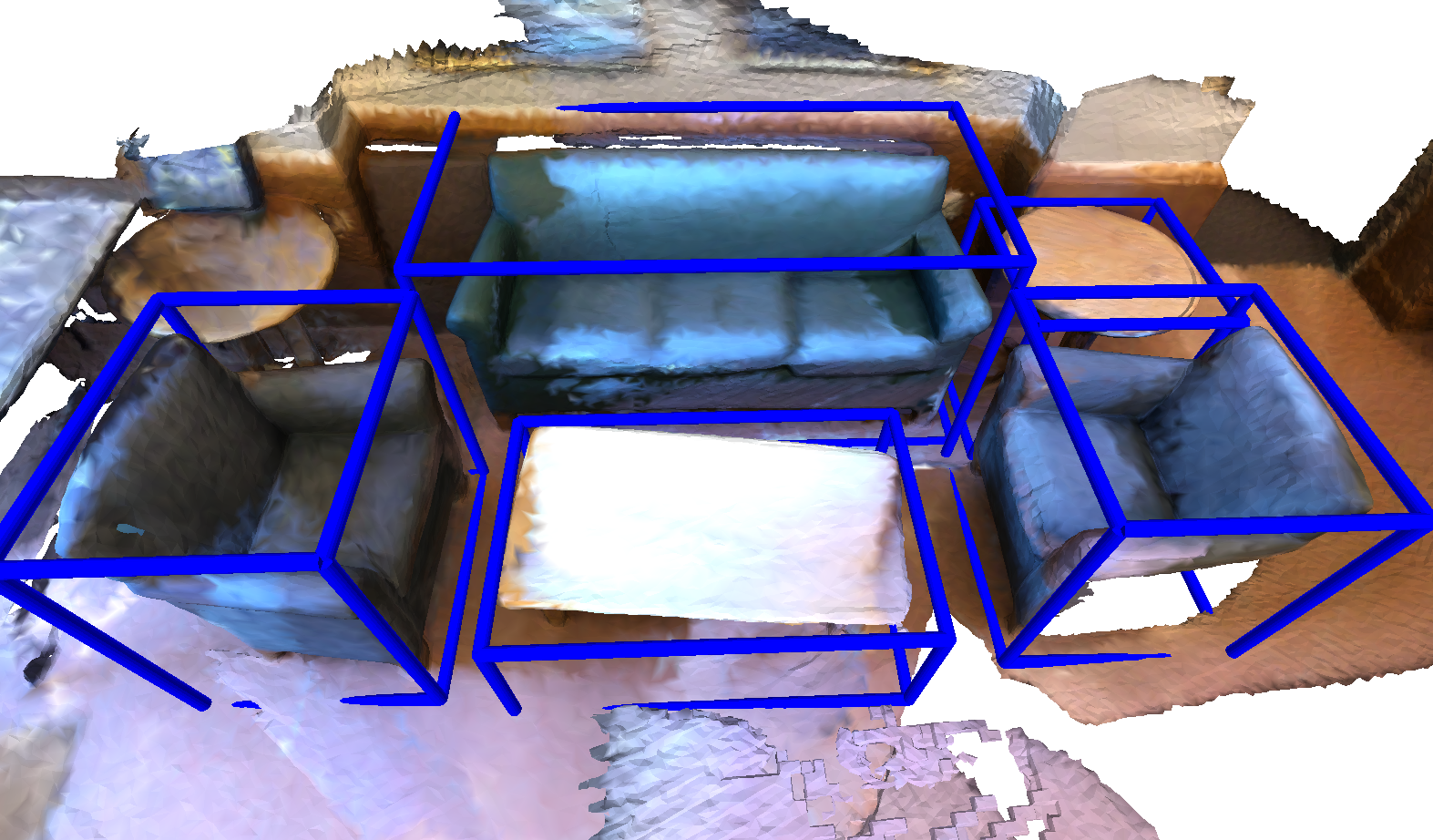} &
  \includegraphics[trim={0cm 2cm 0cm 2.cm},clip,width=0.47\linewidth]{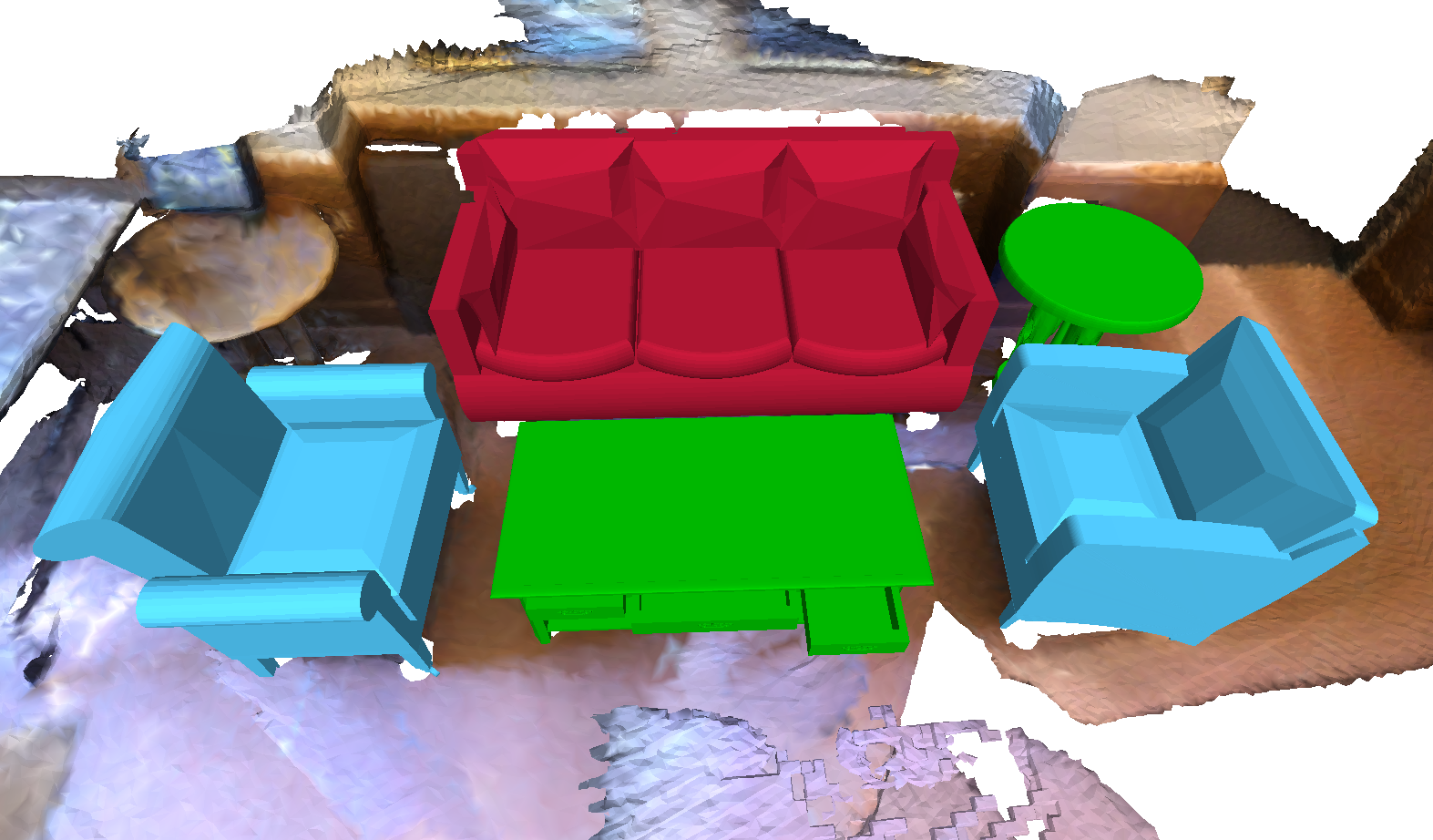} \\
  \begin{tabular}[c]{@{}c@{}}RGB-D scan with \\ initial 3D boxes \end{tabular} &
  \begin{tabular}[c]{@{}c@{}}CAD models retrieved \\ with exh. search \end{tabular} \\
  \includegraphics[trim={0cm 2cm 0cm 2.4cm},clip,width=0.47\linewidth]{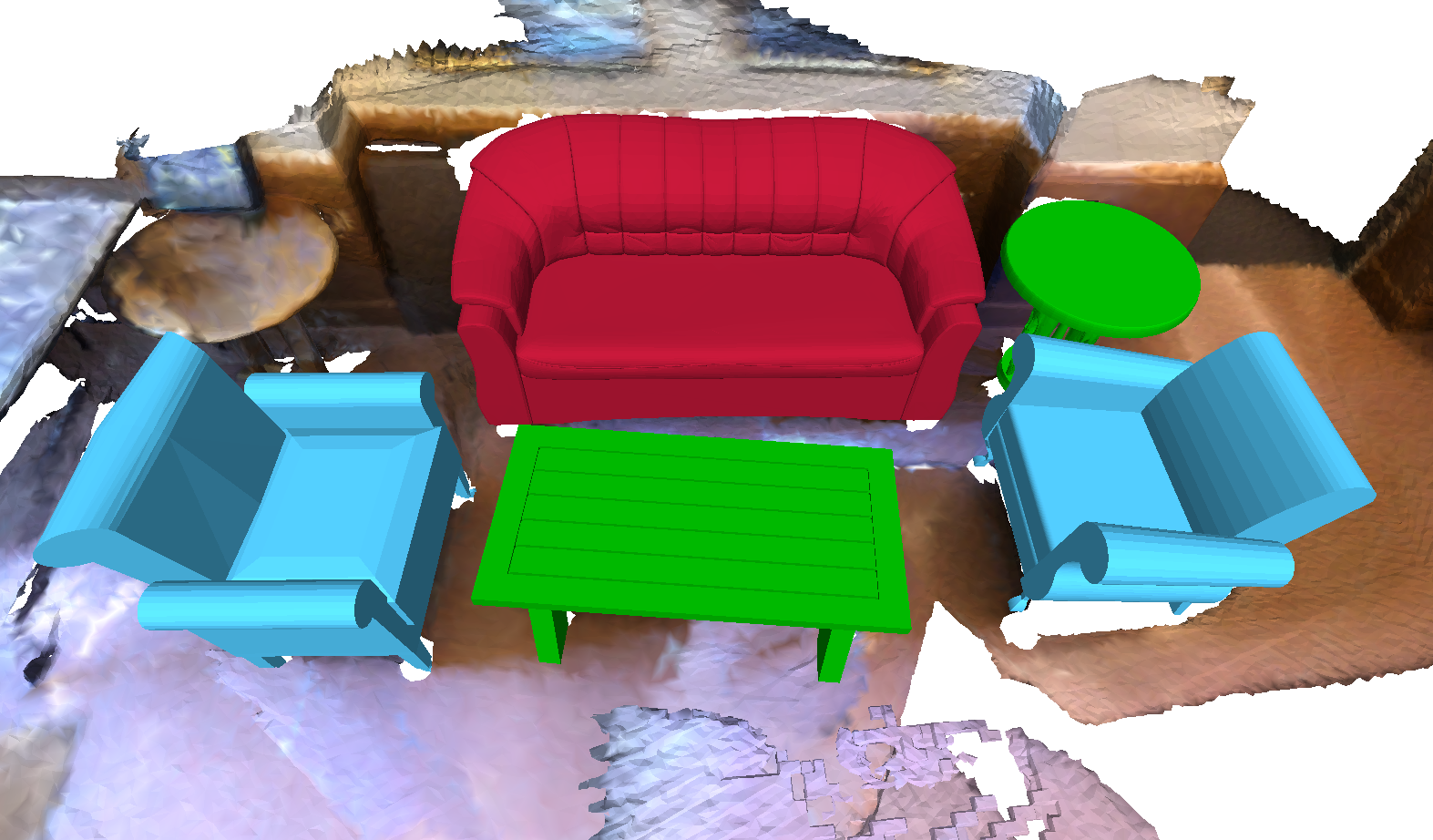} &
  \includegraphics[trim={0cm 2cm 0cm 2.15cm},clip,width=0.47\linewidth]{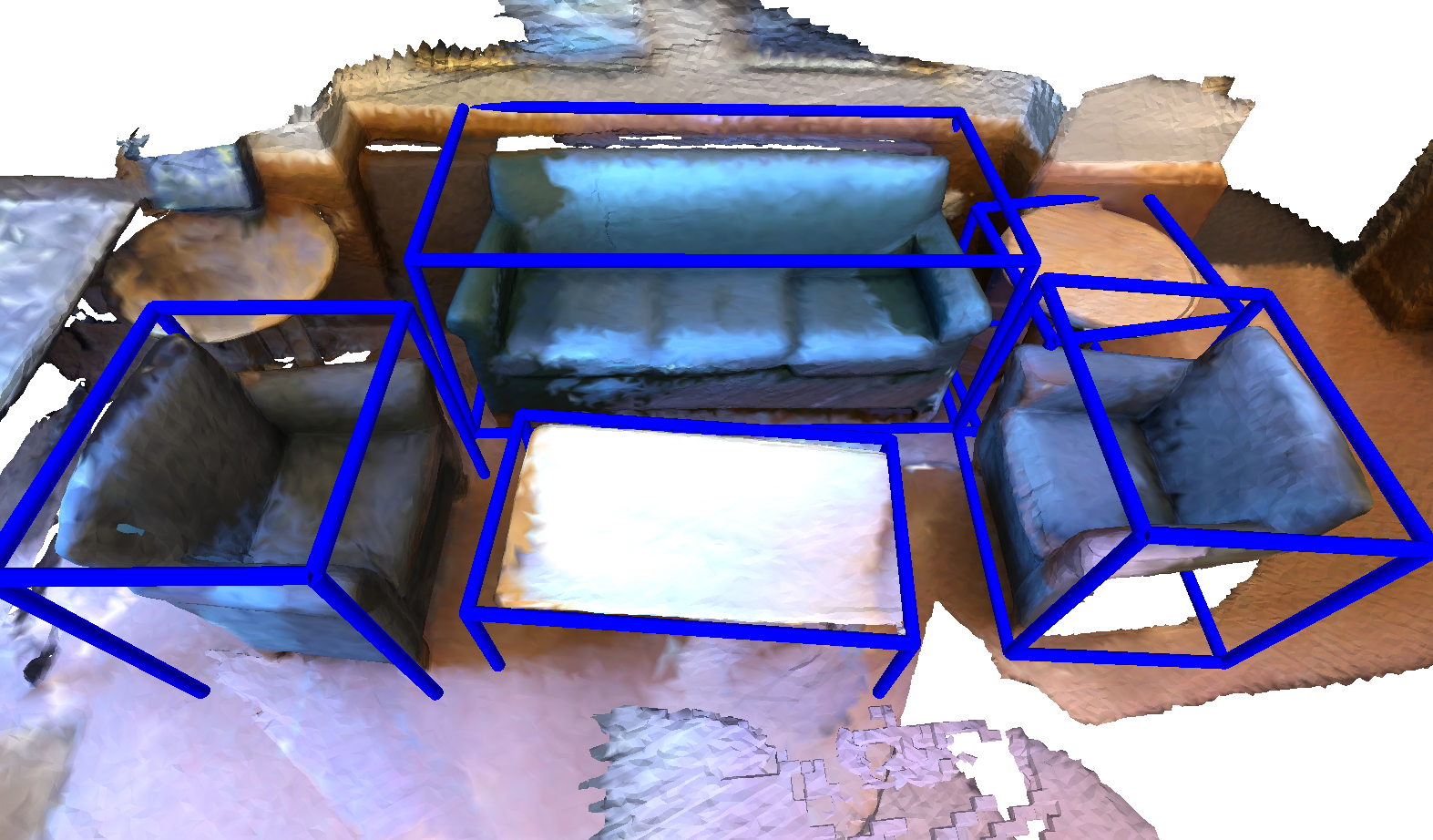} \\
  \begin{tabular}[c]{@{}c@{}}CAD models from \\ HOC-Search + refinement \end{tabular} & 
  \begin{tabular}[c]{@{}c@{}}RGB-D scan with \\ refined 3D boxes \end{tabular}
\end{tabular}
}
\vspace{-0.2cm}
\caption{Results for HOC-Search \textbf{with} refinement. Due to the inaccuracy of the initial 3D boxes, exhaustive search can fail to retrieve a good CAD model, see the front table. Compared to HOC-Search, the simultaneous refinement makes it possible to retrieve suitable CAD models from inaccurate boxes. We observe that the refined 3D boxes fit the target objects much better. More visualizations are provided in the supplementary material.% in Section~\ref{sec:supp_softroup_exp}.
%The supplementary material provides additional visualizations.
}
    \label{fig:quali_results_eval}
\end{figure}

We also evaluate the performance of our HOC-Search with pose refinement for scenarios when manually annotated object poses are not available. As shown in Figure~\ref{fig:teaser}~(bottom), we use the SoftGroup method~\cite{Vu_2022_CVPR} to obtain 3D semantic instance segmentation, which we use to extract axis-aligned 3D bounding boxes for the target objects. Note that this setup makes our approach for CAD model retrieval from RGB-D scans completely automated.

%\paragraph{3D Box proposal generation from instance segmentation.}
\textbf{3D box proposal generation from instance segmentation.} The 3D bounding boxes obtained from the SoftGroup segmentation  are less accurate than the ground-truth Scan2CAD annotations used in previous experiments. For exhaustive search, we opt to rotate the 3D box of the target object around the vertical axis by $[0^\circ,45^\circ,90^\circ,135^\circ,180^\circ,225^\circ,270^\circ,315^\circ]$. For our approach, we keep the four pose nodes with $[0^\circ,90^\circ,180^\circ,270^\circ]$ and perform a rotation around $45^\circ$ directly on the selected CAD model. We empirically observed that this ensures high retrieval accuracy, while adding only minimal computation complexity.

%\paragraph{Evaluation and results.}
\textbf{Evaluation and results.} %For evaluation, a
As SoftGroup detections might differ significantly to Scan2CAD annotations, we first associate SoftGroup predictions to Scan2CAD annotations for evaluation. A retrieved CAD model is assigned to a Scan2CAD annotation if the IoU of their 3D boxes is larger than 0.6 and if they have the same category label. 
% We then measure the chamfer distance between the retrieved model and its assigned annotation. 
Results in Table~\ref{tab:eval_softgroup} confirm our expectations. Performance of the exhaustive search baseline drops, as due to incorrect initial poses, it is difficult to retrieve correct models without pose refinement. By using HOC-Search with refinement, we are even able to outperform exhaustive search in terms of retrieval accuracy. This is also highlighted in Figure~\ref{fig:quali_results_eval}, where our approach is able to retrieve a better CAD model for the front table and further refines its pose. Note that for the exhaustive search it is computationally infeasible to perform refinement for every candidate CAD model in the database.

\vincentrmk{let's see what you get:} \stefanrmk{I added the results to the supplementary material}

\vincent{To show the generalization capabilities of our approach, we also used our pipeline on two scenes from \cite{hampali2021monte}; results are provided in the supplementary material.
}

\subsection{HOC-Search with Category Search}
\vincent{
We can also use our approach to estimate the object category, using the complete tree structure shown in Figure~\ref{fig:fulltree}.
Qualitative results are shown in Figure~\ref{fig:supp_unknown_cat_v3}. For this experiment, the exhaustive search would have to perform $24054 \times 4 = 96216$ iterations of render-and-compare ($24054$ is the number of all CAD models from the considered categories, and the 4x factor is due to the four pose nodes) to retrieve one CAD model. For comparison, $800$ iterations of HOC-Search already lead to CAD models that fit the target objects very well. This results in approx.~\textbf{120x} less render-and-compare iterations. We observe that, due to geometric similarities between different categories,
%\eg Bookshelf and Cabinet, the exhaustive search and our 
HOC-Search sometimes retrieves objects from different categories compared to the ground-truth but of similar geometry.}

% , we use property nodes for category labels as an additional level in the HOC-Tree. 

% %We use the main category labels from ScanNet, namely $\{$\textit{Table, Chair, Sofa, Cabinet, Bed, Bookshelf, Lamp, Display}$\}$. 
% We combine HOC-Trees for individual categories into a single large HOC-Tree as explained in Section~\ref{sec:hoc_eff_eval}. Again, we use the initial 3D bounding box from Scan2CAD annotations, without knowing the orientation of the object inside this box, and we perform a category-agnostic CAD model retrieval.

% Note that HOC-Search does not always retrieve the same category label as in the ground-truth, however, this is mainly due to ambiguities of category labels (e.g. the categories Chair and Sofa share very similar CAD models, the same applies to Bookshelf and Cabinet).

%\input{supp_unknown_catFigure}

\definecolor{chair_color}{rgb}{0.301, 0.745, 0.933}
\definecolor{cabinet_color}{rgb}{0.494, 0.184, 0.556}
\definecolor{bed_color}{rgb}{0.466, 0.674, 0.188}
\definecolor{sofa_color}{rgb}{0.635, 0.078, 0.184}
\definecolor{table_color}{rgb}{0.000, 0.667, 0.000}
\definecolor{bookshelf_color}{rgb}{0.000, 0.000, 1.000}
\definecolor{lamp_color}{rgb}{1.000, 0.333, 0.500}
\definecolor{display_color}{rgb}{0.333, 0.667, 1.000}

\begin{figure}
\centering
\scalebox{0.99}{
\begin{tabular}{ccc}
%[trim={left bottom right top}
\hspace{-0.4cm}
 \includegraphics[trim={8cm 3cm 8cm 4cm},clip,width=0.33\linewidth]{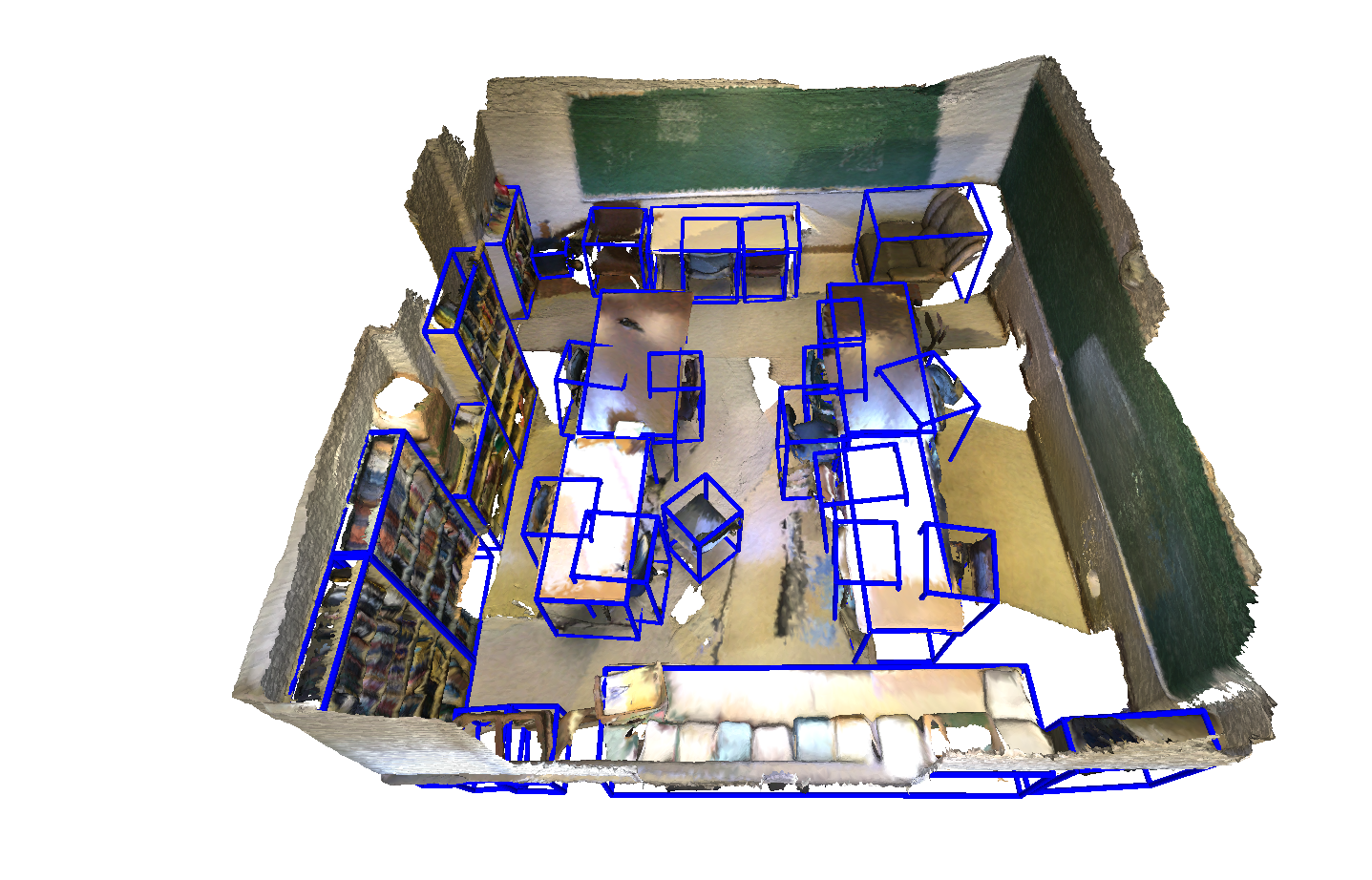} & \hspace{-0.4cm}
 \includegraphics[trim={8cm 3cm 8cm 4cm},clip,width=0.33\linewidth]{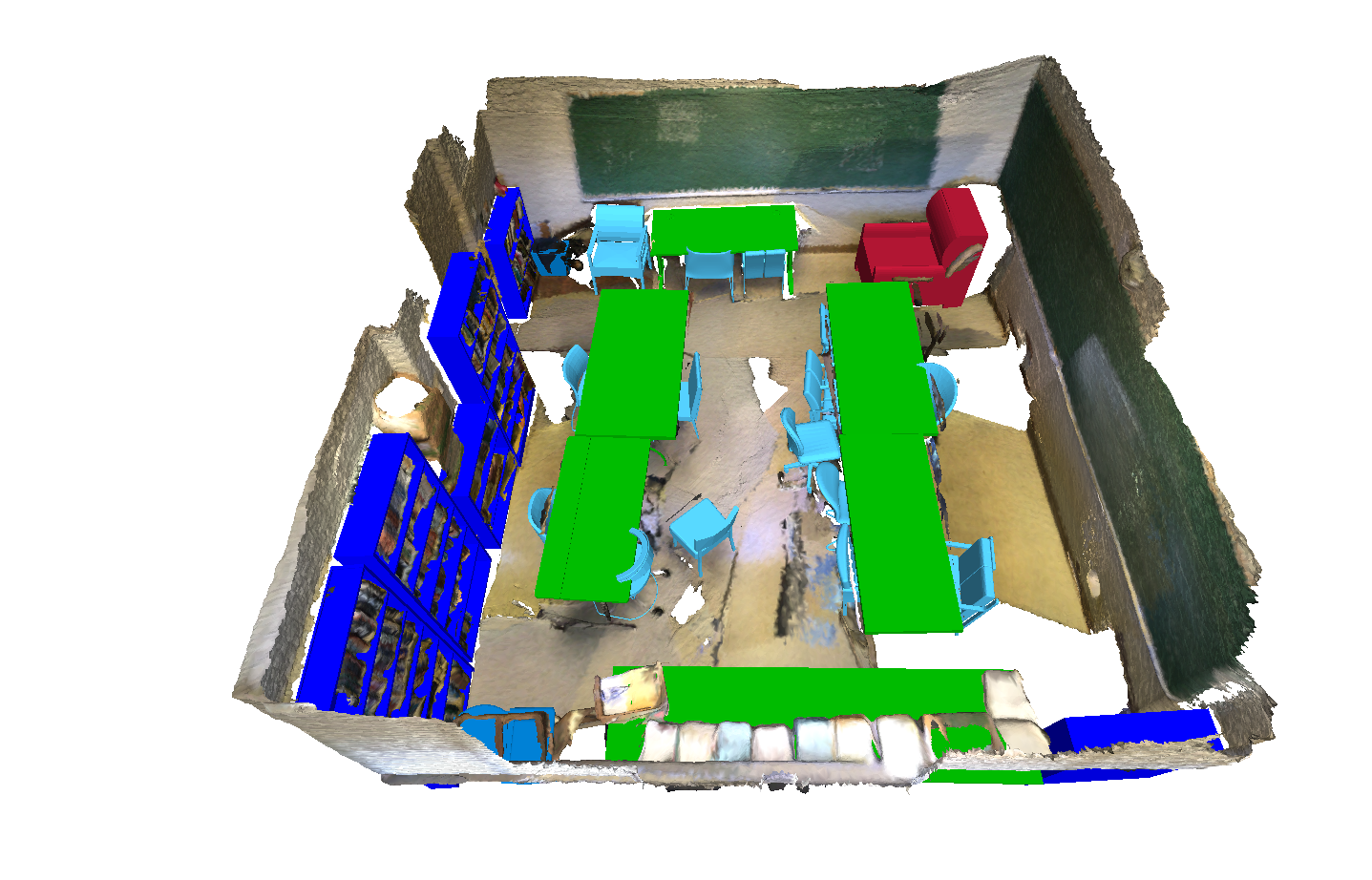} &  \hspace{-0.4cm}
 \includegraphics[trim={8cm 3cm 8cm 4cm},clip,width=0.33\linewidth]{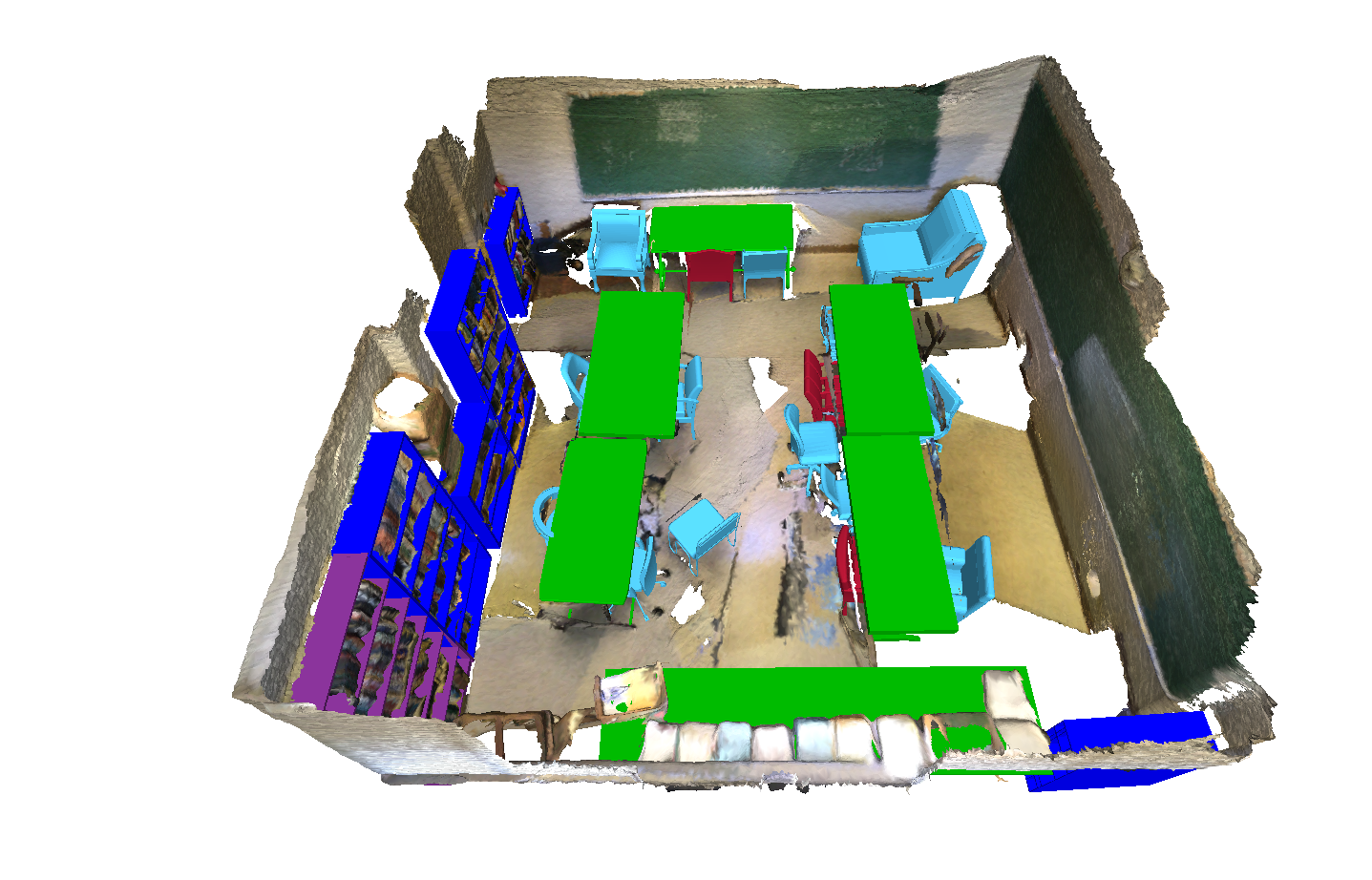}  \\\hspace{-0.4cm}
\includegraphics[trim={8cm 4cm 10cm 5cm},clip,width=0.33\linewidth]{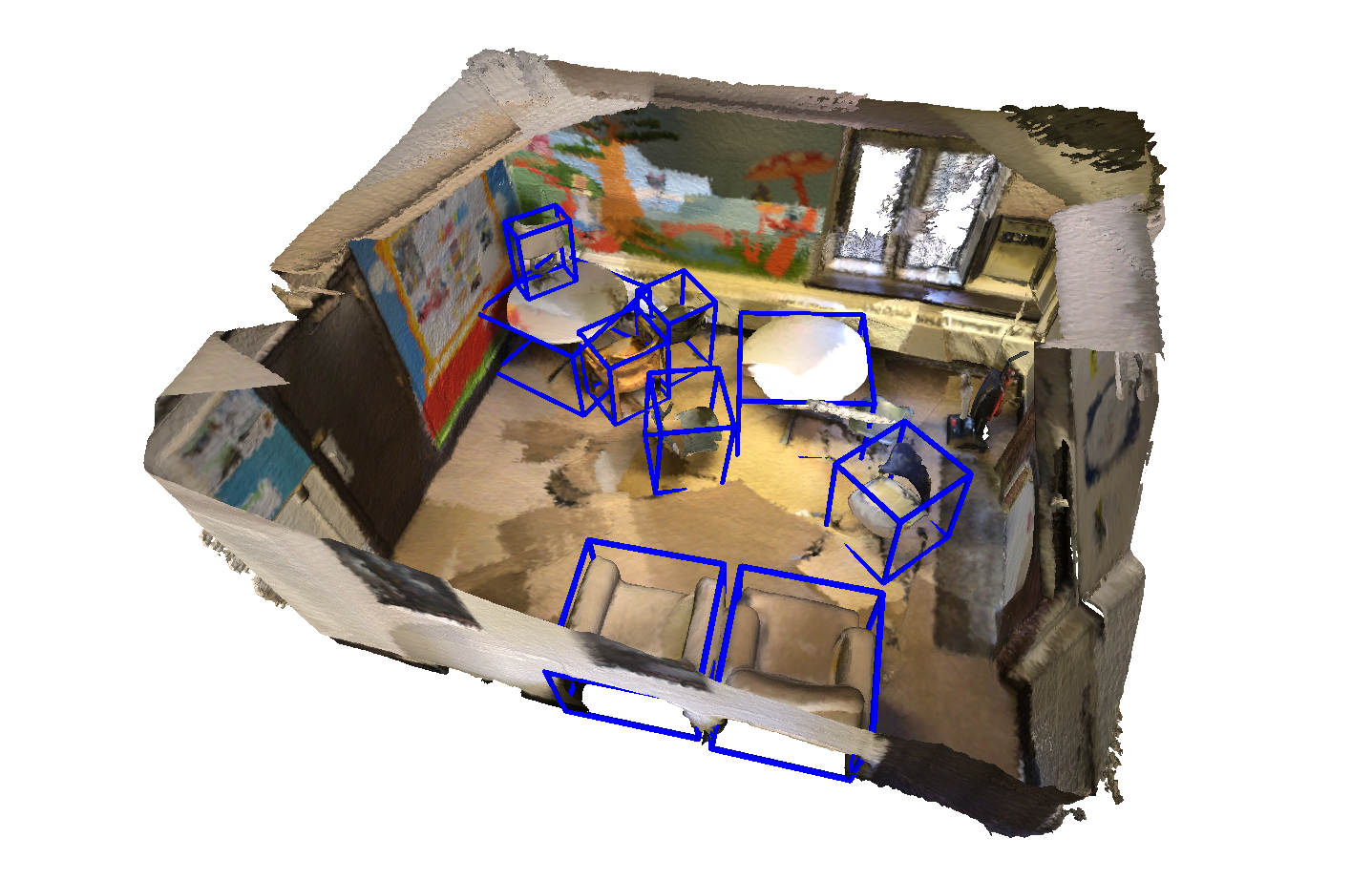} &  \hspace{-0.4cm}
  \includegraphics[trim={8cm 4cm 10cm 5cm},clip,width=0.33\linewidth]{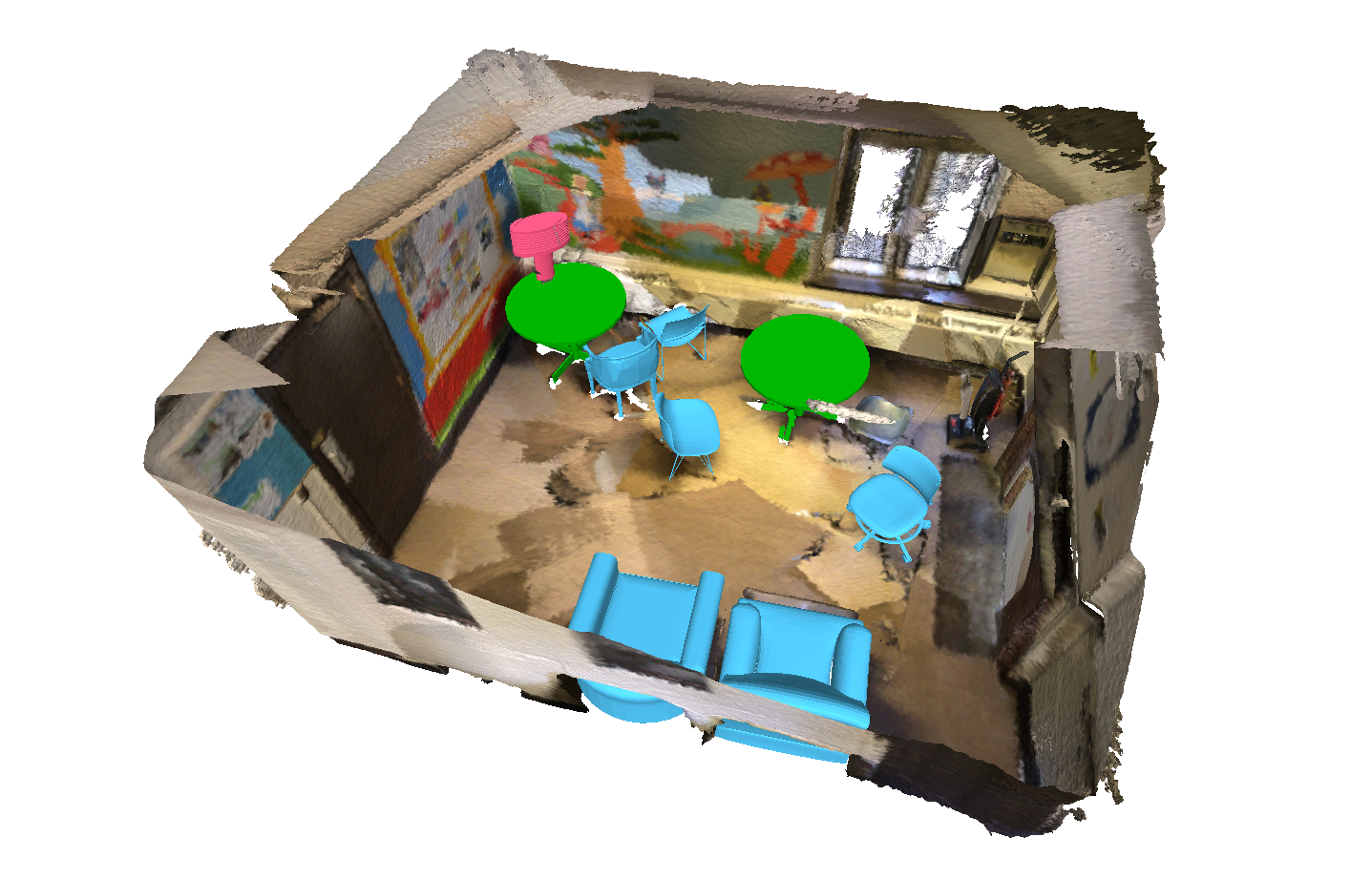} &  \hspace{-0.4cm}
   \includegraphics[trim={8cm 4cm 10cm 5cm},clip,width=0.33\linewidth]{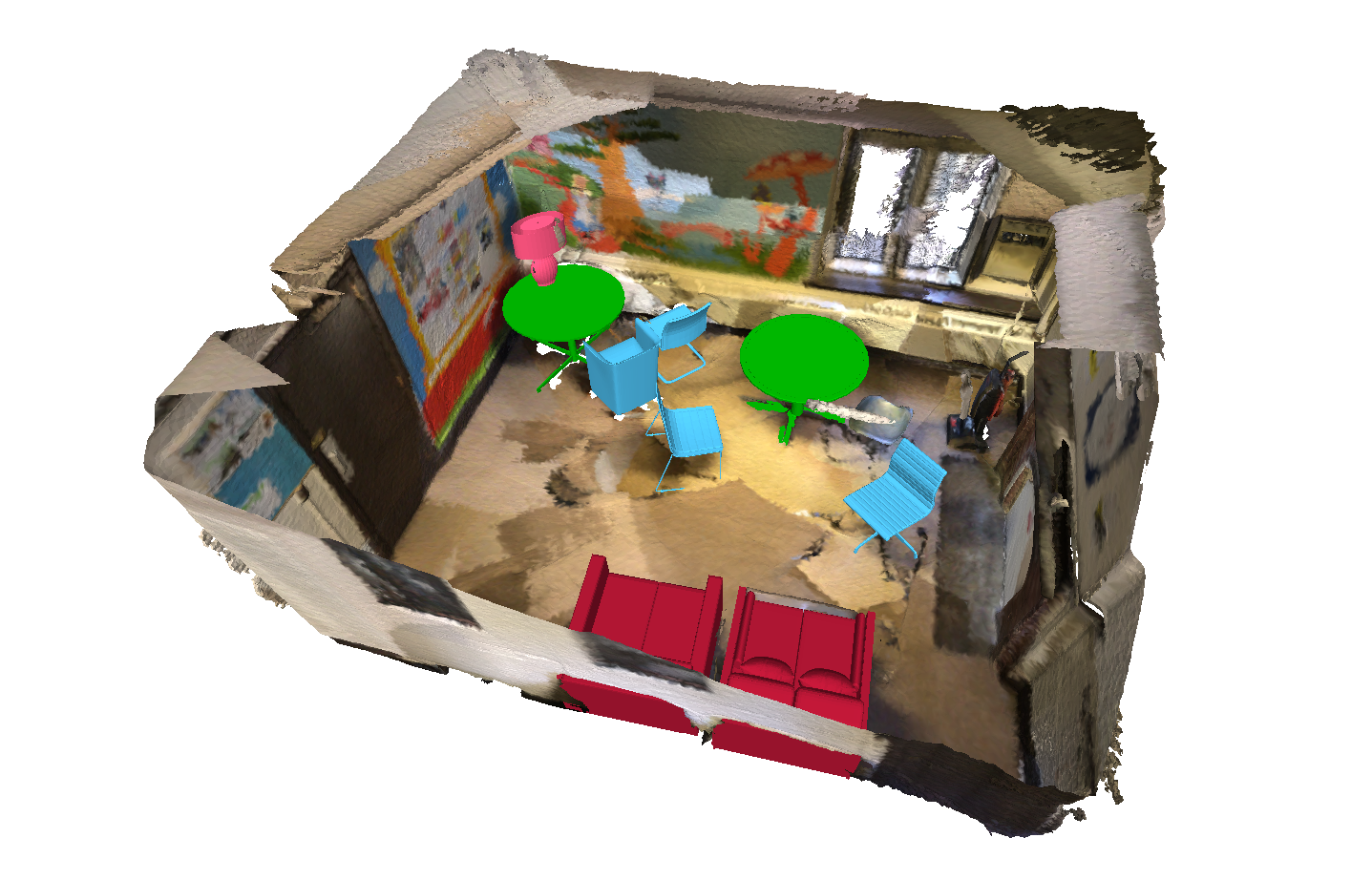} \\

% \hspace{-0.4cm} \includegraphics[trim={7cm 3cm 7cm 4cm},clip,width=0.33\linewidth]{figures/results_HOC_unknown_cat/scene0435_00/scene_boxes.png} & \hspace{-0.4cm}
%\includegraphics[trim={7cm 3cm 7cm 4cm},clip,width=0.33\linewidth]{figures/results_HOC_unknown_cat/scene0435_00/exh_search.png} & \hspace{-0.4cm}
%\includegraphics[trim={7cm 3cm 7cm 4cm},clip,width=0.33\linewidth]{figures/results_HOC_unknown_cat/scene0435_00/hoc_800.png} \\

% \hspace{-0.4cm}
%\includegraphics[trim={8cm 4cm 8cm 4.5cm},clip,width=0.33\linewidth]{figures/results_HOC_unknown_cat/scene0609_01/scene_boxes.png} & \hspace{-0.4cm}
%\includegraphics[trim={8cm 4cm 8cm 4.5cm},clip,width=0.33\linewidth]{figures/results_HOC_unknown_cat/scene0609_01/exh_search.png} & \hspace{-0.4cm}
%\includegraphics[trim={8cm 4cm 8cm 4.5cm},clip,width=0.33\linewidth]{figures/results_HOC_unknown_cat/scene0609_01/hoc_800.png} \\

%     RGB-D Scan with &  Exh. search \textbf{with} & HOC-Search w\/o  \\
%   initial 3D boxes         &   \textbf{category label}   &   category label\\
   
\end{tabular}
}
\vspace{-.3cm}
\caption{
Results for CAD model, pose, and category retrieval.
\textbf{Left}: Input scan with 3D boxes from Scan2CAD. \textbf{Middle}: CAD models retrieved with exhaustive search \textbf{and} using the ground-truth category label. \textbf{Right}: Results of HOC-Search for unknown category labels for 800 iterations. Although the complexity for retrieval highly increases with unknown category labels, HOC-Search finds suitable CAD models and predicts correct object poses. Note that HOC-Search does not always retrieve the same category label as in the ground truth, however, this is mainly due to ambiguities of category labels (e.g., categories Chair and Sofa share very similar CAD models). The color of the objects refers to the category label: \textcolor{chair_color}{Chair}, \textcolor{cabinet_color}{Cabinet}, %\textcolor{bed_color}{Bed},
\textcolor{sofa_color}{Sofa}, \textcolor{table_color}{Table}, \textcolor{bookshelf_color}{Bookshelf}, \textcolor{lamp_color}{Lamp}.
%, \textcolor{display_color}{Display}.
}
\label{fig:supp_unknown_cat_v3}
\end{figure}

\section{Conclusion}
\vincent{Our method offers a practical solution for efficiently retrieving CAD models with pose and category label, for objects in point clouds. By combining our method with off-the-shelf object detection techniques like SoftGroup, we achieve precise and fully automated annotations.

We see a strong potential in the concept of property nodes: They could include finer search for initial 3D bounding boxes, object symmetries and scales to perform more complex retrieval. Moreover, we believe that our approach to fast retrieval is general, as our method does not rely on training and could be applied to other problems involving complex objective functions.
}

{
    \small
    \bibliographystyle{ieeenat_fullname}
    \bibliography{cleaned}
}

\clearpage
\setcounter{page}{1}
\maketitlesupplementary

%\textvars{dpt,Sil,CD,MSCD,cad,msh,sns}
%\textvars{cad,msh,sns,Sil}
\textvars{cad,msh,sns}

\section{CAD Model and Pose Retrieval in the Wild}
\label{sec:supp_retrieval_custom_scenes}

\vincent{
We show the generalization capabilities of our approach by applying it together with SoftGroup on the two scans provided by the authors of \cite{hampali2021monte} and which were captured independently from ScanNet. Since we used SoftGroup for getting the 3D bounding boxes, our pipeline is fully automated.

Results are shown in Figure~\ref{fig:supp_softgroup_results_custom}. The retrieved CAD models fit the captured scan very well. These results show that our method generalizes well for common indoor scenes, hence it can easily be used for CAD model and pose retrieval using custom data.

\begin{figure*}
\centering
\scalebox{0.9}{
\begin{tabular}{cccc}
%[trim={left bottom right top}

\includegraphics[trim={4.5cm 1cm 1.5cm 0.5cm},clip,width=0.25\linewidth]{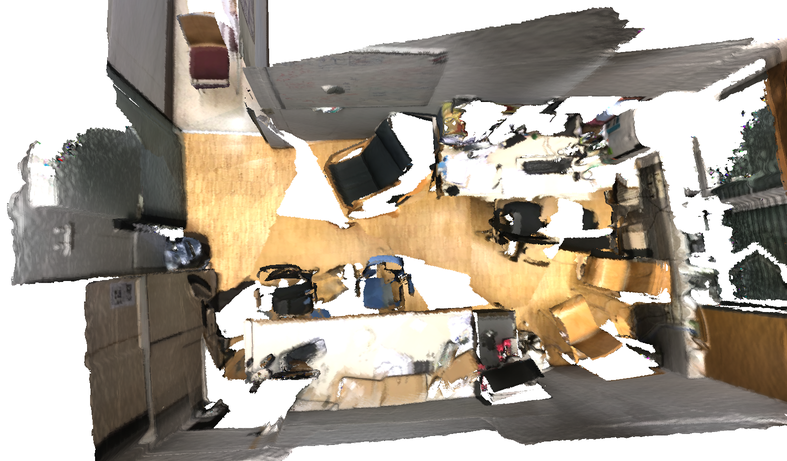} &
 \includegraphics[trim={5.5cm 2.5cm 5.5cm 2.5cm},clip,width=0.25\linewidth]{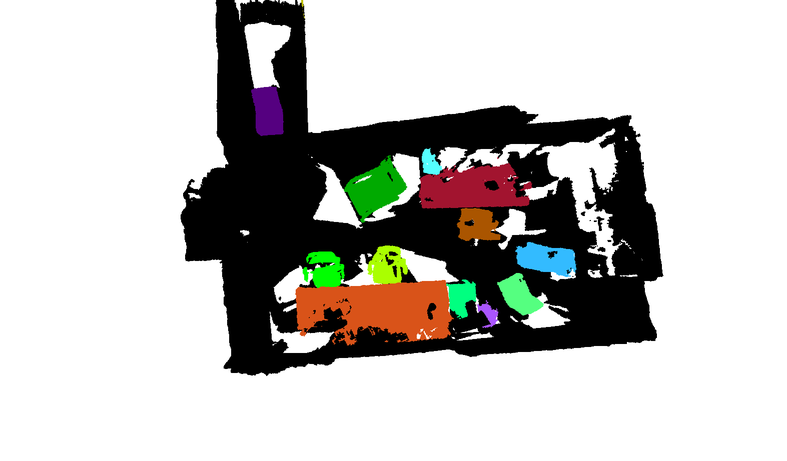} &
 \includegraphics[trim={4.5cm 1cm 1.5cm 0.5cm},clip,width=0.25\linewidth]{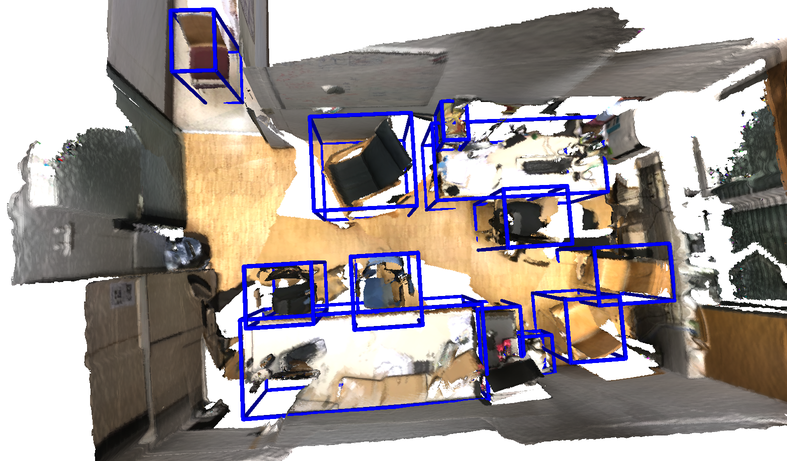}  &
  \includegraphics[trim={4.5cm 1cm 1.5cm 0.5cm},clip,width=0.25\linewidth]{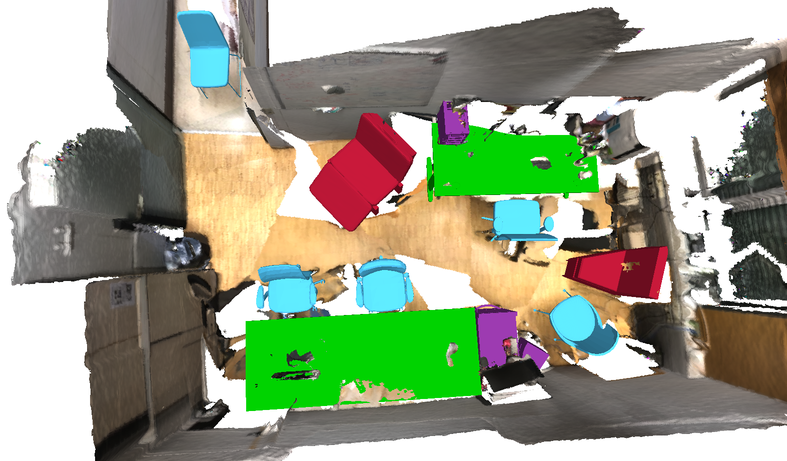} \\

\includegraphics[trim={4.5cm 0.5cm 4.5cm 0.5cm},clip,width=0.25\linewidth]{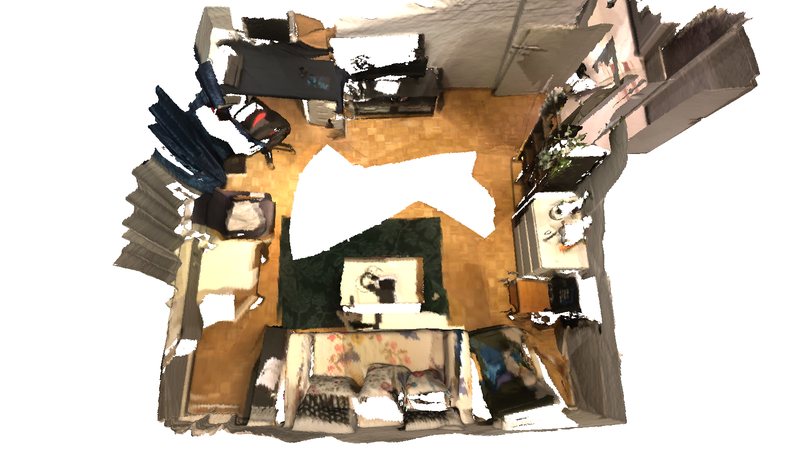} &
 \includegraphics[trim={4.5cm 0.5cm 4.5cm 0.5cm},clip,width=0.25\linewidth]{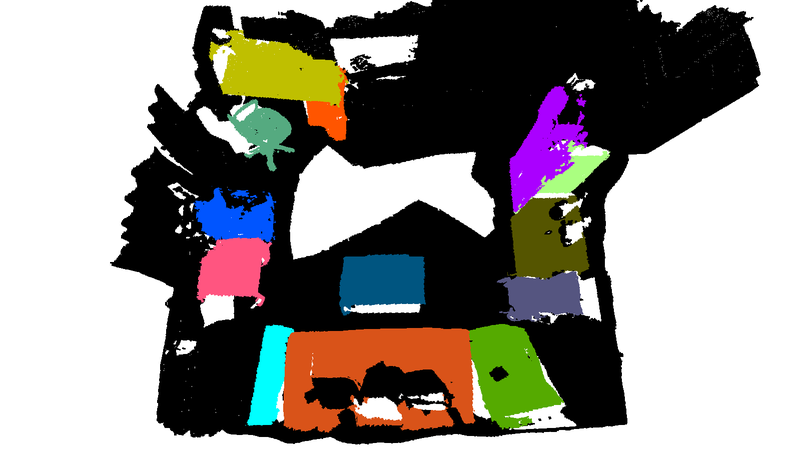} &
 \includegraphics[trim={4.5cm 0.5cm 4.5cm 0.5cm},clip,width=0.25\linewidth]{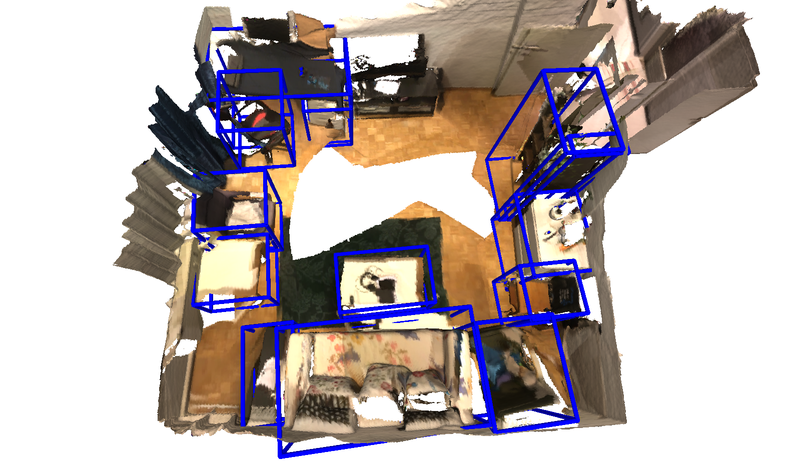}  &
  \includegraphics[trim={4.5cm 0.5cm 4.5cm 0.5cm},clip,width=0.25\linewidth]{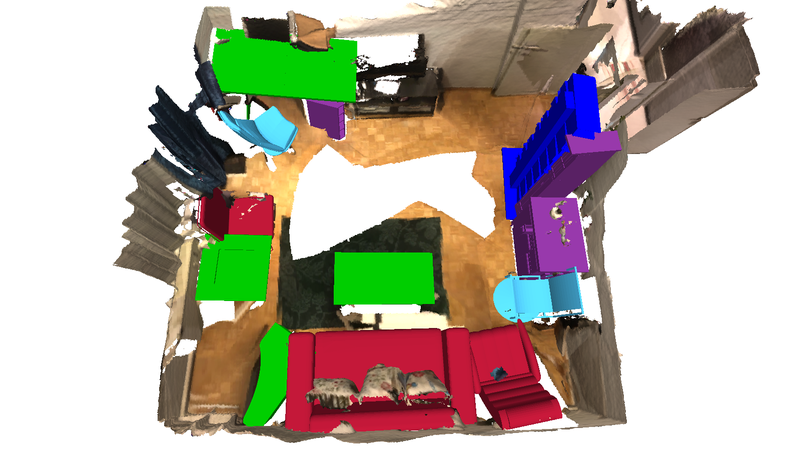} \\

  RGB-D Scan  &  Segmentation & Axis-aligned  & CAD models from %& RGB-D Scan with
  \\
              &   by SoftGroup   &      bounding boxes    & HOC-Search + refinement %& refined boxes
              \\
              &      &          & (800 iterations)% & 
              \\
\end{tabular}
}
\caption{
\stefan{Qualitative results for CAD model and pose retrieval using HOC-Search with refinement on scans from \protect\cite{hampali2021monte}.  SoftGroup generalizes well to common indoor scenes, which enables us to calculate an approximation for the initial 3D bounding boxes of the objects in the scene. Our results for CAD model and pose retrieval fit the target objects very well, as our refinement step helps to significantly improve the initial pose, and hence also the accuracy of CAD model retrieval.}
}
\label{fig:supp_softgroup_results_custom}
\end{figure*}

}

\section{Additional Information About the Objective Functions}
\label{sec:supp_details_objective_function}
\stefan{
Hereafter, we provide additional information about the objective functions considered for our experiments.
}

\subsection{Render-and-Compare}
As proposed in~\cite{ainetter2023automatically}, the render-and-compare objective function is defined as 
\begin{equation}
\calL_{RC} = \calL_\dpt + \lambda_\Sil \calL_\Sil + \lambda_\CD \calL_\CD \>,
\end{equation}
where $\calL_\dpt$ is a depth matching term defined as the L1-distance between depth maps before and after replacing the target object with a CAD model. $\calL_\Sil$ defines the Intersection-over-Union between silhouettes of target object and CAD model and $\calL_\CD$ defines the single-direction chamfer distance, with $\lambda_\Sil$ and $\lambda_\CD$ the corresponding weights.

\paragraph{Depth matching term.} The depth matching term $\calL_\dpt$ is defined as
\begin{equation}
  \label{eq:Ldpt}
\begin{aligned}
  \calL_\dpt = 
  \frac{1}{N_{T}} \sum_t \Bigl( &\frac{\lambda_m}{V_m^t} |M_m^t\cdot(D_\cad^t - D_\msh^t)|_1 \;+ \\
& \frac{\lambda_s}{V_s^t} |M_s^t\cdot(D_\cad^t - D_\sns^t) |_1 \Bigr) \> ,
\end{aligned}
\end{equation}
where $N_{T}$ defines the number of selected frames of the RGB-D scan for the target object.
Depth maps $D_\msh^t$ and $D_\cad^t$ are respectively the depth maps rendered from the 3D scan before and after replacing the target object with the CAD model for frame $t$. $D_\sns^t$ denotes the captured sensor depth map for frame $t$.
$M_m^t$ and $M_s^t$ denote the valid pixel maps for the depth maps $D_\msh^t$ and $D_\sns^t$, respectively. The L1 norm is used to compare the depth maps, and the norms are then divided by the numbers of valid pixels $V_m^t$ and $V_s^t$ for normalization. $\lambda_m$ and $\lambda_s$ are specific weights depending on the quality of the captured depth maps and the 3D scan.
\paragraph{Silhouette matching term.}
$\calL_\Sil$ denotes the Intersection-over-Union between the silhouettes of the target object and the CAD model:
\begin{equation}
\calL_\Sil = \frac{1}{N_{T}} \sum_t (1-\text{IoU}(S_\msh^t,  S_\cad^t)) \> ,
\label{eq:supp_loss_sil}
\end{equation}
where $S_\msh^t$ and $S_\cad^t$ are the rendered silhouettes for the target object and the CAD model, respectively, for frame $t$. 

\paragraph{Single-direction chamfer distance.}
The single-direction chamfer distance from the points of the target object to the CAD model is defined as
\begin{equation}
\calL_\CD = \frac{1}{|P|} \sum_{p \in P} \min_{q \in Q}  \|p - q\| \> , 
\end{equation}
where $P$ is the point cloud of the target object and $Q$ is a set of 3D points sampled on the surface of the  CAD model.

\paragraph{Implementation details.} All renderings are generated using the rendering pipeline of~\cite{ravi2020pytorch3d}. We used the same procedure as~\cite{ainetter2023automatically} to predict the selected frames $N_{T}$ for each target object, to replace the target object in the scene with a CAD model, and to generate the renderings for the objective function. Throughout our experiments, we used $N_{T}=14$ frames, with $\lambda_m=0.6$, $\lambda_s=1.0$, $\lambda_\Sil=0.5$ and $\lambda_\CD=2.0$. For calculating $\calL_\CD$, we uniformly sampled 10k points from the surfaces of the target object and the surfaces of the CAD model. 

\subsection{Modified Single-direction Chamfer Distance}
The authors of \cite{wei2022accurate} proposed the Modified Single-direction Chamfer Distance (MSCD) as objective function for CAD model retrieval which is defined as
\begin{equation}
\MSCD = \frac{1}{|P|} \sum_{p \in P} \min_{q \in Q}  \|p - q\|_2 \> .
\end{equation}

MSCD considers only the distance of points from a scanned object $P$ to the points from the CAD model $Q$, which increases robustness for incomplete point clouds by focusing on visible parts of the scanned object and ignoring unobserved/missing parts.

\subsection{Nearest Neighbor Search in Embedding Space}
Learned 3D shape descriptors can be used to efficiently encode information about geometric properties of objects in the embedding space.
Let $E(\cdot)$ denote an encoder network to extract a feature embedding for a given object. Using this encoder it is possible to calculate a feature embedding for each CAD model $x_{i}$ from the CAD model database $X = \{x_{i}, 1 \leq i \leq S\}$, where $S$ defines the number of CAD models. Nearest neighbor search in embedding space for a given target object $y$ can then be defined as 
\begin{equation}
    NN = \min_{x_{i} \in X}  \|E(y) - E(x_{i})\|_2 \> .
\end{equation}
For our experiments, we use the encoder network from~\cite{cai2020learning} which is based on PointNet~\cite{qi2017pointnet}. The network is trained on ShapeNet~\cite{chang2015shapenet} for the task of 3D shape reconstruction for point clouds. Note that several other learned feature embeddings can be effectively used for extracting discriminative features of shapes of objects, as shown in~\cite{wei2022accurate}.

\section{Qualitative Results for Comparison of Objective Functions}
\label{sec:supp_results_objective_function}
\begin{figure*}
\centering
\scalebox{0.84}{
\begin{tabular}{ccccc}
%[trim={left bottom right top}

  \includegraphics[trim={4cm 2.5cm 0cm .5cm},clip,width=0.22\linewidth]{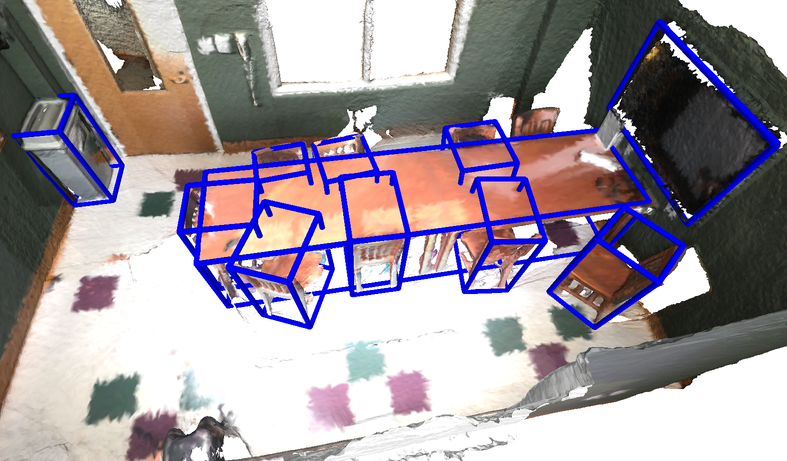} &
  \includegraphics[trim={4cm 2.5cm 0cm .5cm},clip,width=0.22\linewidth]{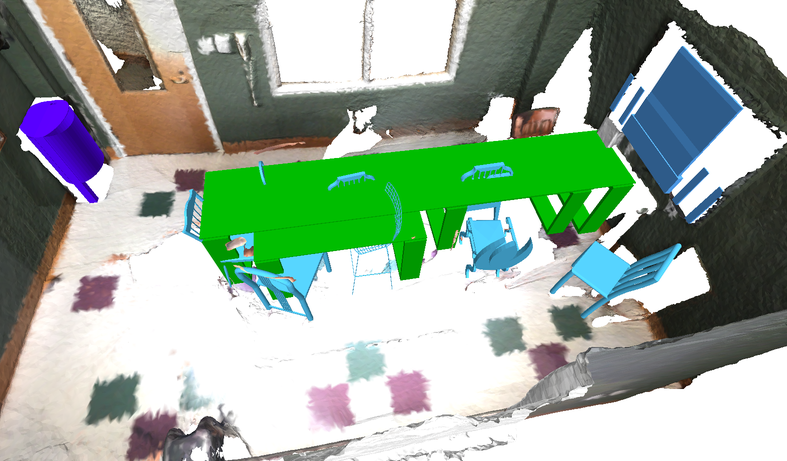} &
  \includegraphics[trim={4cm 2.5cm 0cm .5cm},clip,width=0.22\linewidth]{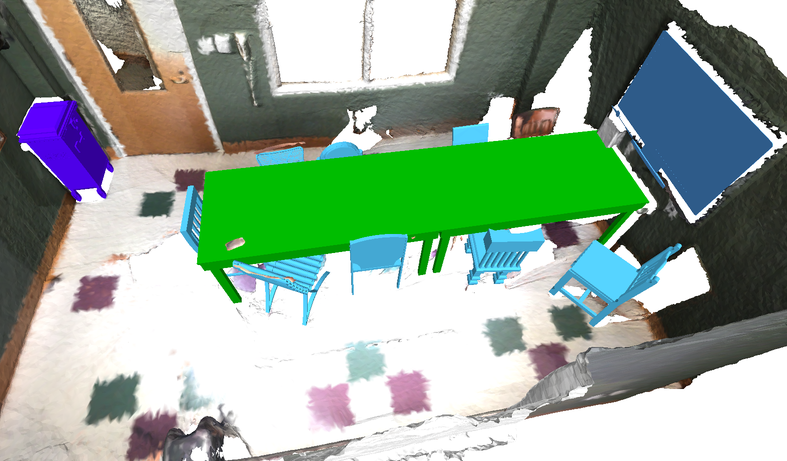} &
  \includegraphics[trim={4cm 2.5cm 0cm .5cm},clip,width=0.22\linewidth]{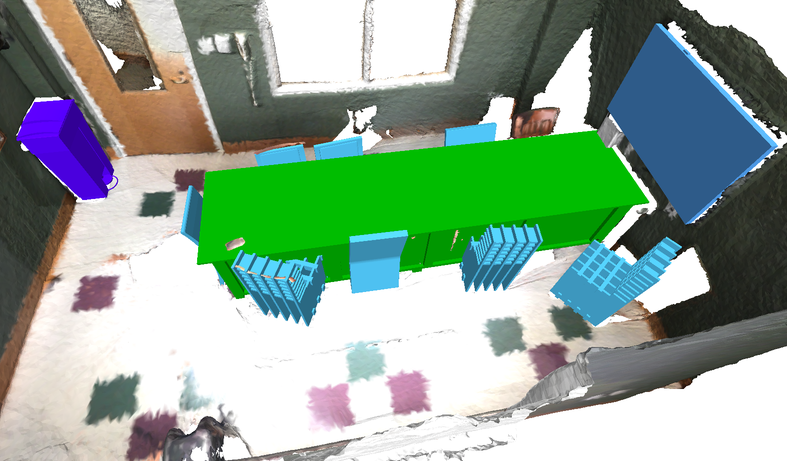} &
  \includegraphics[trim={4cm 2.5cm 0cm .5cm},clip,width=0.22\linewidth]{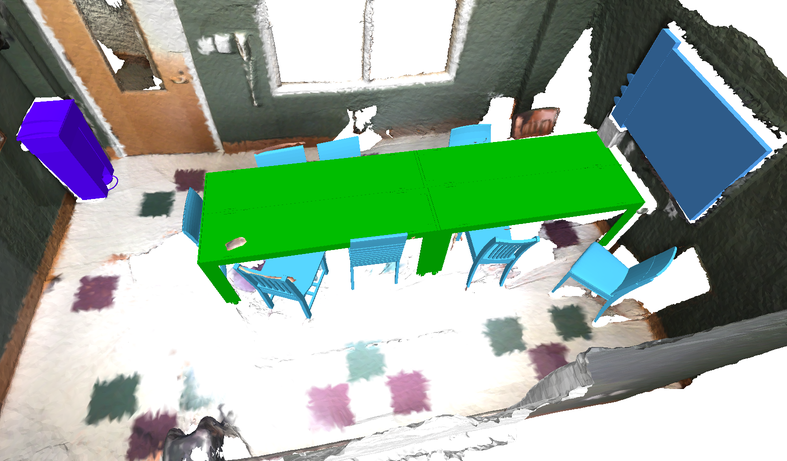} \\

  \includegraphics[trim={3.4cm 1cm 2.5cm 0.cm},clip,width=0.22\linewidth]{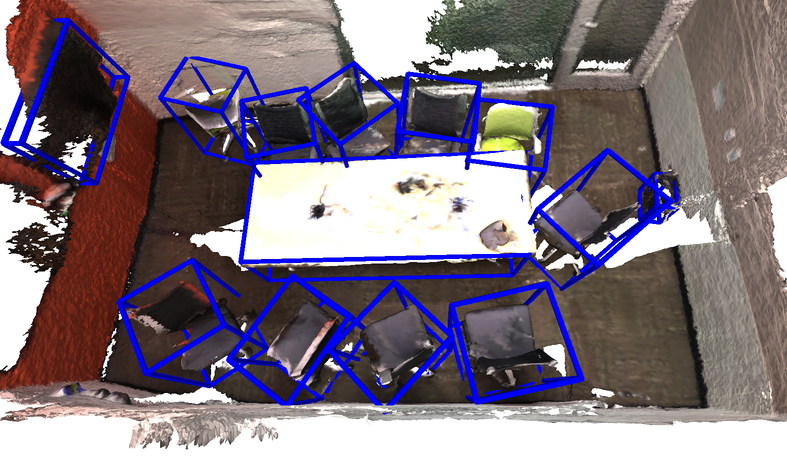} &
  \includegraphics[trim={3.4cm 1cm 2.5cm 0.cm},clip,width=0.22\linewidth]{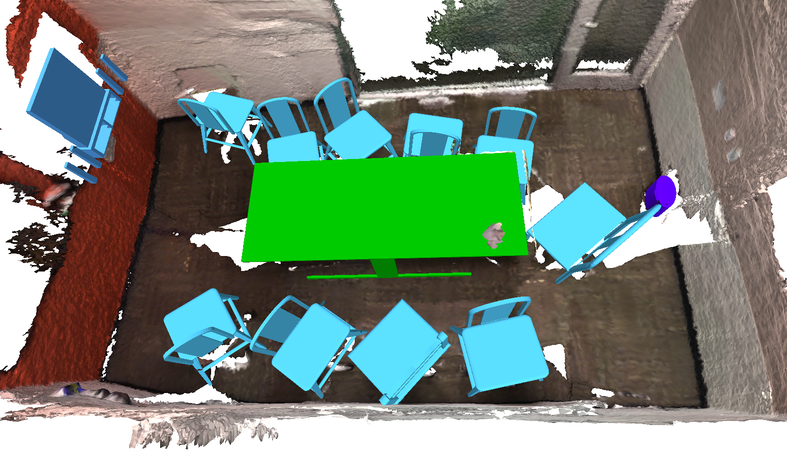} &
  \includegraphics[trim={3.4cm 1cm 2.5cm 0.cm},clip,width=0.22\linewidth]{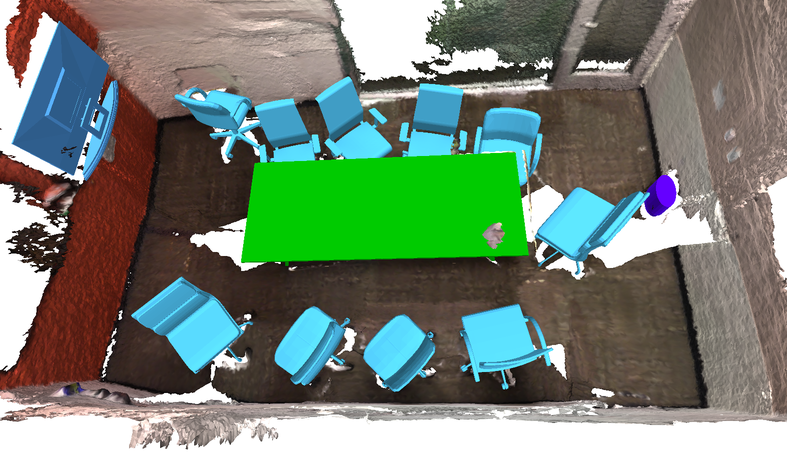} &
  \includegraphics[trim={3.4cm 1cm 2.5cm 0.cm},clip,width=0.22\linewidth]{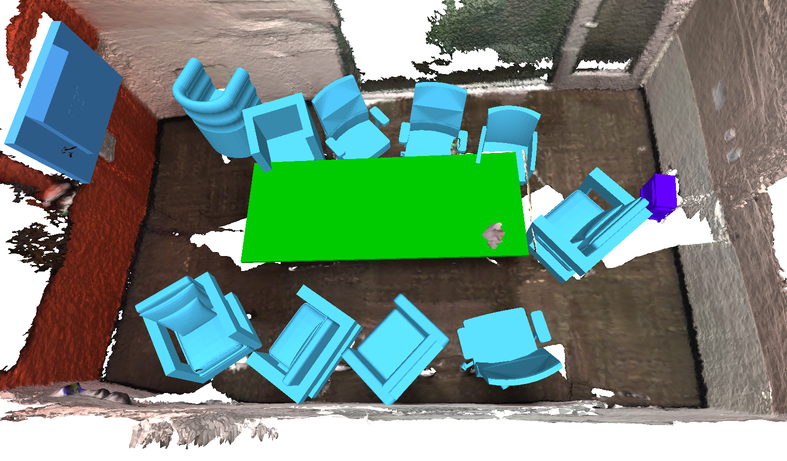} &
  \includegraphics[trim={3.4cm 1cm 2.5cm 0.cm},clip,width=0.22\linewidth]{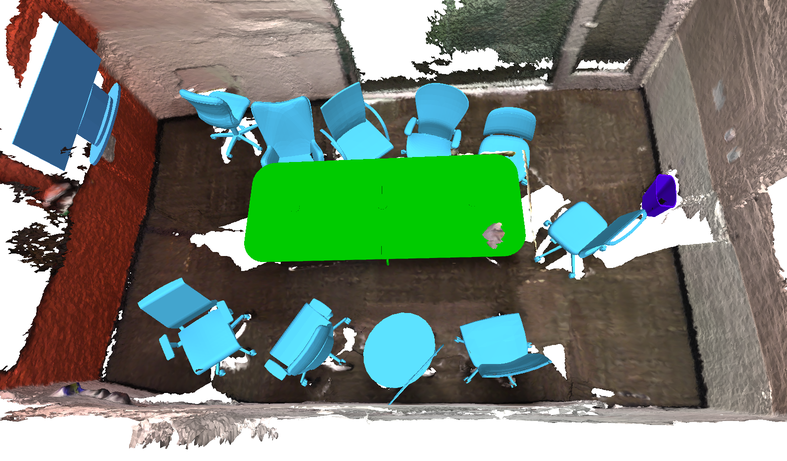} \\

  \includegraphics[trim={4.5cm 3cm 4cm .5cm},clip,width=0.22\linewidth]{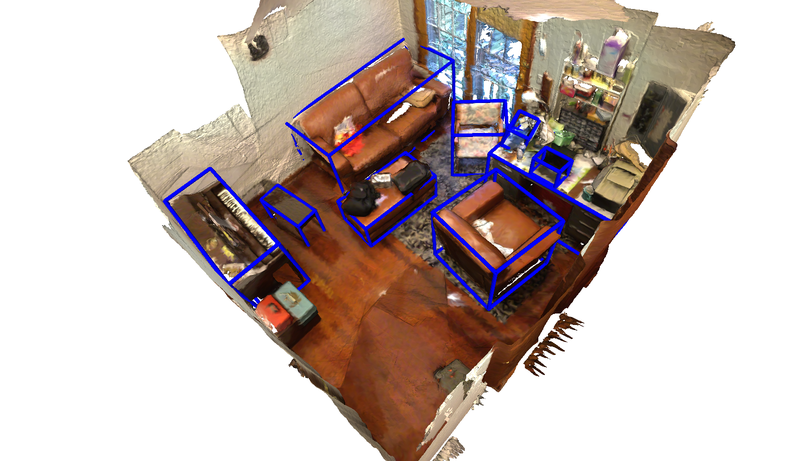} &
  \includegraphics[trim={4.5cm 3cm 4cm .5cm},clip,width=0.22\linewidth]{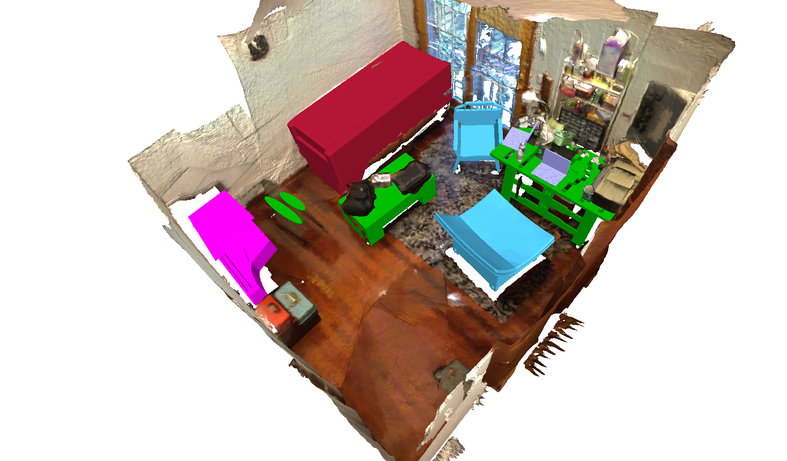} &
  \includegraphics[trim={4.5cm 3cm 4cm .5cm},clip,width=0.22\linewidth]{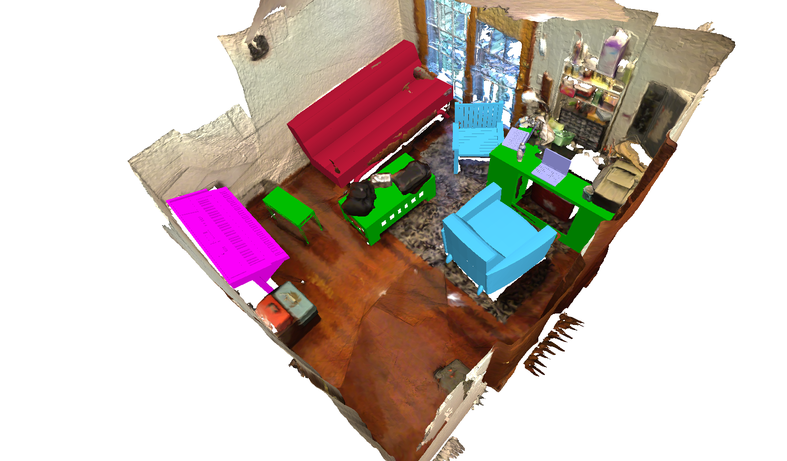} &
  \includegraphics[trim={4.5cm 3cm 4cm .5cm},clip,width=0.22\linewidth]{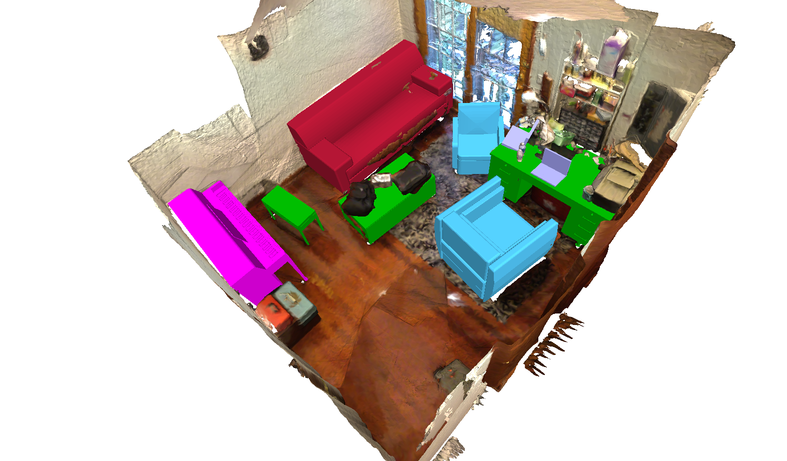} &
  \includegraphics[trim={4.5cm 3cm 4cm .5cm},clip,width=0.22\linewidth]{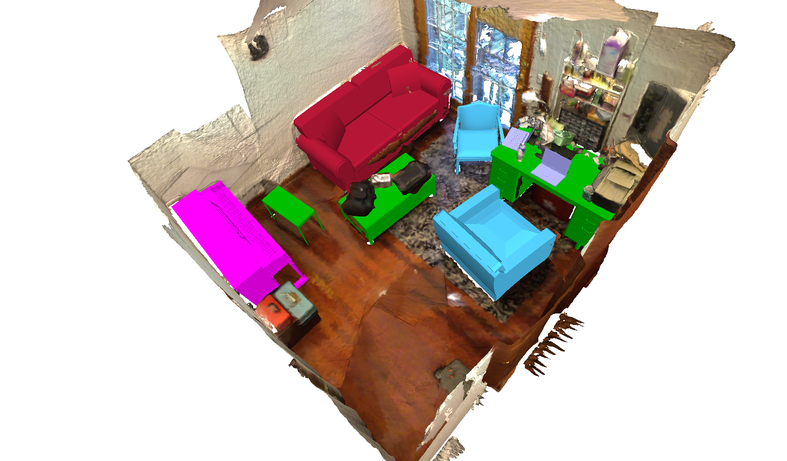} \\
  
  RGBD Scan &  NN embeddings & Chamfer Distance & MSCD & Render-and-Compare\\
\end{tabular}
}
\caption{Visualization of CAD retrieval results for ScanNet using 3D boxes from Scan2CAD annotations. Using nearest neighbor search in embedding space delivers the lowest quality, where the orientation of the retrieved CAD model is often incorrect. Strengths and weaknesses of using the chamfer distance and MSCD are also clearly visible: Chamfer distance delivers good results if the target object scans are very complete, whereas it delivers bad results for incomplete scans. MSCD, on the other hand, copes well with incomplete scans, but is not able to use the advantage of highly complete object scans due to the fact that the chamfer distance from CAD model to scanned target points is ignored.}
\label{fig:supp_results_obj_func}
\end{figure*}

Qualitative comparison for retrieved CAD models for different objective functions are shown in Figure~\ref{fig:supp_results_obj_func}. Visualizations show that using chamfer distance delivers good results if the target object scans are very complete, whereas it delivers bad results for incomplete scans. MSCD~\cite{wei2022accurate}, on the other hand, copes well with incomplete scans, but is not able to use the advantage of highly complete object scans as shown in the top two rows in Figure~\ref{fig:supp_results_obj_func}, due to the fact that the chamfer distance from CAD model to scanned target points is ignored. Render-and-Compare appears to be the most robust objective function for incomplete 3D data, and delivers on average the most accurate results.

\section{Additional Results for HOC-Search Efficiency Evaluation}
\label{sec:supp_results_hoc_efficiency}
Per-category Top-1/Top-5 retrieval accuracy is shown in Table~\ref{tab:supp_eval_topk_acc}. For categories Cabinet and Table, results show a below-average Top-1/Top-5 retrieval accuracy. For the category Cabinet, most objects have a very similar box-like shape, which makes it hard for HOC-Search to focus on a specific branch of the HOC-Tree, and consequently, makes it difficult to retrieve the exact same model as the exhaustive search baseline. In the case of the category Table, we consider the huge amount of 8437 CAD models as the reason for the below-average accuracy, as there are multiple CAD models with similar shape. 
 
Figure~\ref{fig:supp_top5} shows qualitative results for Top-5 candidates retrieved with exhaustive search and HOC-Search. In a large scale CAD model database~(more than 32k CAD models in total, e.g., 8437 Tables models and the 6779 Chairs models), it is highly likely that there exist multiple suitable CAD models. Often the 3D shapes of the Top-5 candidates of HOC-Search and exhaustive search are very similar compared to the target object. Our method possibly retrieves a CAD model that is not the same model as from the exhaustive search but which fits the real object as well.
 
Failure cases are shown in Figure~\ref{fig:supp_top5_fail}. One can see that for highly incomplete scans, the render-and-compare objective function as proposed in \cite{ainetter2023automatically} delivers inaccurate results. A reason for this is that the 2D silhouettes used in Equation~\ref{eq:supp_loss_sil} are obtained by rendering the 3D instance segmentation of the target objects, which leads to an over-reliance on 3D data. Using the RGB images to obtain the silhouettes could improve the quality of CAD model retrieval for objects where the majority of 3D points is missing.

\begin{table*}[hpt!]
\centering
\scalebox{1.}{
\begin{tabular}{@{}ccccccccc@{}}
\toprule

%\# iterations      & Top-1 Accuracy &\begin{tabular}[c]{@{}c@{}} Chamfer Distance\\ of Top1 Candidates\end{tabular} & Top-5 Accuracy & Top-10 Accuracy & Runtime in (sec) & Speed-up Factor\\

\# iterations      & 
Bookshelf &
Cabinet &
Chair &
Display &
Sofa &
Table &
Lamp &
Others\\
\midrule
200     & 36.2/69.1 & 5.2/21.3 & 19.1/44.0  & 14.2/47.0  & 20.9/45.5  & 5.6/17.4 & 24.0/64.0 & 45.8/82.3\\
300     &  43.2/82.1 & 10.4/36.9 & 26.6/55.7  & 21.6/60.3  & 30.0/60.0  & 8.4/25.3 & 24.0/68.0 & 56.2/88.5\\
400     &  55.1/86.4 & 16.5/48.7 & 31.5/62.0  & 31.2/74.0  & 34.5/68.1  & 11.5/33.6 & 28.0/72.0 & 59.1/90.7\\
600     &  61.0/87.0 & 27.3/69.1 & 38.9/72.6  & 51.8/88.3  & 46.3/76.3  & 18.0/44.1 & 36.0/76.0 & 62.0/92.2\\
800     &  61.0/87.0 & 35.6/80.4 & 44.8/77.5  & 60.3/93.1  & 52.7/80.9  & 21.6/50.7 & 44.0/84.0 & 63.5/92.3\\
1000     &  61.1/87.0 & 46.5/88.2 & 48.4/78.9  & 68.2/95.2  & 60.0/86.3  & 26.6/57.1 & 48.0/88.0 & 63.7/92.3\\
\bottomrule
\end{tabular}
}
% \vspace{-0.2cm}
\caption{Mean Top-1/Top-5 per-category retrieval accuracy in $[\%]$ for the categories present in the ScanNet validation set using 3D box annotations from Scan2CAD. Retrieval accuracy is calculated in relation to the exhaustive search baseline. 'Others' includes the following categories: \{Dishwasher, Keyboard, Pillow, Motorbike, Bathtub, Bench, Bowl, Clock, Faucet, File Cabinet, Flowerpot, Guitar, Bed, Laptop, Microwaves, Piano, Printer, Stove, Trash Bin, Washer\}.}
\label{tab:supp_eval_topk_acc}
\end{table*}

%\begin{figure*}
%\centering
%\includegraphics[width=1\linewidth]{figures/comparison_top5_results/full_figure/top5candidatesv2.png}\\
%\begin{tabular}{cccc}
%Target $\quad\quad$ &
%top-5 candidates retrieved by exhaustive search $\quad\quad\quad\quad$ &
%top-5 candidates retrieved by our method& \\
%object  $\quad\quad$ & & &\\
%\end{tabular}
%\caption{Visualization of top5 candidates (ascending order) from exhaustive search compared to MCTS-based search for 800 iterations for randomly selected target objects. Left shows the target object, middle column are the results from exhaustive search, and right column shows the results using MCTS. Although not always the same model is found, one can see that the overall quality of the results is very high. The two methods retrieve very similar CAD models; on the last row, the 5 candidates are even exactly the same.
%We refer to the supplementary material for additional visualizations. }
%    \label{fig:vis_topk_acc}
%\end{figure*}
\newcommand{\widthtopfv}{0.07}

\begin{figure*}
\centering
\scalebox{0.85}{
\begin{tabular}{c|ccccc|ccccc}
%[trim={left bottom right top}

 \includegraphics[trim={15cm 2.5cm 15cm 5.cm},clip,width=\widthtopfv\linewidth]{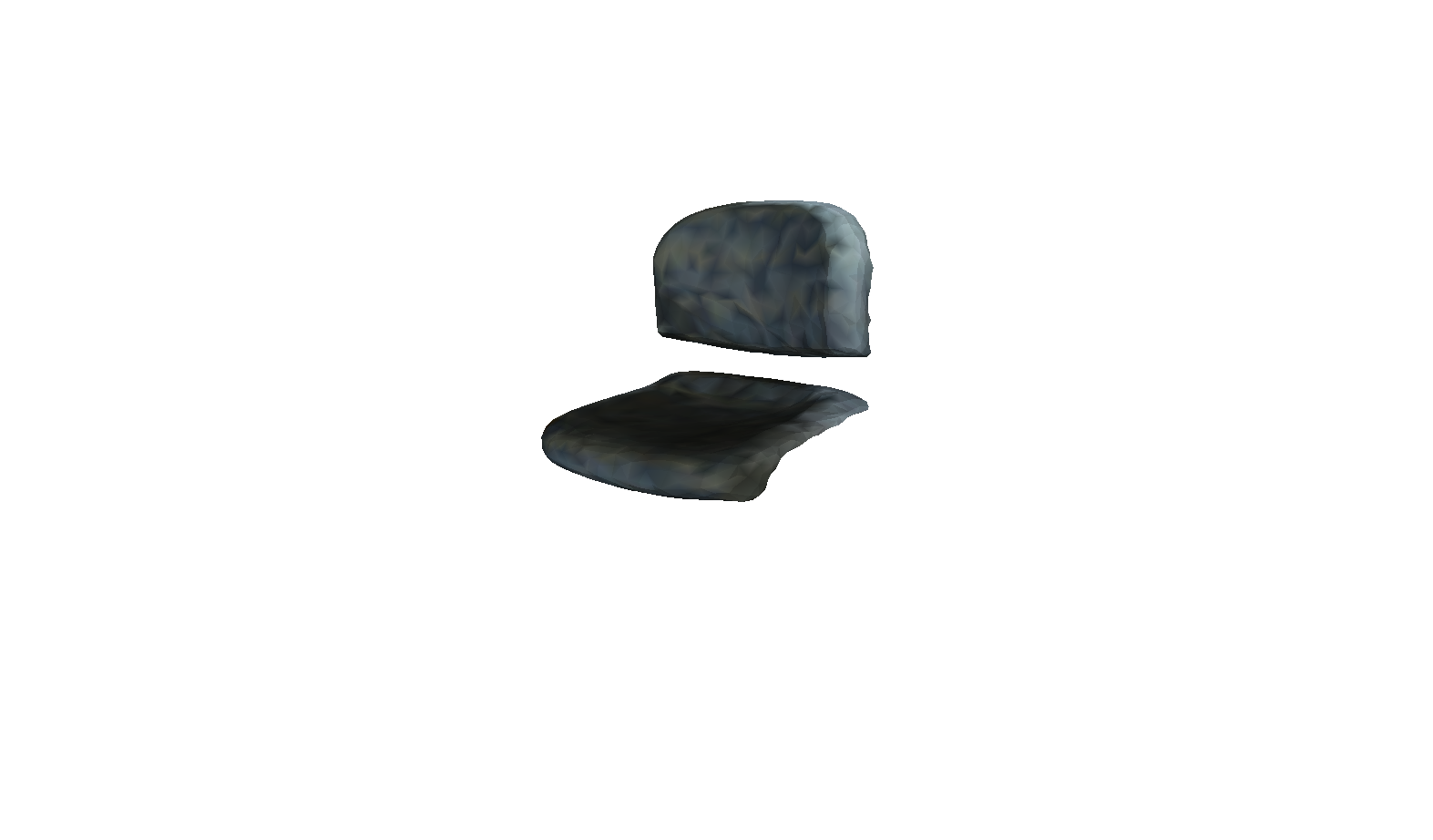} &
  \includegraphics[trim={15cm 2.5cm 15cm 5.cm},clip,width=\widthtopfv\linewidth]{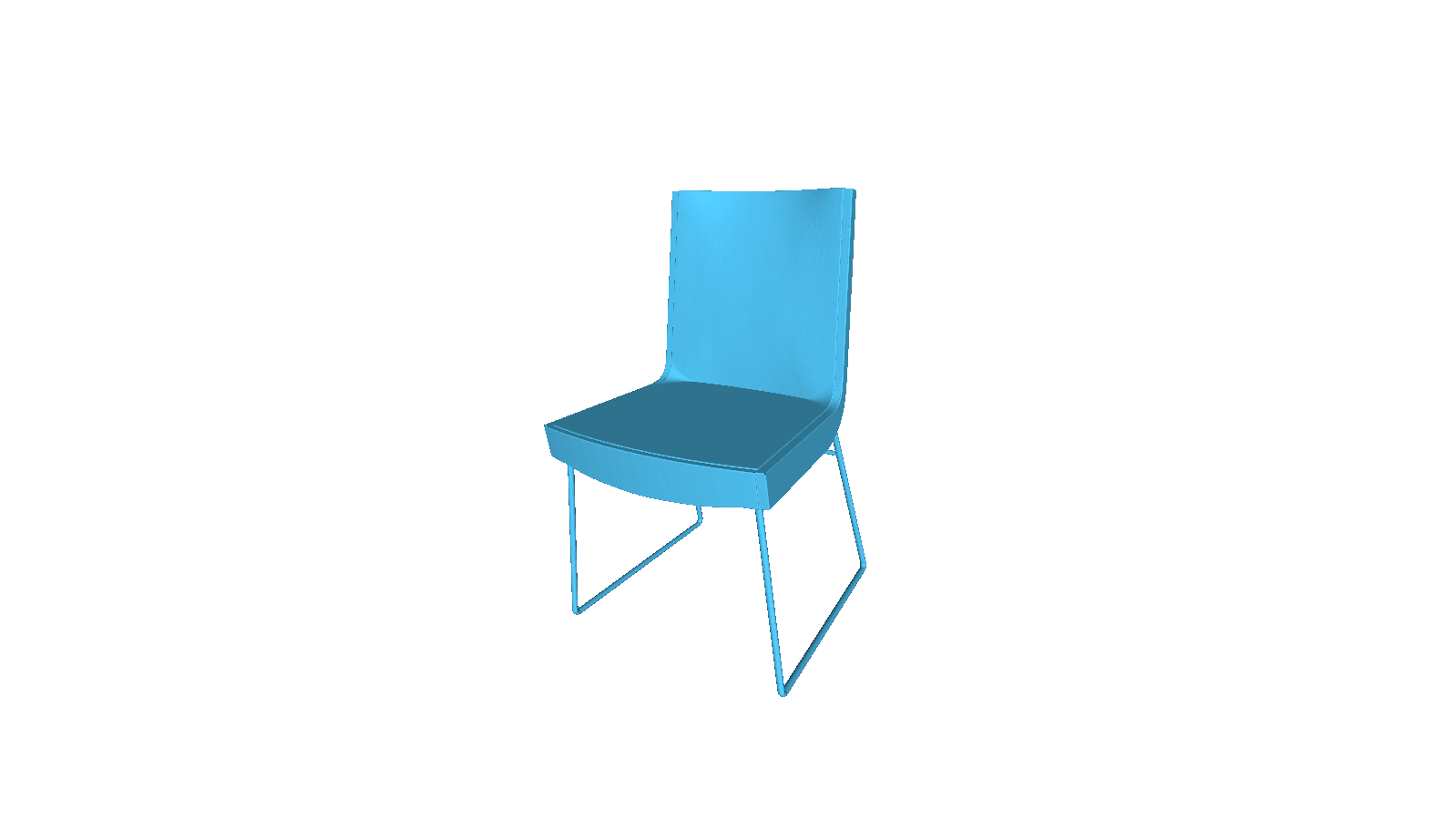} &  \includegraphics[trim={15cm 2.5cm 15cm 5.cm},clip,width=\widthtopfv\linewidth]{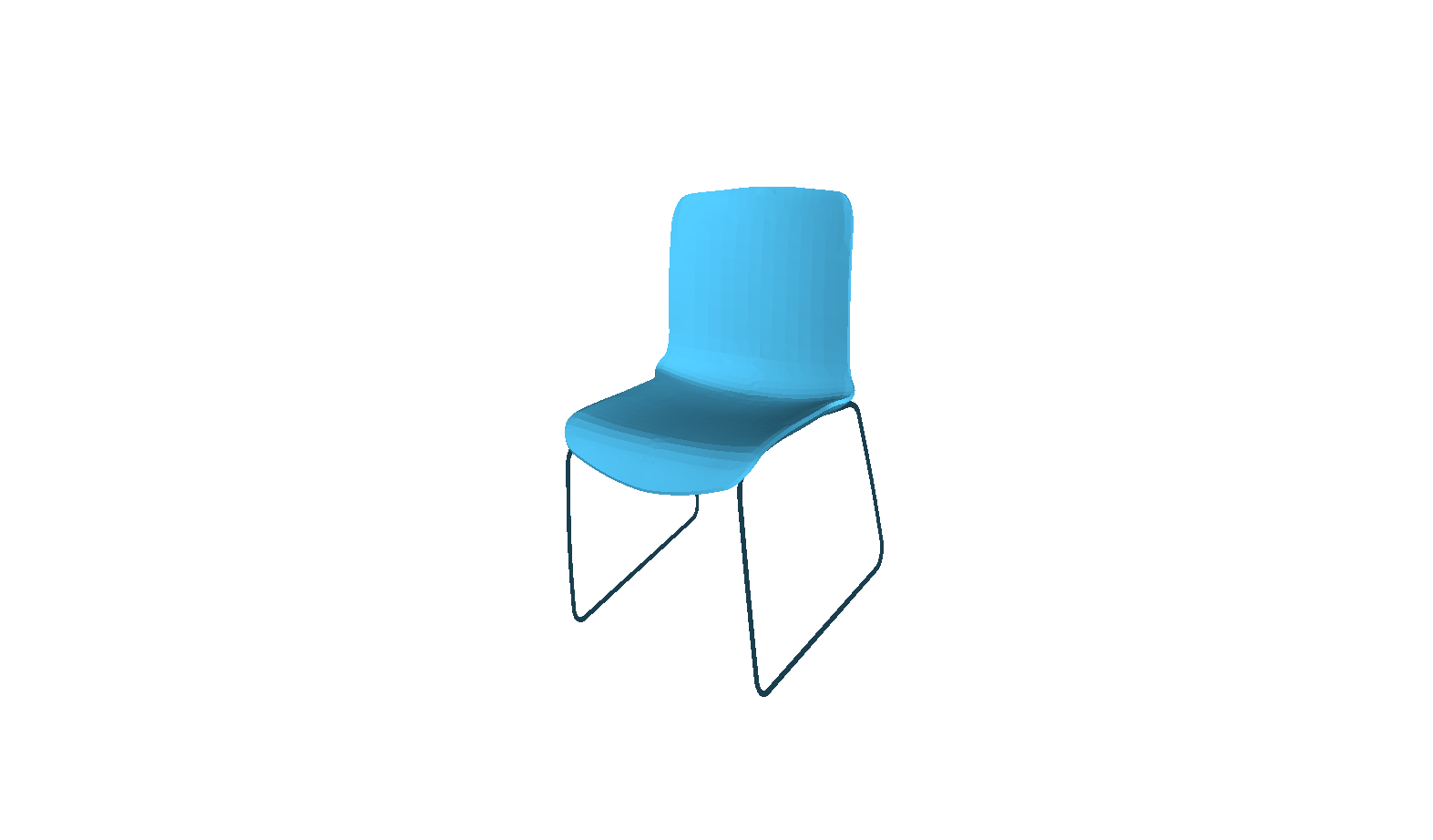} &
  \includegraphics[trim={15cm 2.5cm 15cm 5.cm},clip,width=\widthtopfv\linewidth]{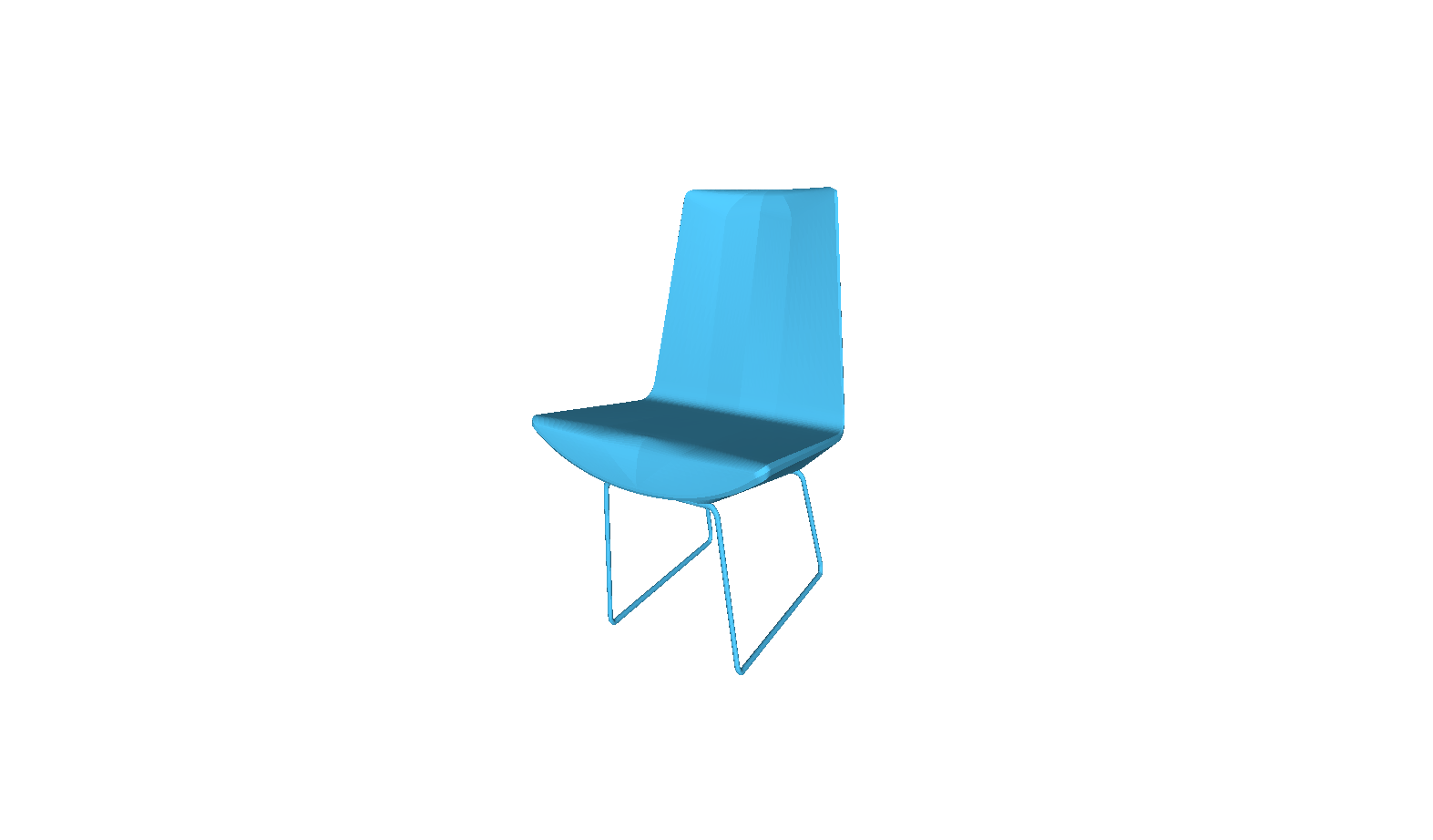} &
  \includegraphics[trim={15cm 2.5cm 15cm 5.cm},clip,width=\widthtopfv\linewidth]{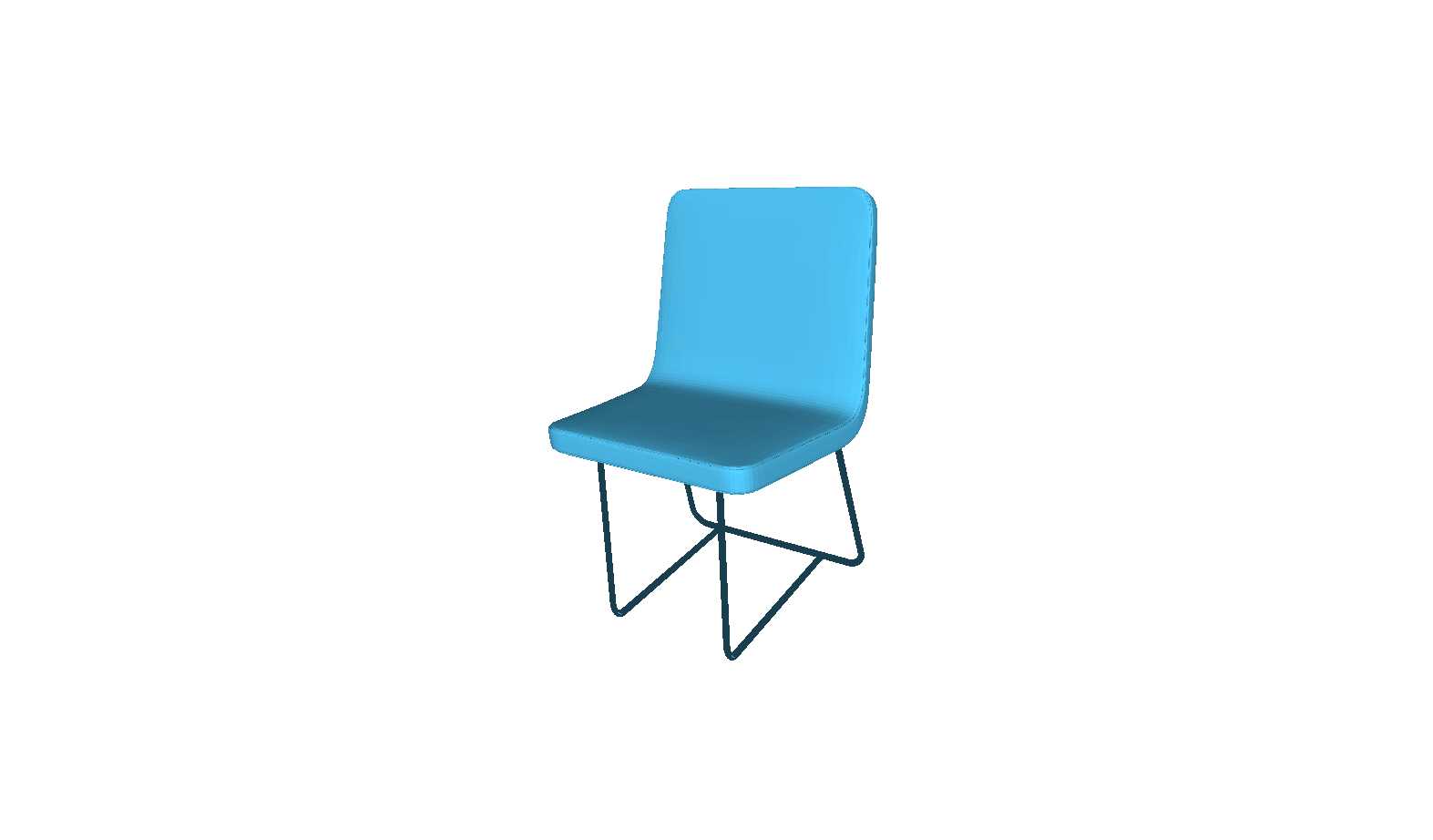} &
  \includegraphics[trim={15cm 2.5cm 15cm 5.cm},clip,width=\widthtopfv\linewidth]{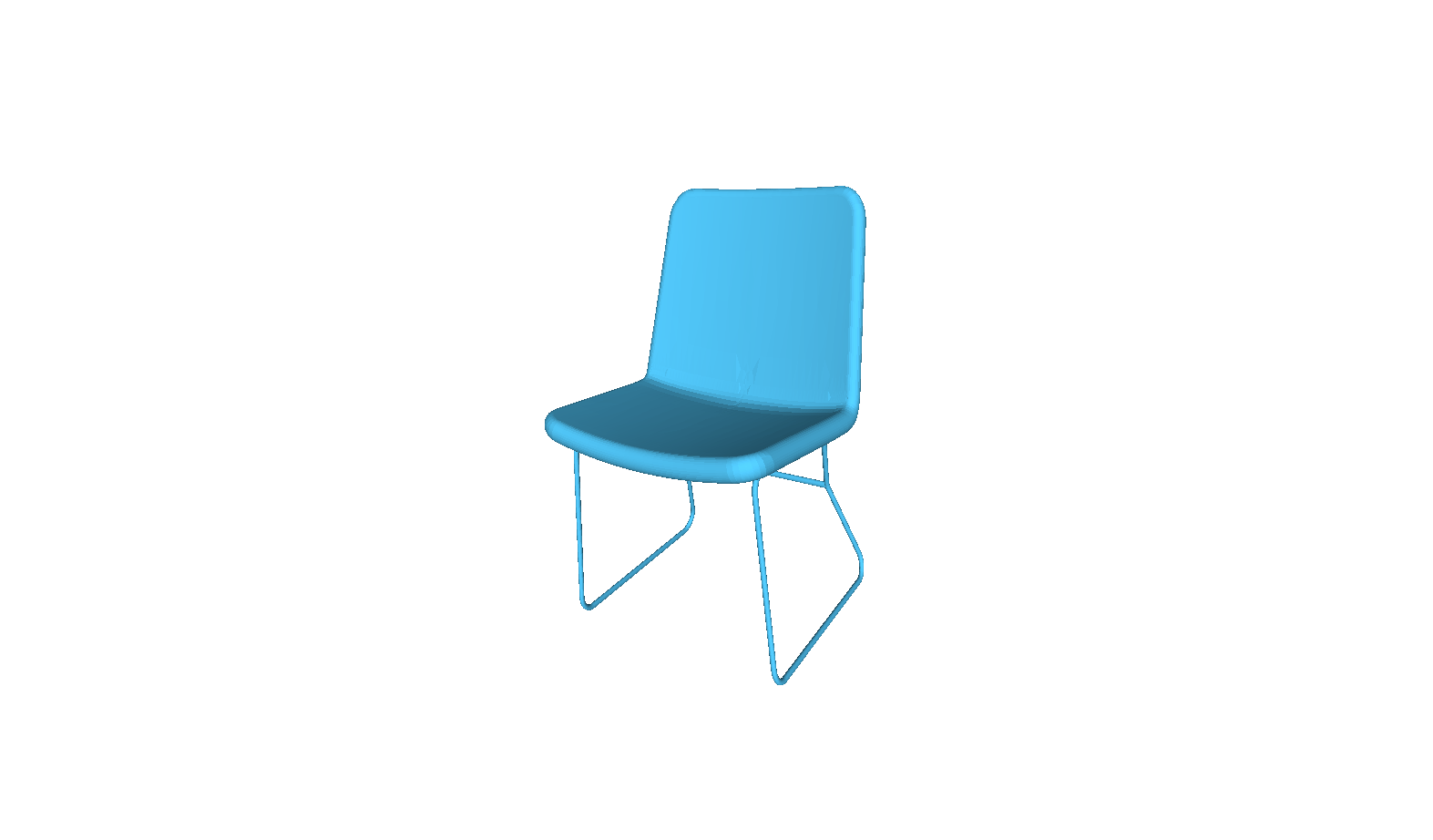} &
  \includegraphics[trim={15cm 2.5cm 15cm 5.cm},clip,width=\widthtopfv\linewidth]{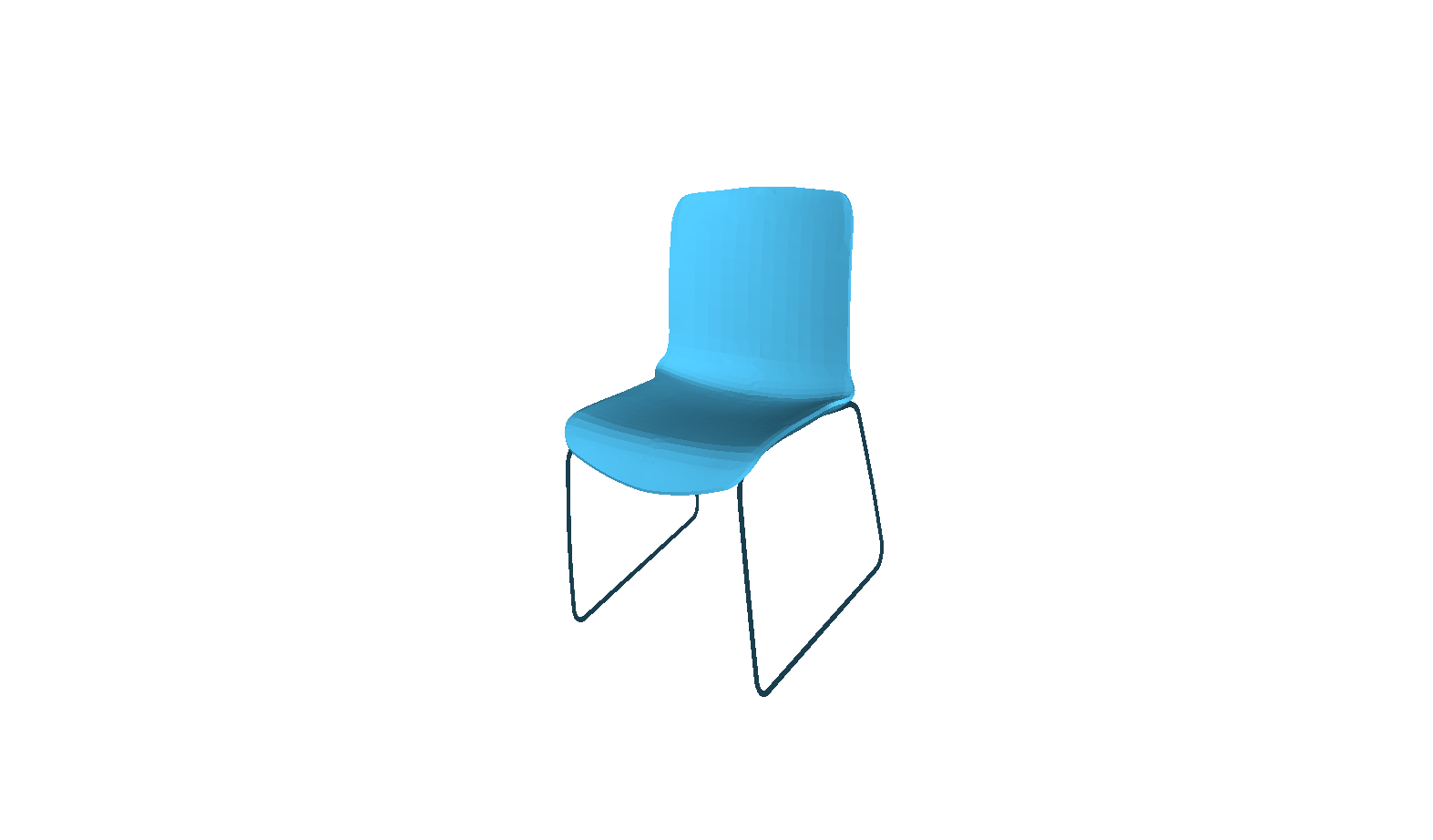} &
  \includegraphics[trim={15cm 2.5cm 15cm 5.cm},clip,width=\widthtopfv\linewidth]{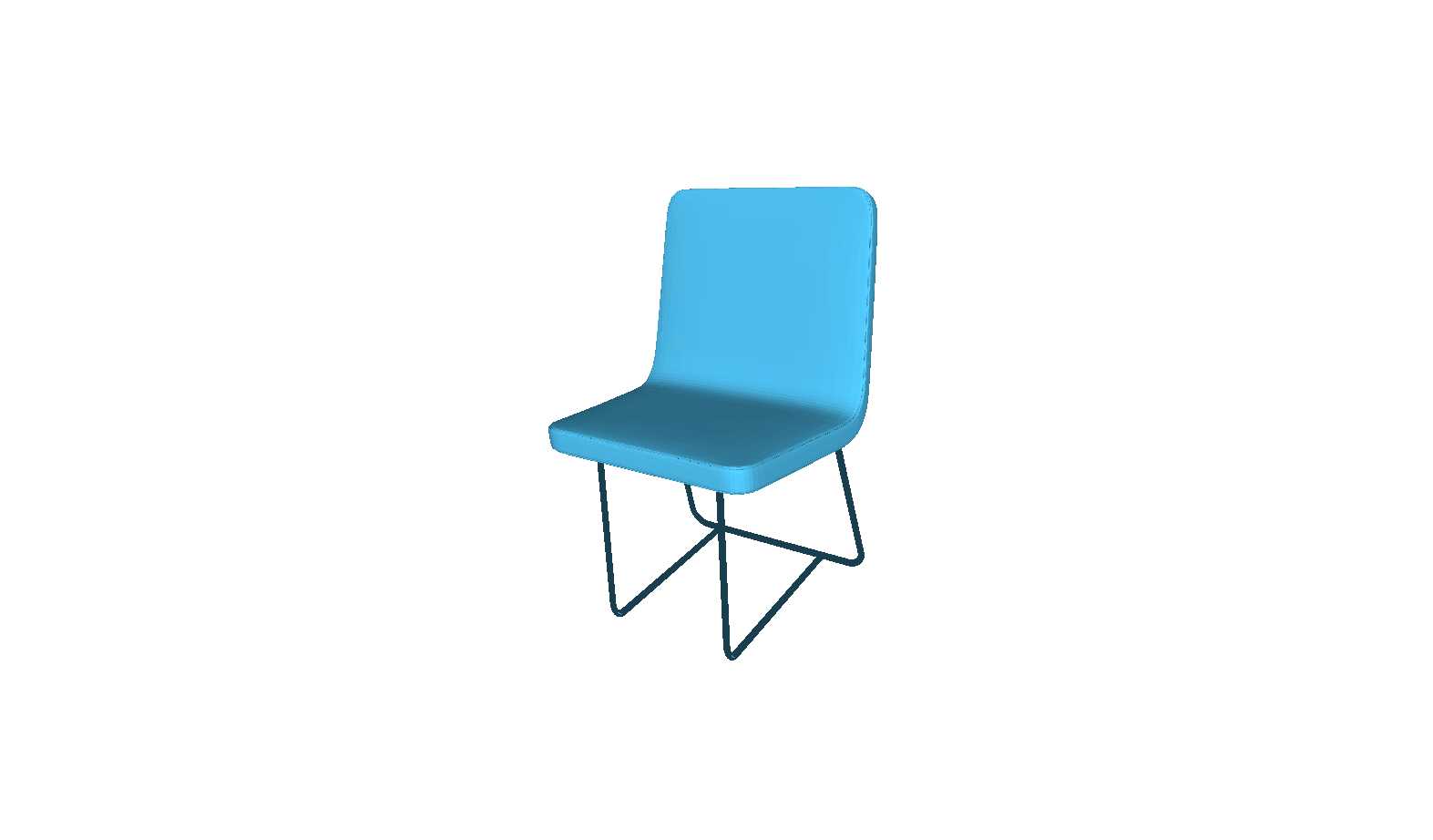} &
  \includegraphics[trim={15cm 2.5cm 15cm 5.cm},clip,width=\widthtopfv\linewidth]{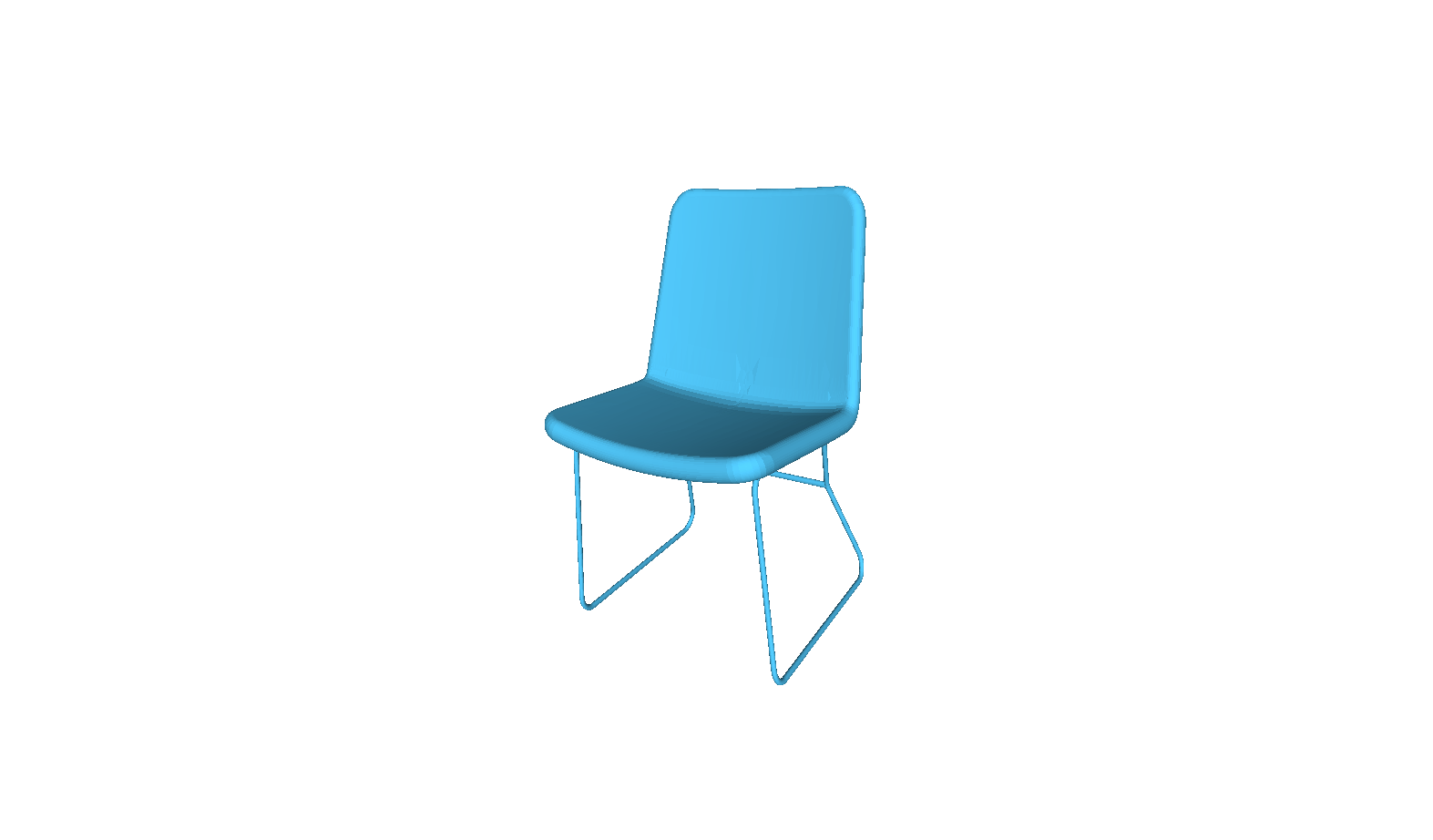} &
  \includegraphics[trim={15cm 2.5cm 15cm 5.cm},clip,width=\widthtopfv\linewidth]{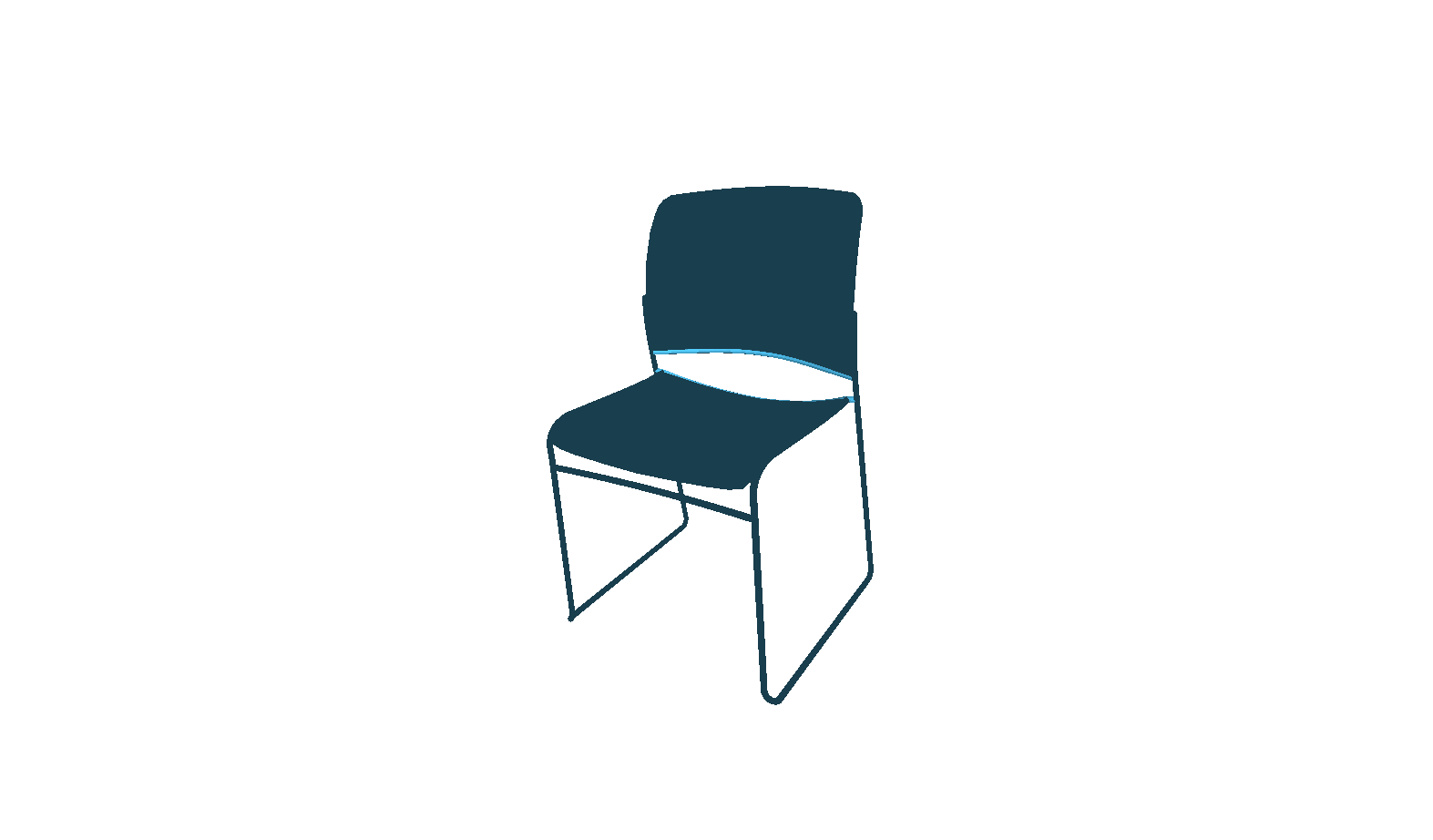} &
  \includegraphics[trim={15cm 2.5cm 15cm 5.cm},clip,width=\widthtopfv\linewidth]{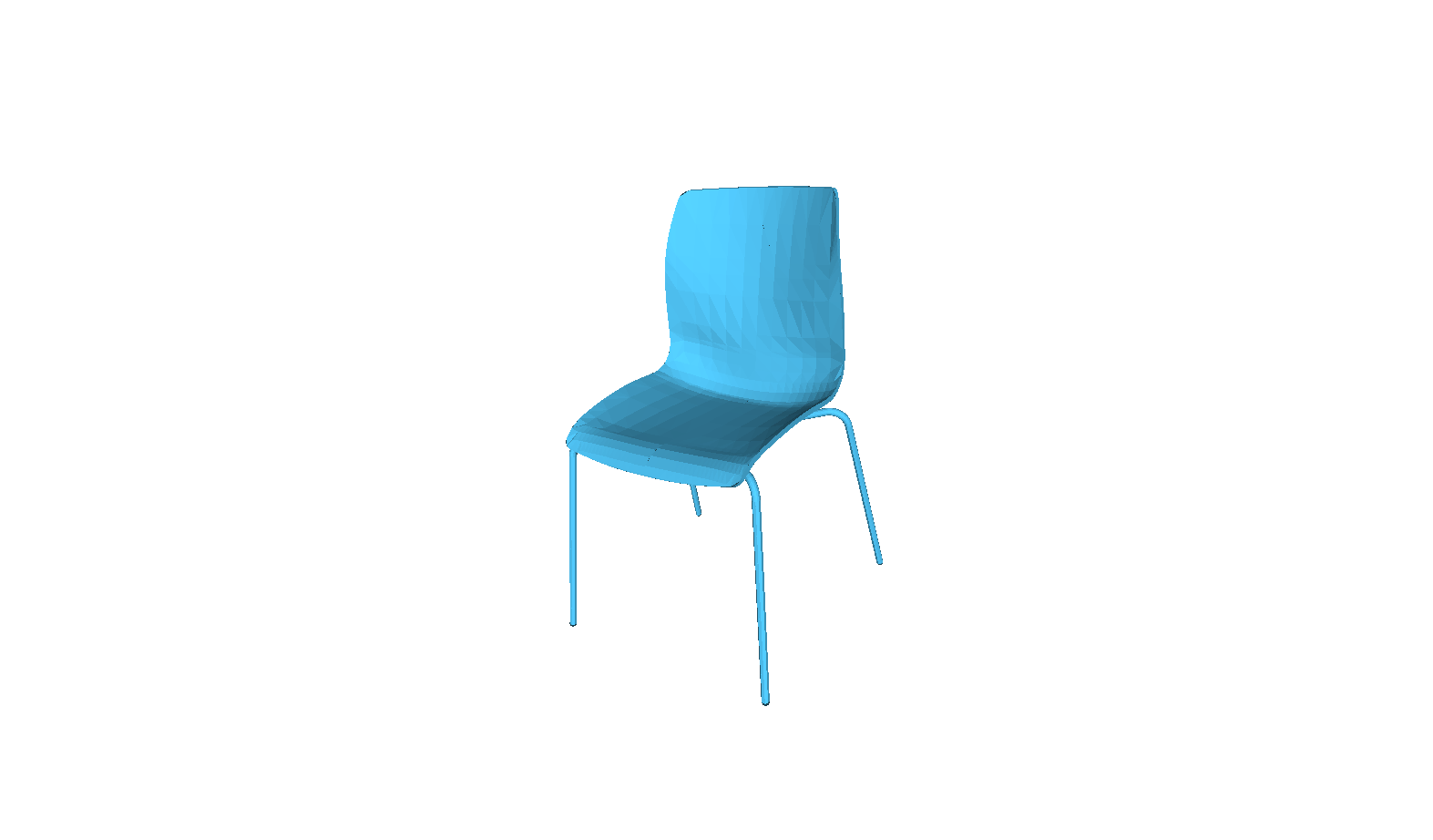} \\

 \includegraphics[trim={12cm 5cm 10cm 5cm},clip,width=\widthtopfv\linewidth]{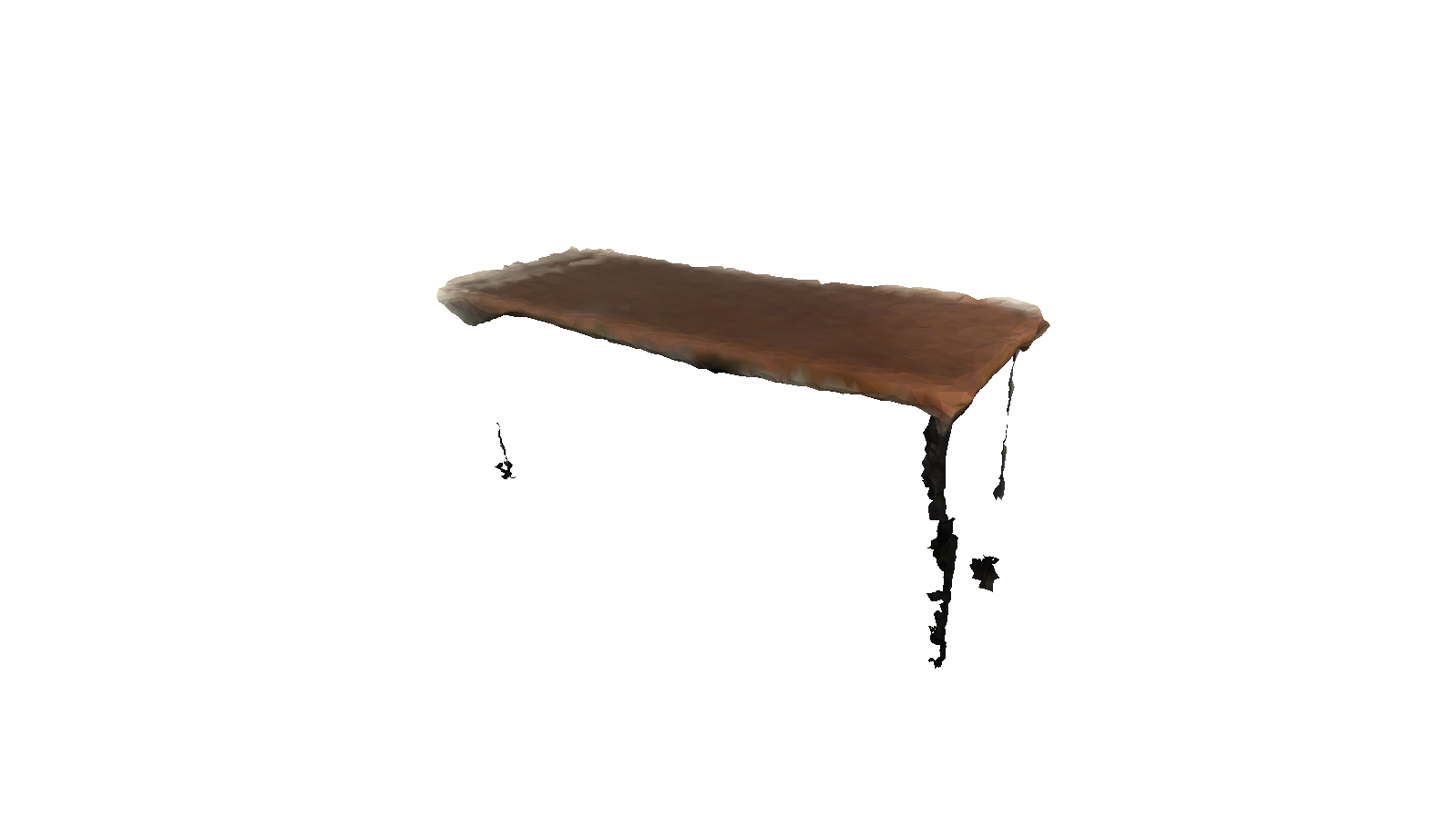} &
 \includegraphics[trim={12cm 5cm 10cm 5cm},clip,width=\widthtopfv\linewidth]{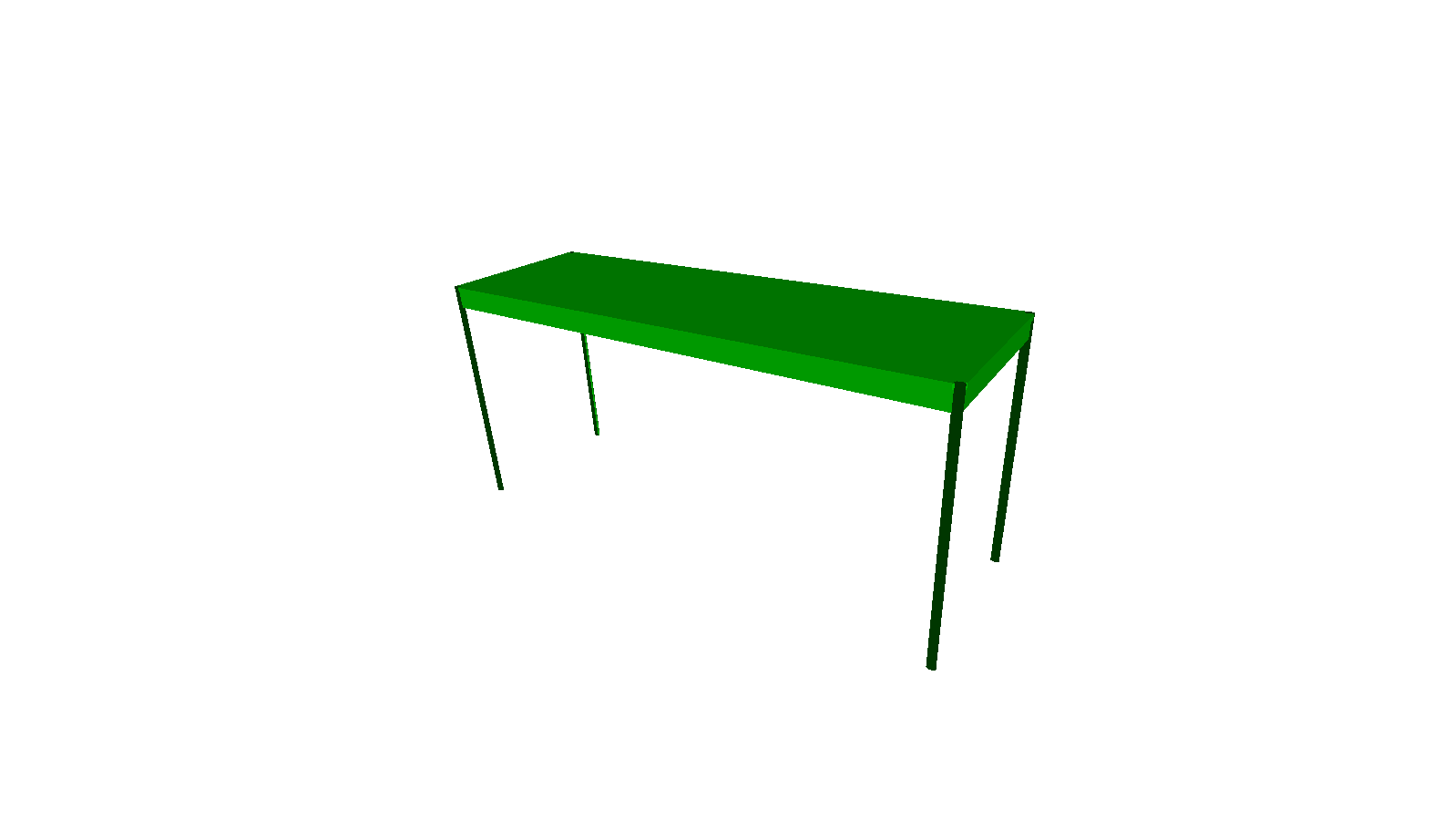} &   \includegraphics[trim={12cm 5cm 10cm 5cm},clip,width=\widthtopfv\linewidth]{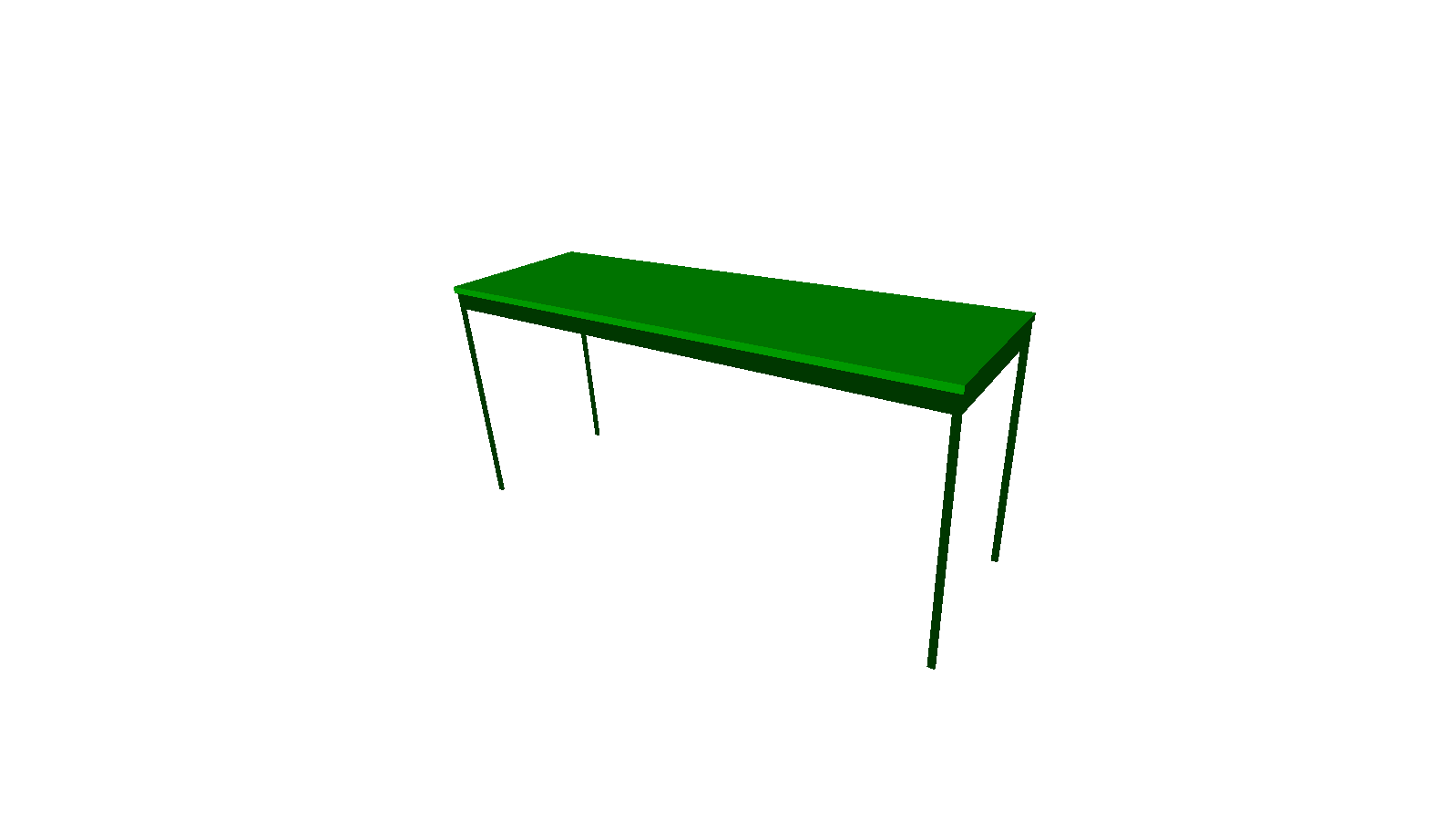} &
 \includegraphics[trim={12cm 5cm 10cm 5cm},clip,width=\widthtopfv\linewidth]{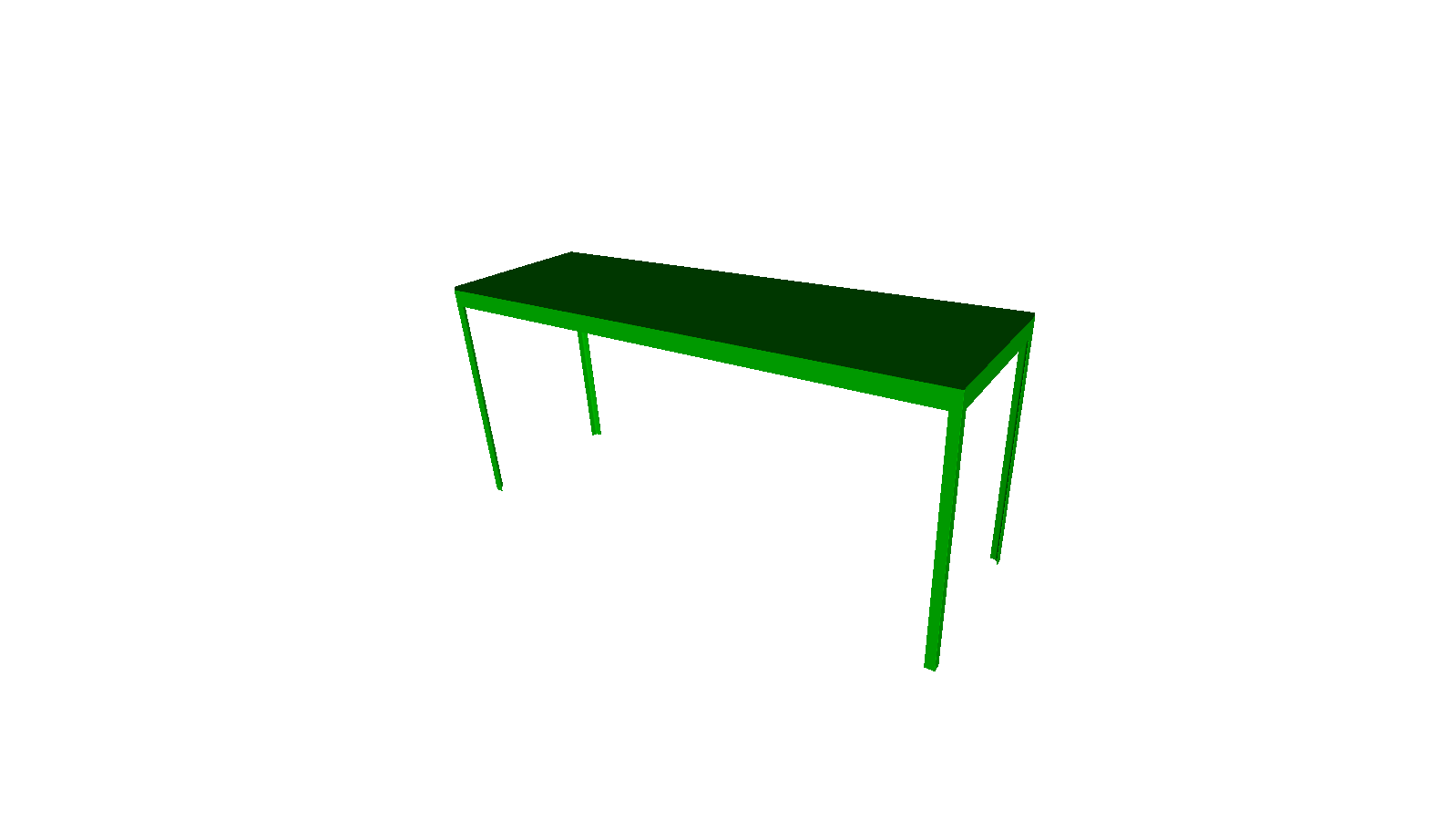} &
 \includegraphics[trim={12cm 5cm 10cm 5cm},clip,width=\widthtopfv\linewidth]{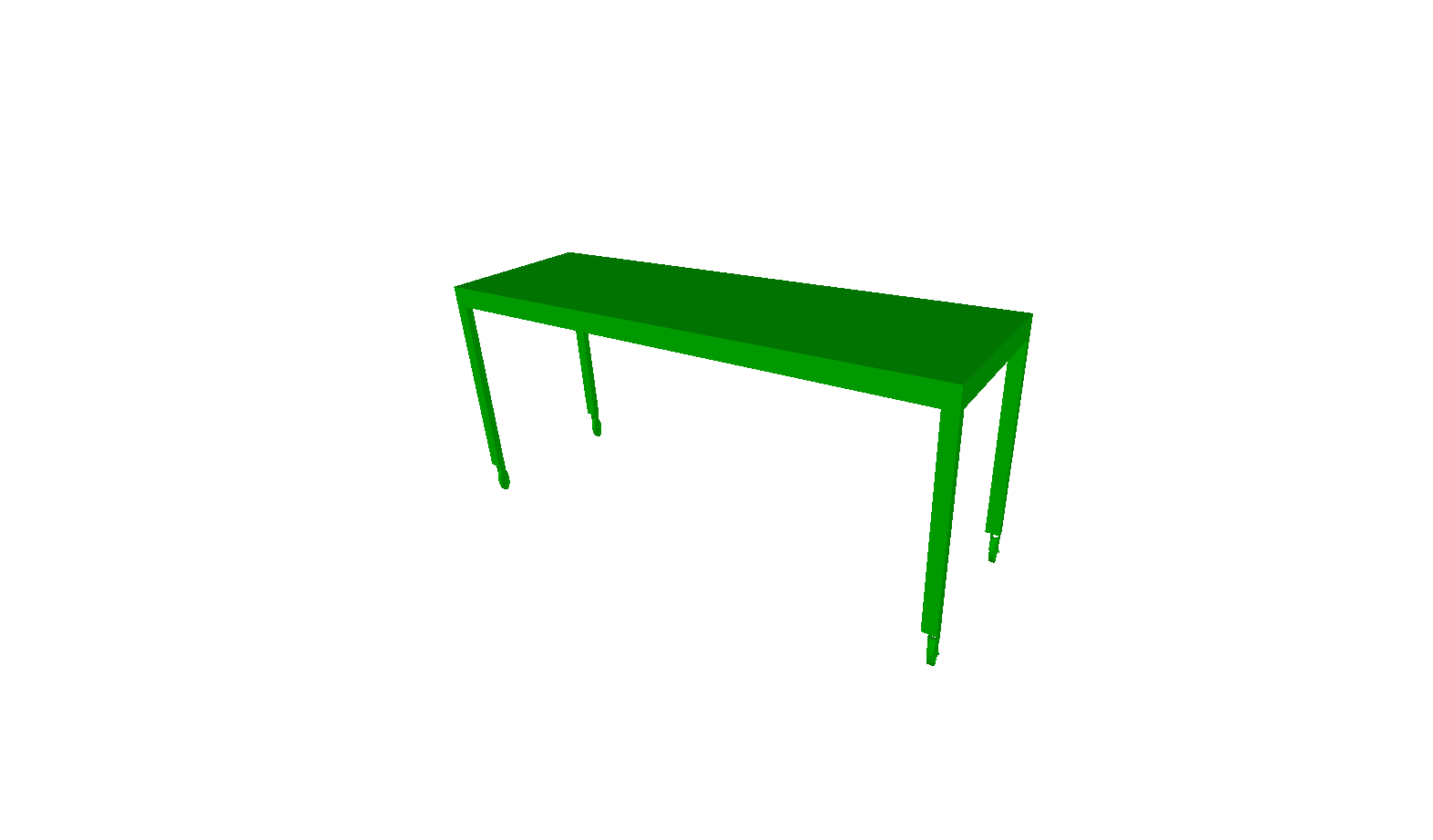} &
 \includegraphics[trim={12cm 5cm 10cm 5cm},clip,width=\widthtopfv\linewidth]{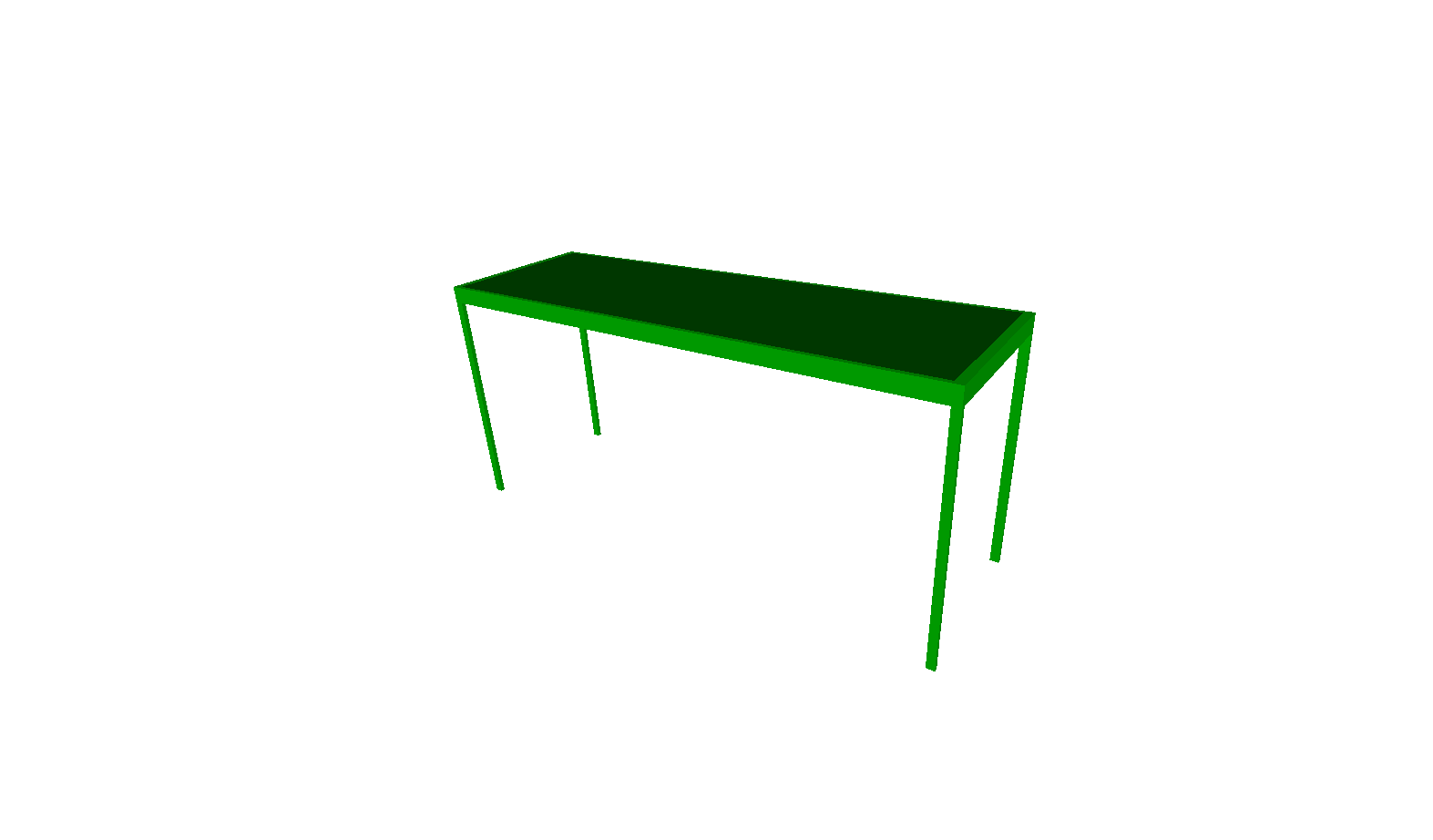} &
 \includegraphics[trim={12cm 5cm 10cm 5cm},clip,width=\widthtopfv\linewidth]{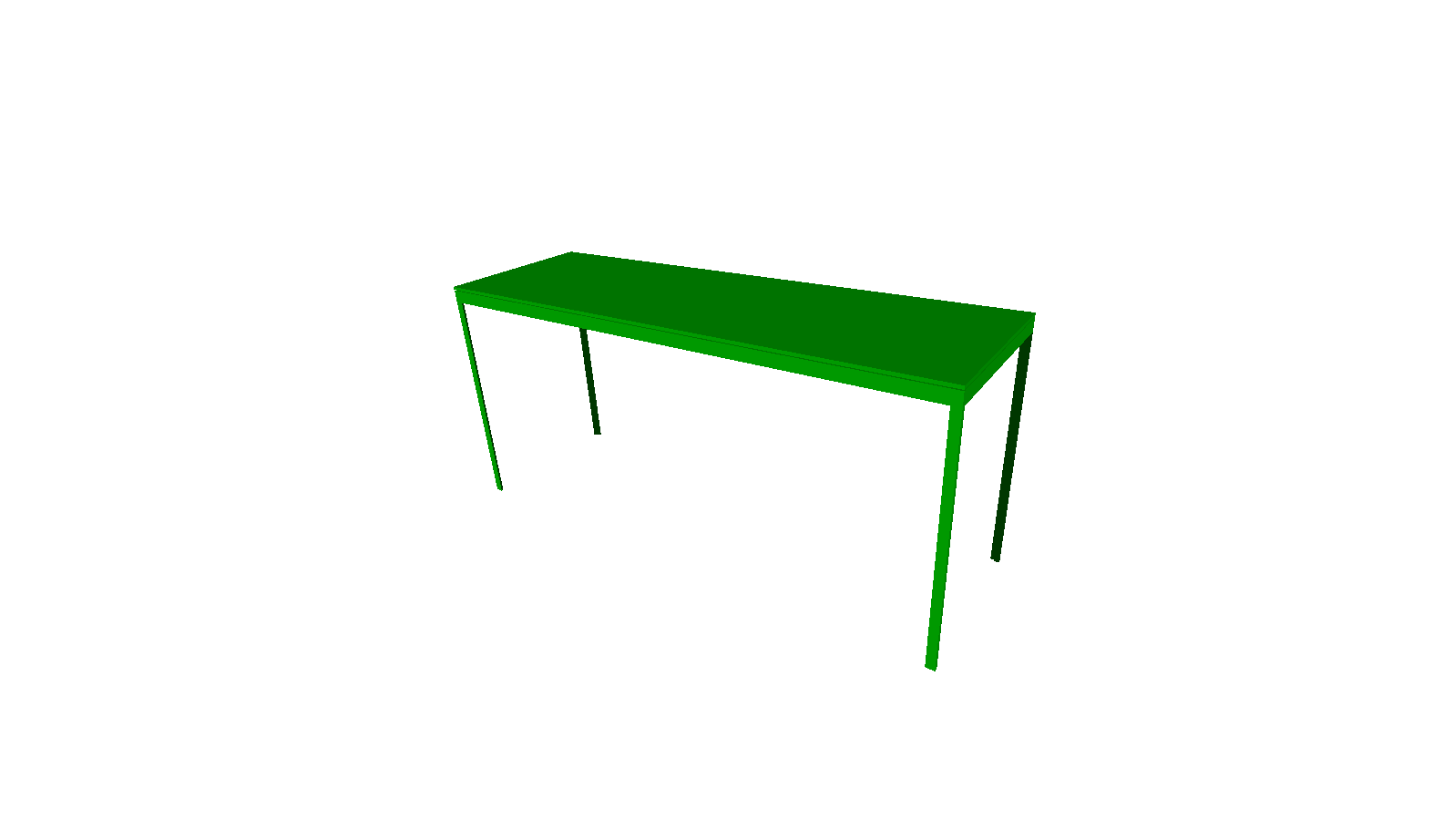} &
 \includegraphics[trim={12cm 5cm 10cm 5cm},clip,width=\widthtopfv\linewidth]{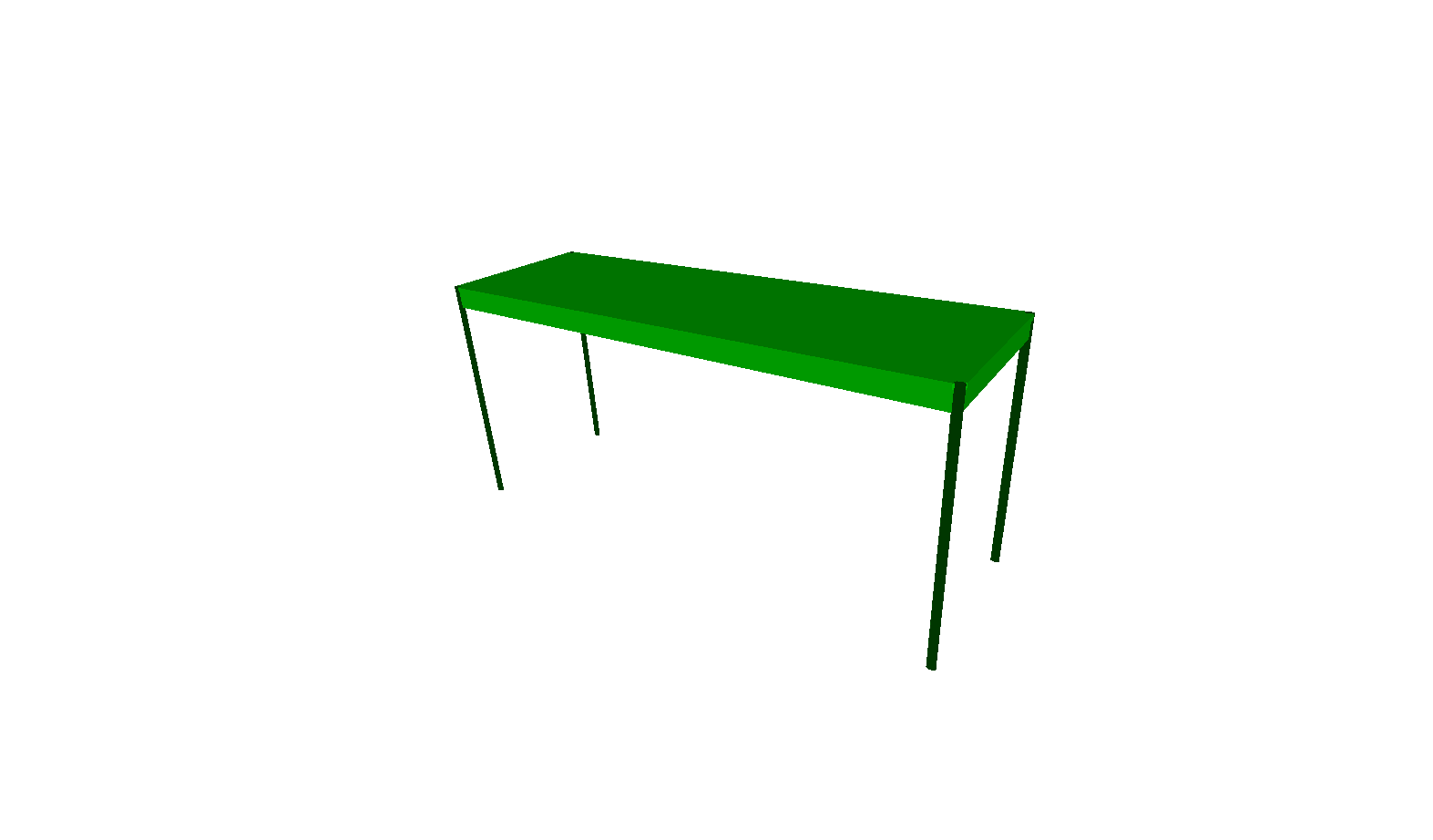} &
 \includegraphics[trim={12cm 5cm 10cm 5cm},clip,width=\widthtopfv\linewidth]{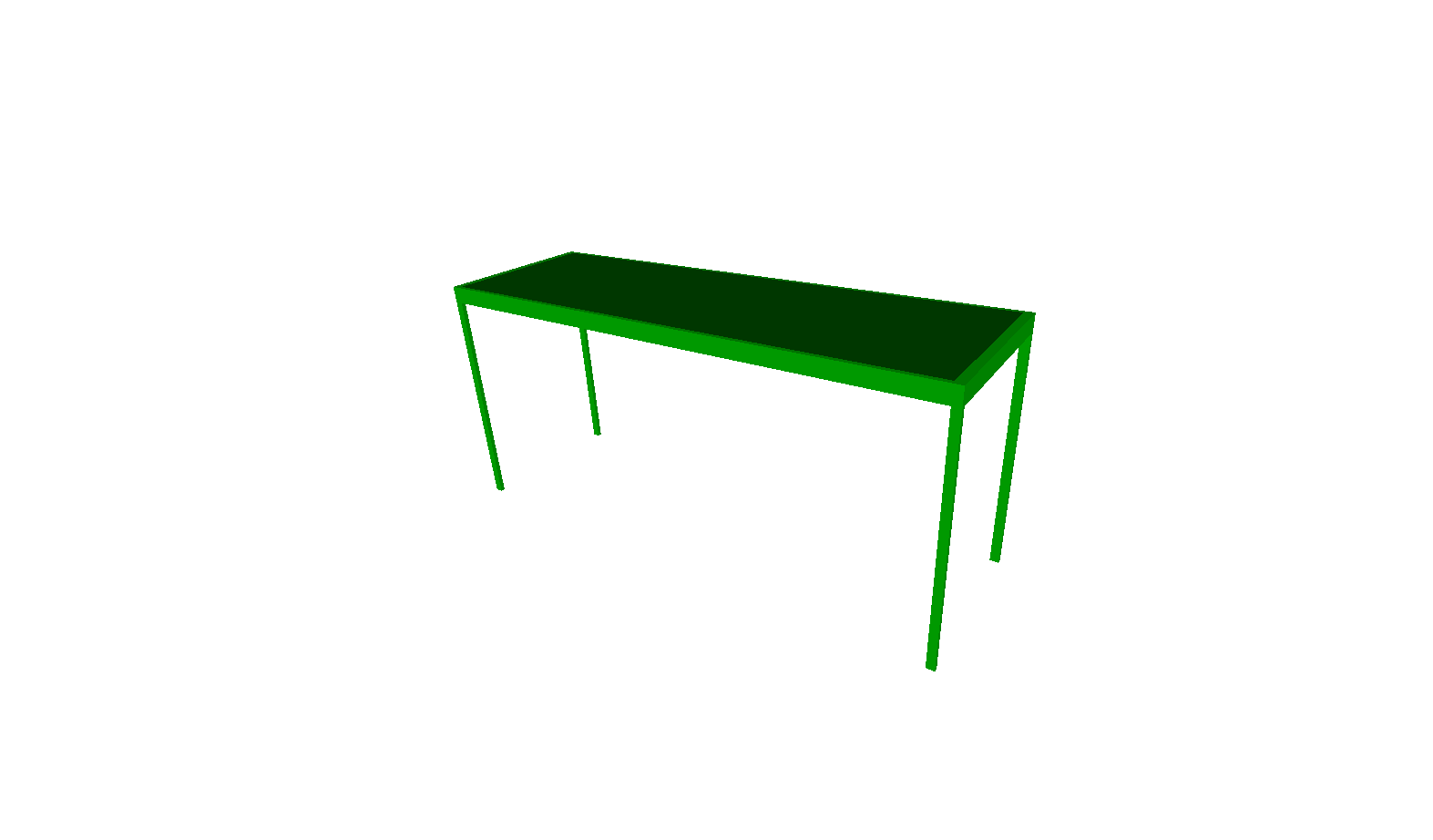} &
 \includegraphics[trim={12cm 5cm 10cm 5cm},clip,width=\widthtopfv\linewidth]{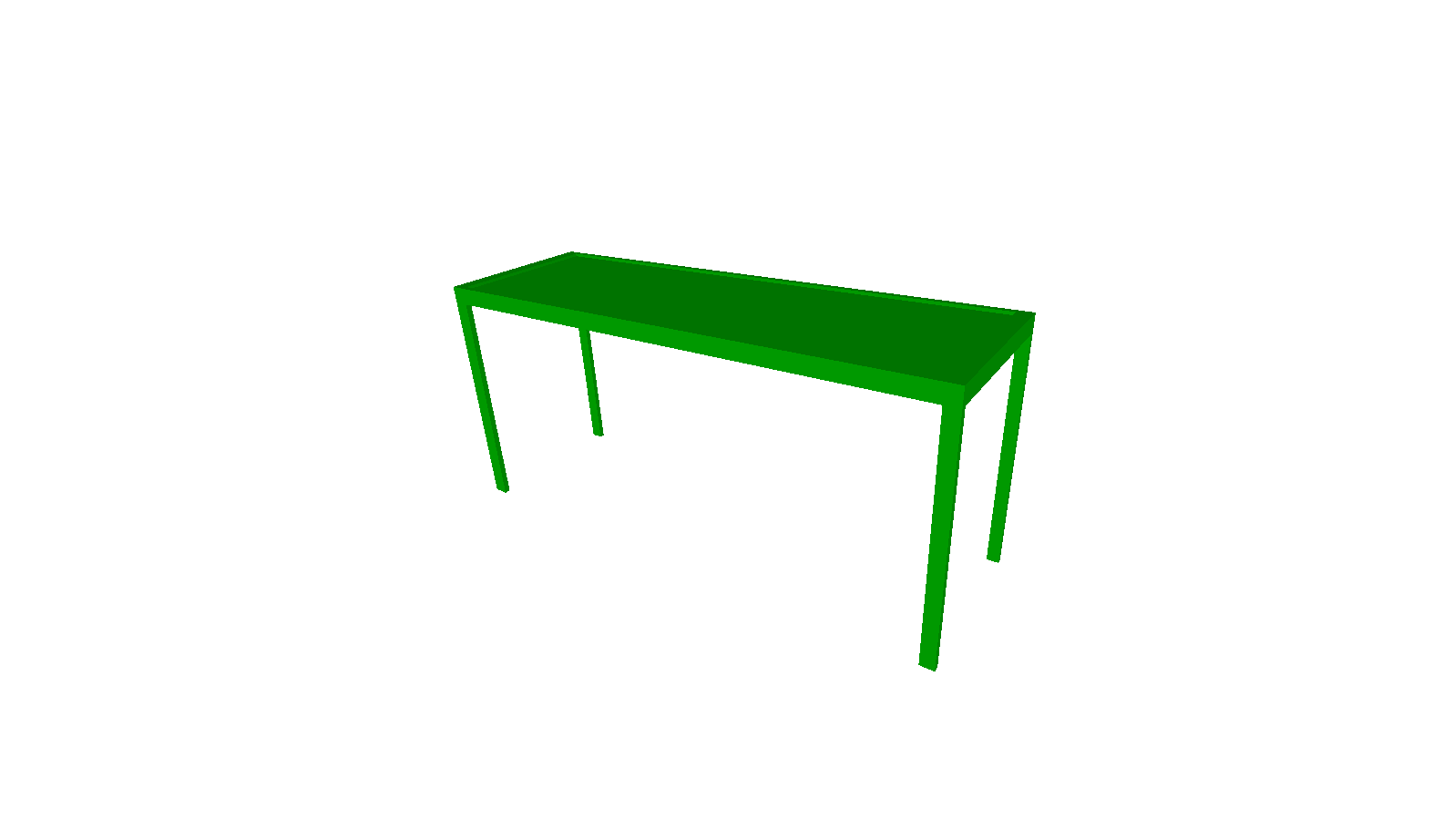} &
 \includegraphics[trim={12cm 5cm 10cm 5cm},clip,width=\widthtopfv\linewidth]{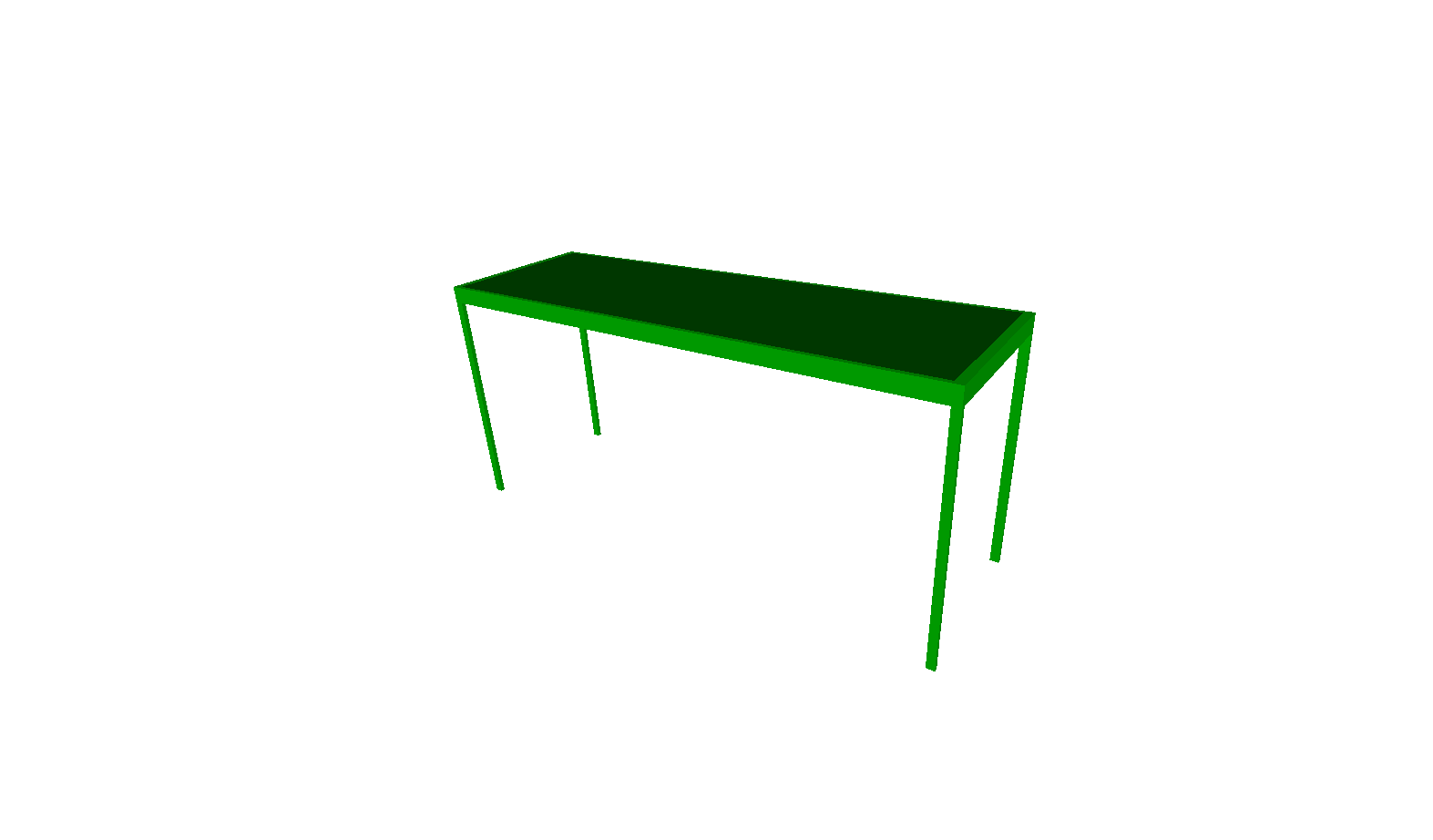} \\

  \includegraphics[trim={15cm 2.5cm 15cm 5.cm},clip,width=\widthtopfv\linewidth]{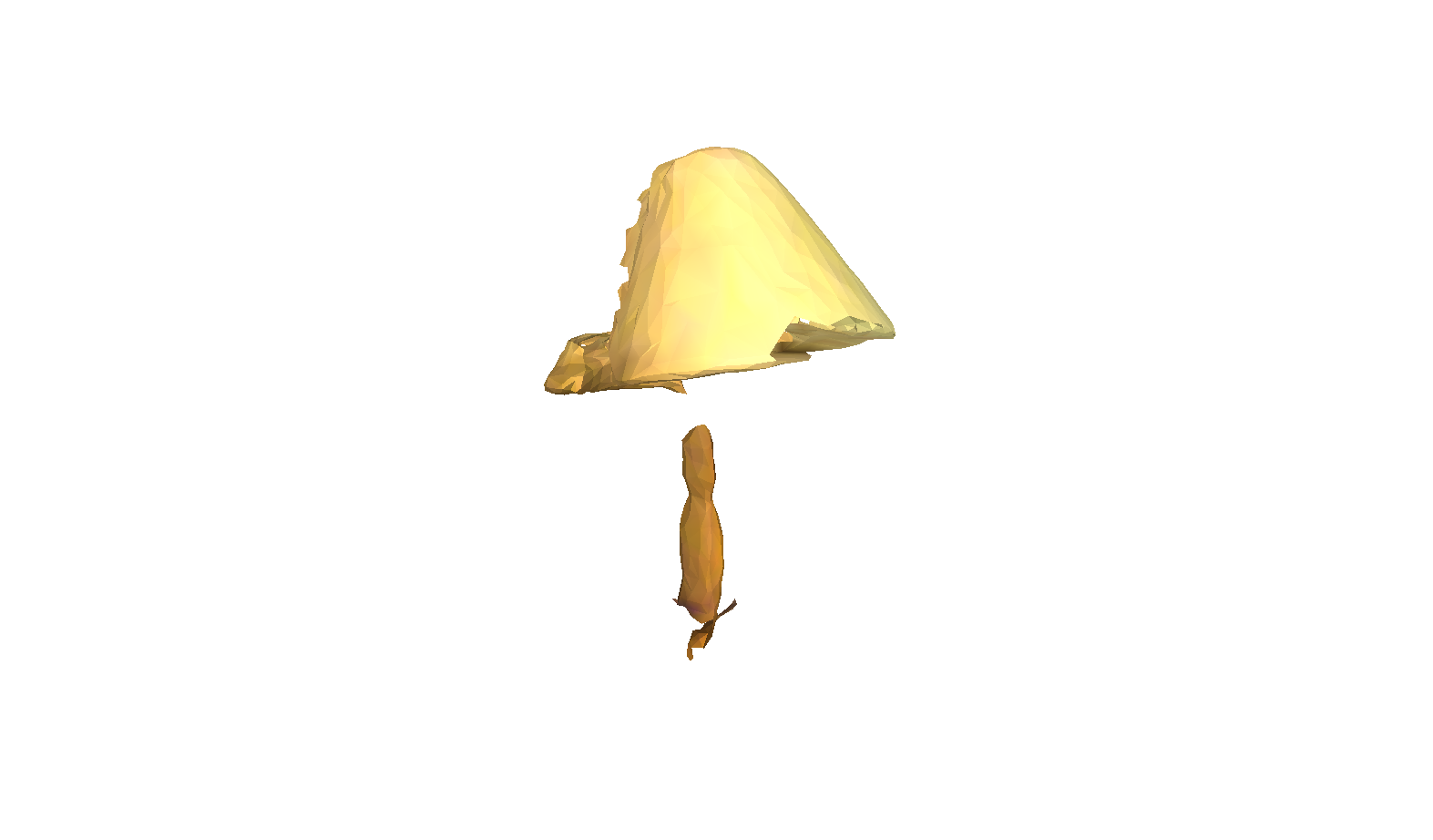} &
 \includegraphics[trim={15cm 2.5cm 15cm 5.cm},clip,width=\widthtopfv\linewidth]{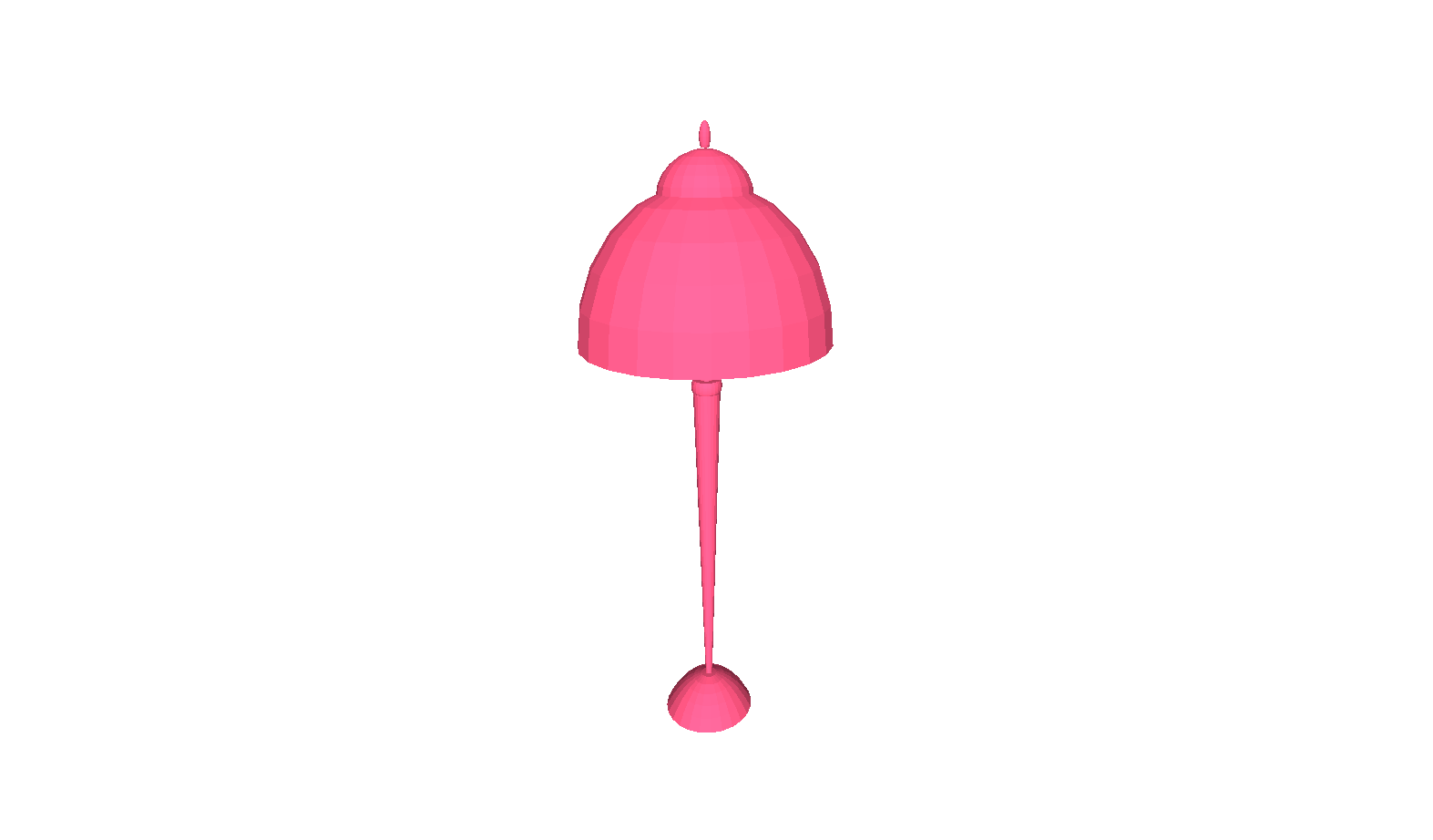} &  
 \includegraphics[trim={15cm 2.5cm 15cm 5.cm},clip,width=\widthtopfv\linewidth]{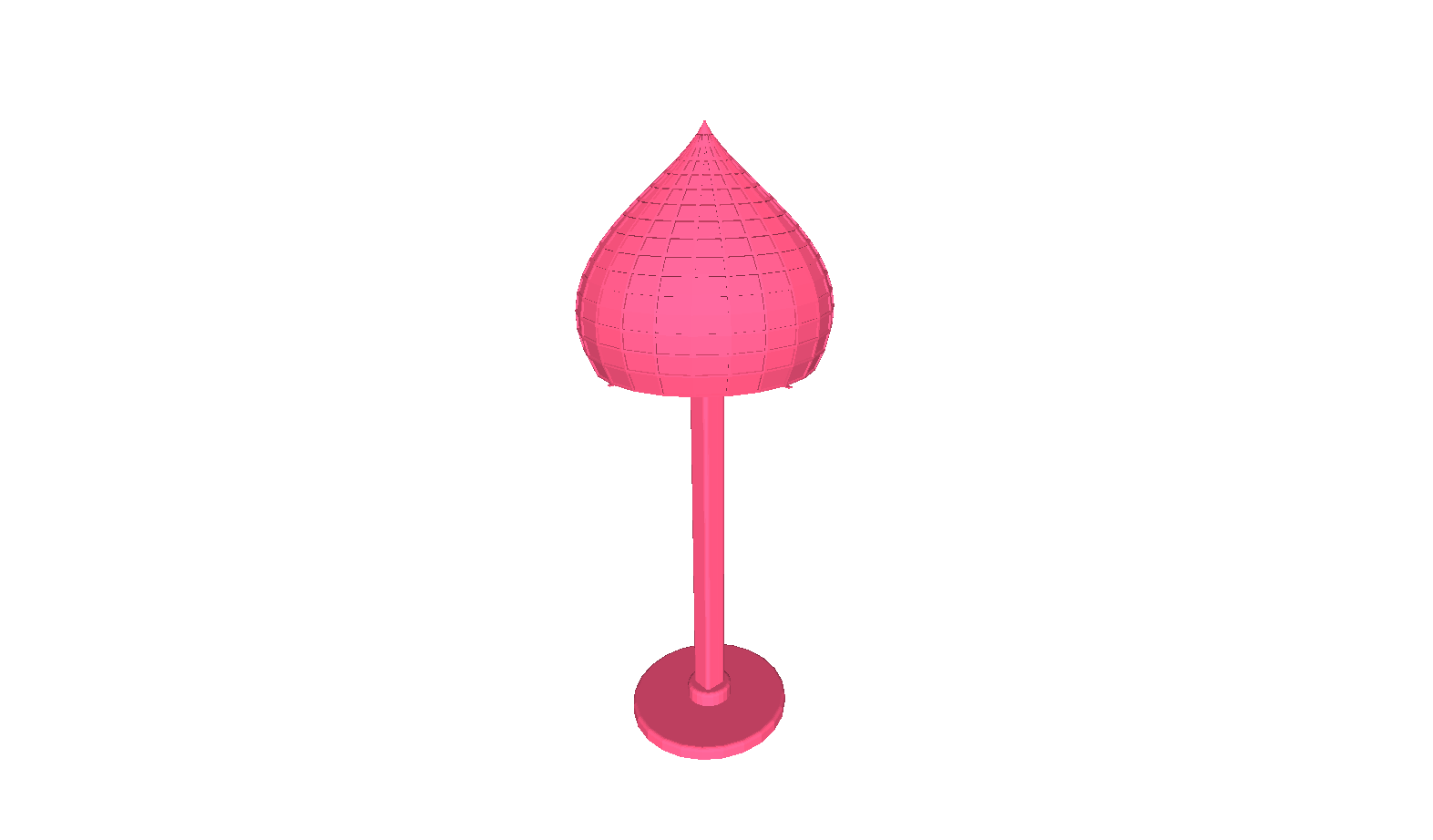} &
  \includegraphics[trim={15cm 2.5cm 15cm 5.cm},clip,width=\widthtopfv\linewidth]{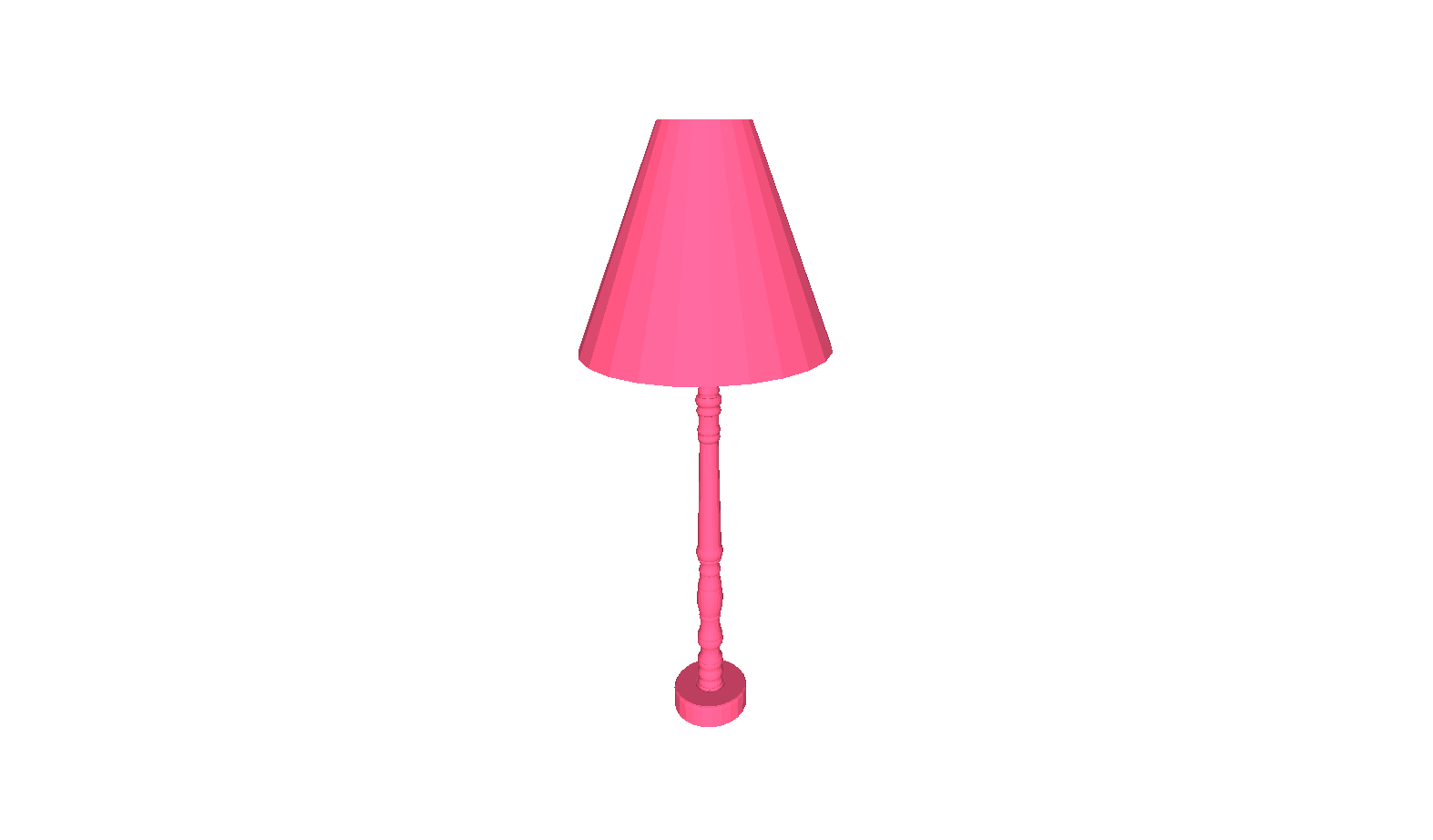} &
  \includegraphics[trim={15cm 2.5cm 15cm 5.cm},clip,width=\widthtopfv\linewidth]{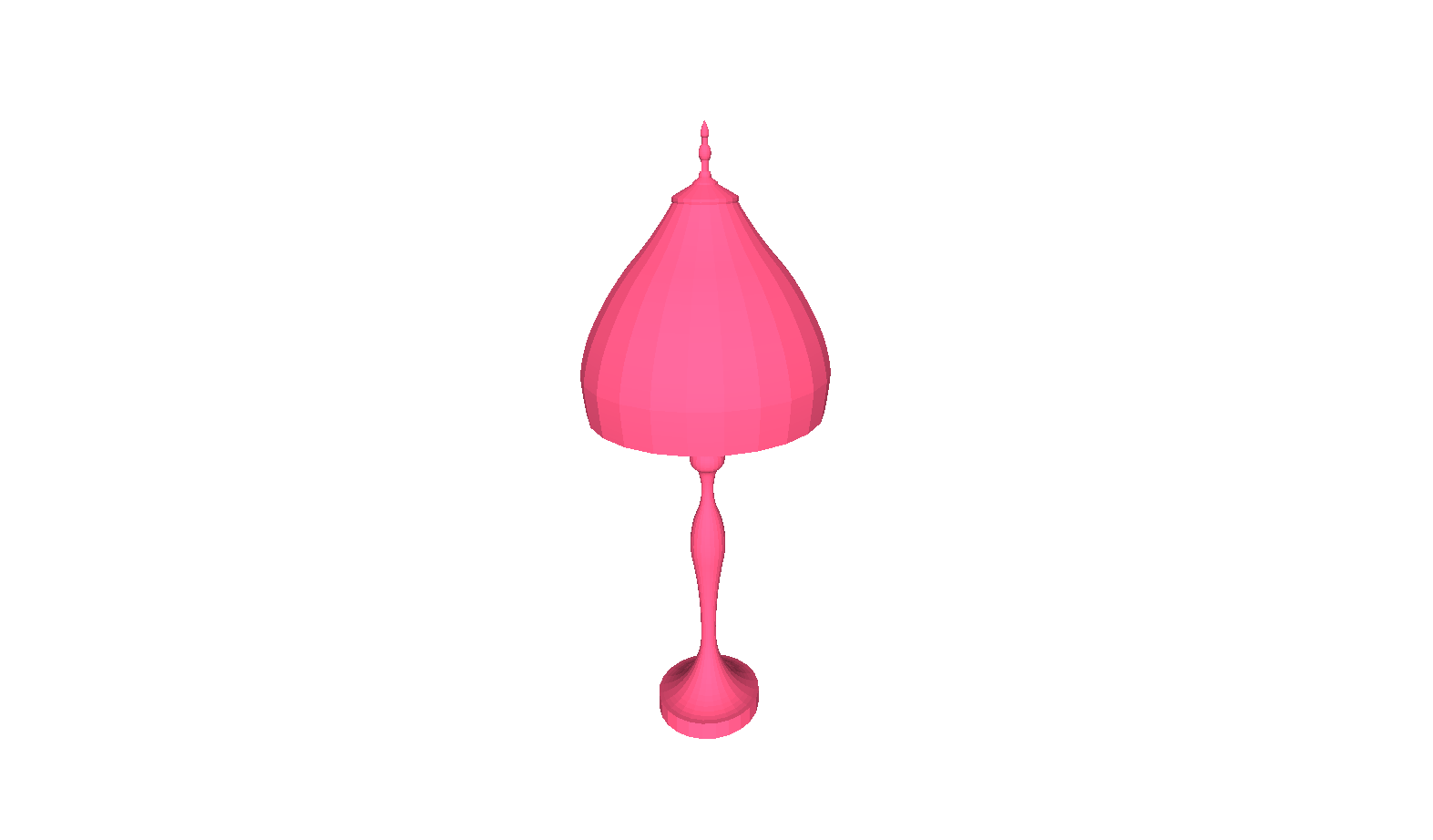} &
  \includegraphics[trim={15cm 2.5cm 15cm 5.cm},clip,width=\widthtopfv\linewidth]{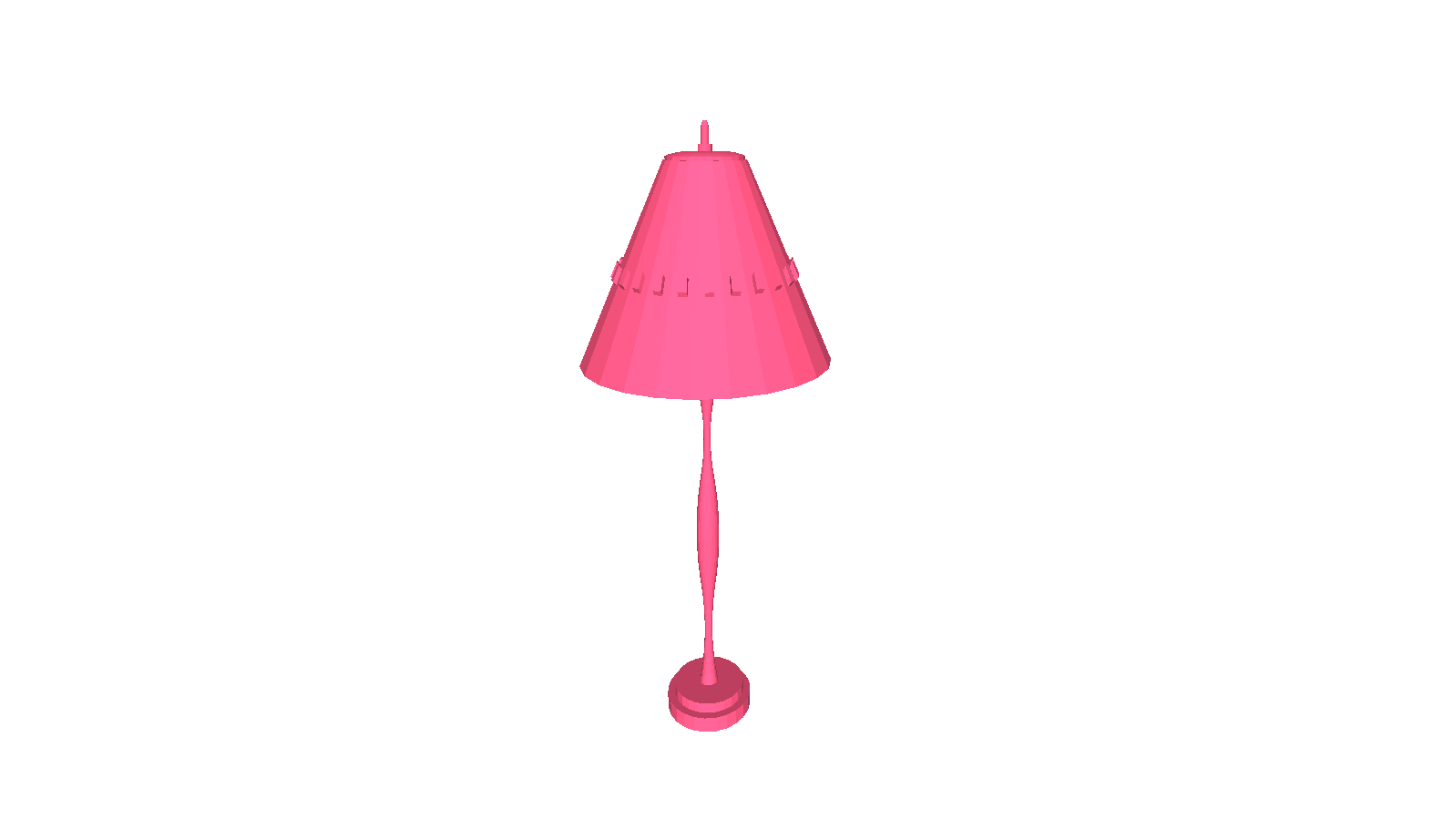} &
  \includegraphics[trim={15cm 2.5cm 15cm 5.cm},clip,width=\widthtopfv\linewidth]{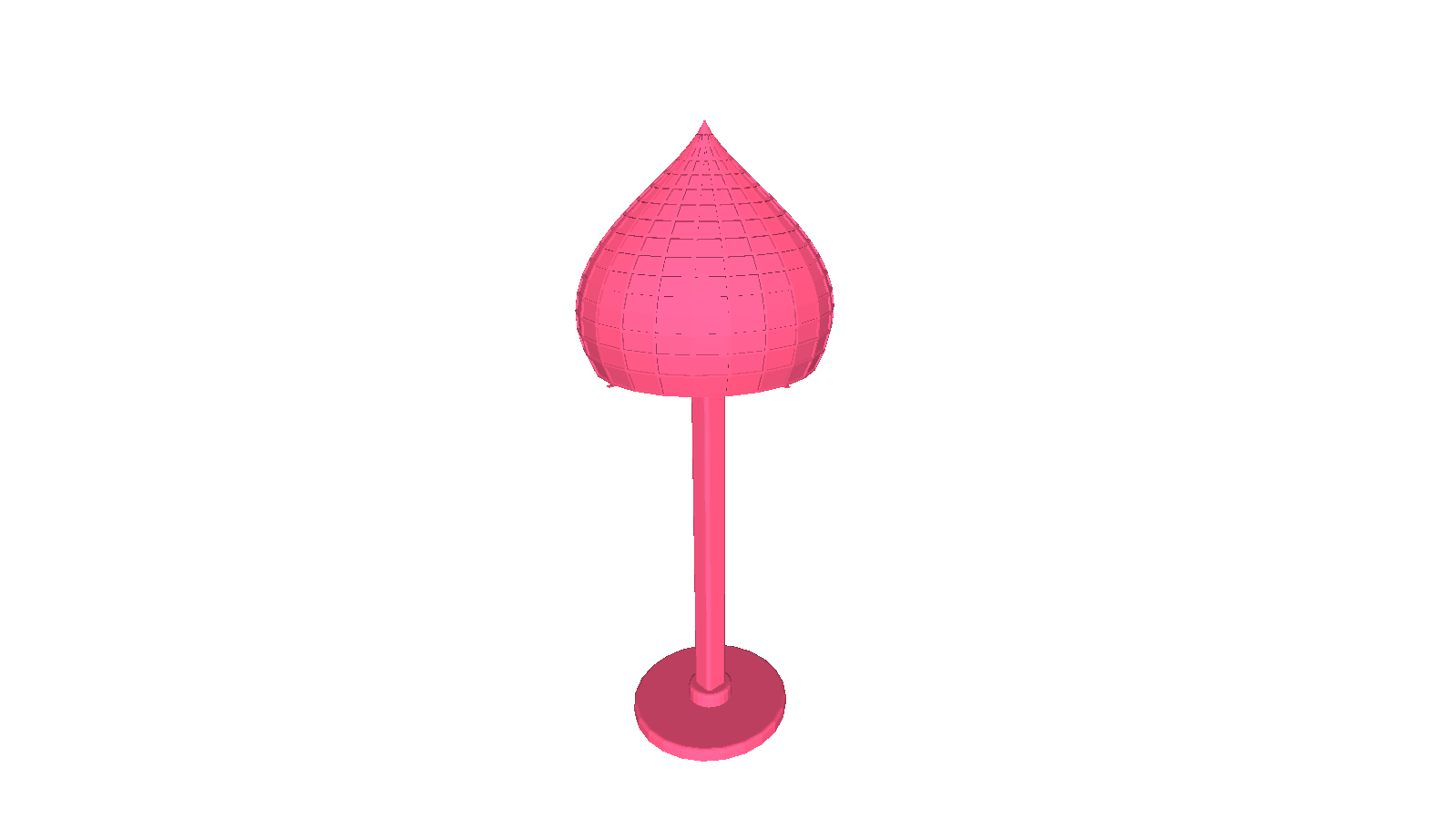} &
  \includegraphics[trim={15cm 2.5cm 15cm 5.cm},clip,width=\widthtopfv\linewidth]{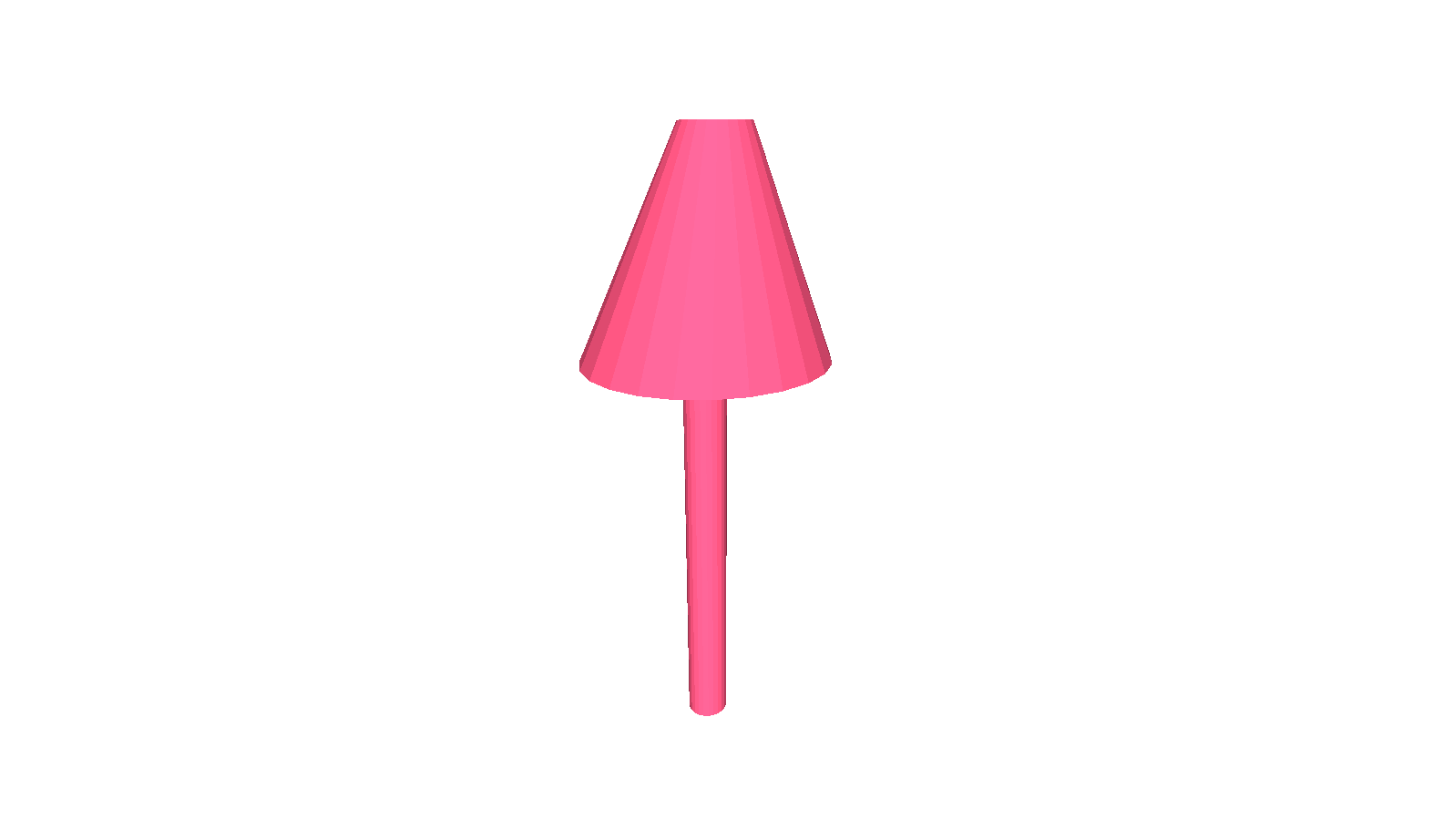} &
  \includegraphics[trim={15cm 2.5cm 15cm 5.cm},clip,width=\widthtopfv\linewidth]{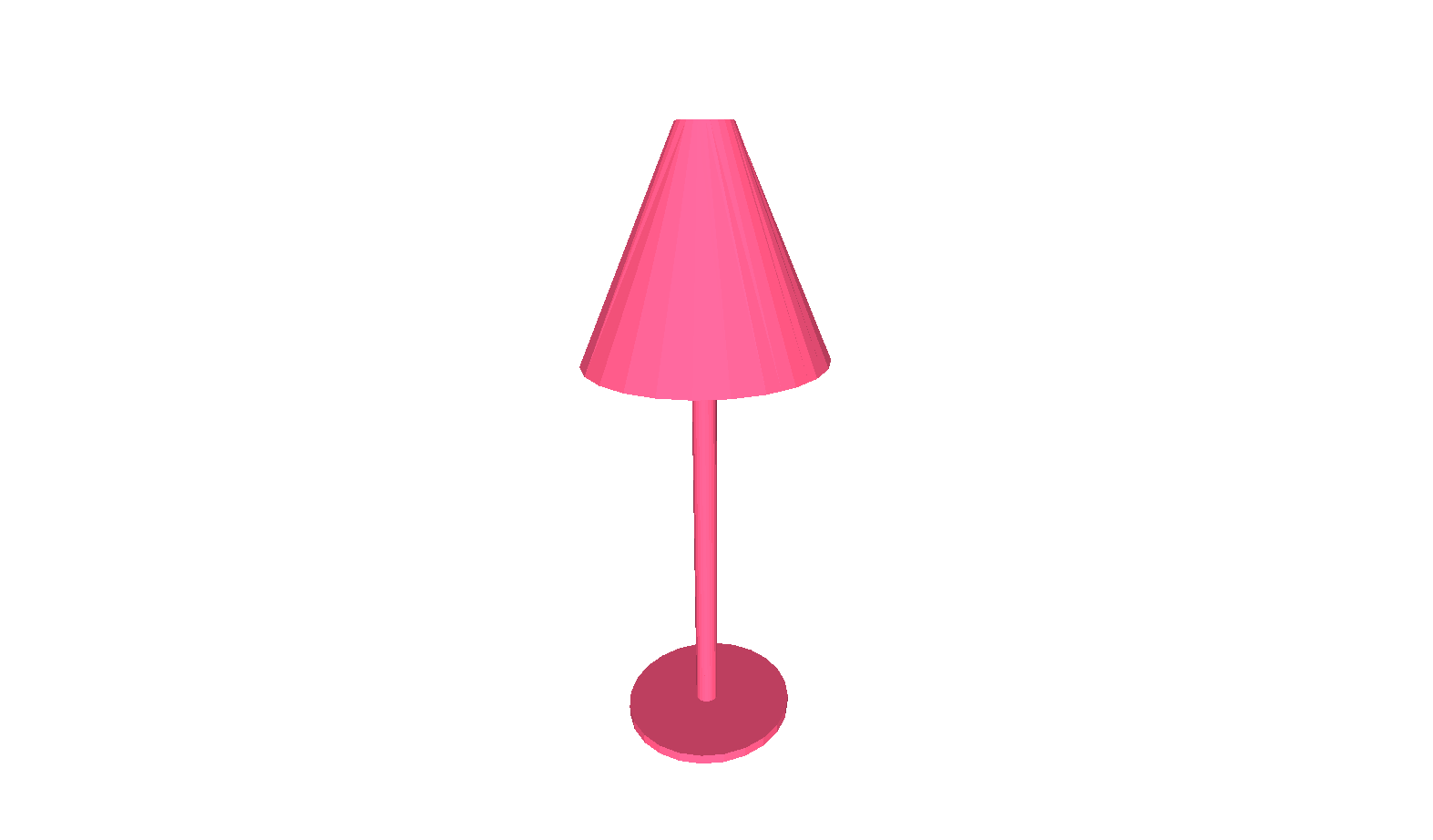} &
  \includegraphics[trim={15cm 2.5cm 15cm 5.cm},clip,width=\widthtopfv\linewidth]{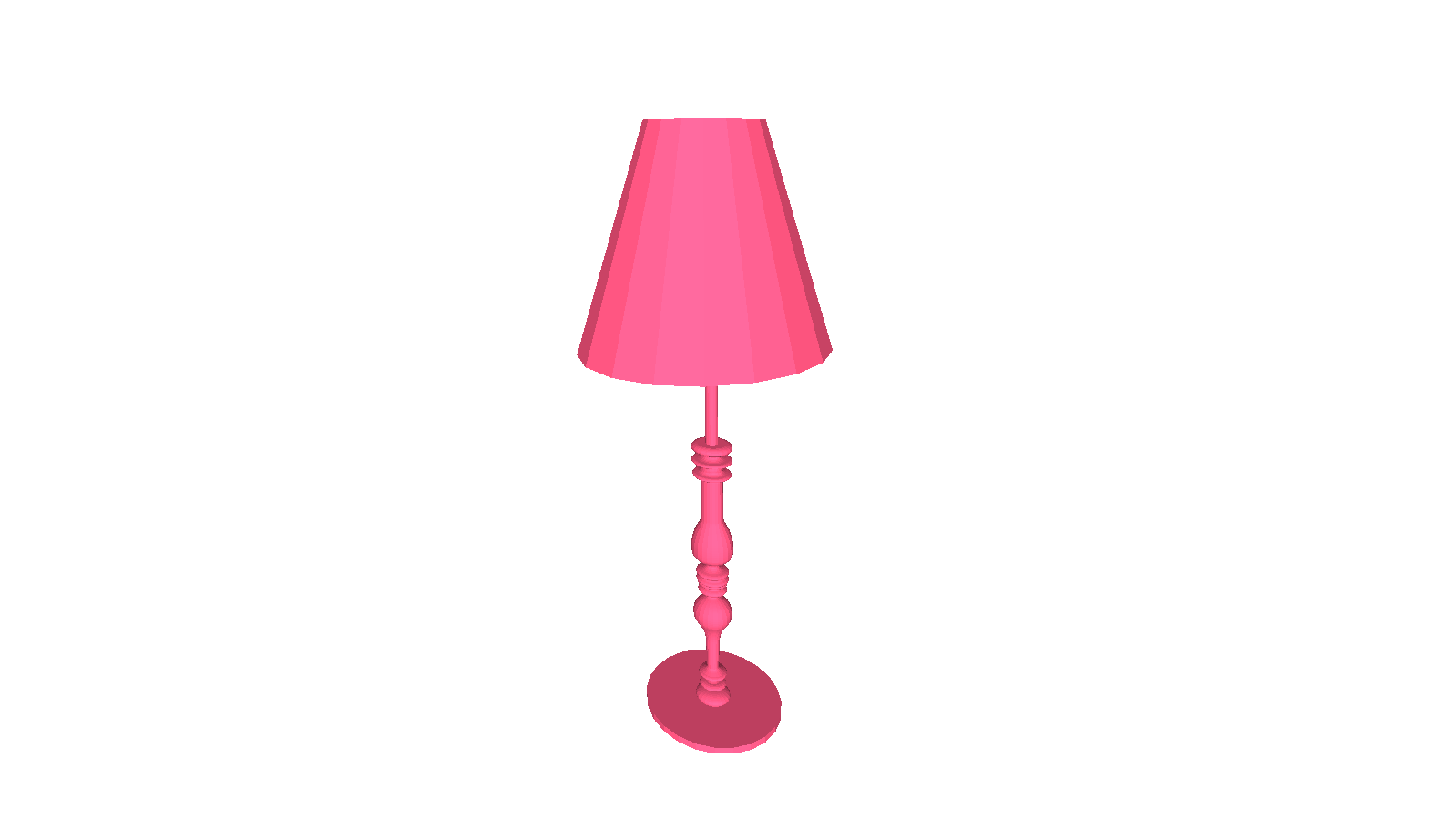} &
  \includegraphics[trim={15cm 2.5cm 15cm 5.cm},clip,width=\widthtopfv\linewidth]{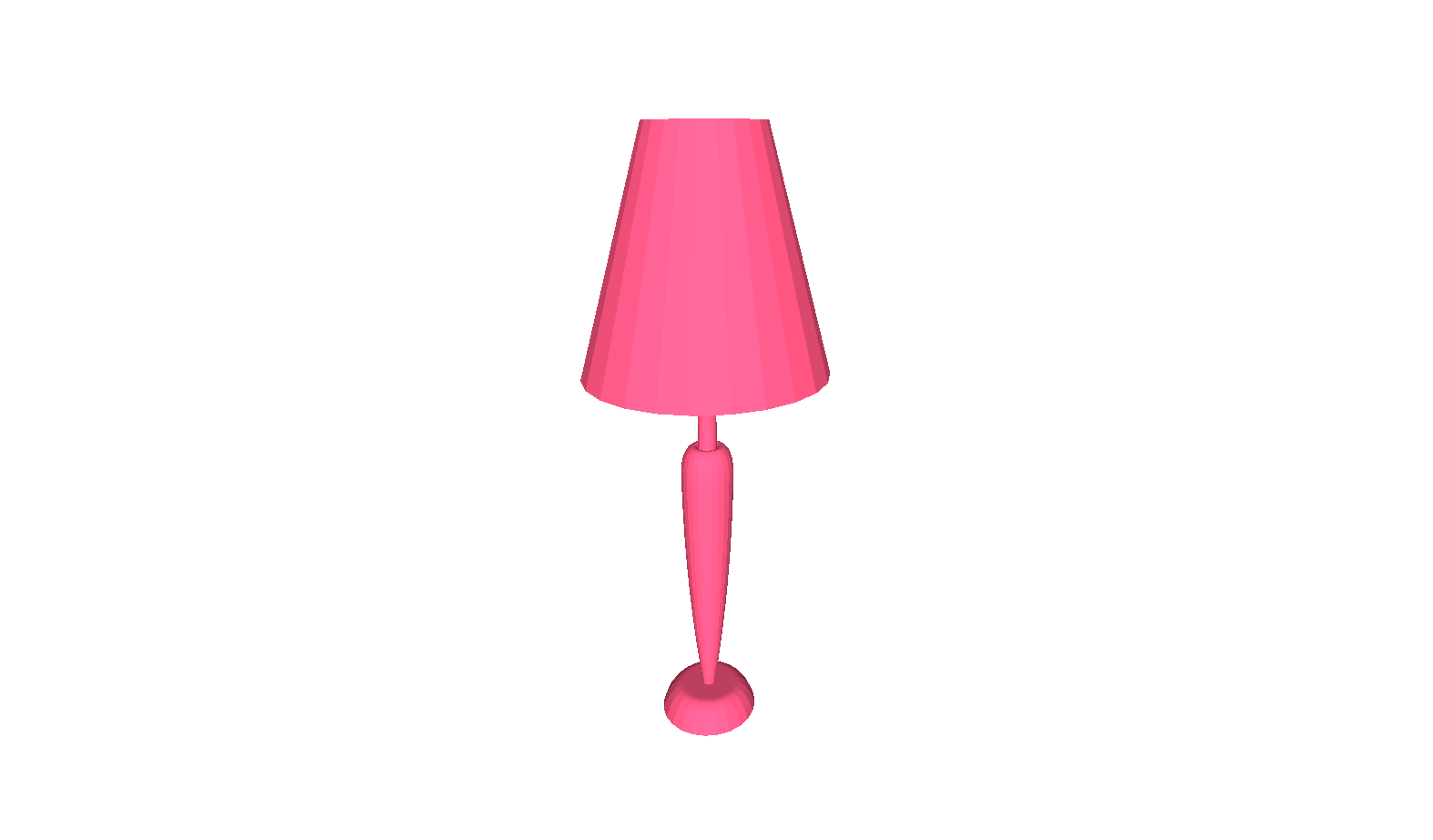} \\

  \includegraphics[trim={15cm 2.5cm 15cm 5.cm},clip,width=\widthtopfv\linewidth]{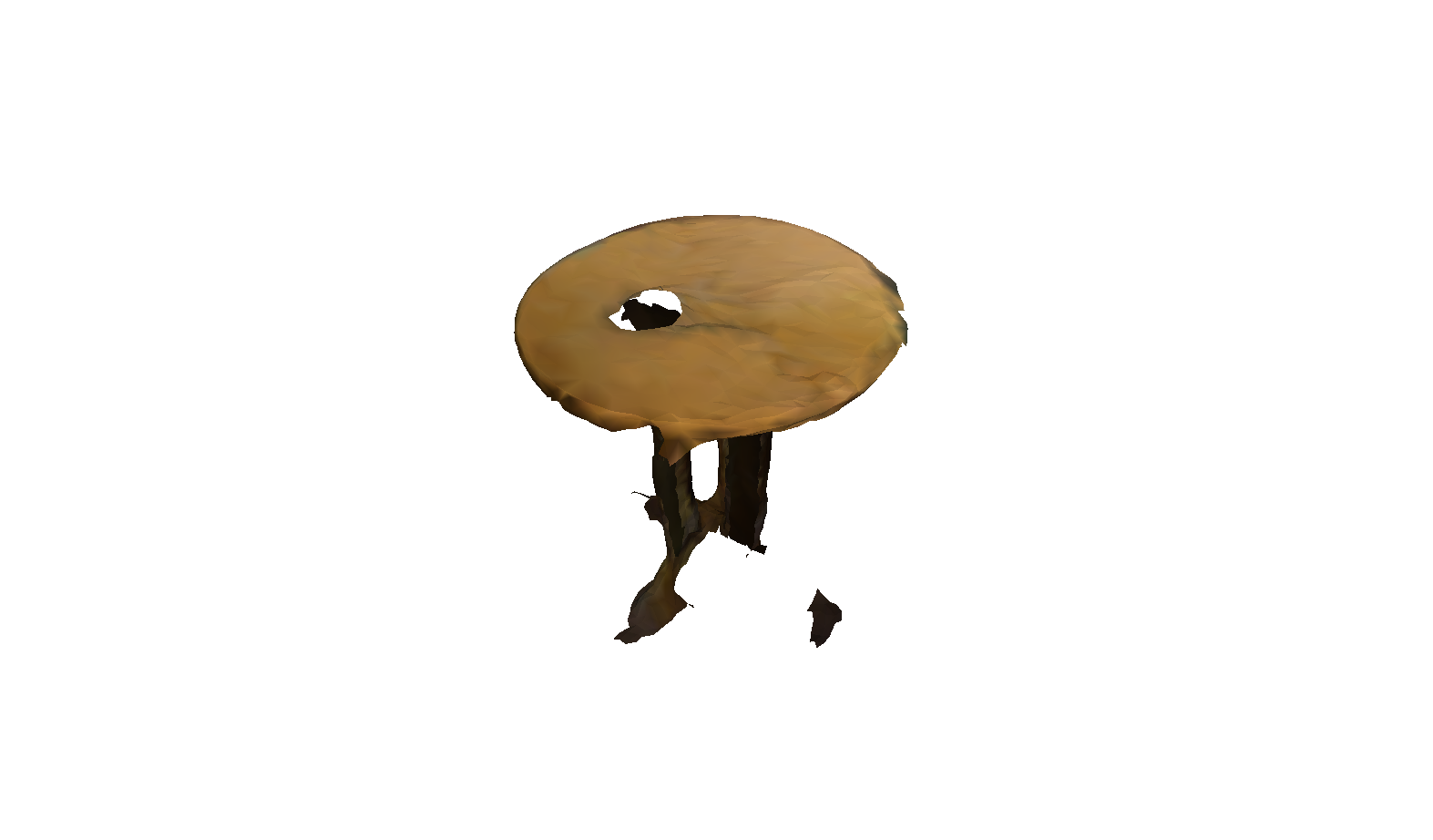} &
  \includegraphics[trim={15cm 2.5cm 15cm 5.cm},clip,width=\widthtopfv\linewidth]{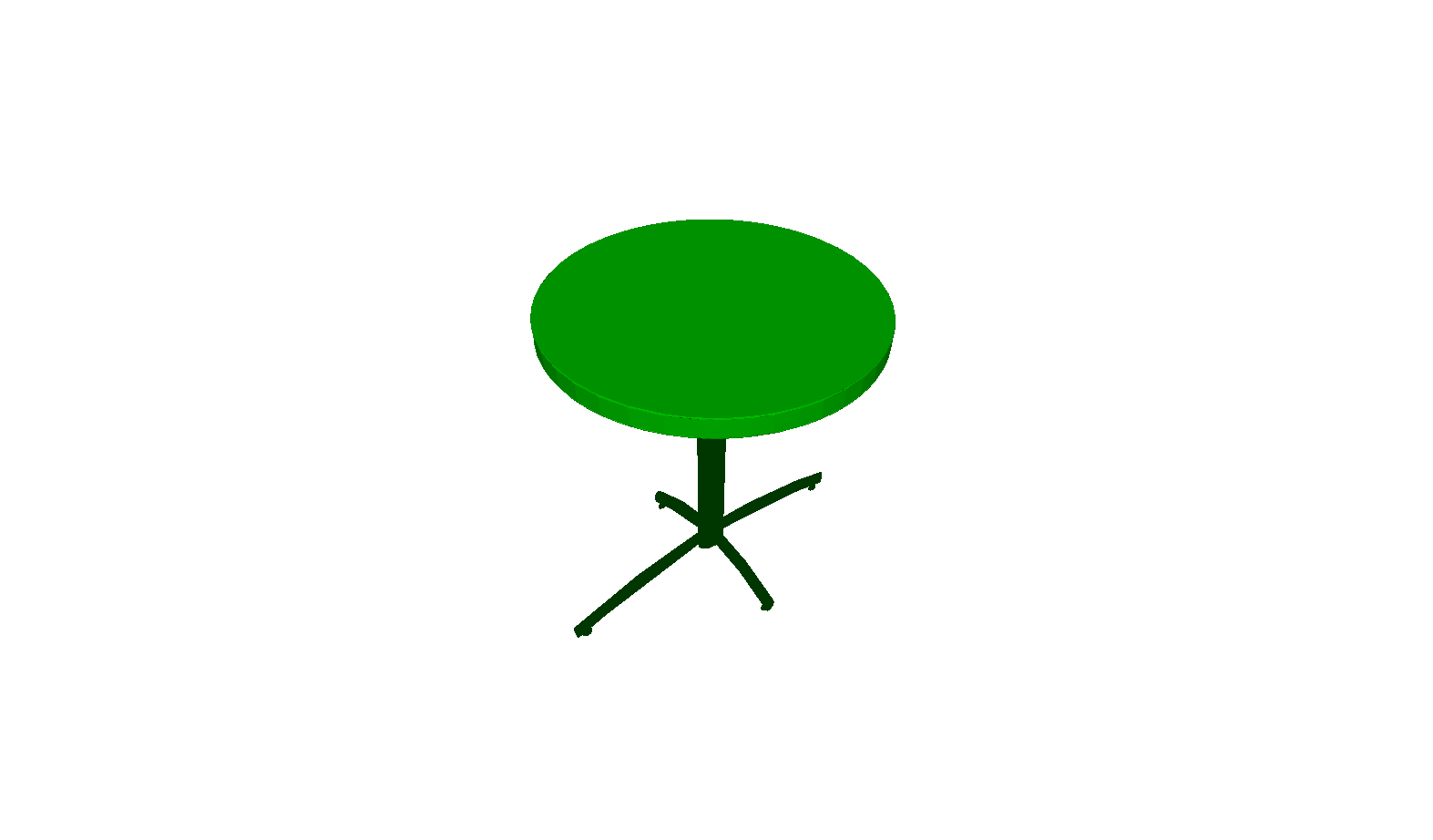} &  \includegraphics[trim={15cm 2.5cm 15cm 5.cm},clip,width=\widthtopfv\linewidth]{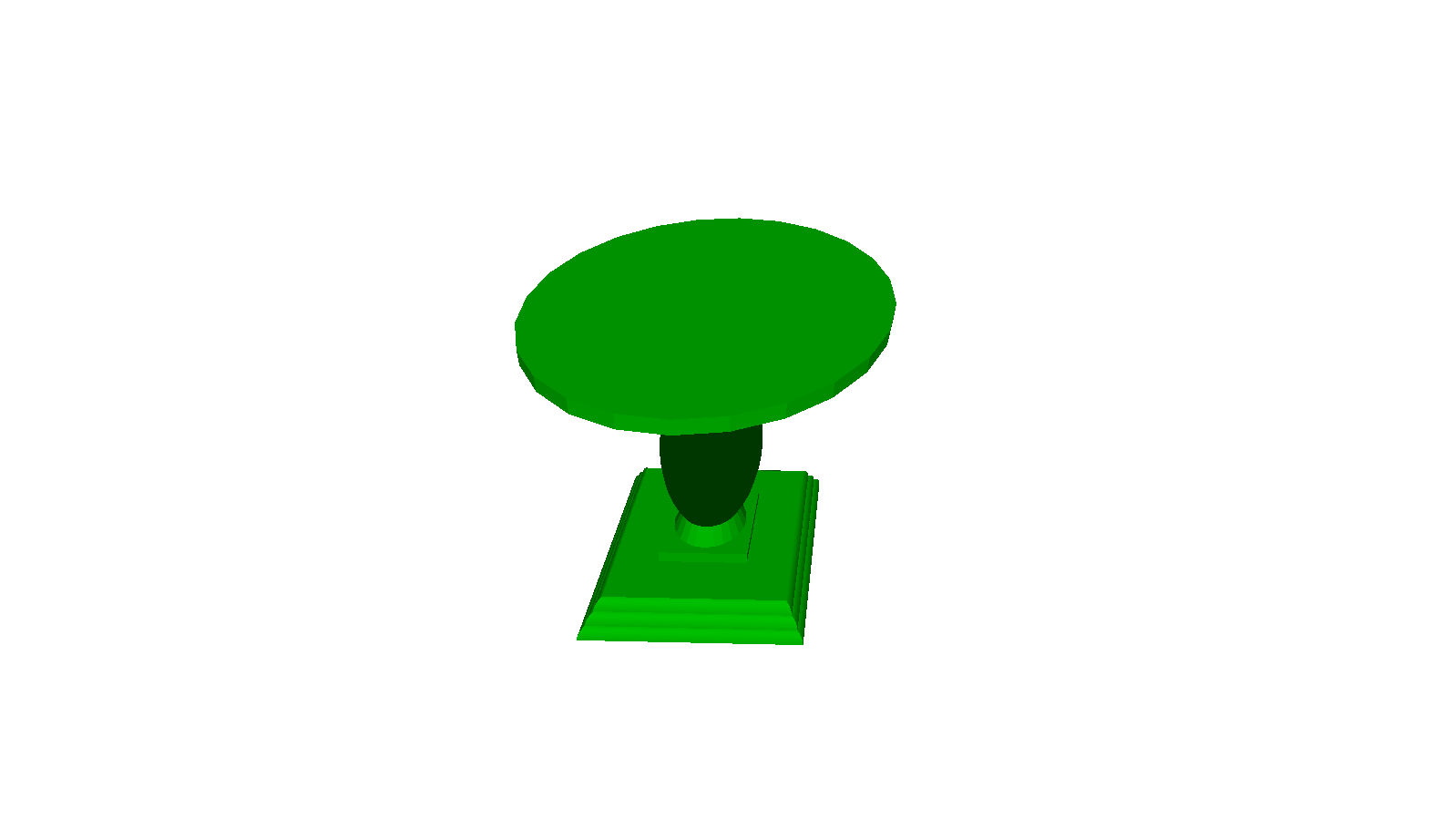} &
  \includegraphics[trim={15cm 2.5cm 15cm 5.cm},clip,width=\widthtopfv\linewidth]{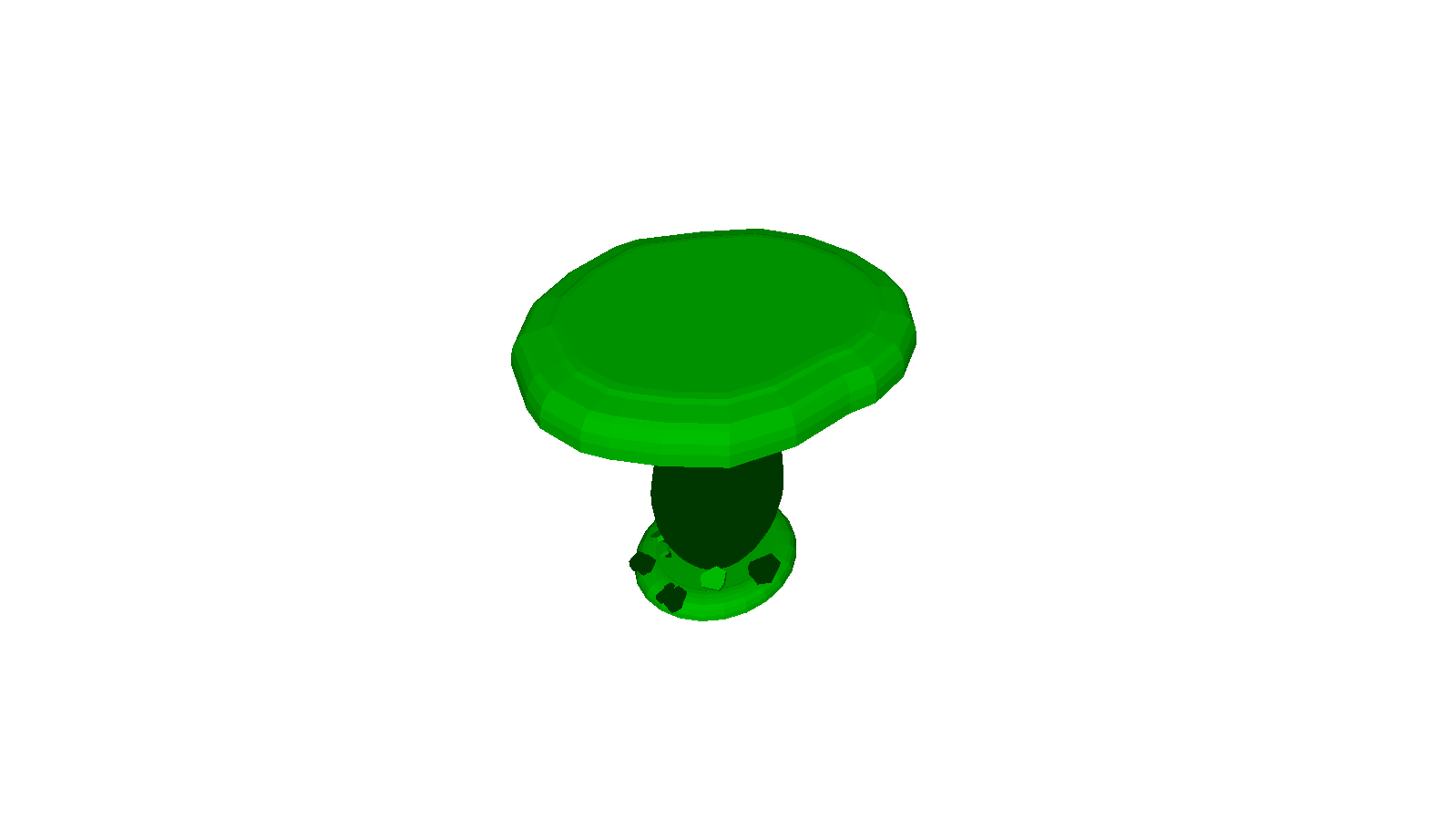} &
  \includegraphics[trim={15cm 2.5cm 15cm 5.cm},clip,width=\widthtopfv\linewidth]{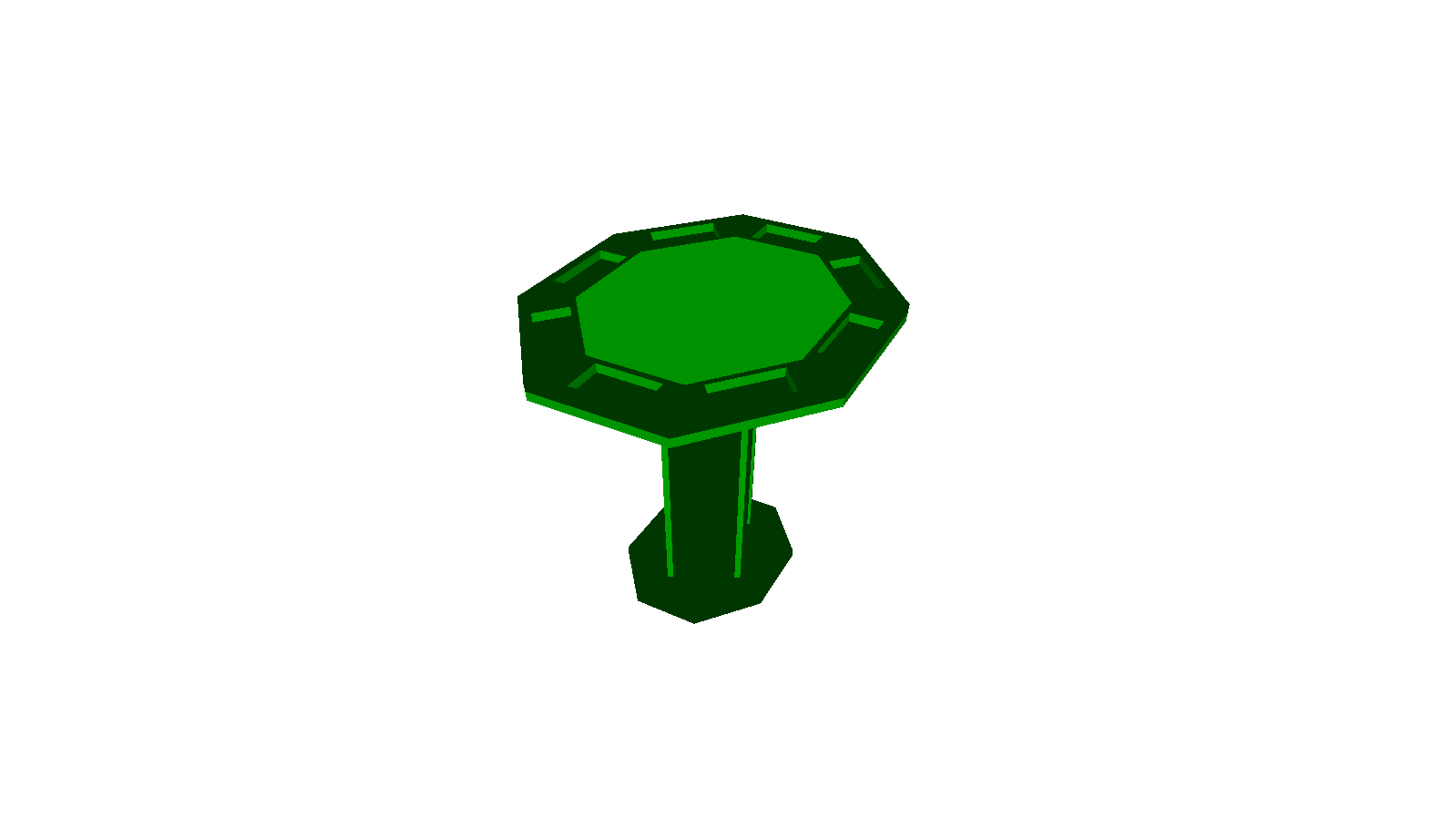} &
  \includegraphics[trim={15cm 2.5cm 15cm 5.cm},clip,width=\widthtopfv\linewidth]{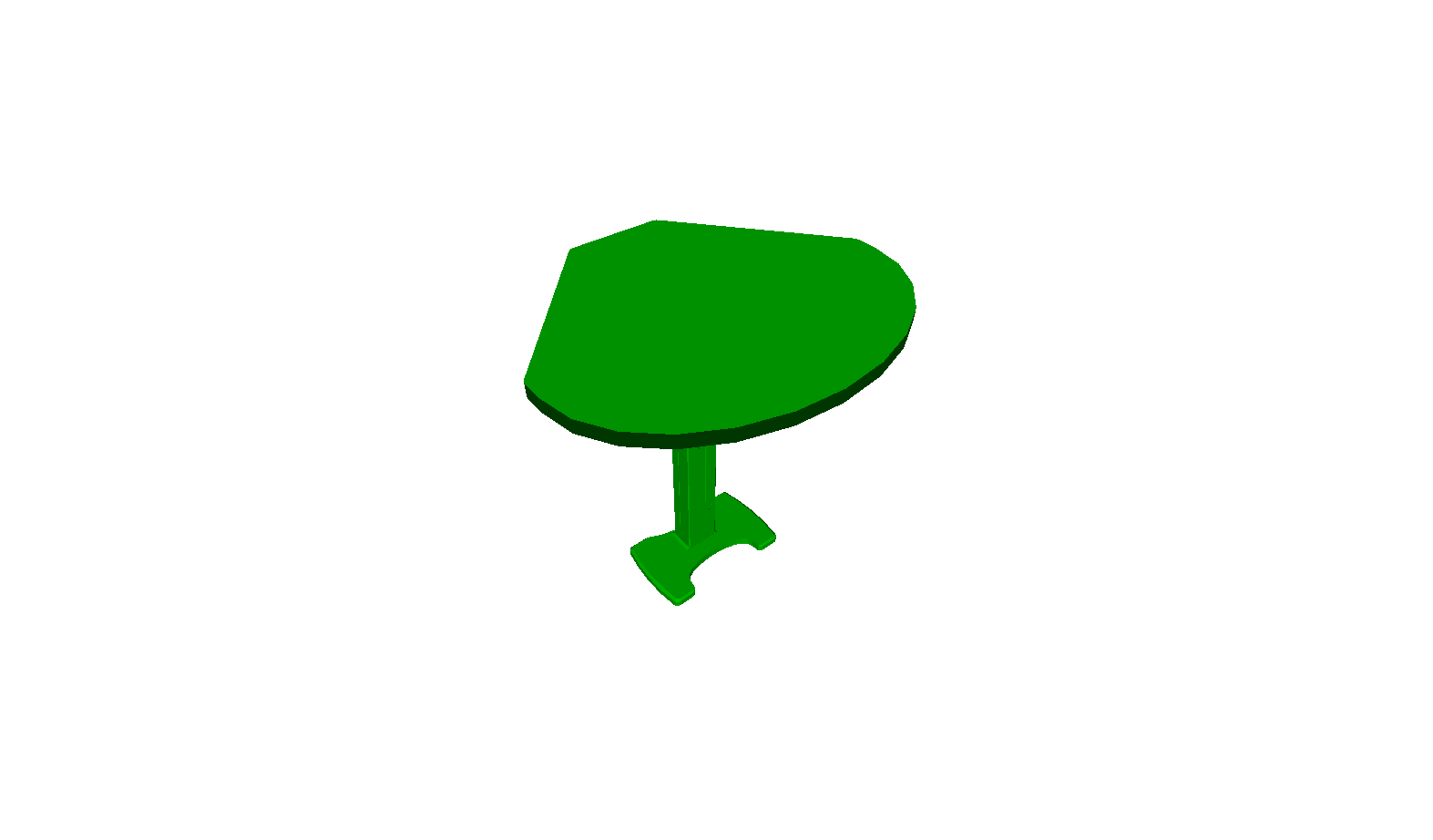} &
  \includegraphics[trim={15cm 2.5cm 15cm 5.cm},clip,width=\widthtopfv\linewidth]{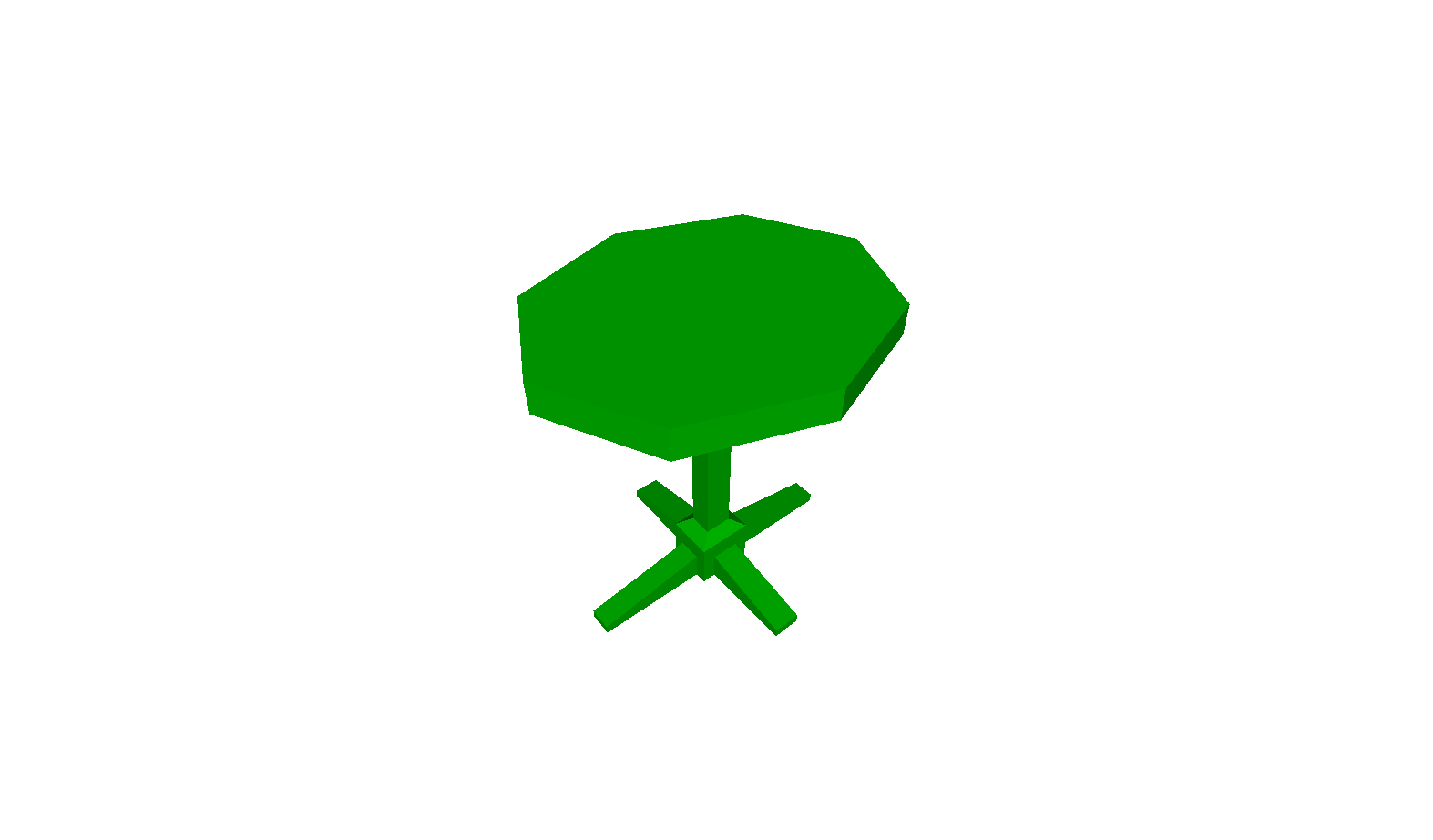} &
  \includegraphics[trim={15cm 2.5cm 15cm 5.cm},clip,width=\widthtopfv\linewidth]{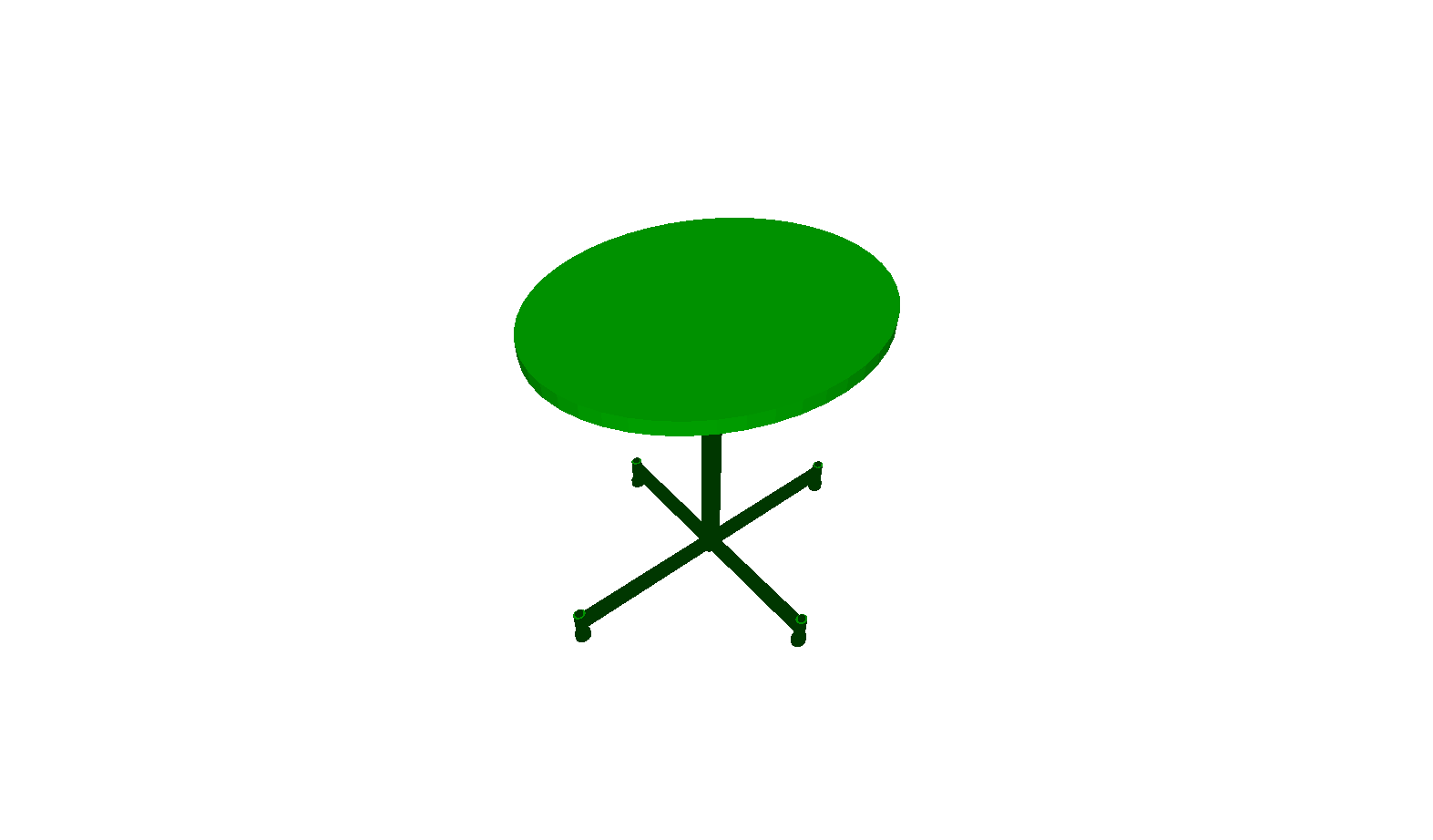} &
  \includegraphics[trim={15cm 2.5cm 15cm 5.cm},clip,width=\widthtopfv\linewidth]{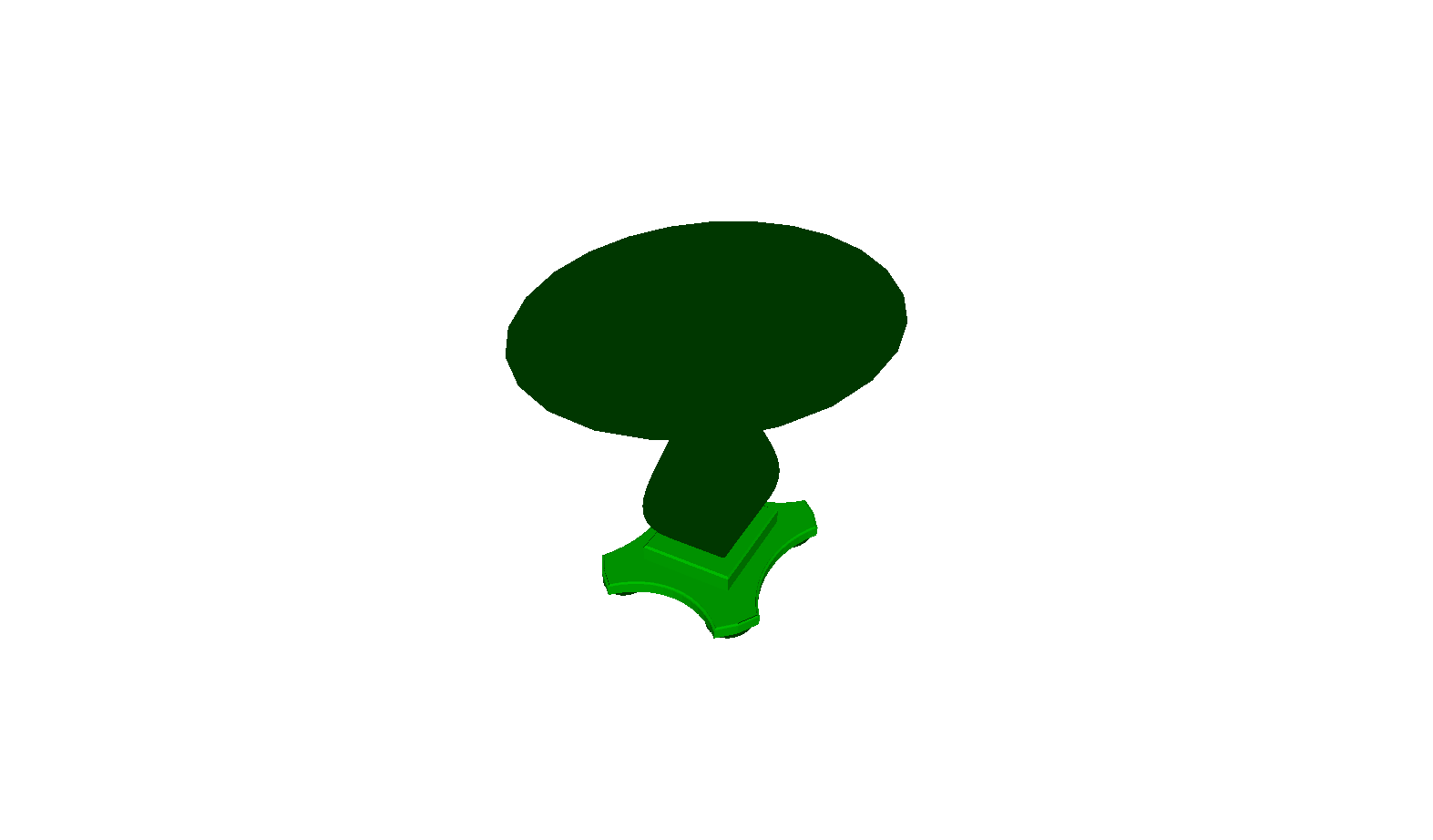} &
  \includegraphics[trim={15cm 2.5cm 15cm 5.cm},clip,width=\widthtopfv\linewidth]{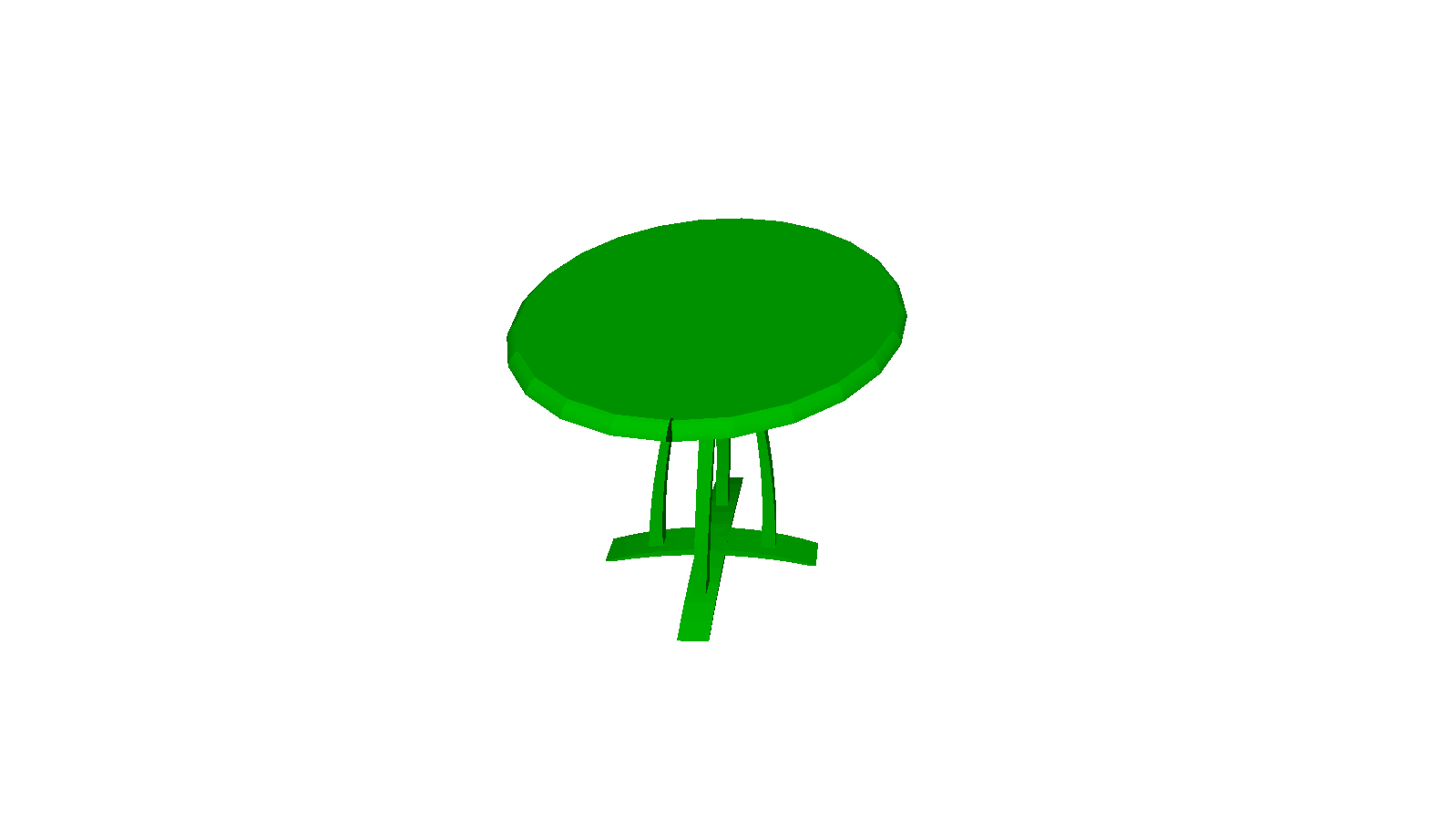} &
  \includegraphics[trim={15cm 2.5cm 15cm 5.cm},clip,width=\widthtopfv\linewidth]{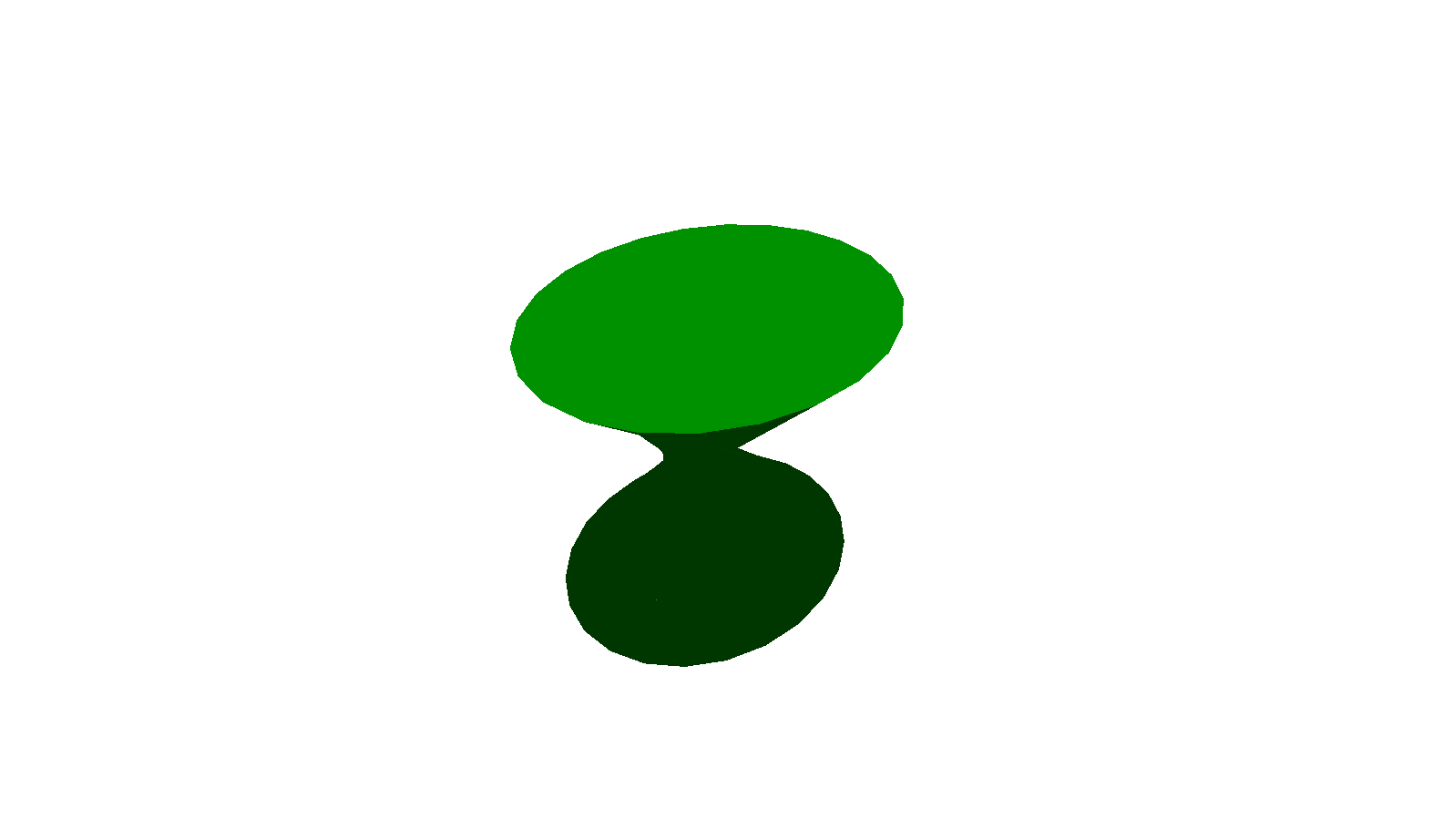} \\

      \includegraphics[trim={13cm 2.5cm 15cm 3.cm},clip,width=\widthtopfv\linewidth]{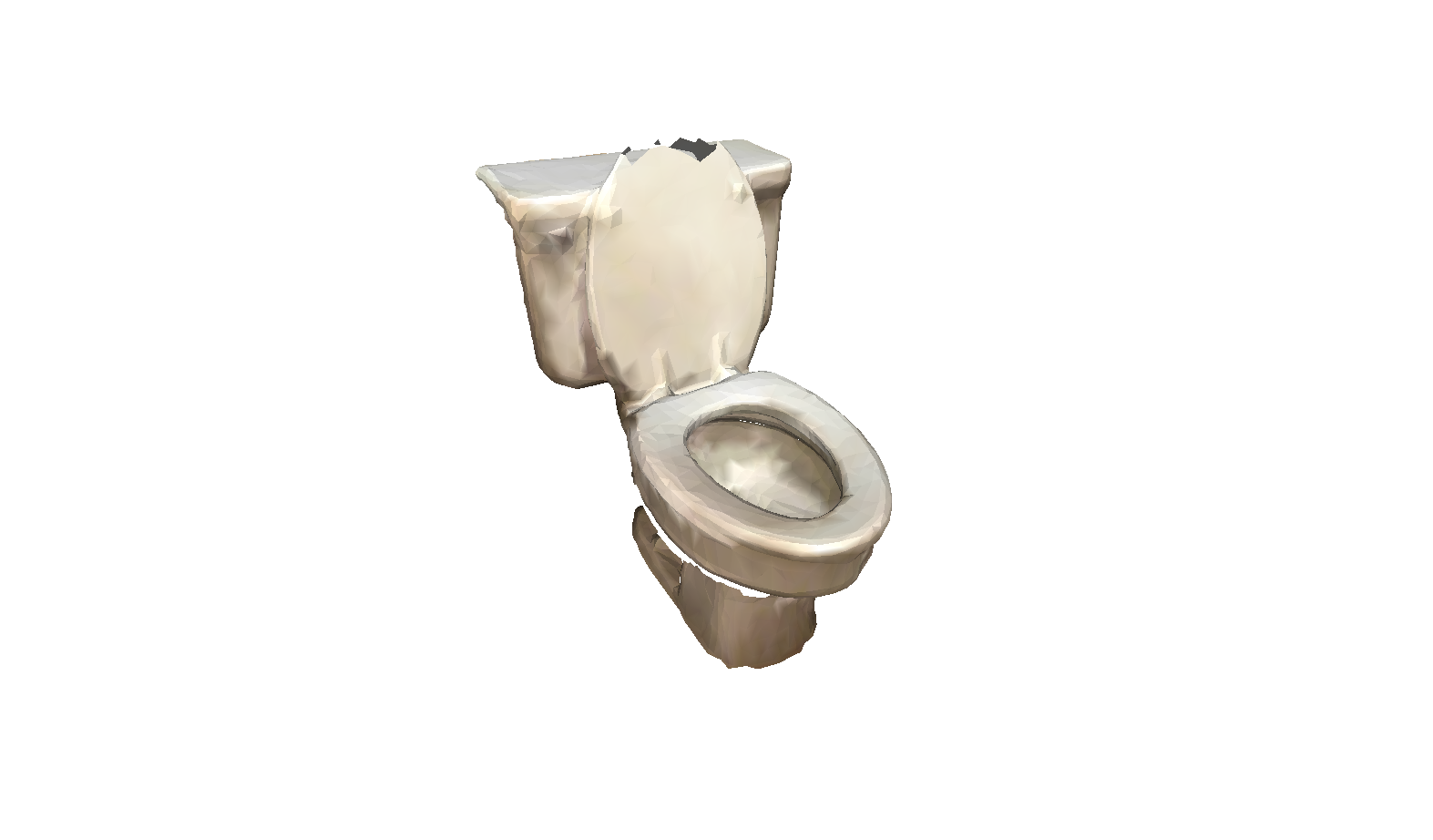} &
  \includegraphics[trim={13cm 2.5cm 15cm 3.cm},clip,width=\widthtopfv\linewidth]{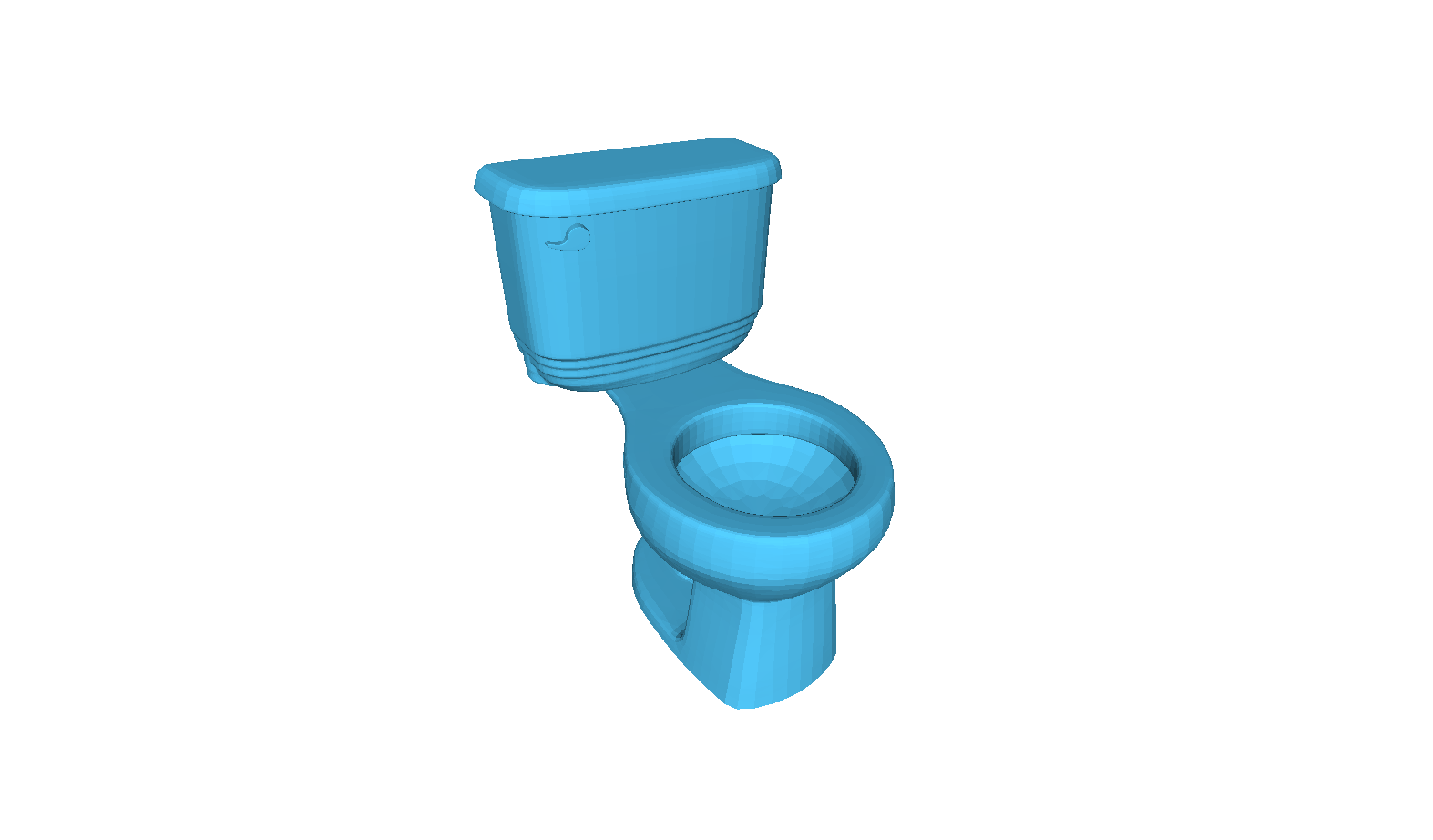} &  
  \includegraphics[trim={13cm 2.5cm 15cm 3.cm},clip,width=\widthtopfv\linewidth]{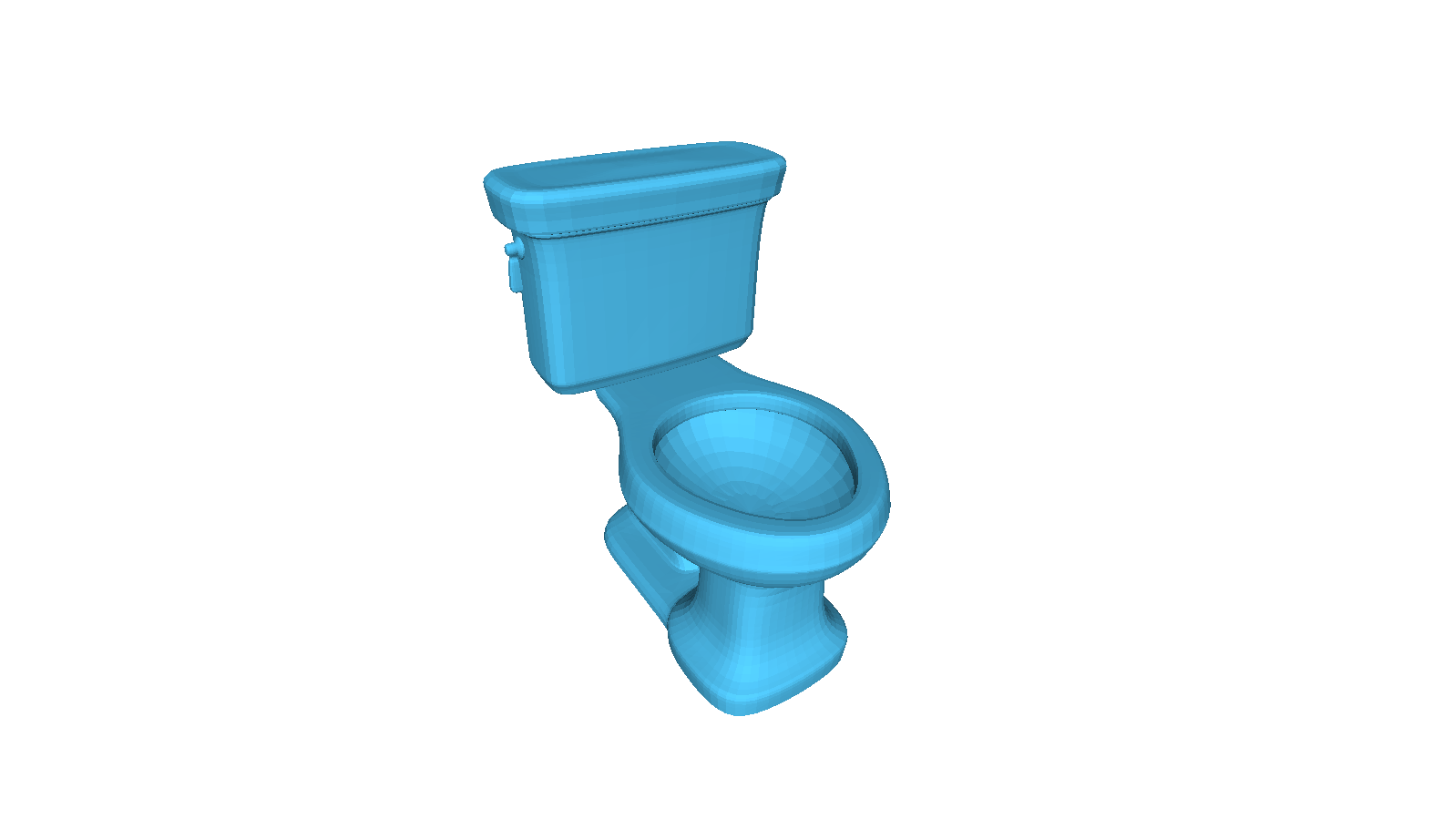} &
  \includegraphics[trim={13cm 2.5cm 15cm 3.cm},clip,width=\widthtopfv\linewidth]{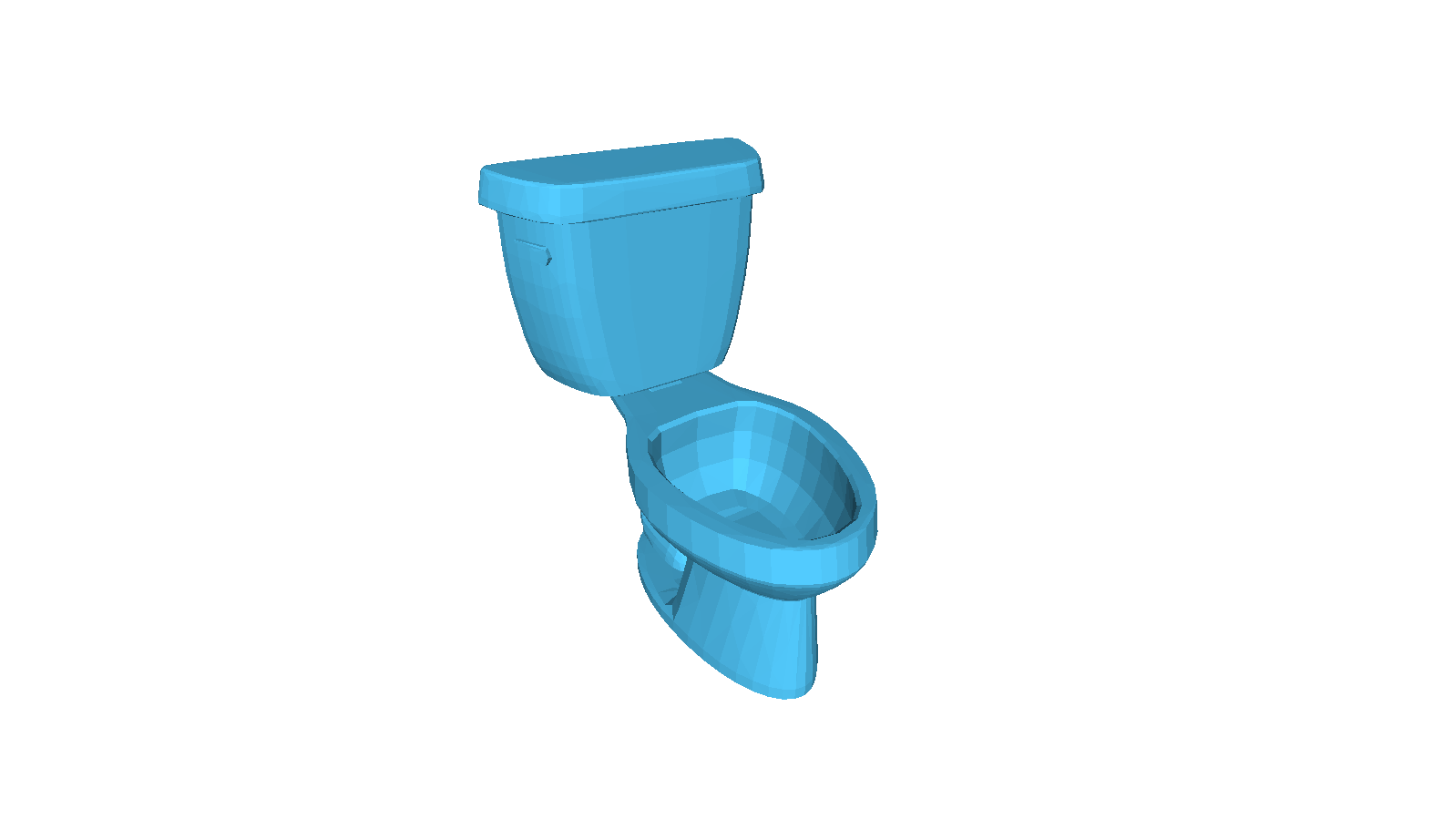} &
  \includegraphics[trim={13cm 2.5cm 15cm 3.cm},clip,width=\widthtopfv\linewidth]{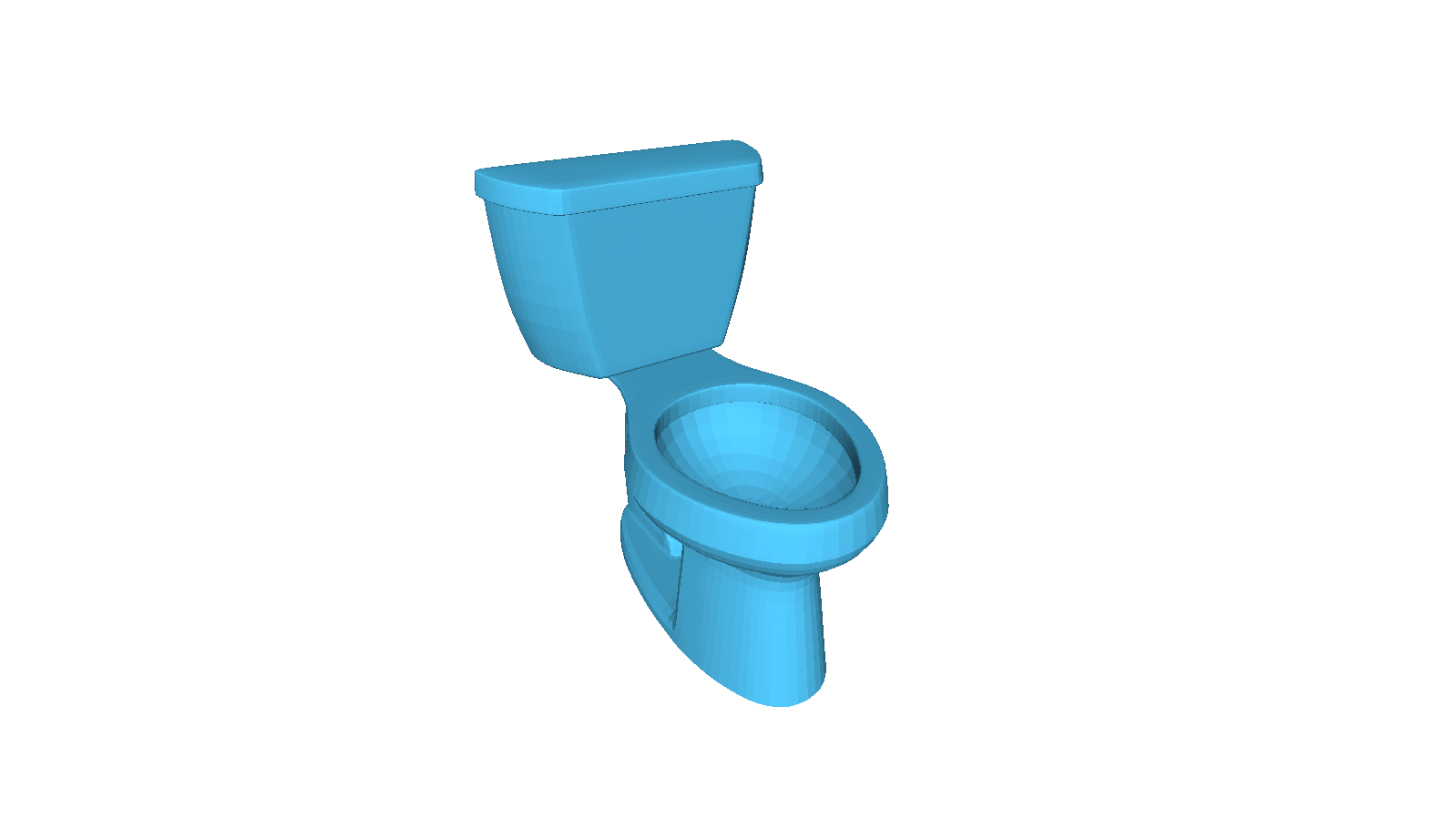} &
  \includegraphics[trim={13cm 2.5cm 15cm 3.cm},clip,width=\widthtopfv\linewidth]{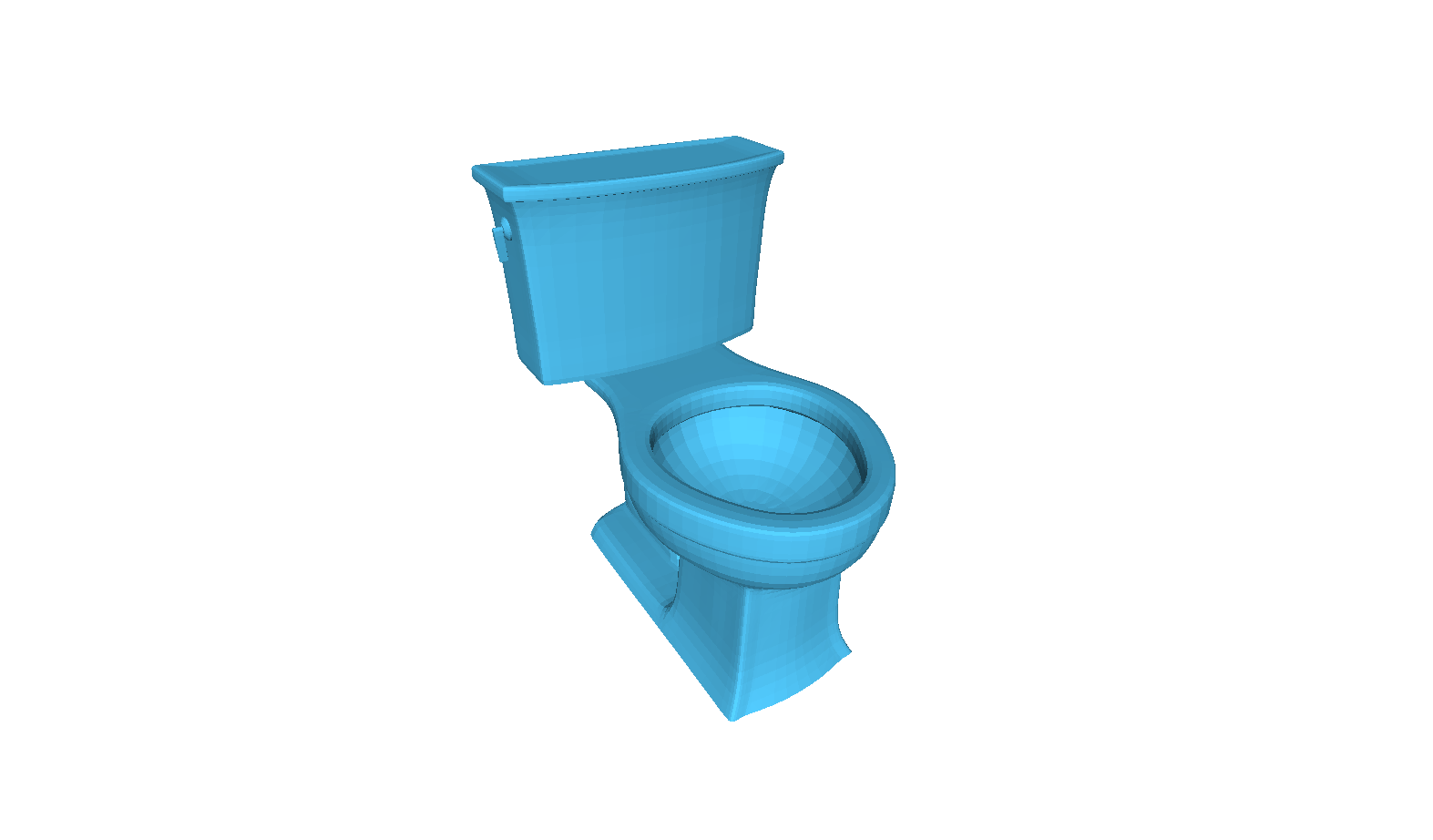} &
  \includegraphics[trim={13cm 2.5cm 15cm 3.cm},clip,width=\widthtopfv\linewidth]{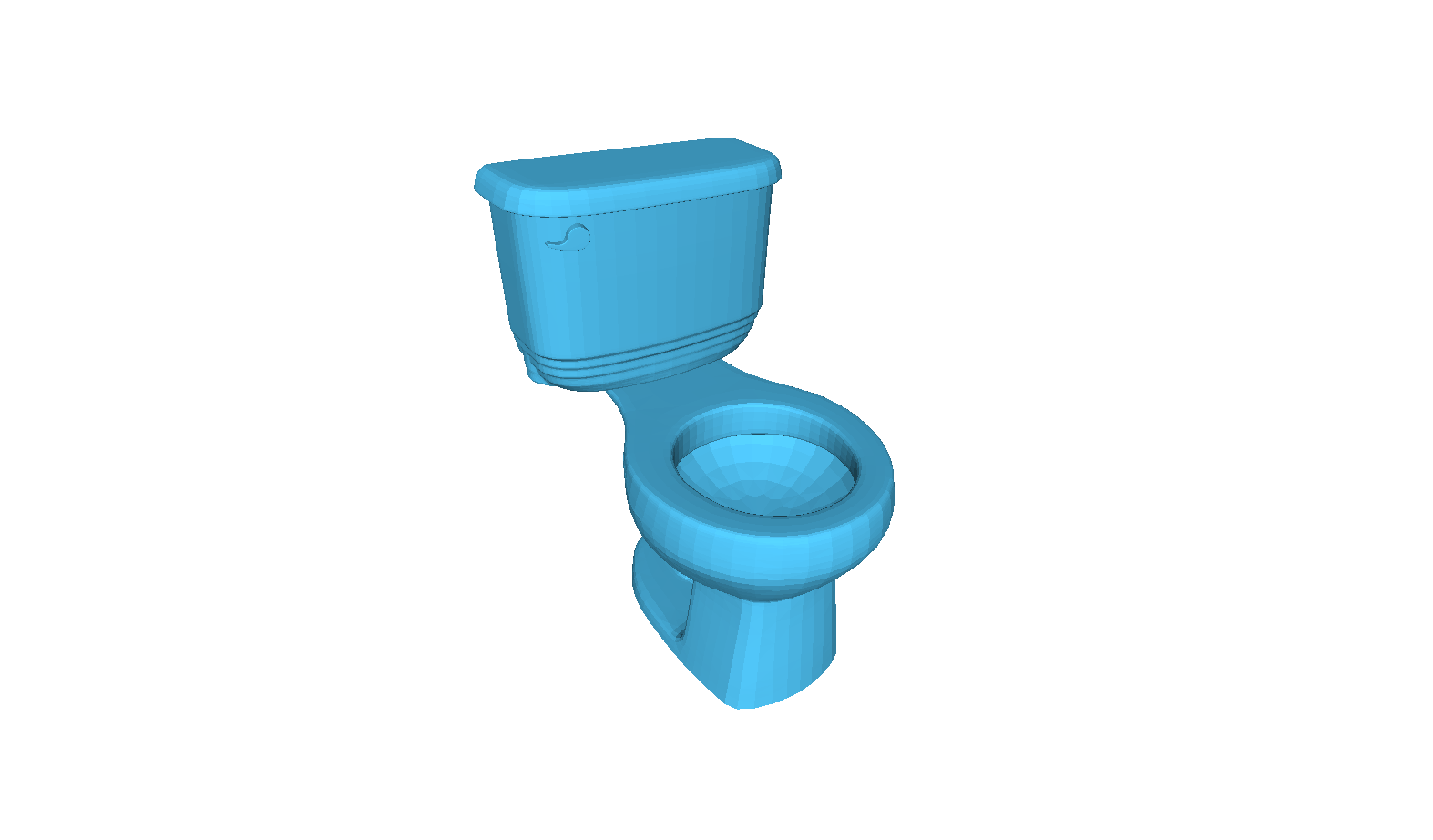} &
  \includegraphics[trim={13cm 2.5cm 15cm 3.cm},clip,width=\widthtopfv\linewidth]{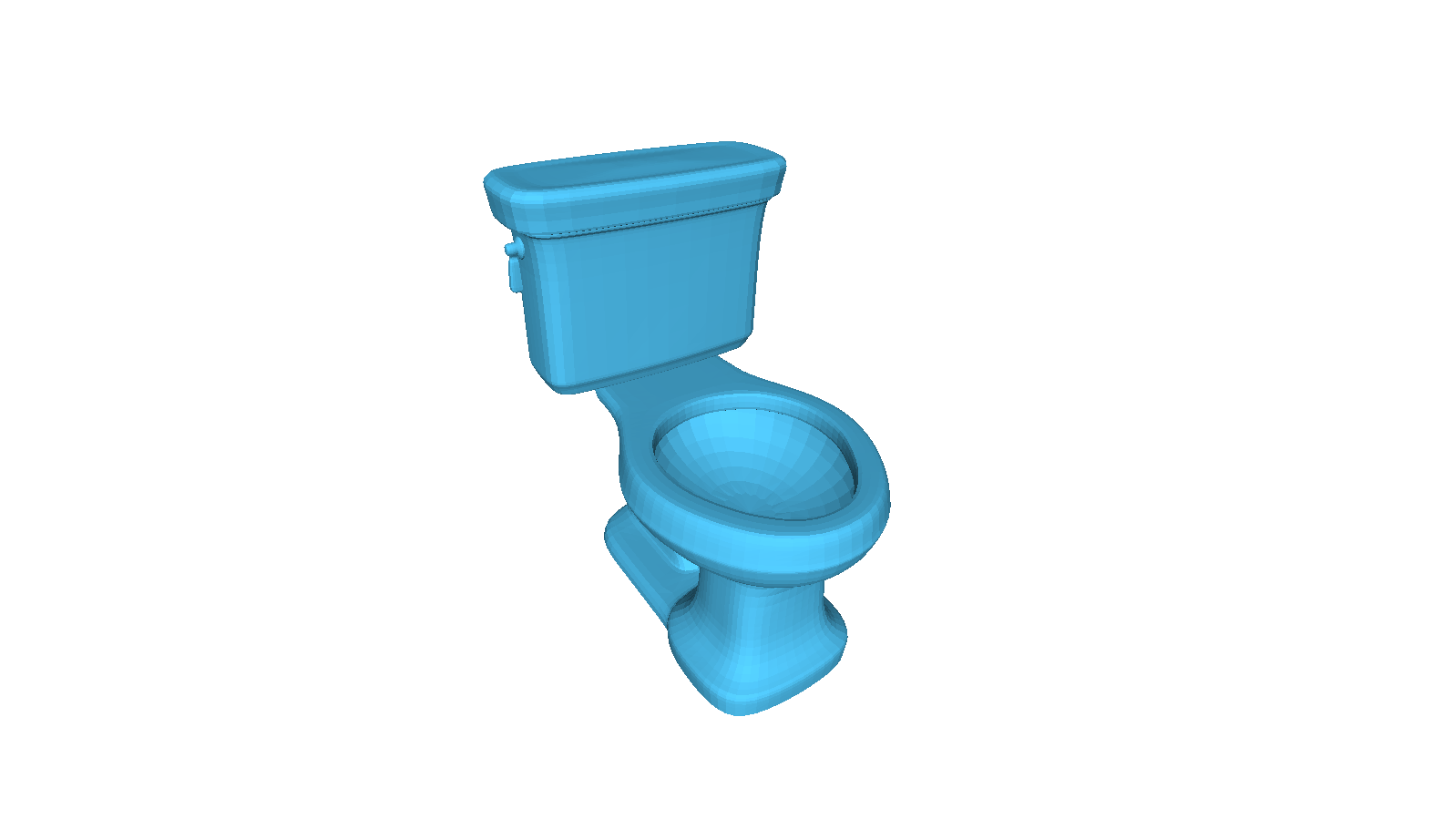} &
  \includegraphics[trim={13cm 2.5cm 15cm 3.cm},clip,width=\widthtopfv\linewidth]{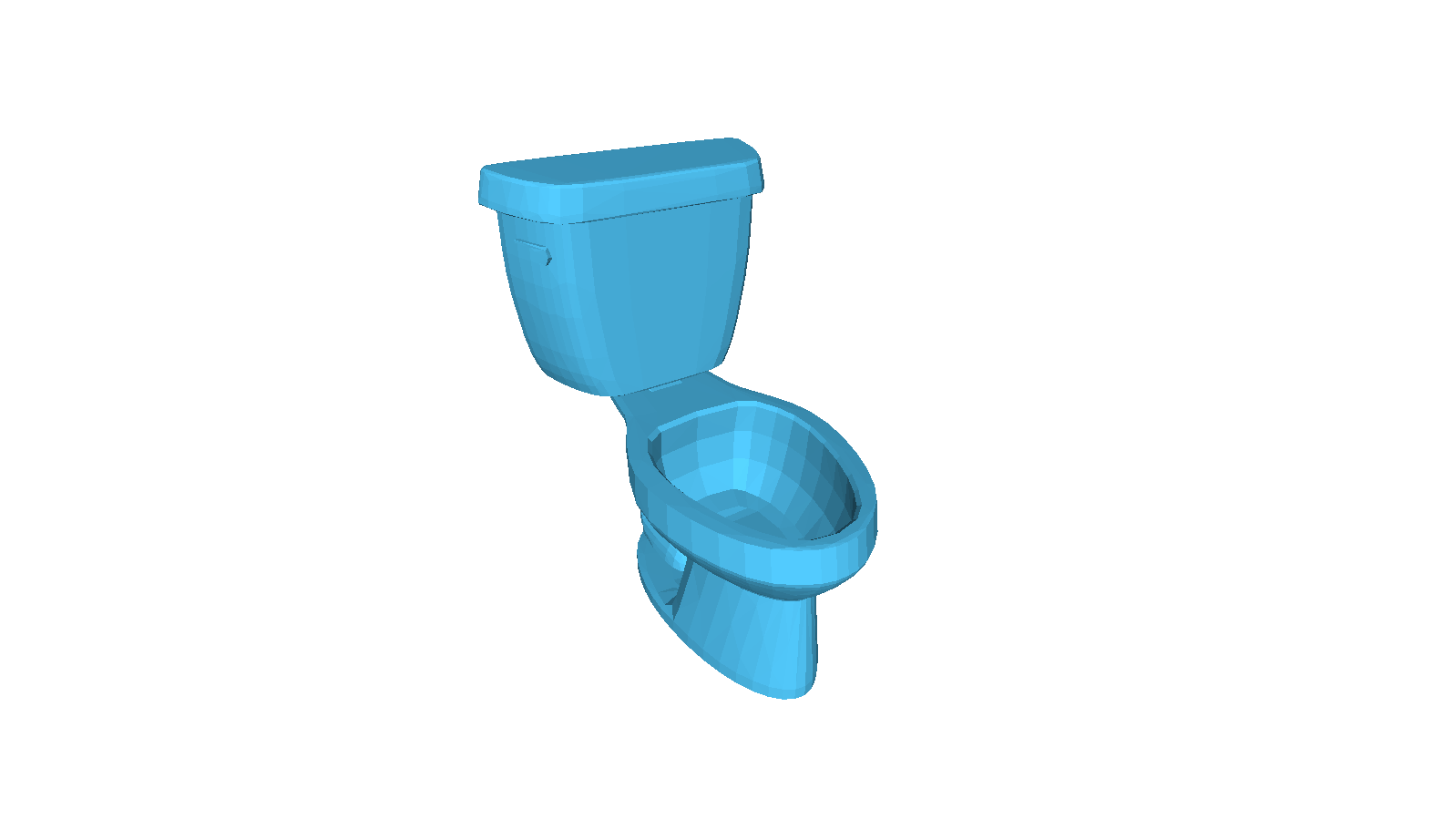} &
  \includegraphics[trim={13cm 2.5cm 15cm 3.cm},clip,width=\widthtopfv\linewidth]{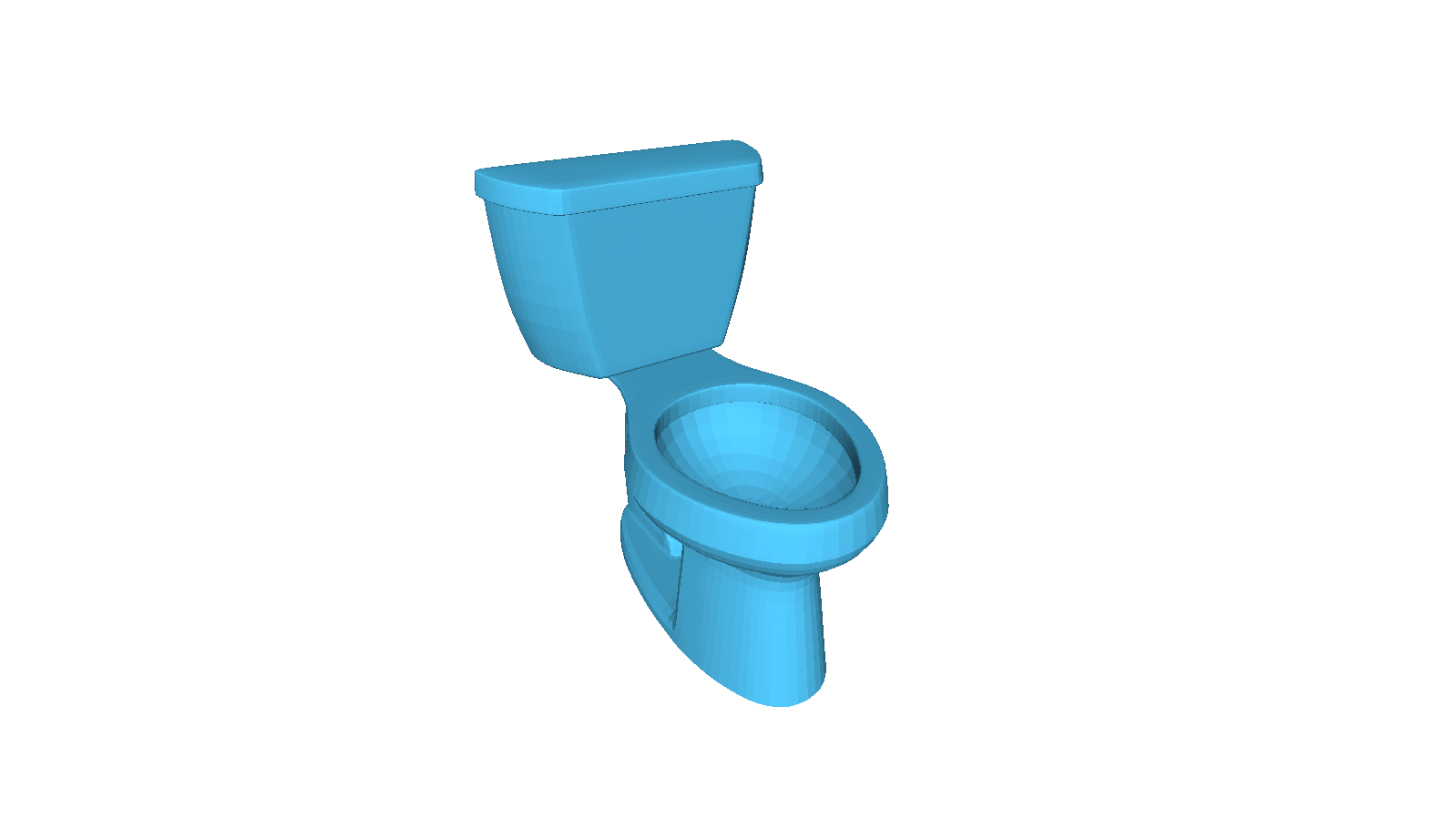} &
  \includegraphics[trim={13cm 2.5cm 15cm 3.cm},clip,width=\widthtopfv\linewidth]{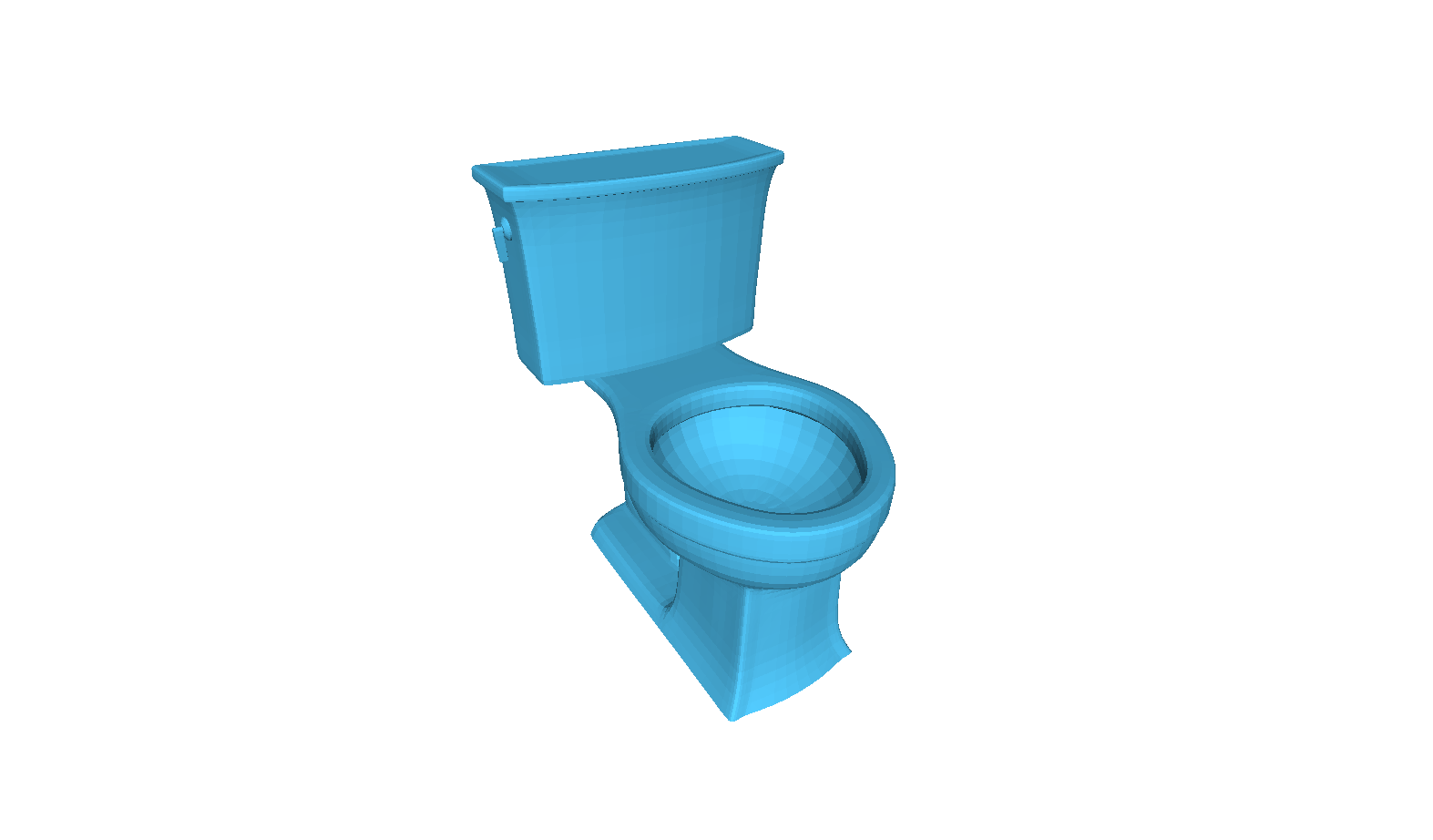} \\

        \includegraphics[trim={15cm 2.5cm 15cm 5.cm},clip,width=\widthtopfv\linewidth]{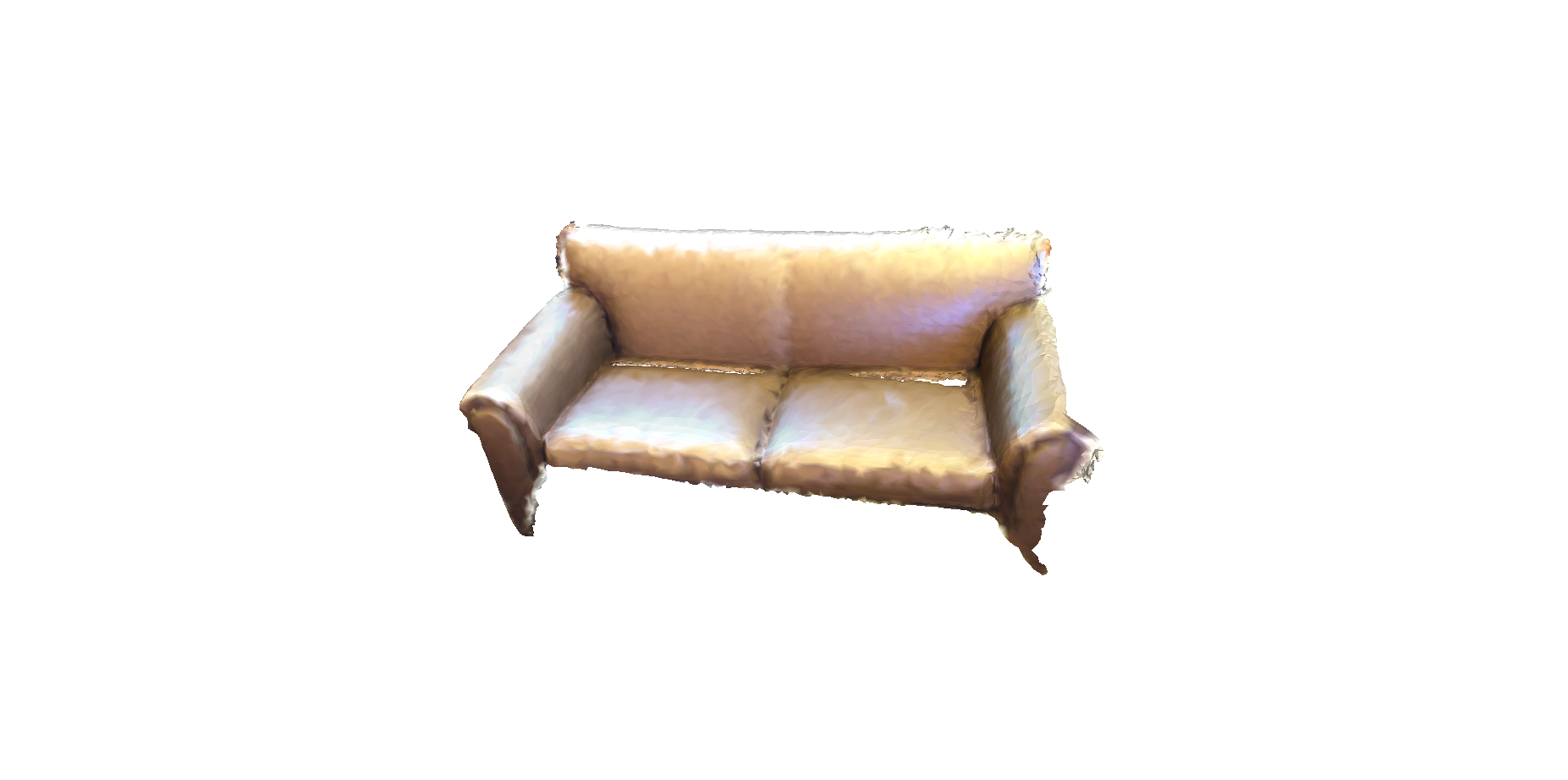} &
  \includegraphics[trim={15cm 2.5cm 15cm 5.cm},clip,width=\widthtopfv\linewidth]{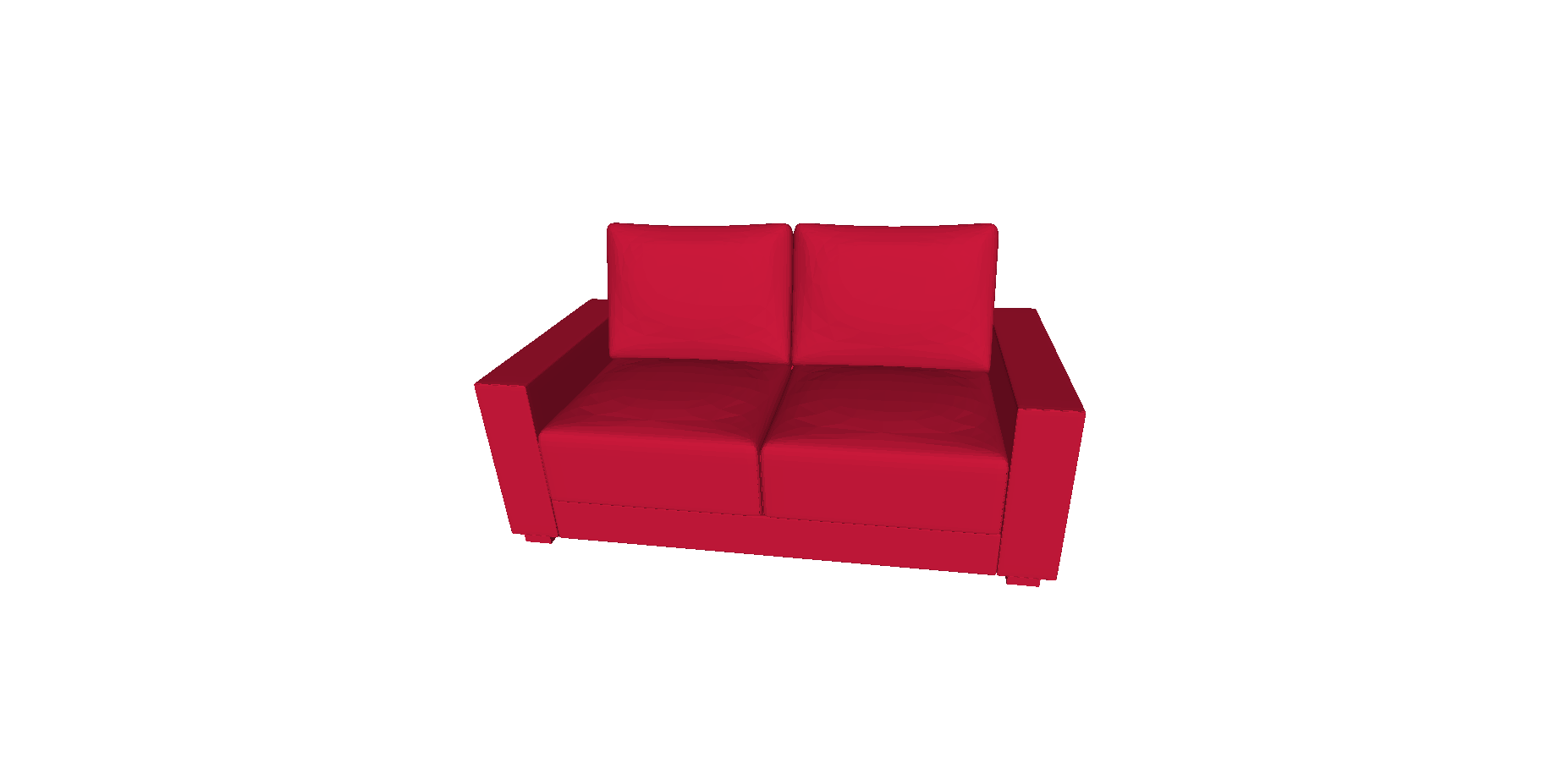} &  
  \includegraphics[trim={15cm 2.5cm 15cm 5.cm},clip,width=\widthtopfv\linewidth]{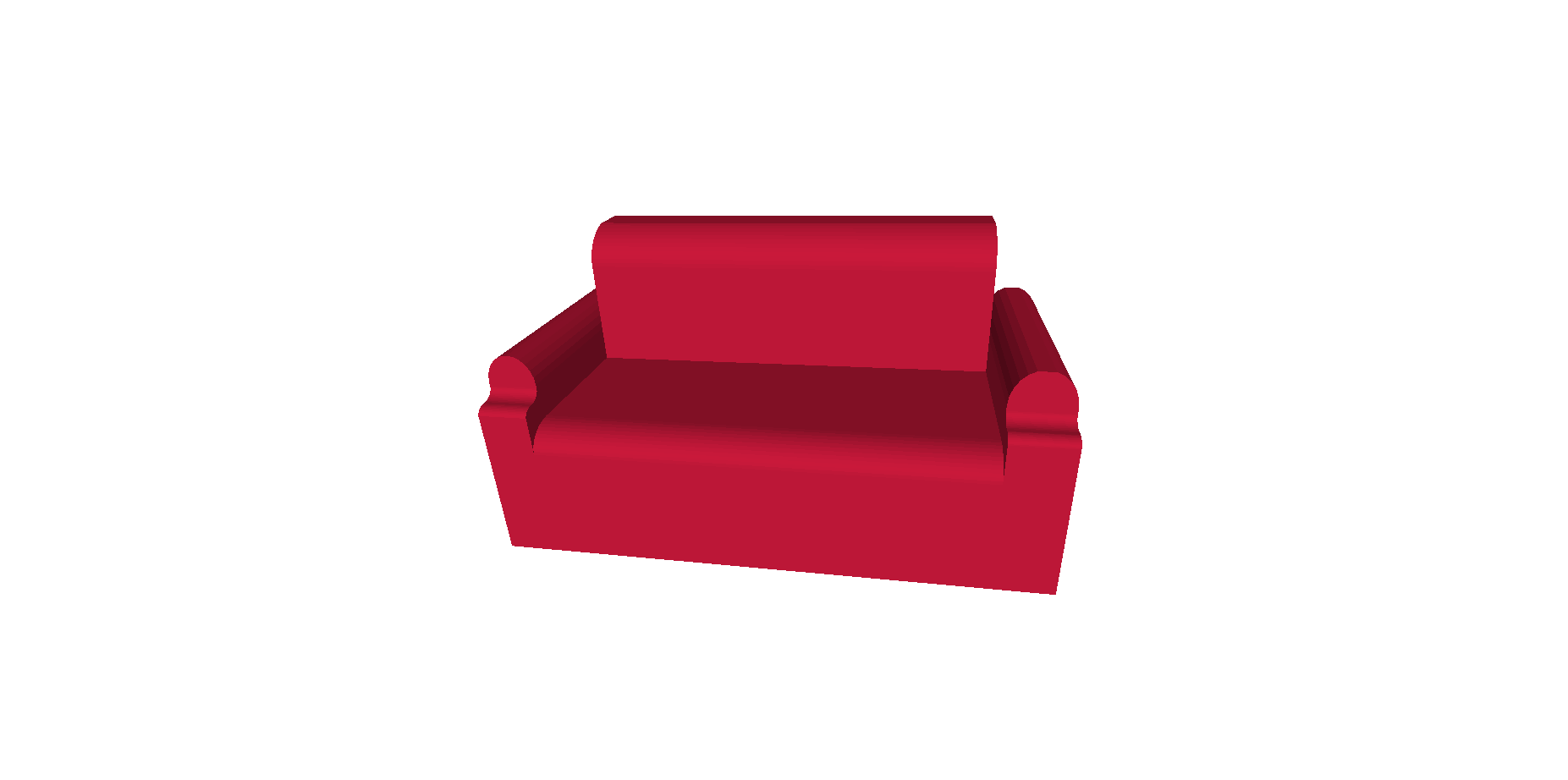} &
  \includegraphics[trim={15cm 2.5cm 15cm 5.cm},clip,width=\widthtopfv\linewidth]{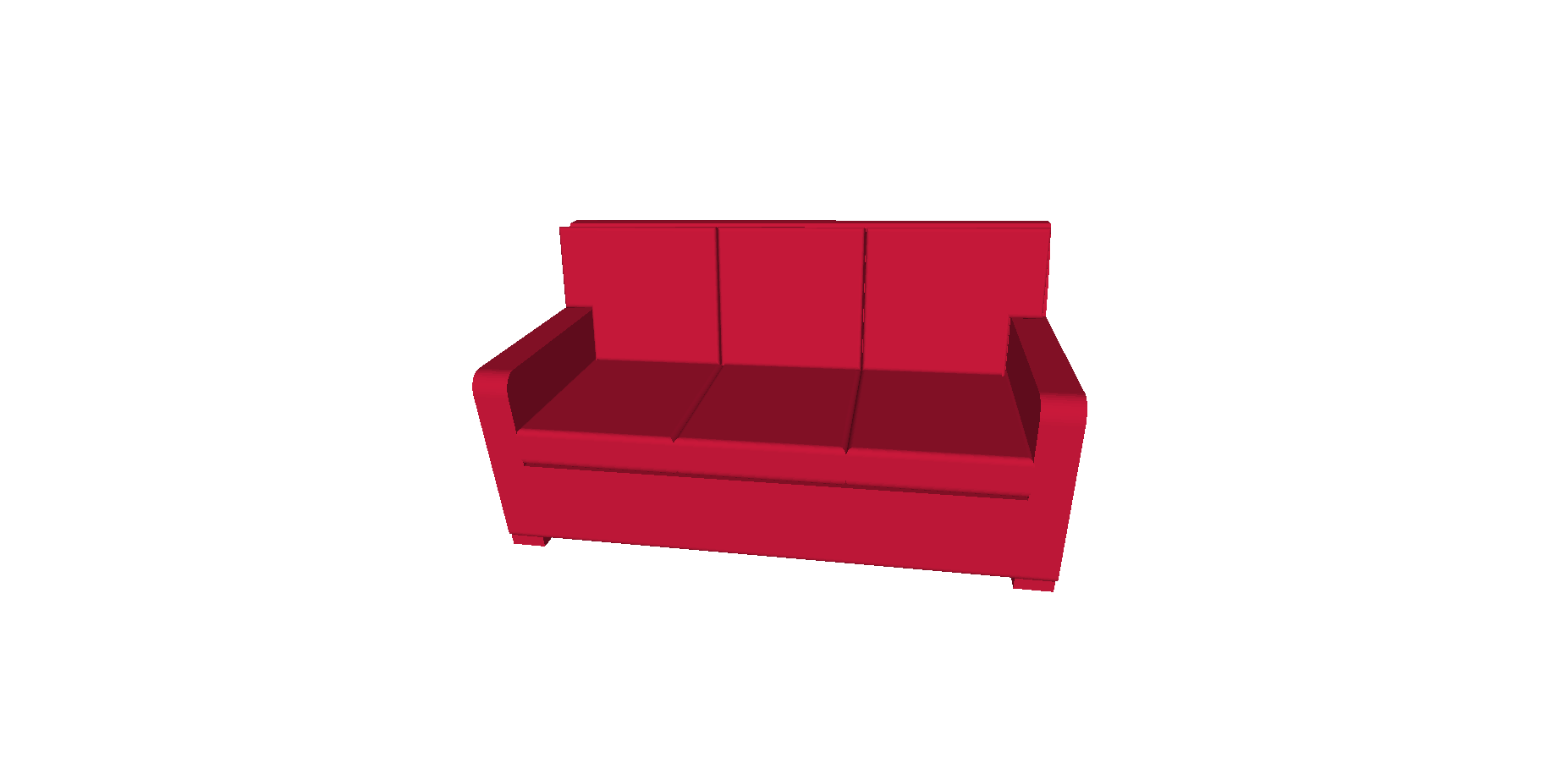} &
  \includegraphics[trim={15cm 2.5cm 15cm 5.cm},clip,width=\widthtopfv\linewidth]{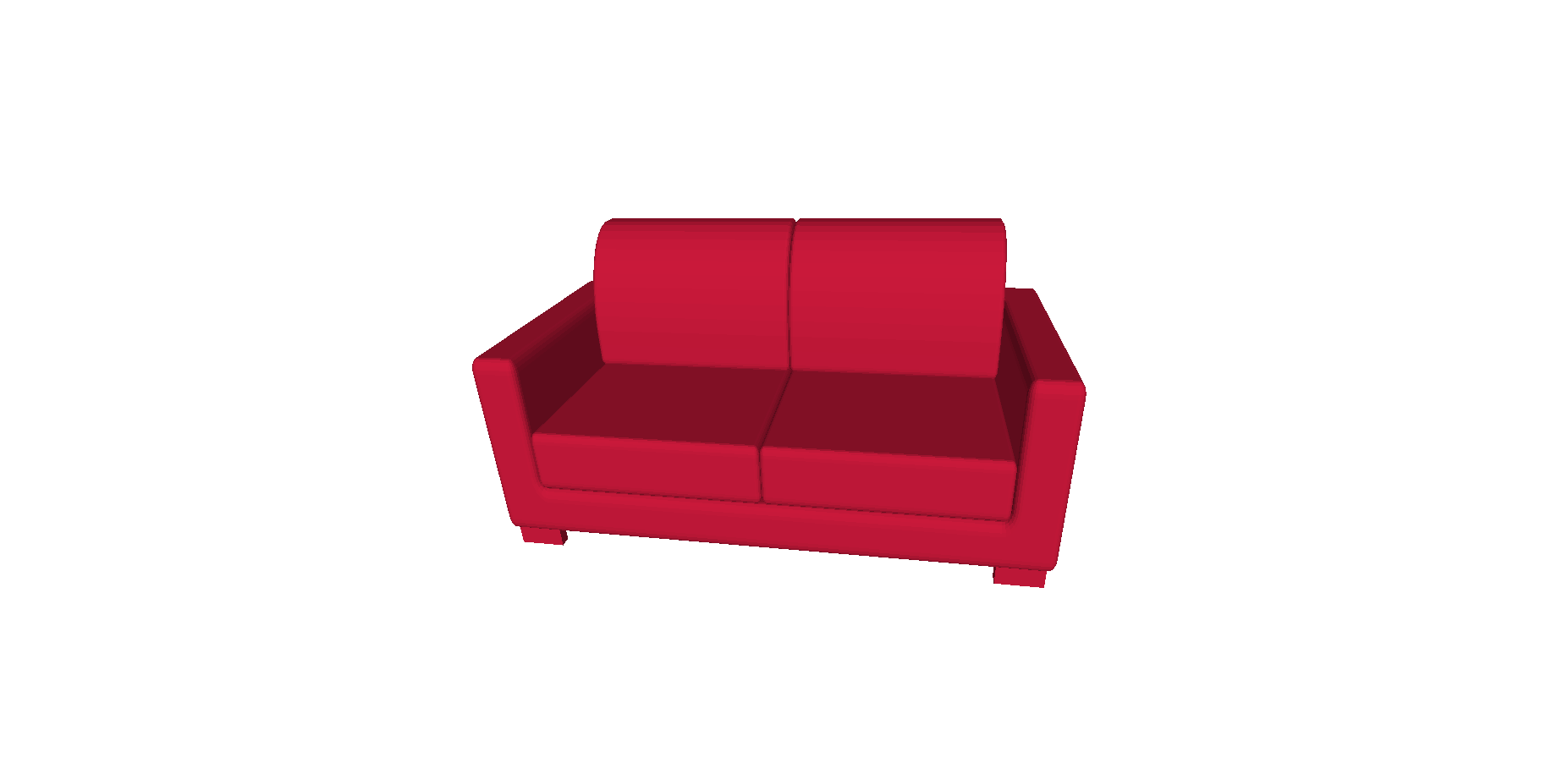} &
  \includegraphics[trim={15cm 2.5cm 15cm 5.cm},clip,width=\widthtopfv\linewidth]{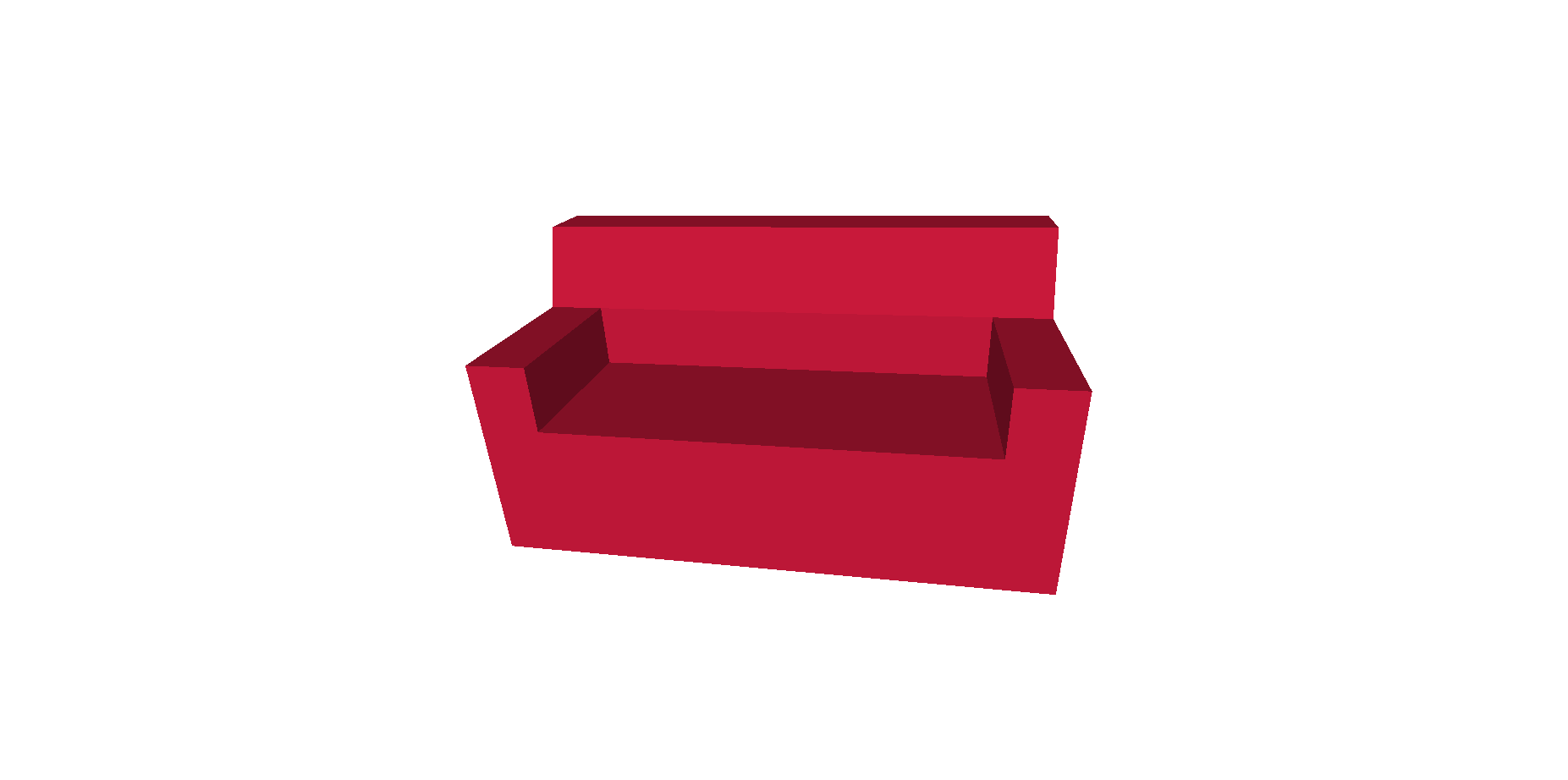} &
  \includegraphics[trim={15cm 2.5cm 15cm 5.cm},clip,width=\widthtopfv\linewidth]{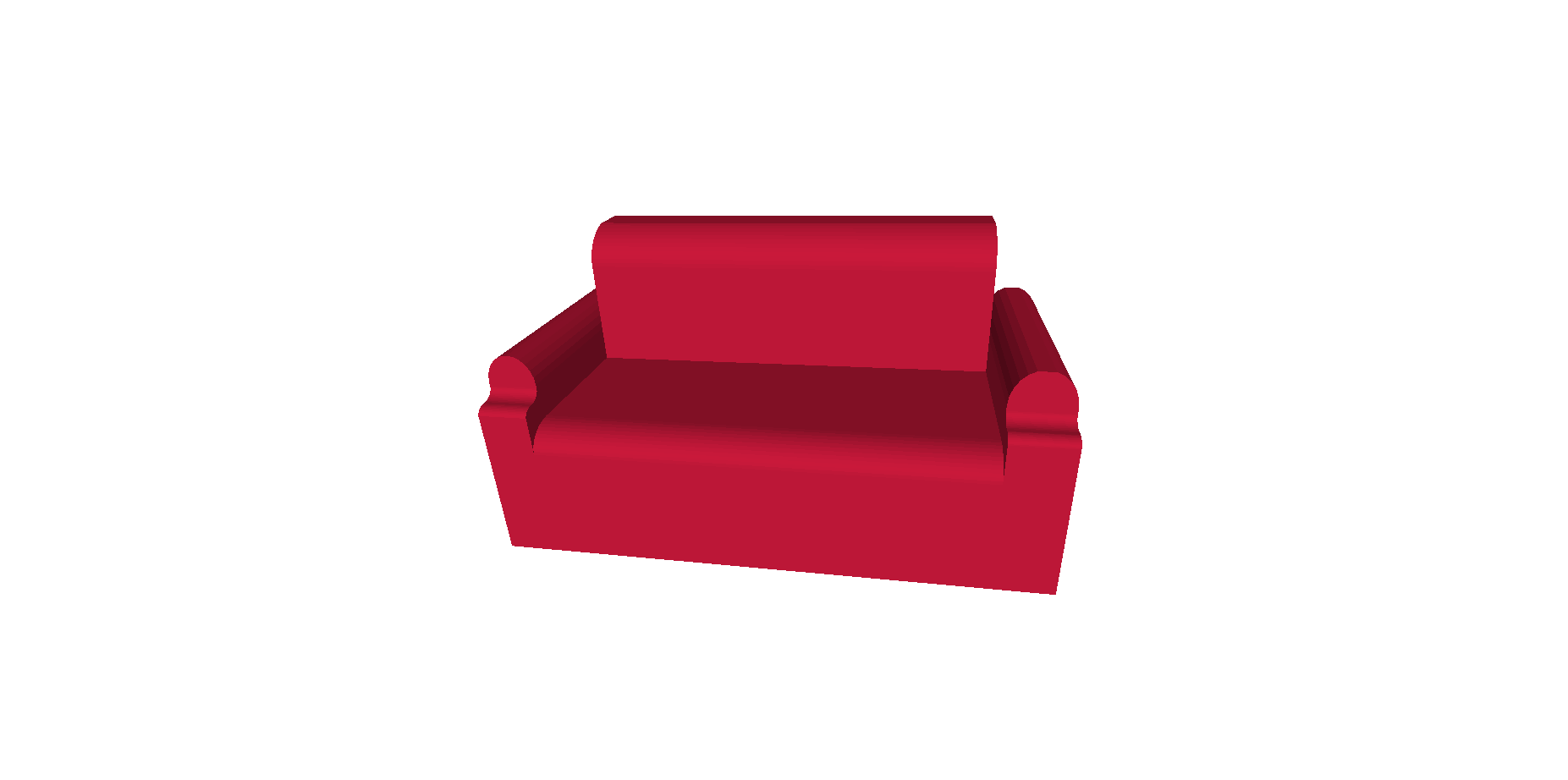} &
  \includegraphics[trim={15cm 2.5cm 15cm 5.cm},clip,width=\widthtopfv\linewidth]{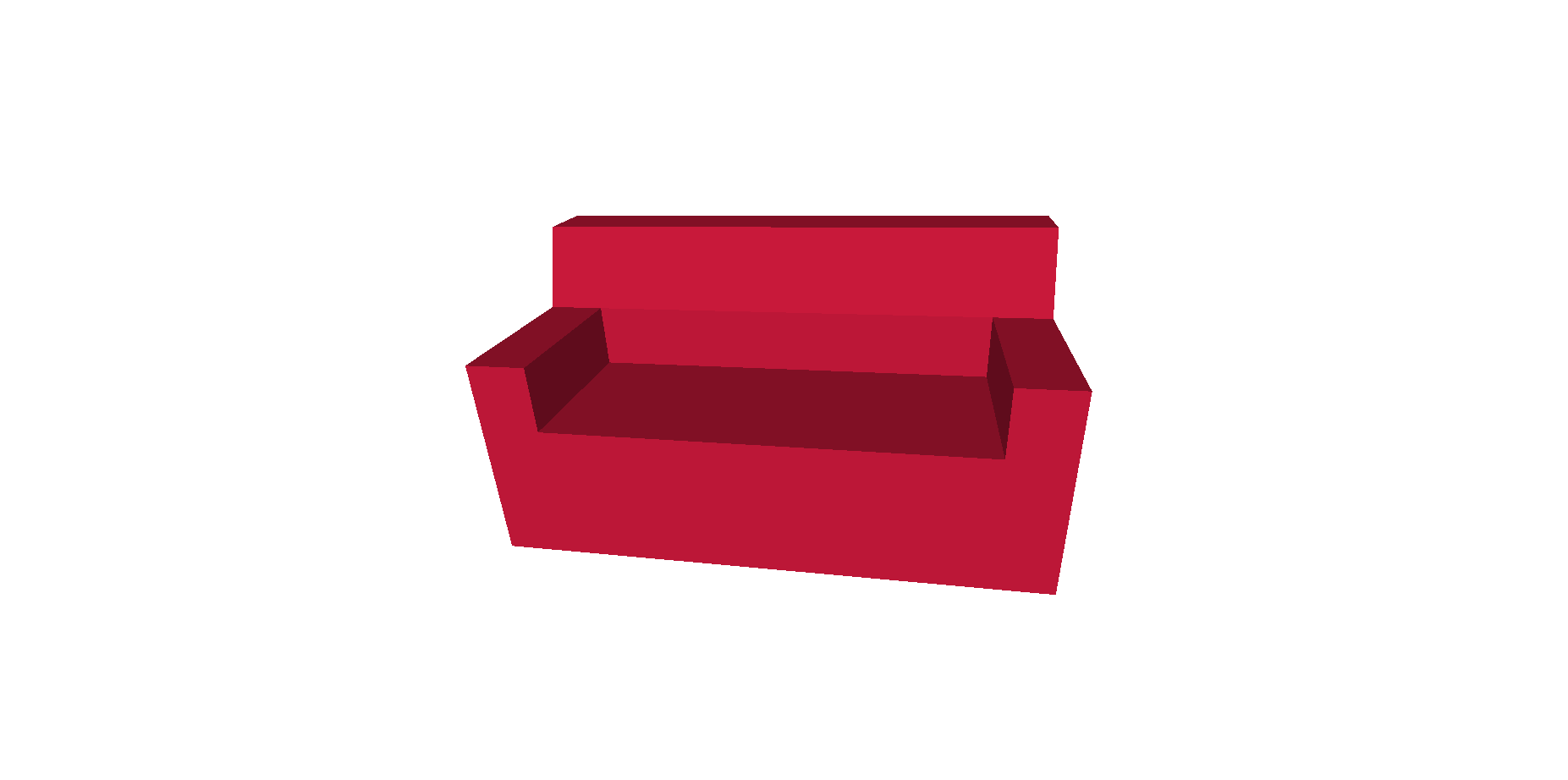} &
  \includegraphics[trim={15cm 2.5cm 15cm 5.cm},clip,width=\widthtopfv\linewidth]{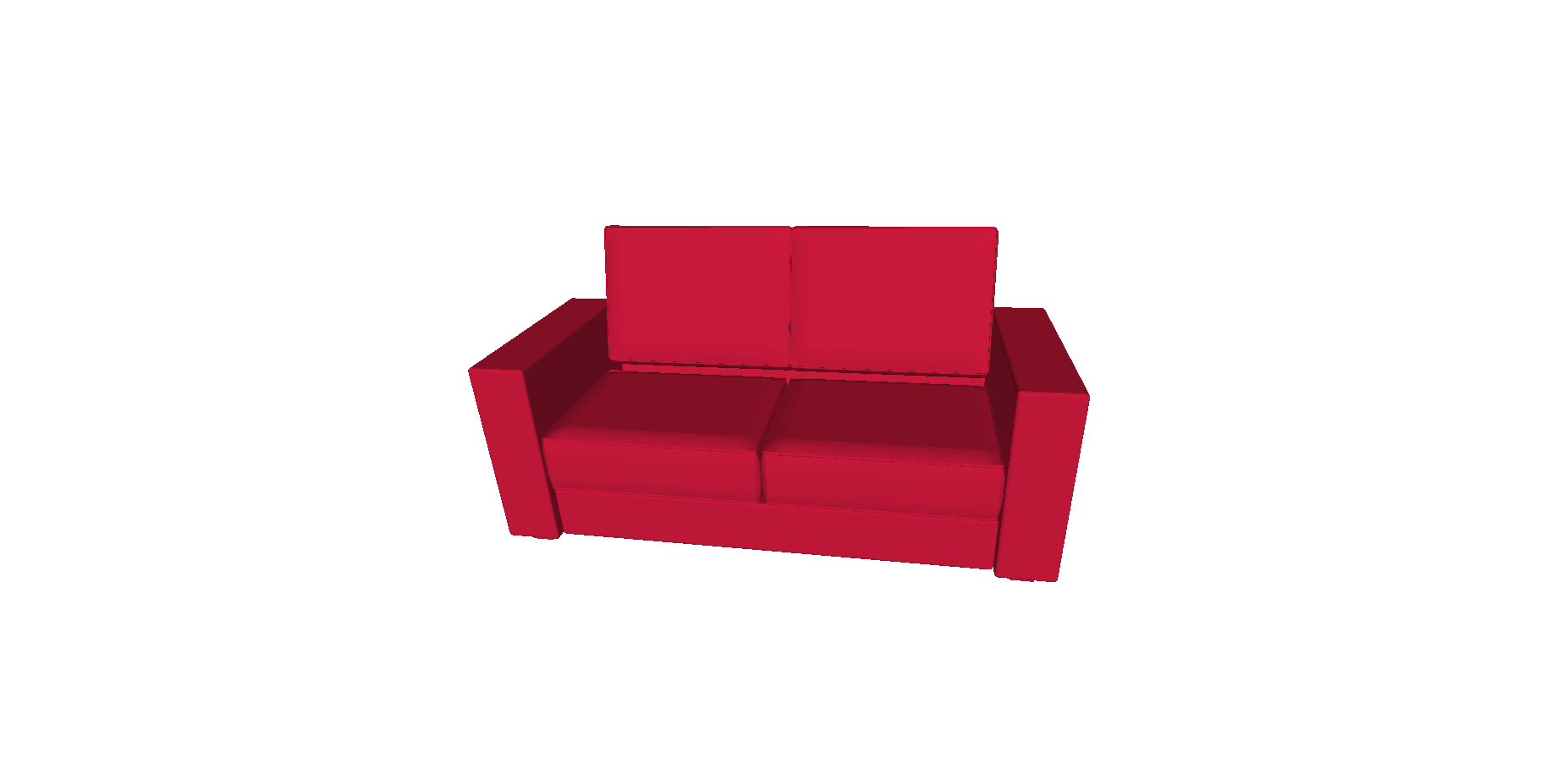} &
  \includegraphics[trim={15cm 2.5cm 15cm 5.cm},clip,width=\widthtopfv\linewidth]{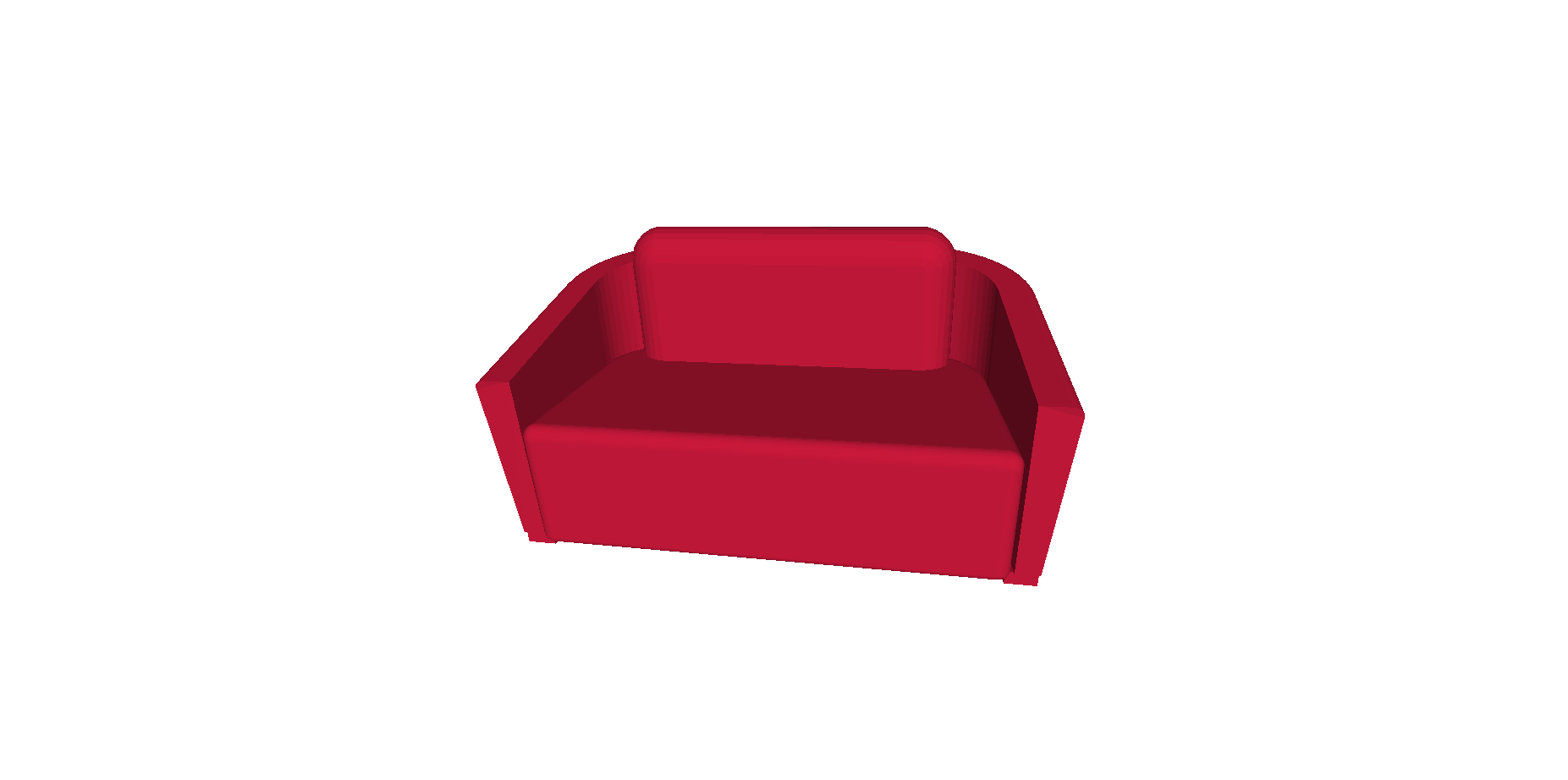} &
  \includegraphics[trim={15cm 2.5cm 15cm 5.cm},clip,width=\widthtopfv\linewidth]{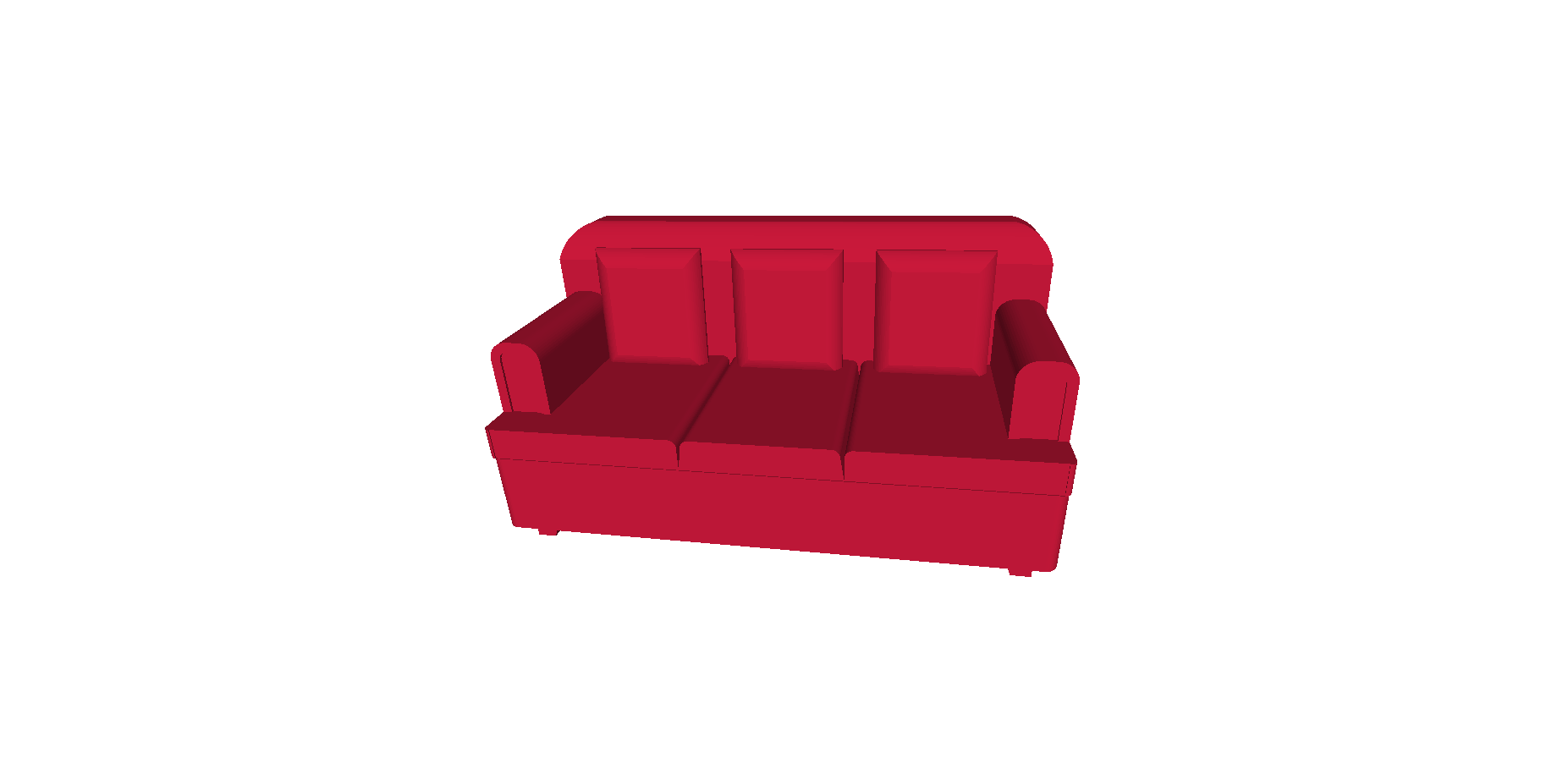} \\

          \includegraphics[trim={14cm 2.5cm 15cm 5.cm},clip,width=\widthtopfv\linewidth]{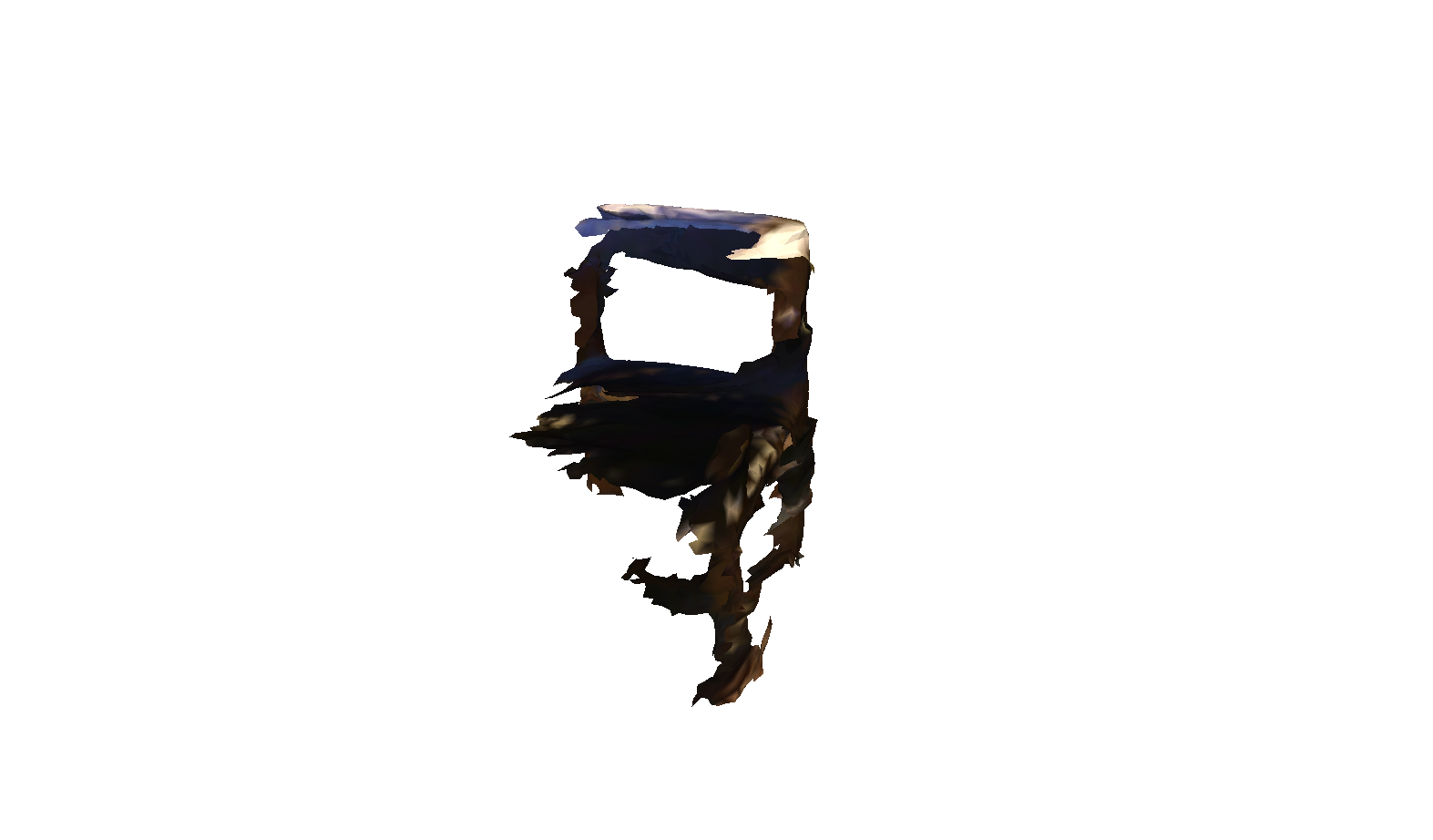} &
  \includegraphics[trim={14cm 2.5cm 15cm 5.cm},clip,width=\widthtopfv\linewidth]{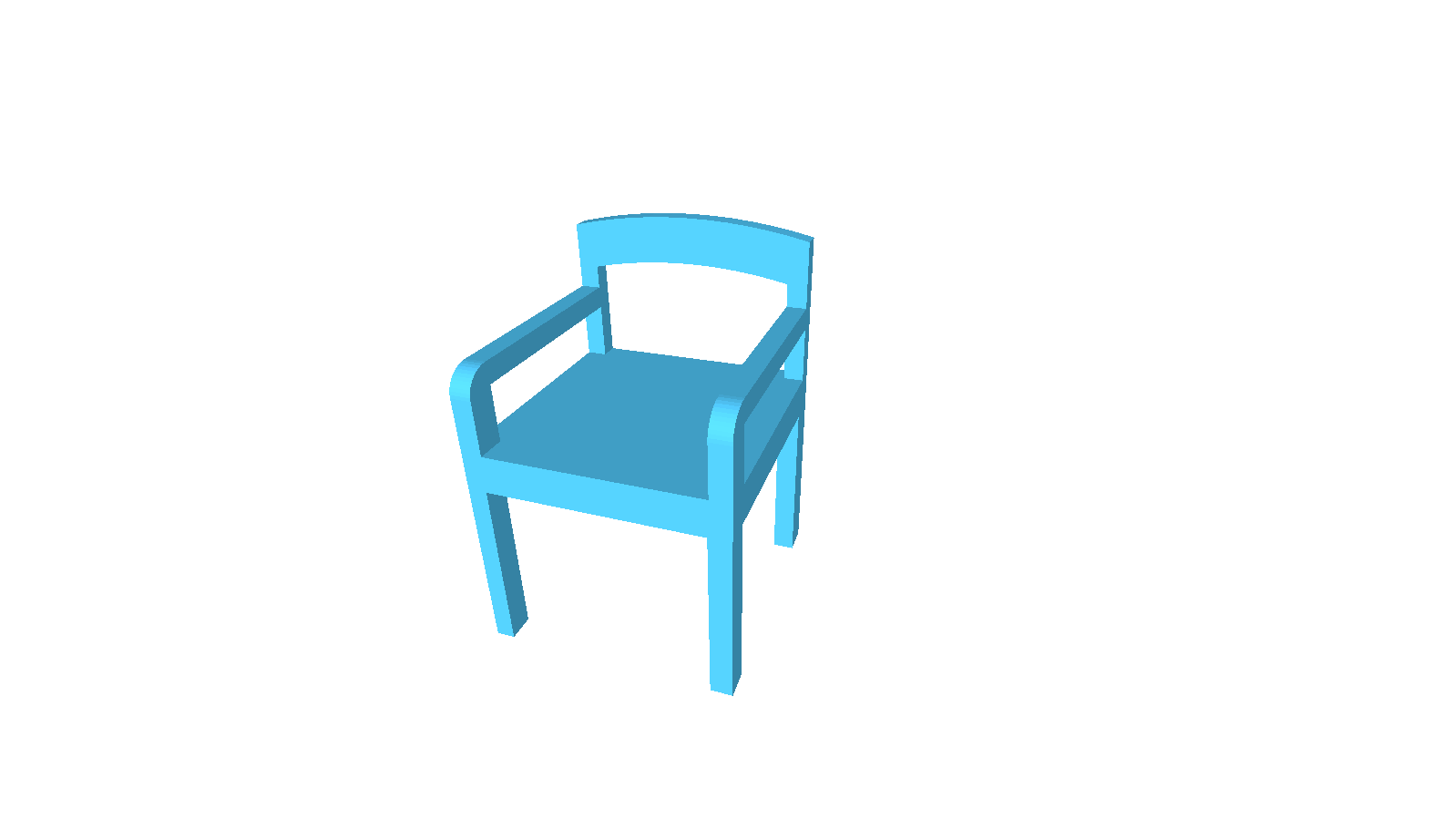} &  
  \includegraphics[trim={14cm 2.5cm 15cm 5.cm},clip,width=\widthtopfv\linewidth]{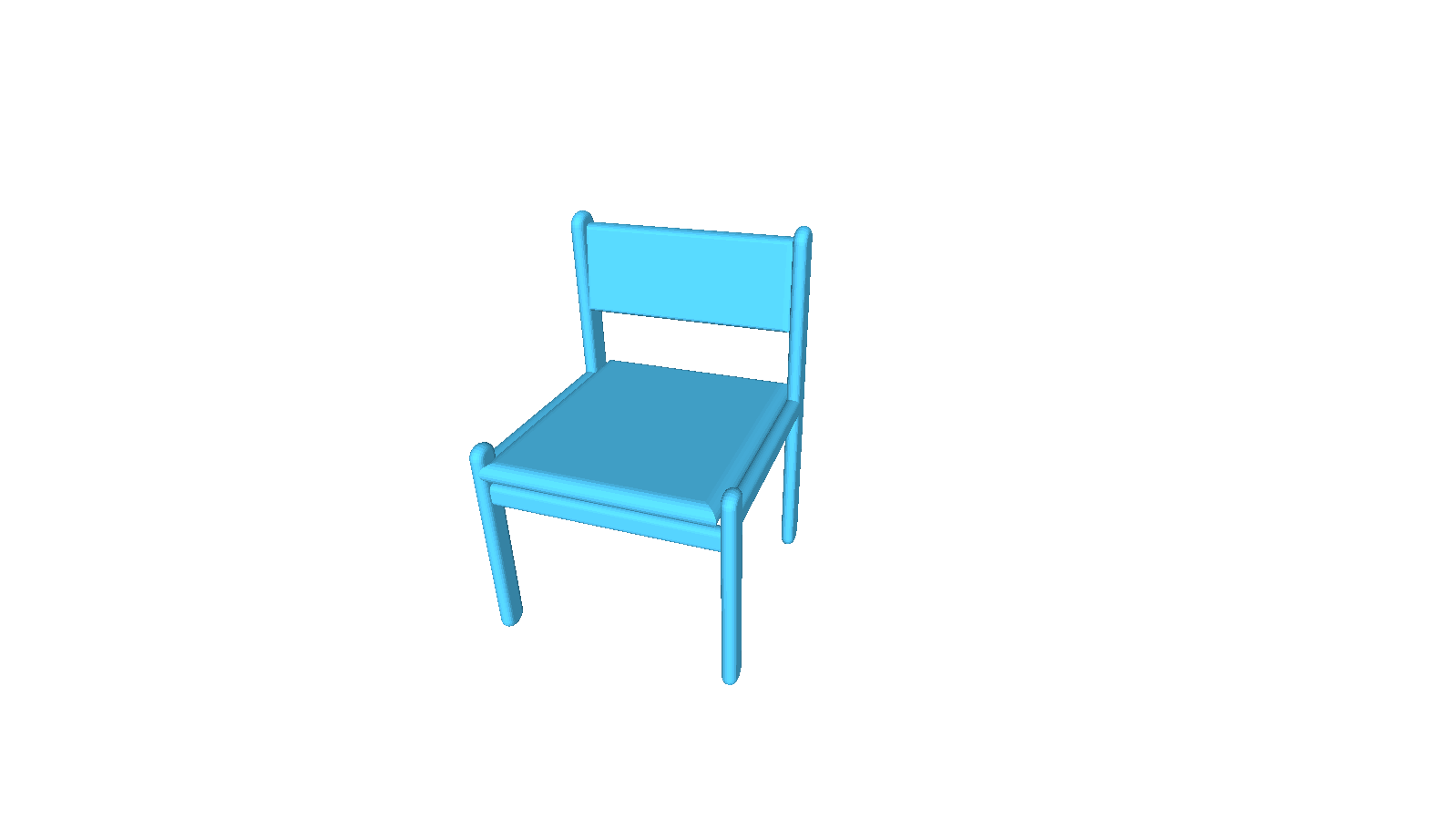} &
  \includegraphics[trim={14cm 2.5cm 15cm 5.cm},clip,width=\widthtopfv\linewidth]{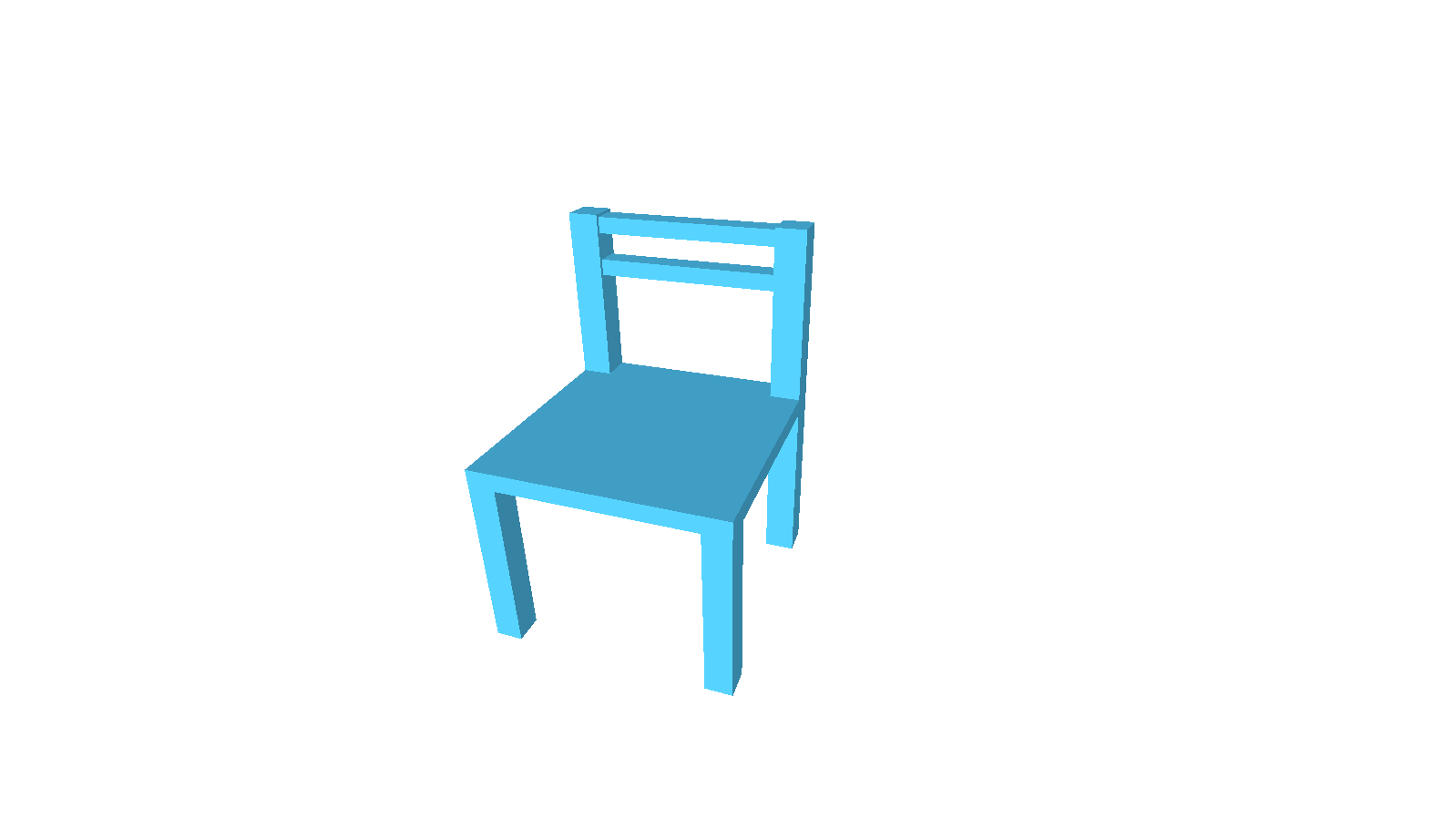} &
  \includegraphics[trim={14cm 2.5cm 15cm 5.cm},clip,width=\widthtopfv\linewidth]{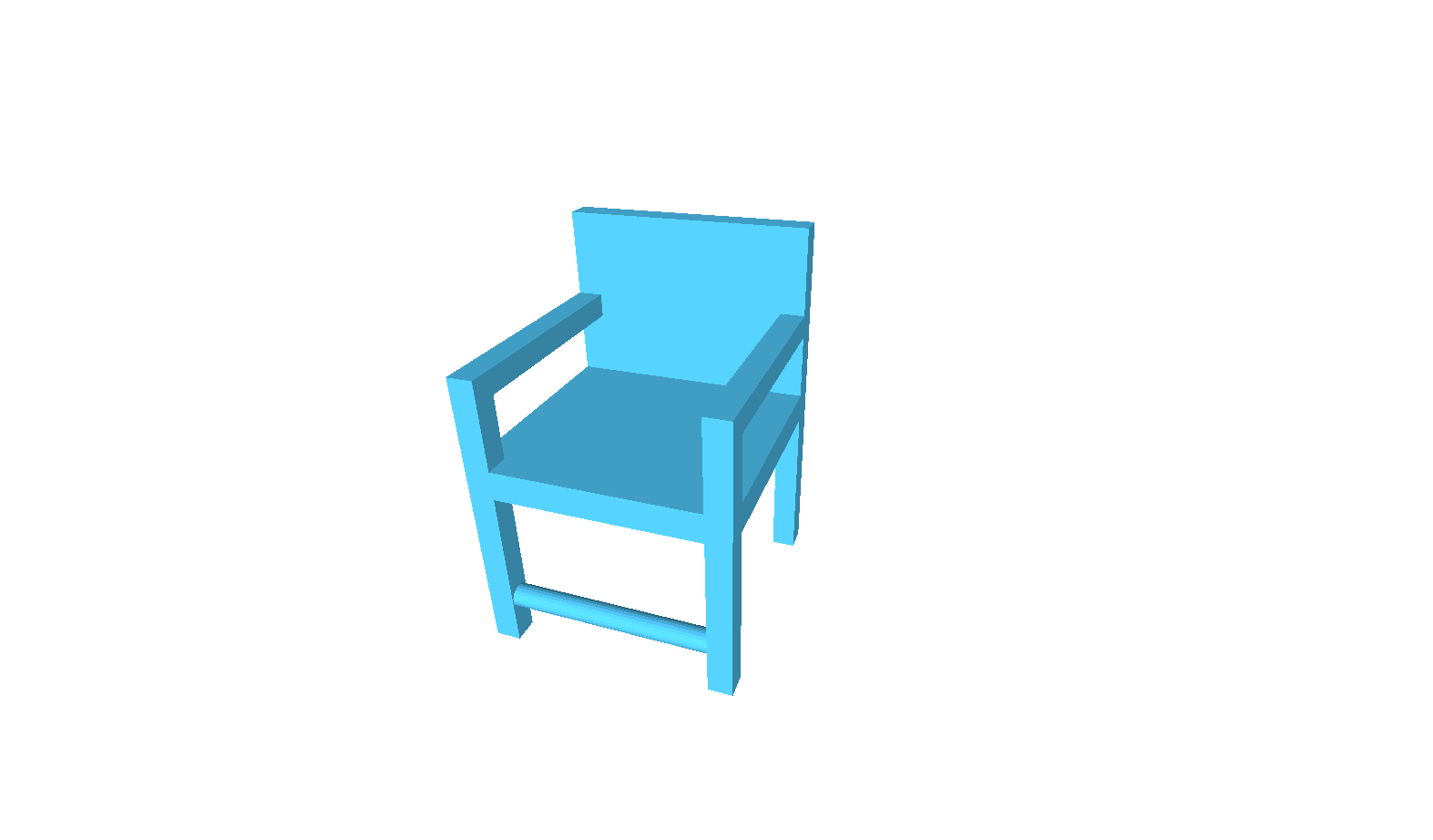} &
  \includegraphics[trim={14cm 2.5cm 15cm 5.cm},clip,width=\widthtopfv\linewidth]{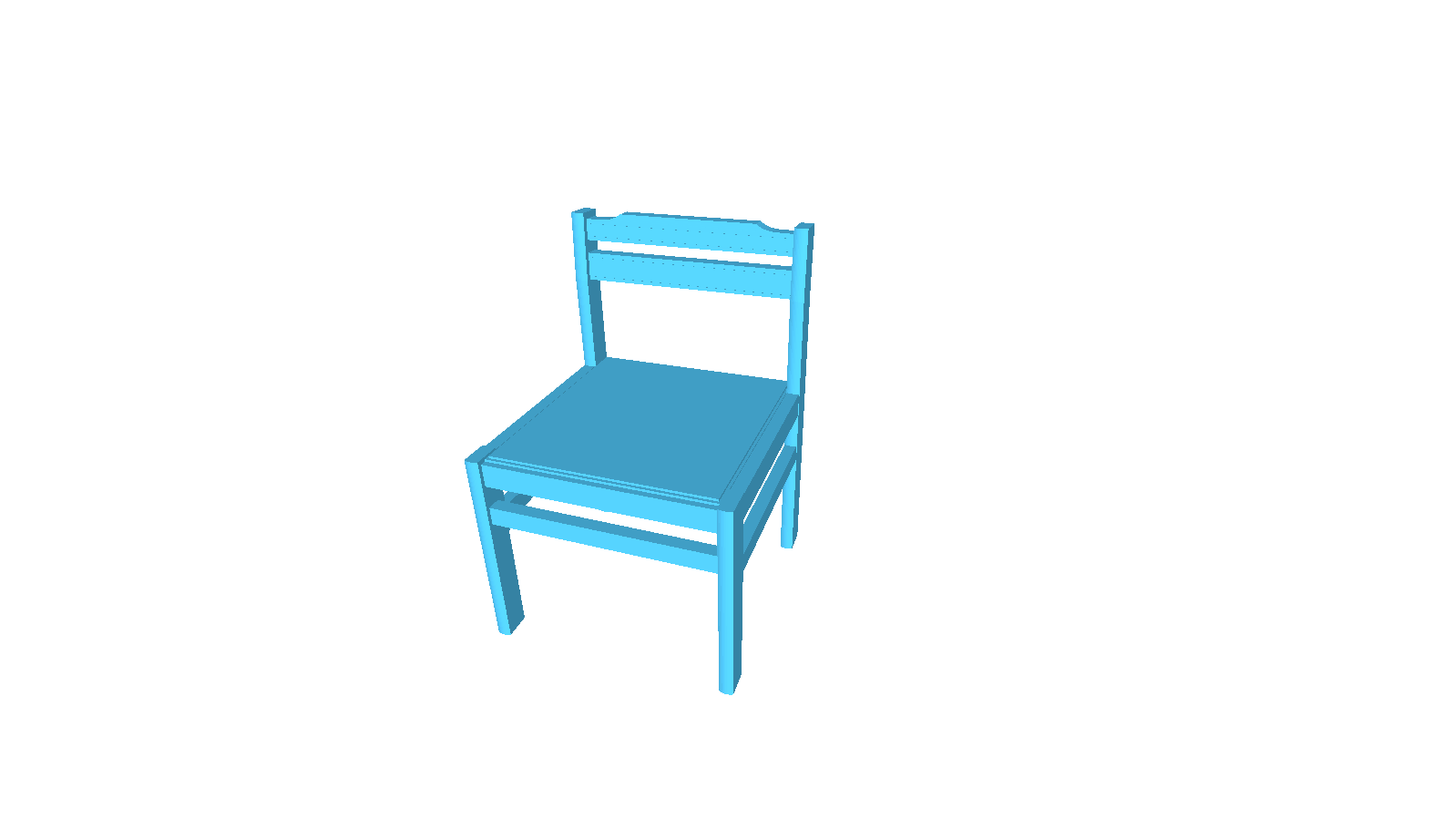} &
  \includegraphics[trim={14cm 2.5cm 15cm 5.cm},clip,width=\widthtopfv\linewidth]{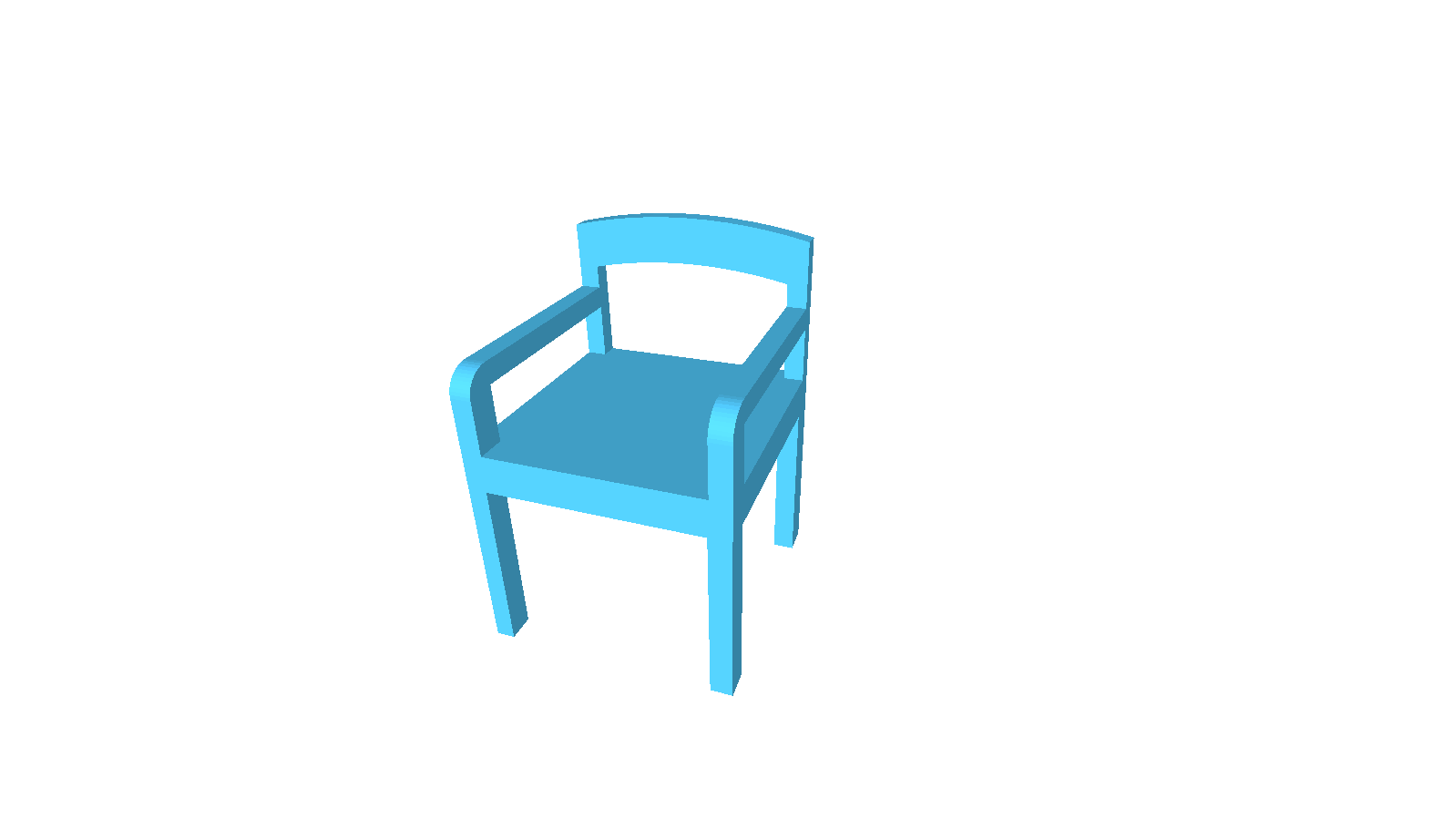} &
  \includegraphics[trim={14cm 2.5cm 15cm 5.cm},clip,width=\widthtopfv\linewidth]{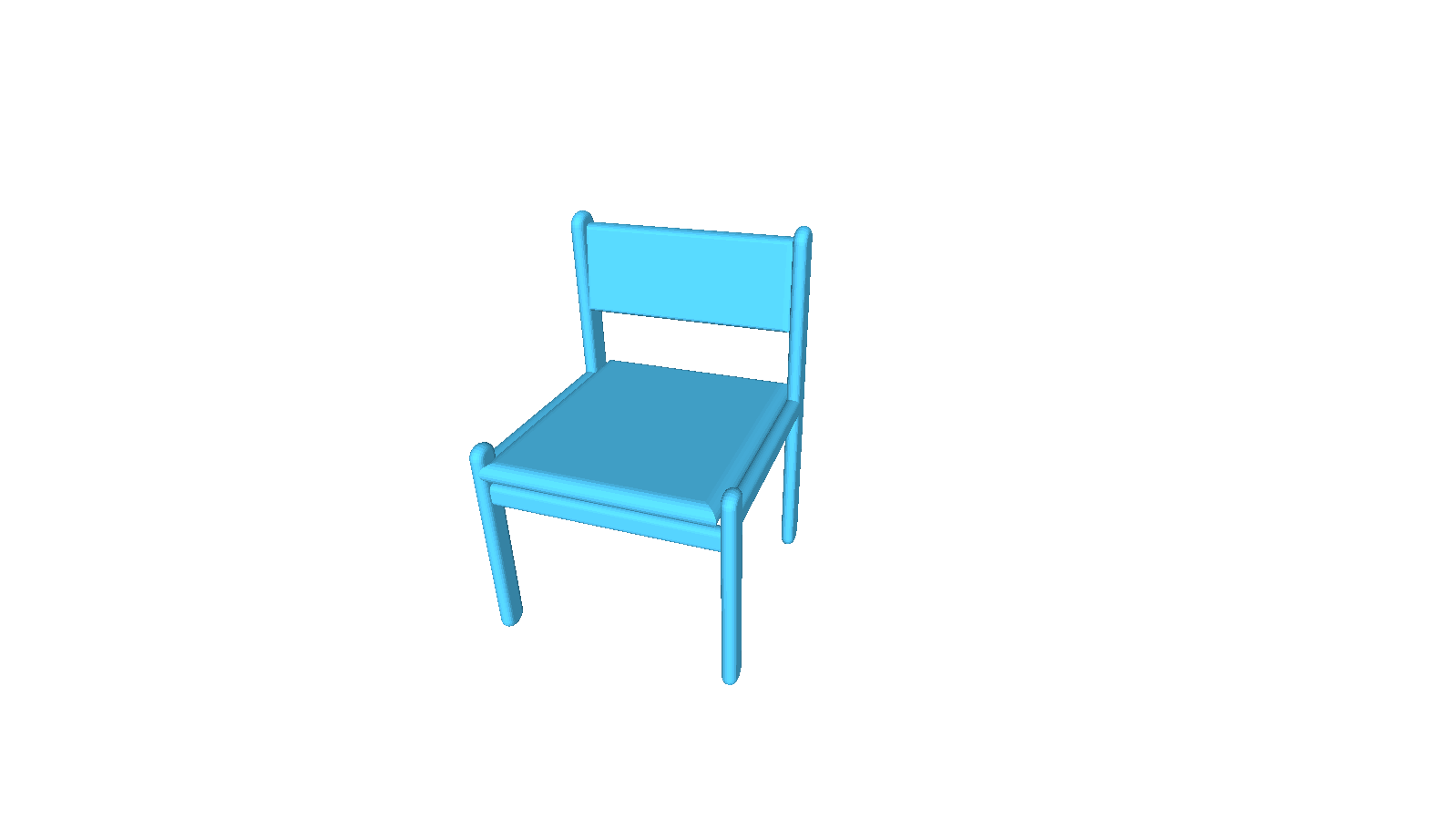} &
  \includegraphics[trim={14cm 2.5cm 15cm 5.cm},clip,width=\widthtopfv\linewidth]{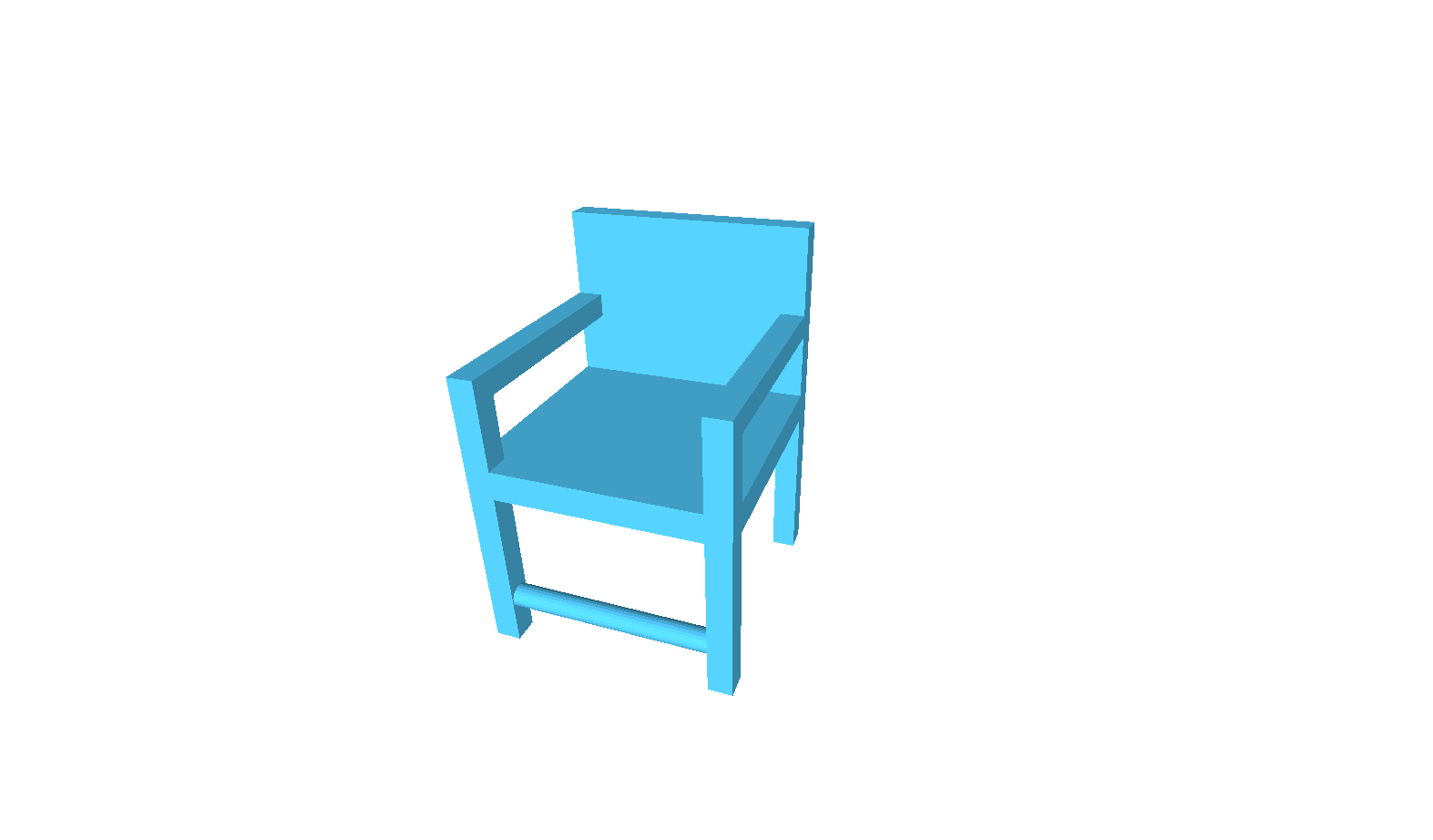} &
  \includegraphics[trim={14cm 2.5cm 15cm 5.cm},clip,width=\widthtopfv\linewidth]{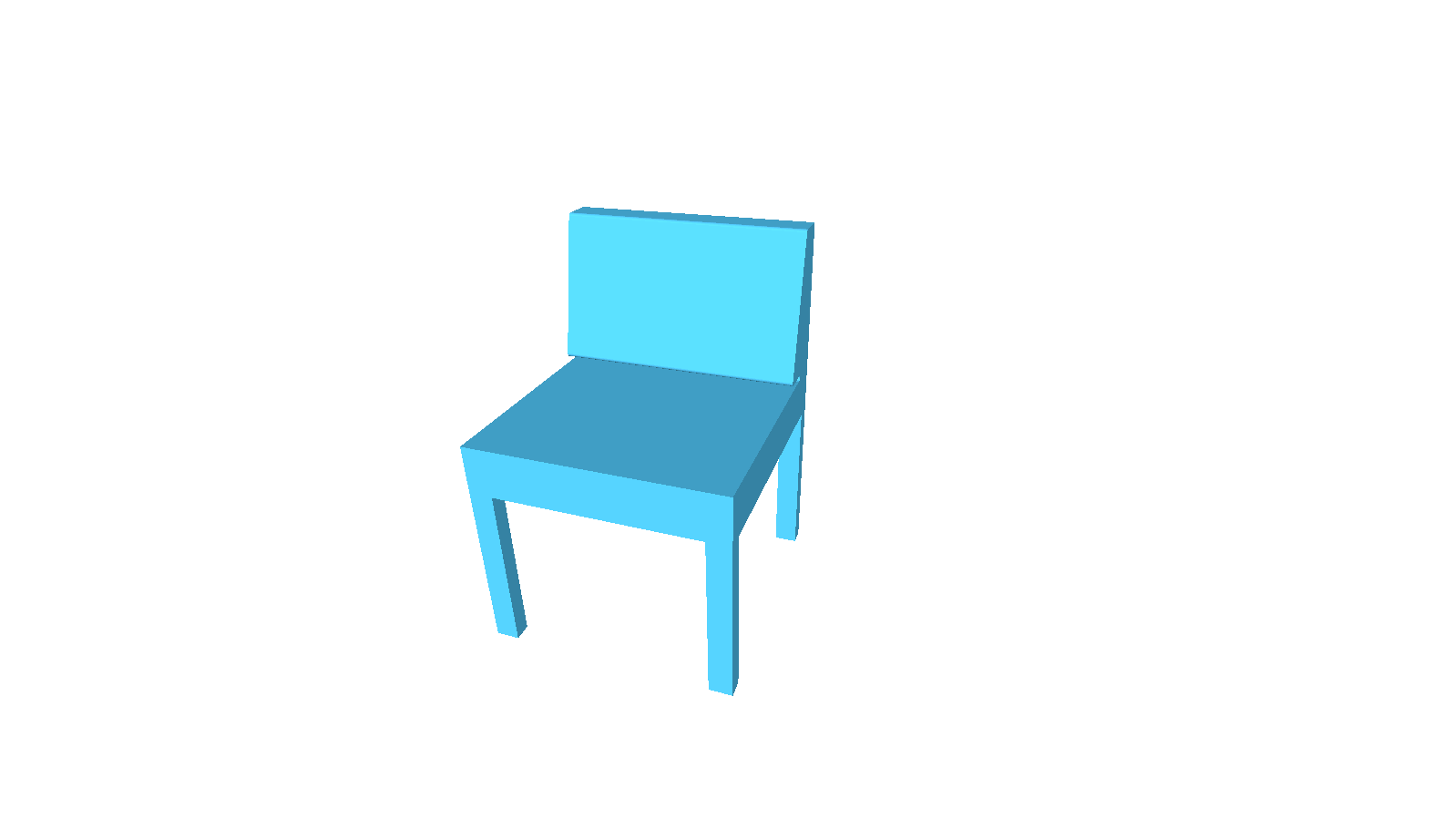} &
  \includegraphics[trim={14cm 2.5cm 15cm 5.cm},clip,width=\widthtopfv\linewidth]{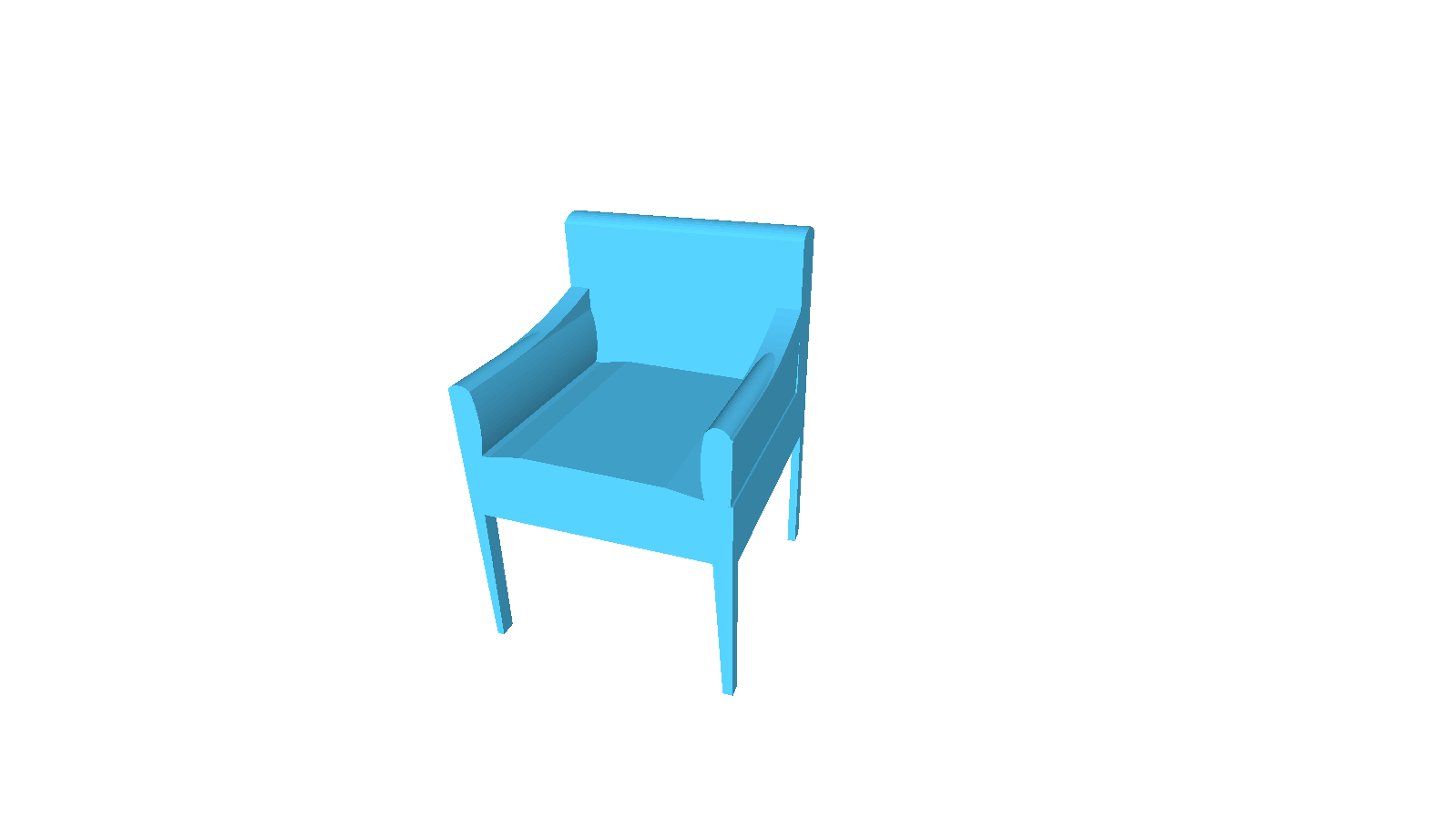} \\

  Target object & 
  \multicolumn{5}{c|}{Top-5 candidates, exhaustive search} &
  \multicolumn{5}{c}{Top-5 candidates, HOC-Search (ours)} \\  
  % object & &&&& & &&&&\\
\end{tabular}
}
% \vspace{-0.2cm}
\caption{Additional visualizations of top-5 candidates (in descending order) from exhaustive search compared to HOC-Search for 800 iterations. Left shows the target object, middle column are the results from the exhaustive search, and right column shows the results using HOC-Search. In row 5, HOC-Search successfully retrieves toilet seats, which is notable, as this is a small sub-category in the Chair object category (there are approx. 30 toilet seats in 6778 Chair models).}
    \label{fig:supp_top5}
\end{figure*}

\begin{figure*}
\centering
\scalebox{0.85}{
\begin{tabular}{c|ccccc|ccccc}
%[trim={left bottom right top}

    \includegraphics[trim={14.5cm 3.5cm 14.5cm 3.5cm},clip,width=\widthtopfv\linewidth]{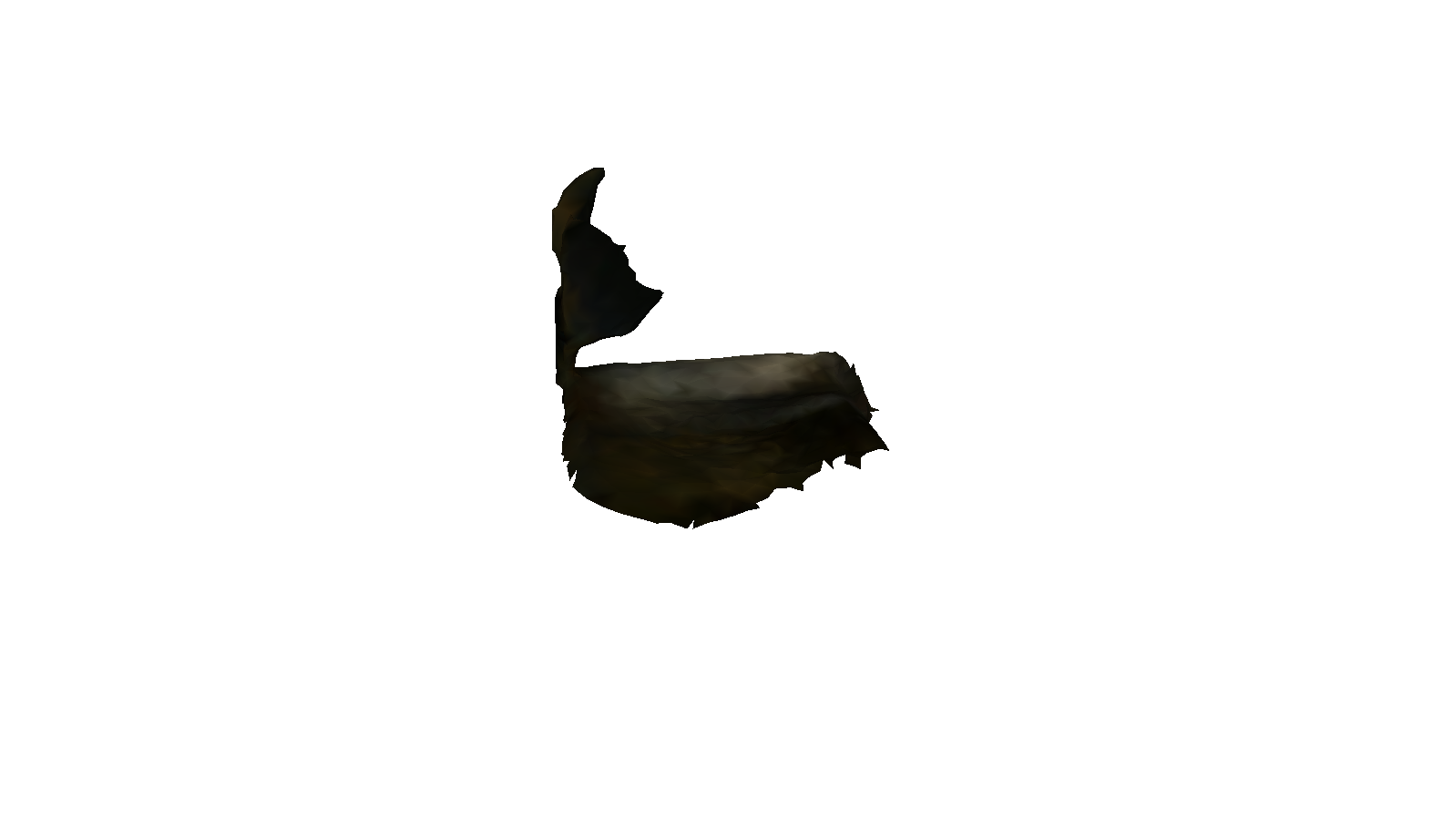} &
  \includegraphics[trim={14.5cm 3.5cm 14.5cm 3.5cm},clip,width=\widthtopfv\linewidth]{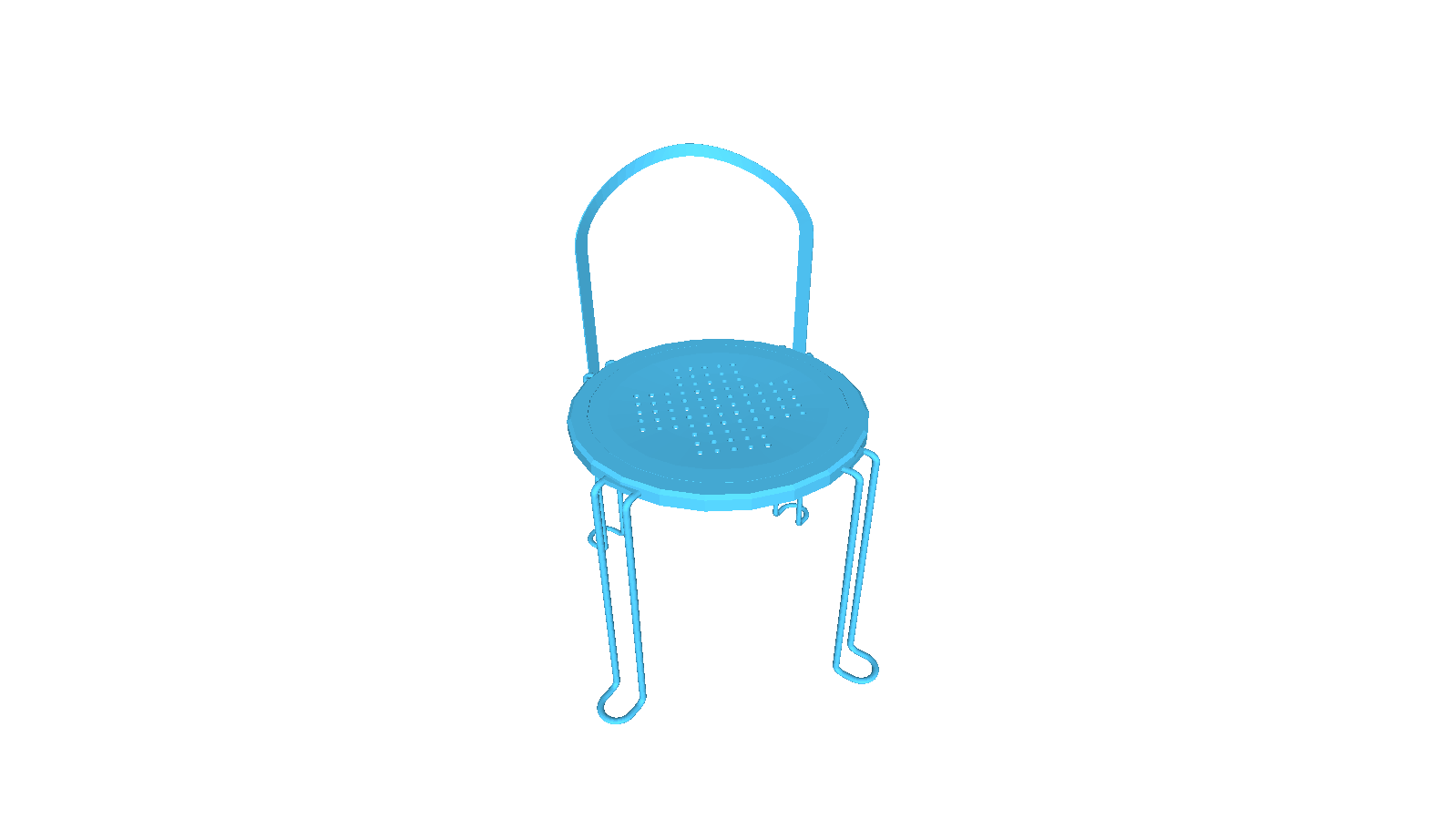} &  
  \includegraphics[trim={14.5cm 3.5cm 14.5cm 3.5cm},clip,width=\widthtopfv\linewidth]{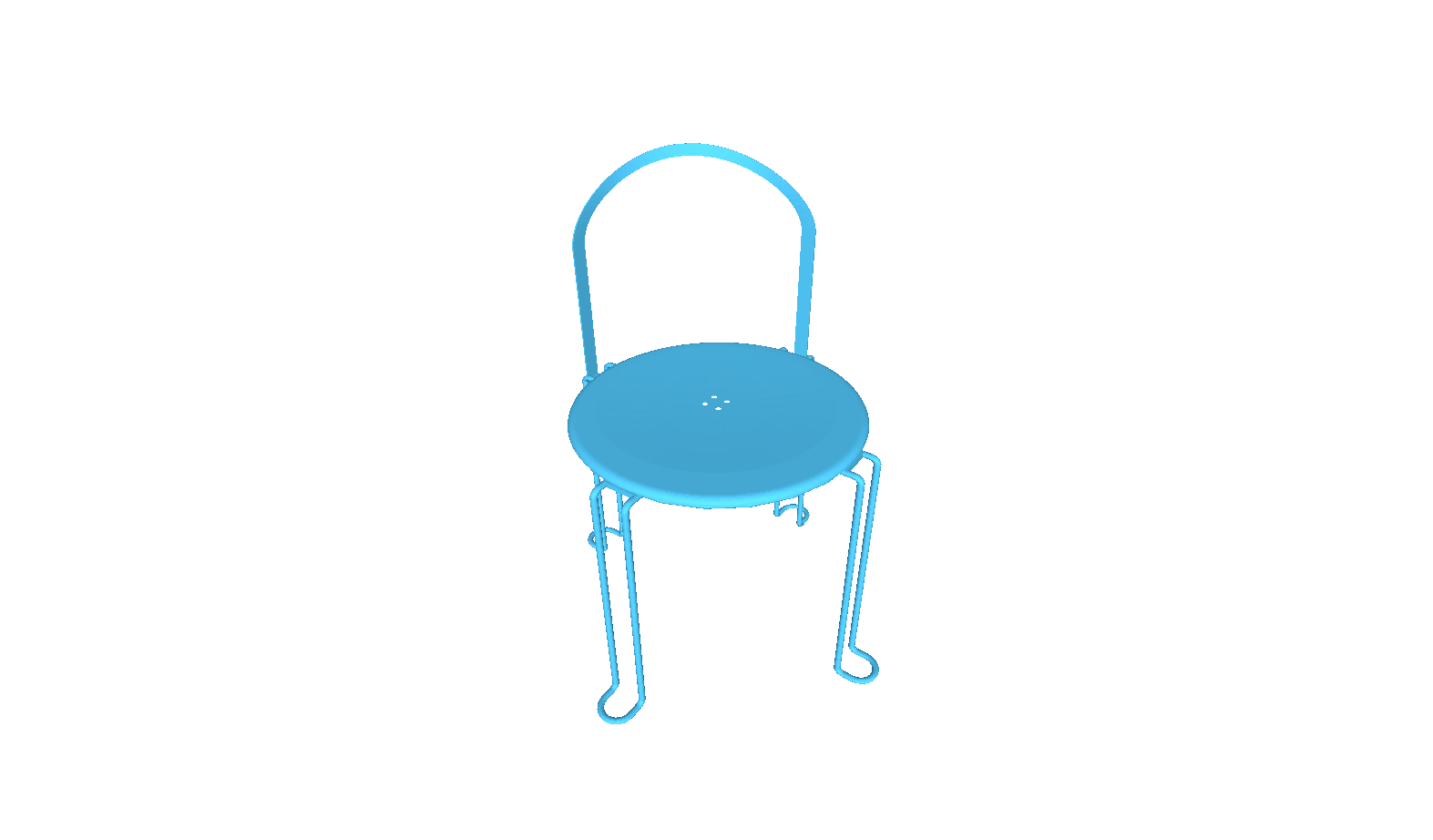} &
  \includegraphics[trim={14.5cm 3.5cm 14.5cm 3.5cm},clip,width=\widthtopfv\linewidth]{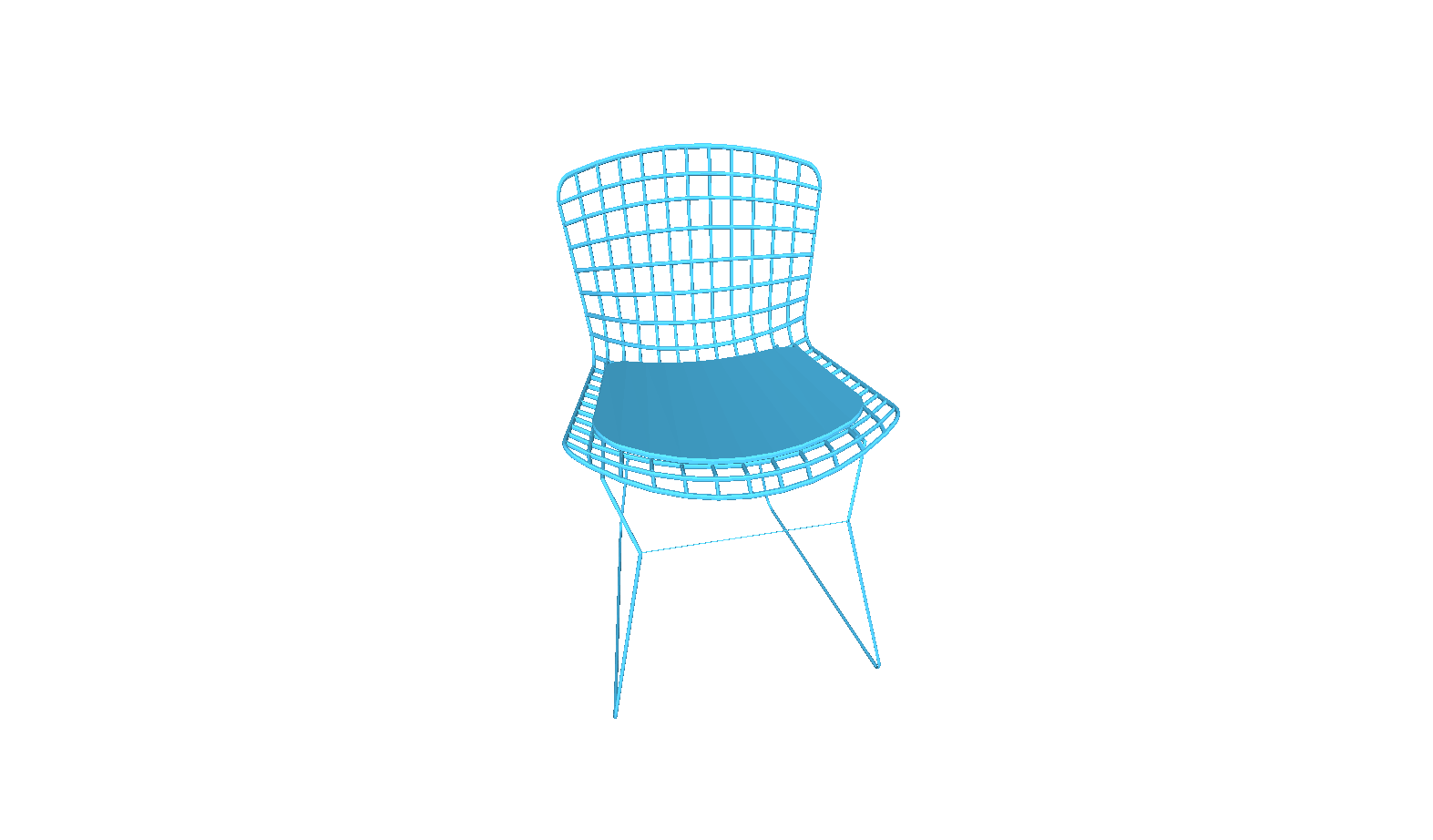} &
    \textcolor{green}{\frame{\includegraphics[trim={14.5cm 3.5cm 14.5cm 3.5cm},clip,width=\widthtopfv\linewidth]{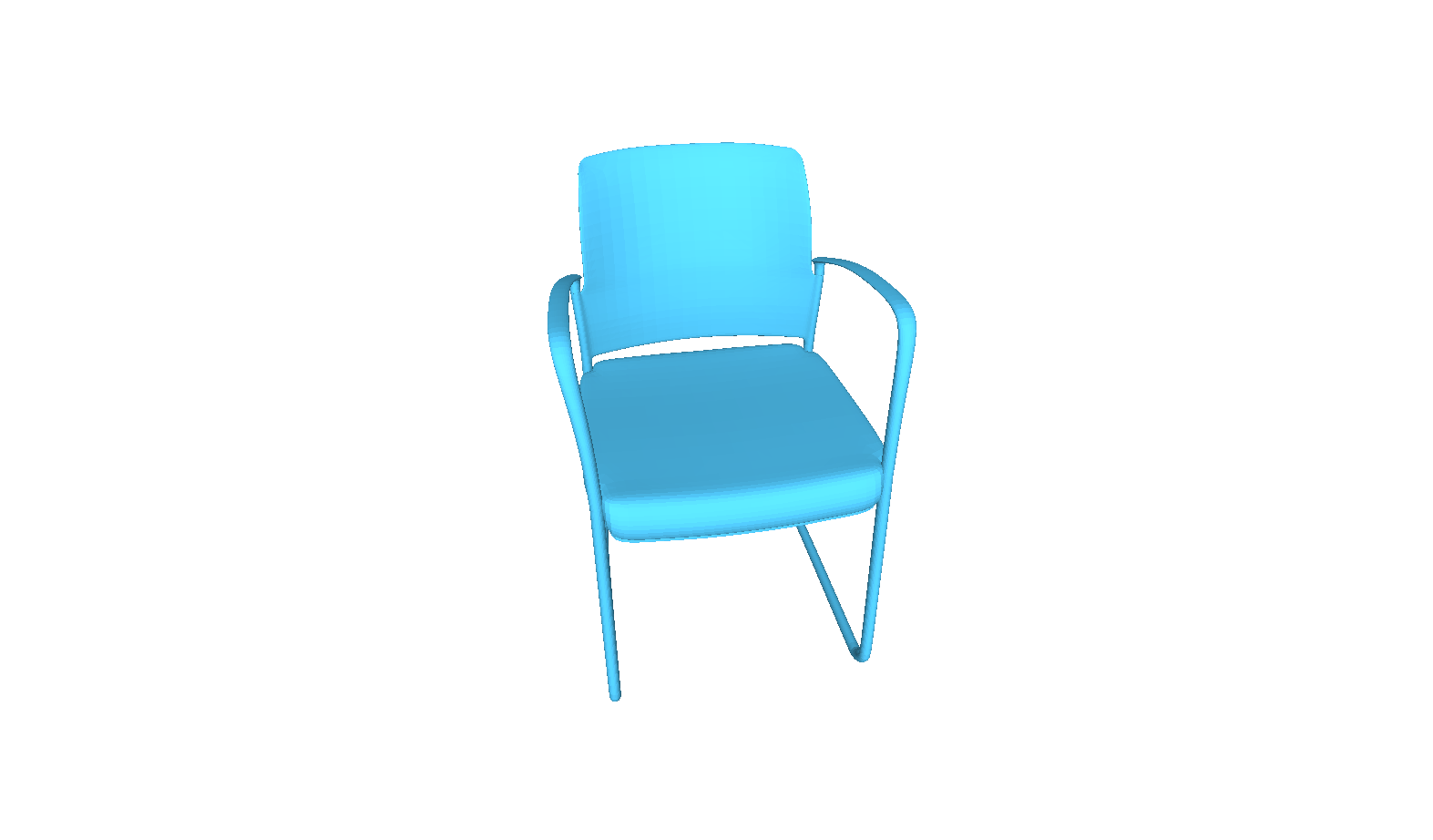}}} &
  \includegraphics[trim={14.5cm 3.5cm 14.5cm 3.5cm},clip,width=\widthtopfv\linewidth]{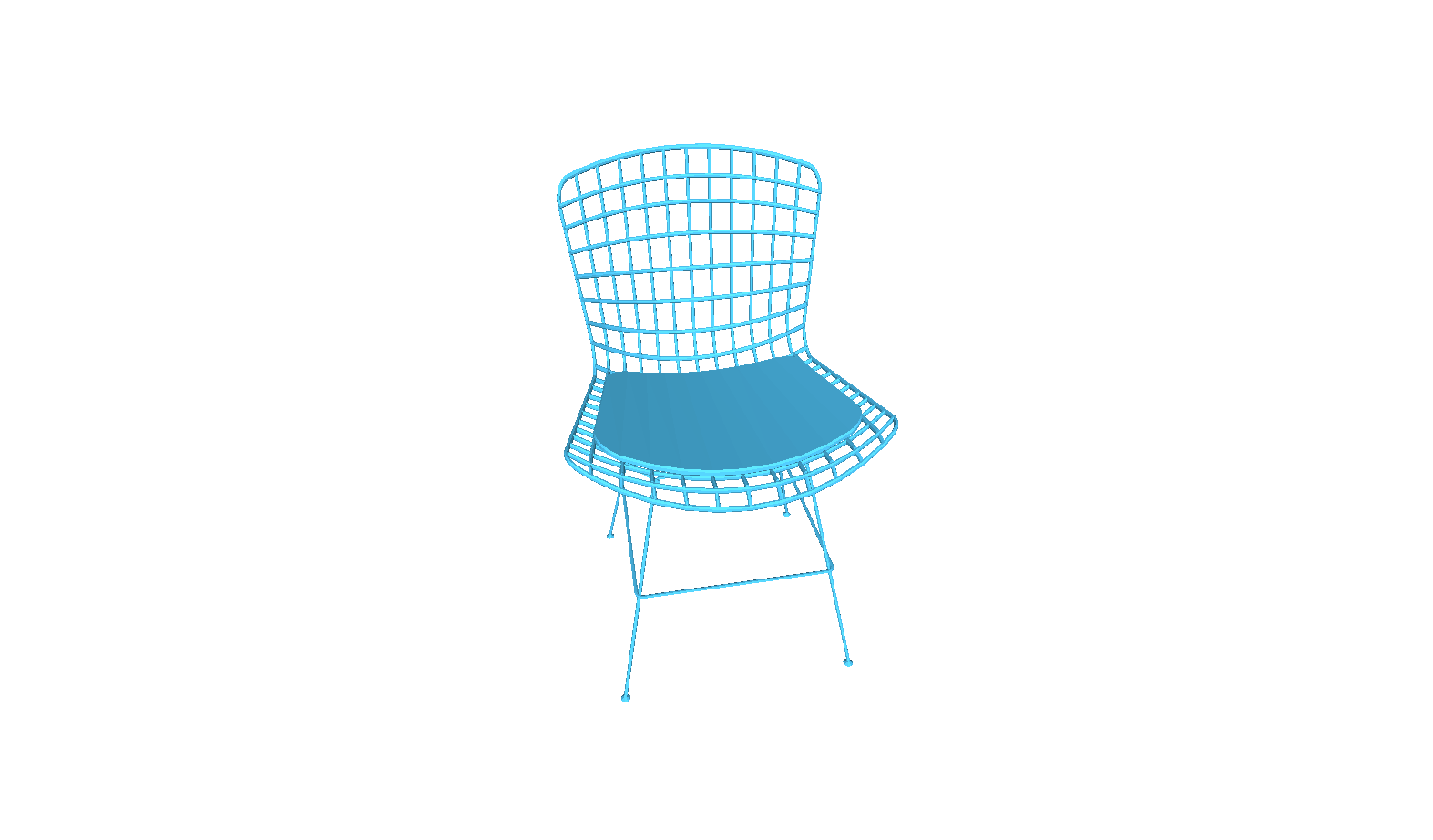} &
  \includegraphics[trim={14.5cm 3.5cm 14.5cm 3.5cm},clip,width=\widthtopfv\linewidth]{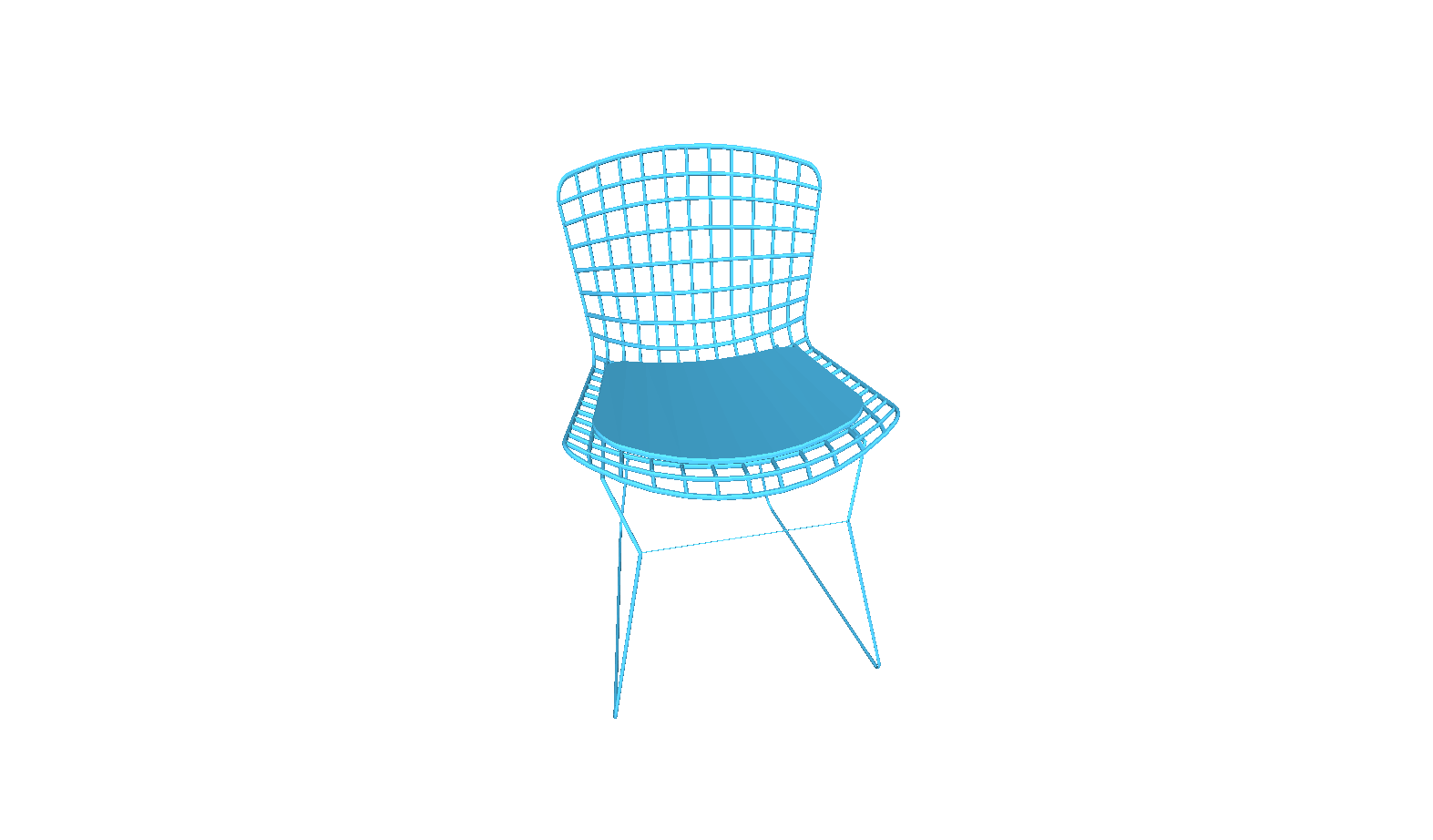} &
  \includegraphics[trim={14.5cm 3.5cm 14.5cm 3.5cm},clip,width=\widthtopfv\linewidth]{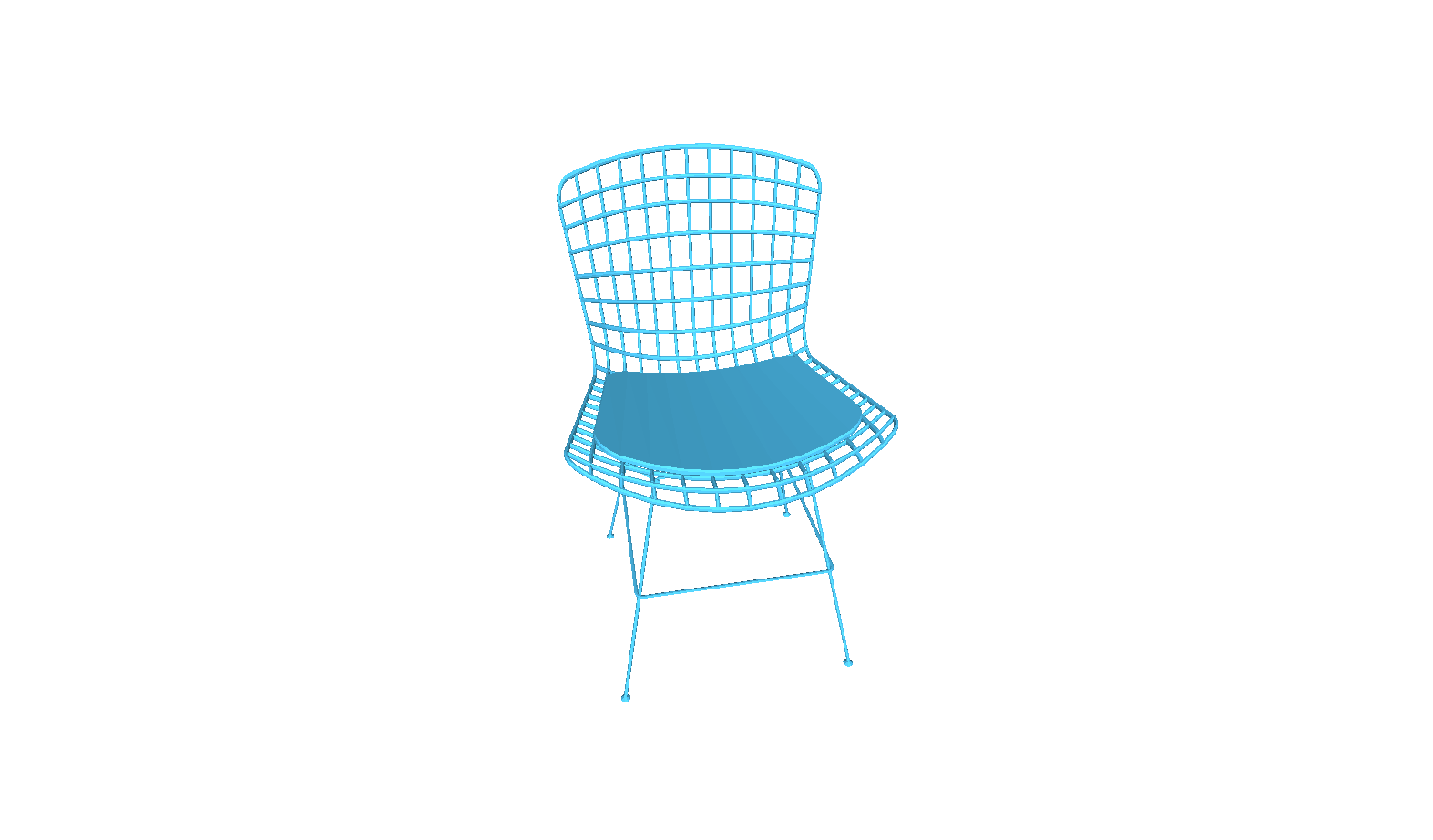} &
  \includegraphics[trim={14.5cm 3.5cm 14.5cm 3.5cm},clip,width=\widthtopfv\linewidth]{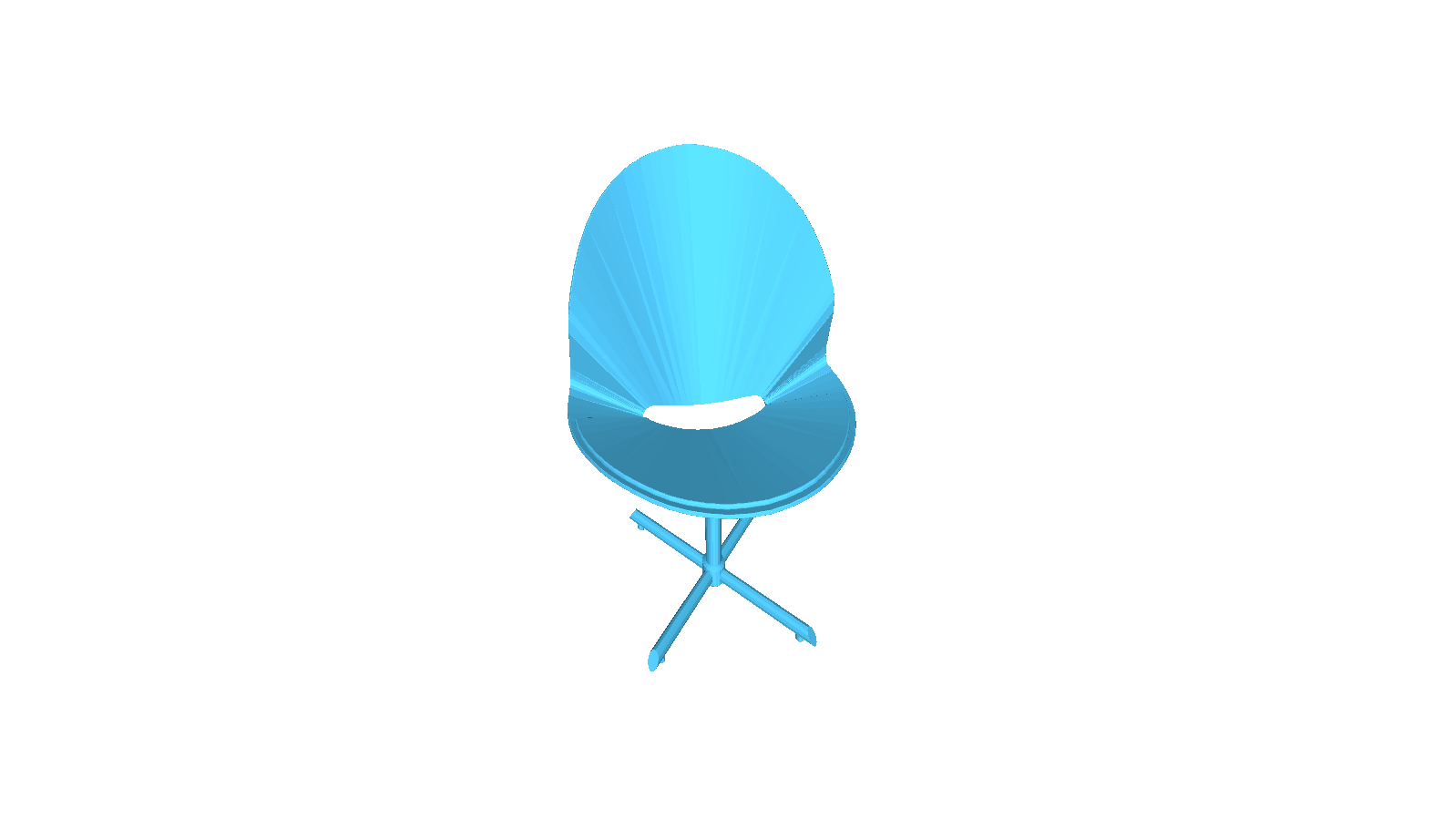} &
  \textcolor{green}{\frame{\includegraphics[trim={14.5cm 3.5cm 14.5cm 3.5cm},clip,width=\widthtopfv\linewidth]{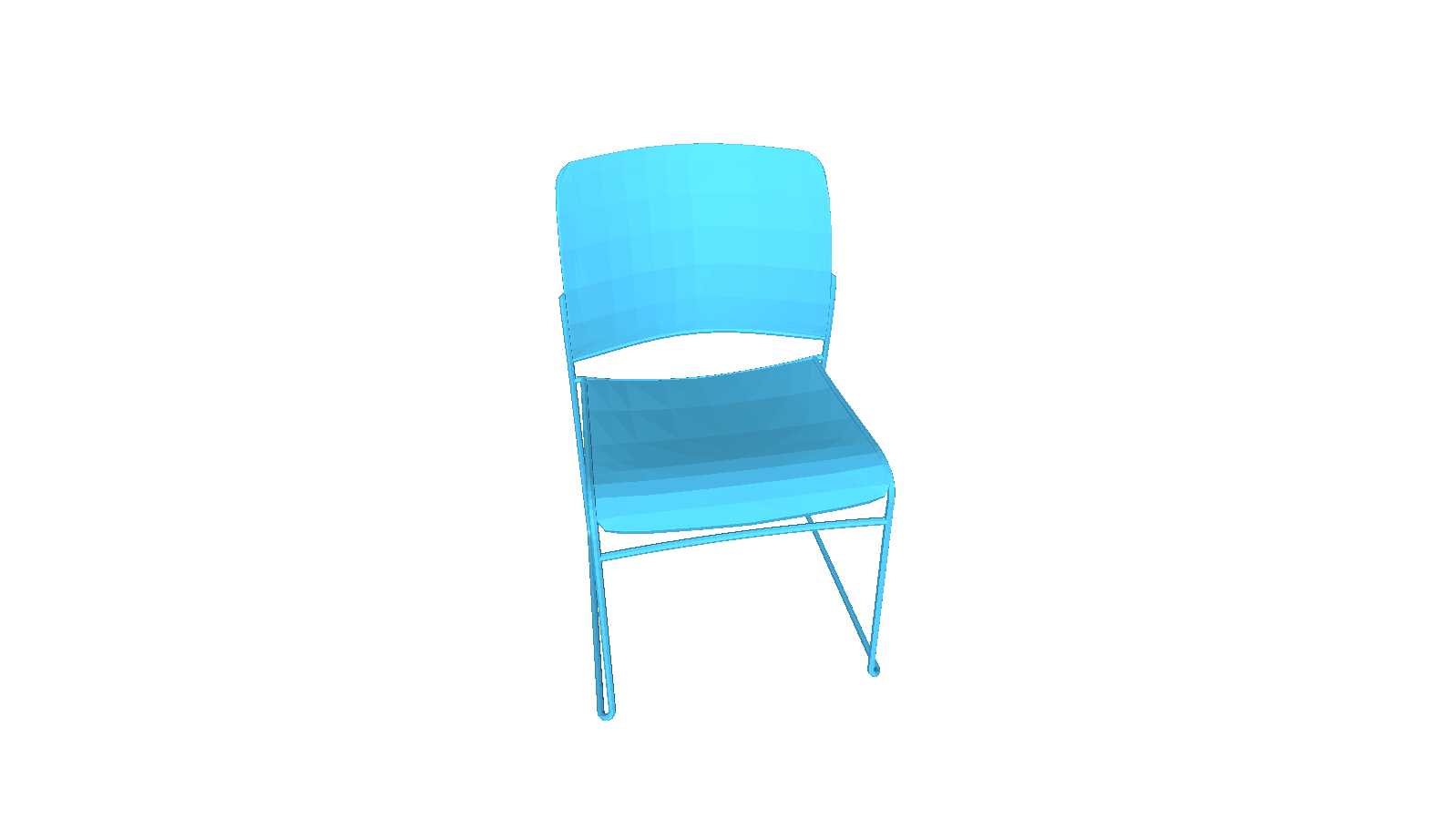}}} &
  \includegraphics[trim={14.5cm 3.5cm 14.5cm 3.5cm},clip,width=\widthtopfv\linewidth]{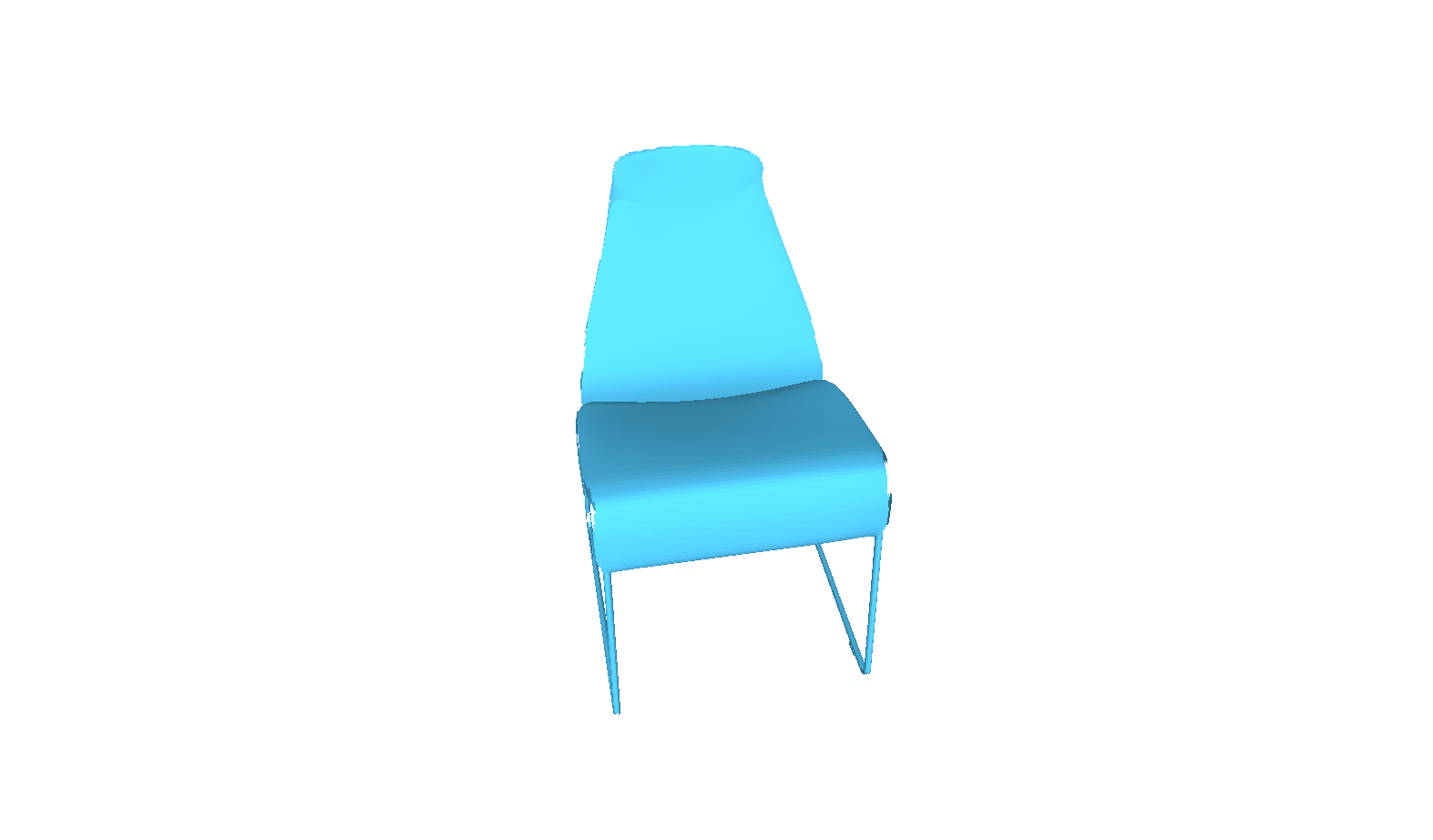} \\

    \includegraphics[trim={15.5cm 3.5cm 15cm 5.5cm},clip,width=\widthtopfv\linewidth]{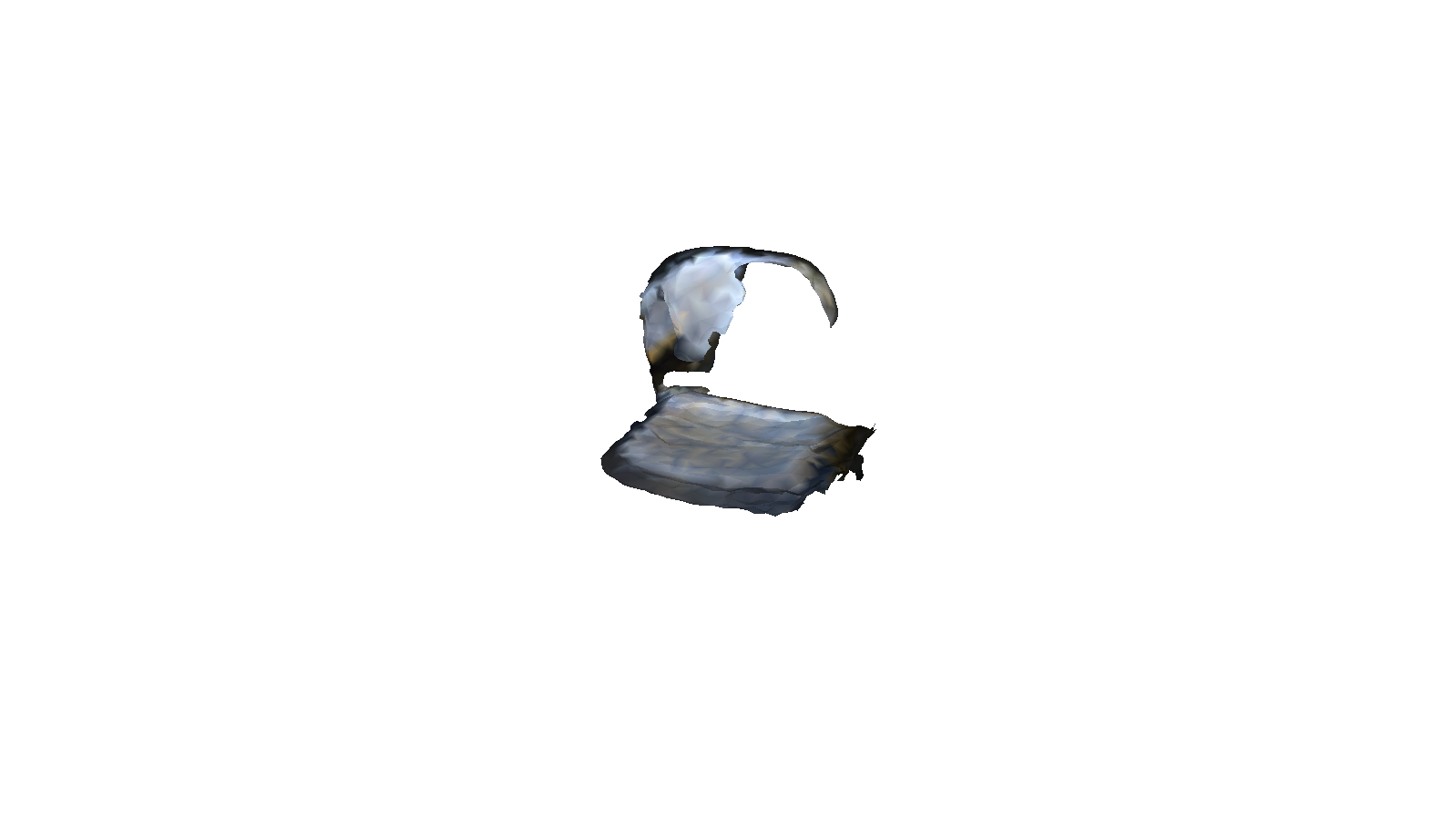} &
  \includegraphics[trim={15.5cm 3.5cm 15cm 5.5cm},clip,width=\widthtopfv\linewidth]{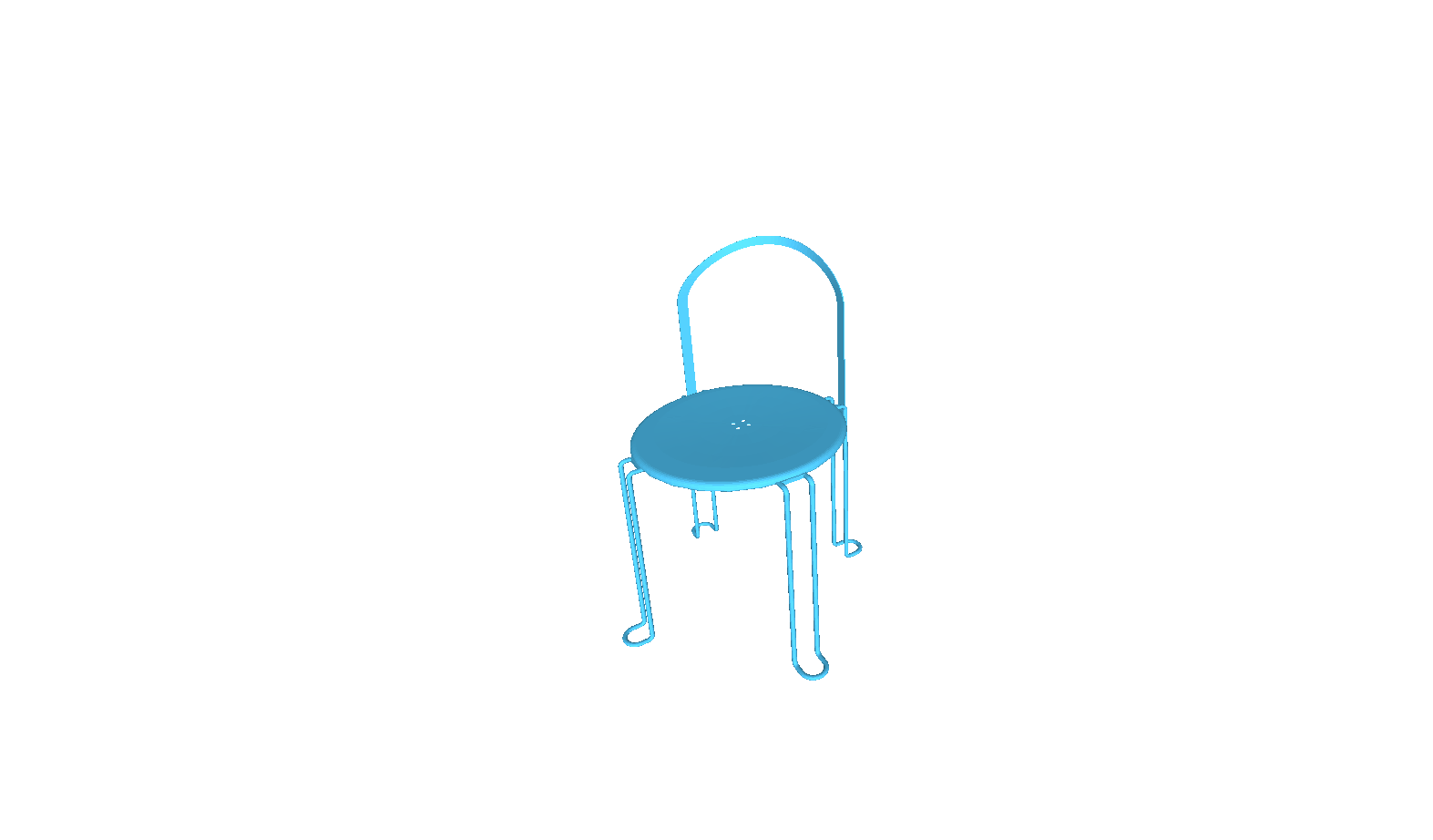} &  
  \includegraphics[trim={15.5cm 3.5cm 15cm 5.5cm},clip,width=\widthtopfv\linewidth]{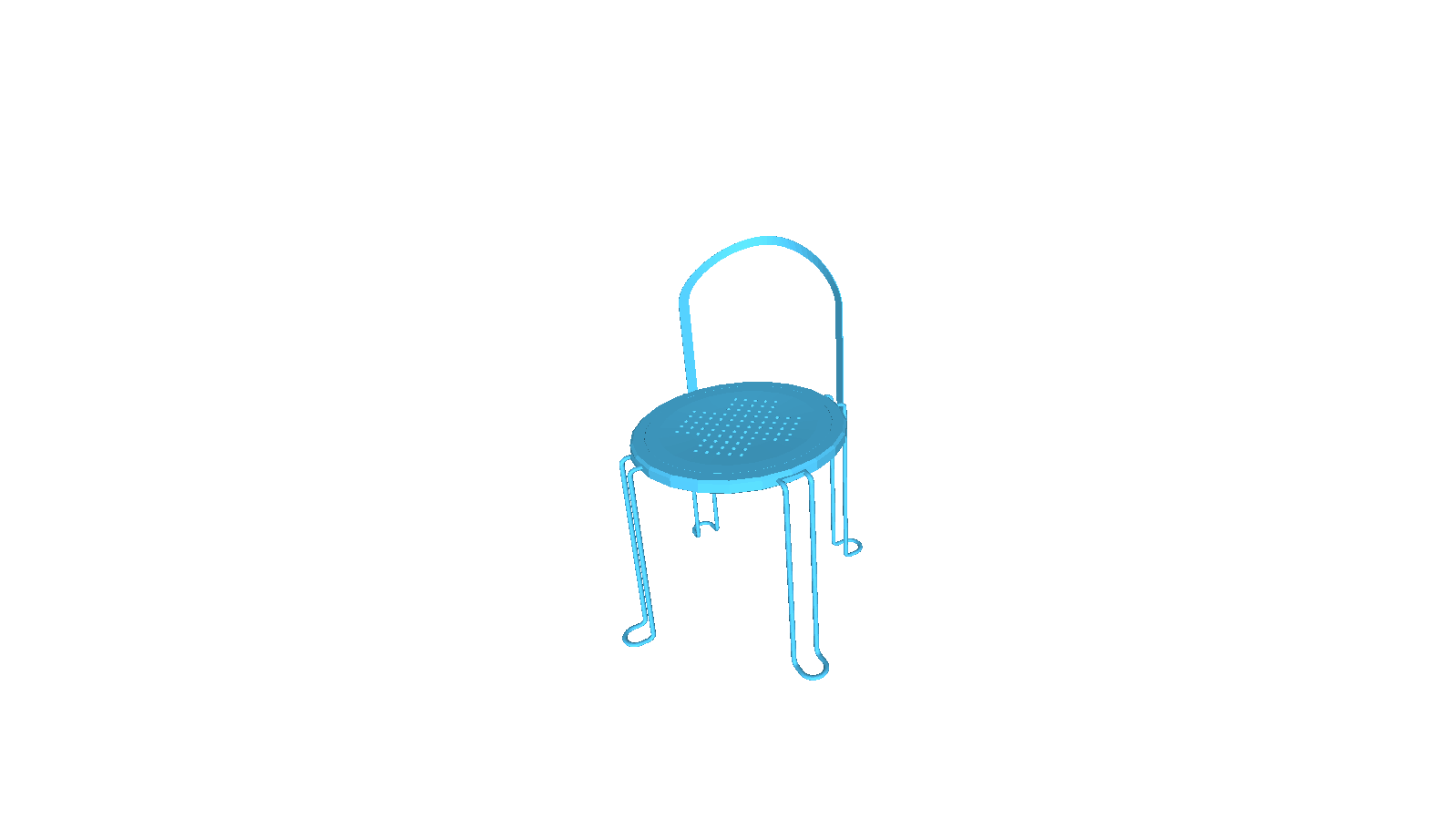} &
  \includegraphics[trim={15.5cm 3.5cm 15cm 5.5cm},clip,width=\widthtopfv\linewidth]{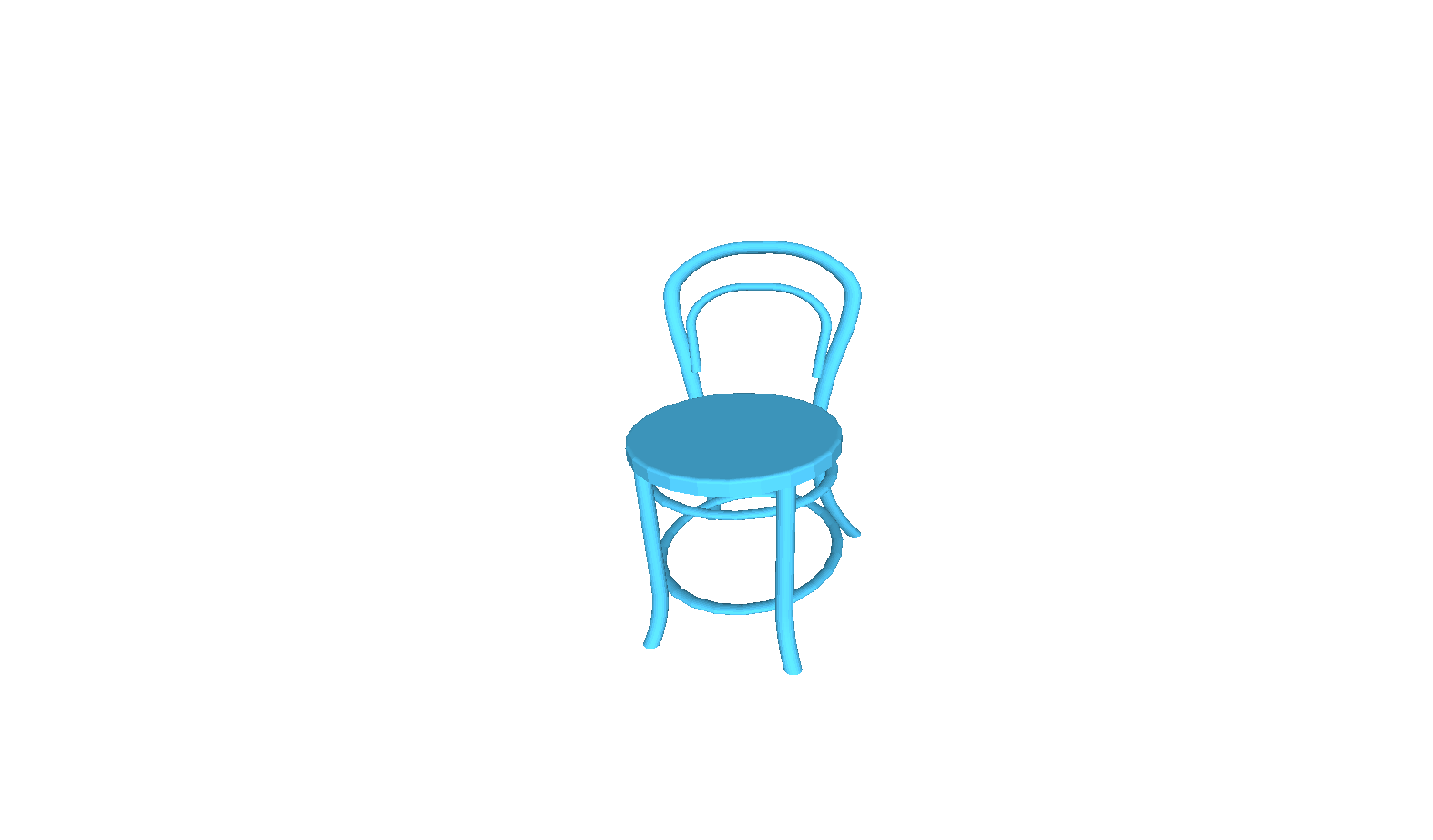} &
  \includegraphics[trim={15.5cm 3.5cm 15cm 5.5cm},clip,width=\widthtopfv\linewidth]{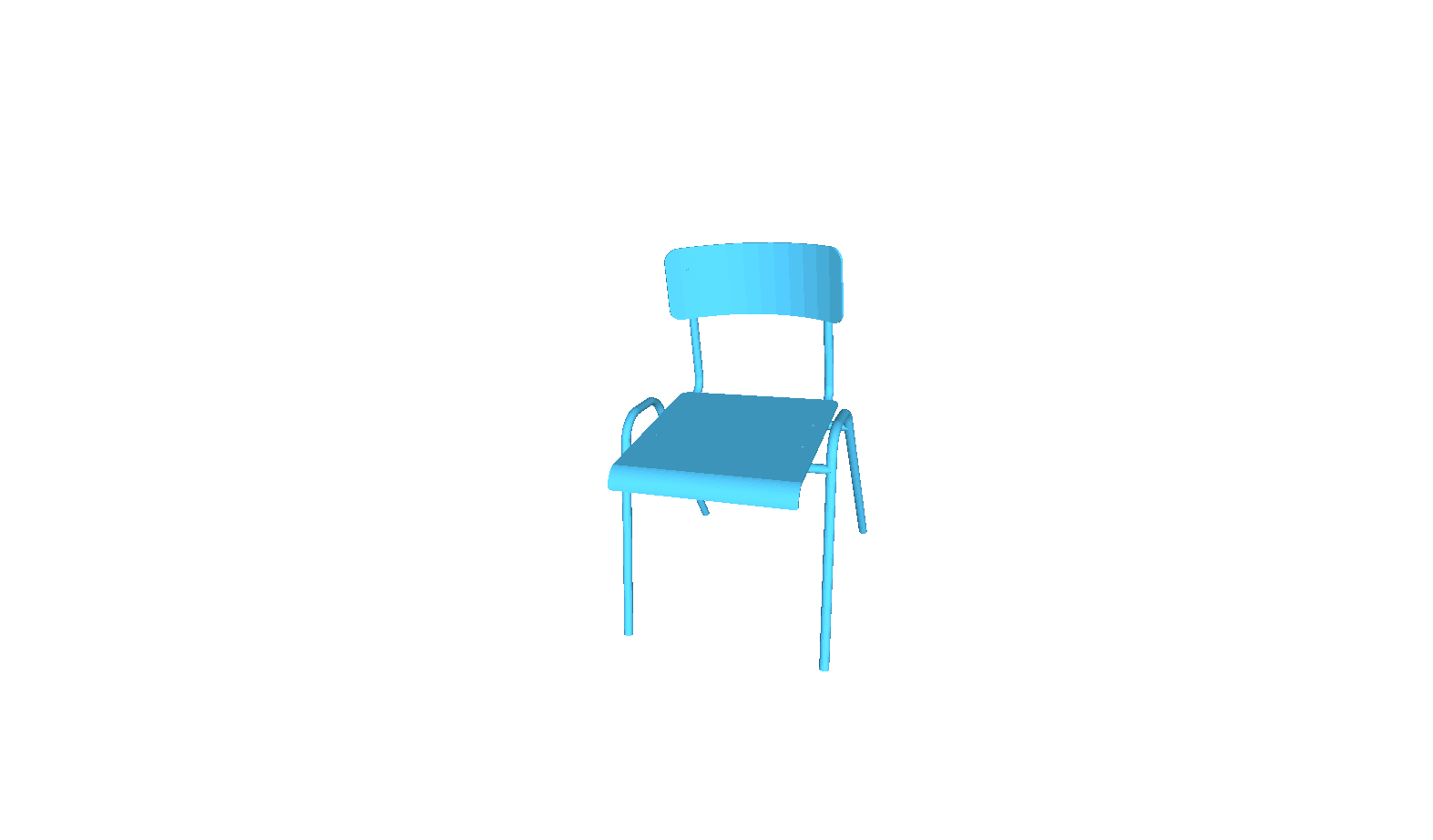} &
  \includegraphics[trim={15.5cm 3.5cm 15cm 5.5cm},clip,width=\widthtopfv\linewidth]{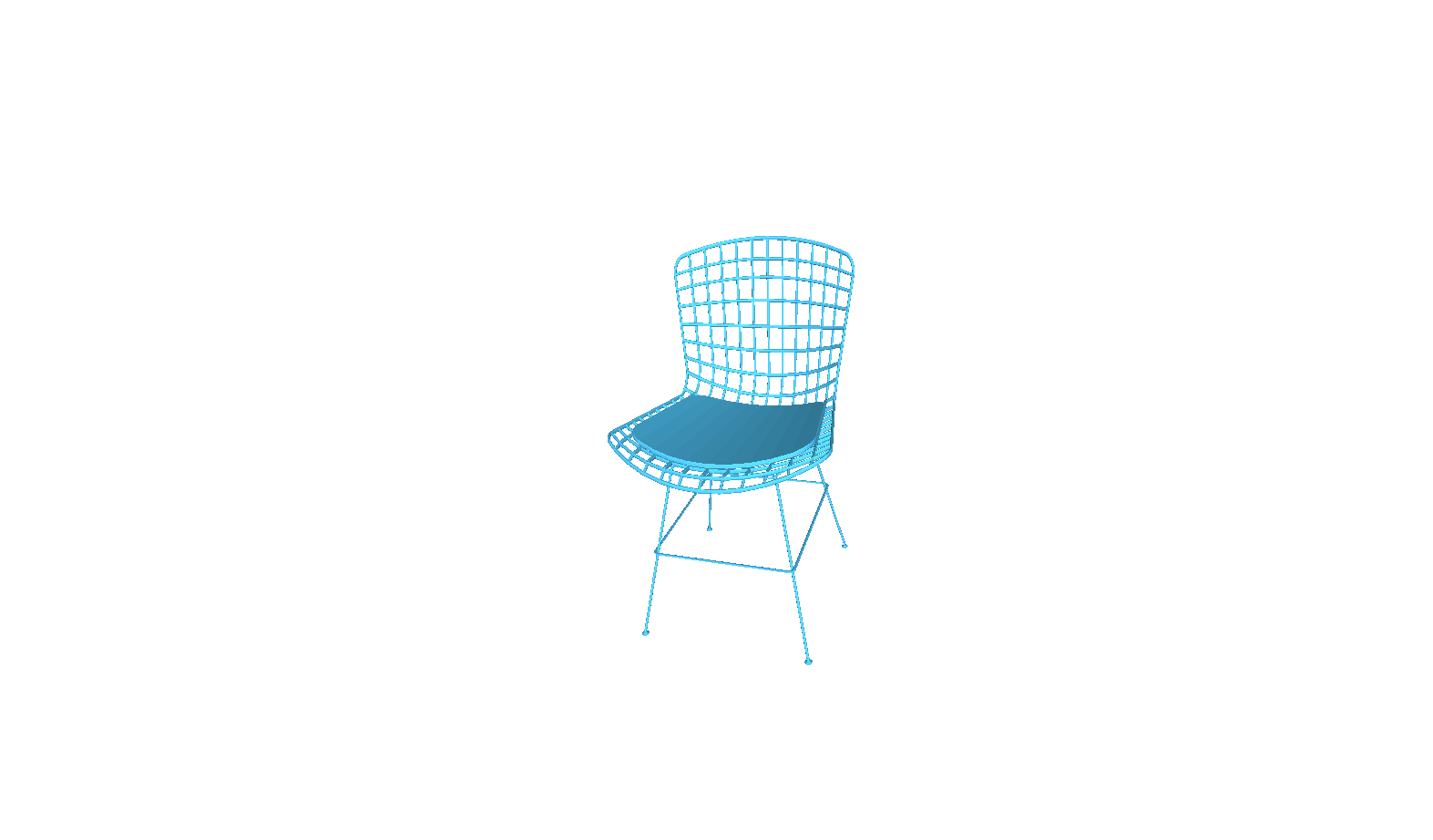} &
  \includegraphics[trim={15.5cm 3.5cm 15cm 5.5cm},clip,width=\widthtopfv\linewidth]{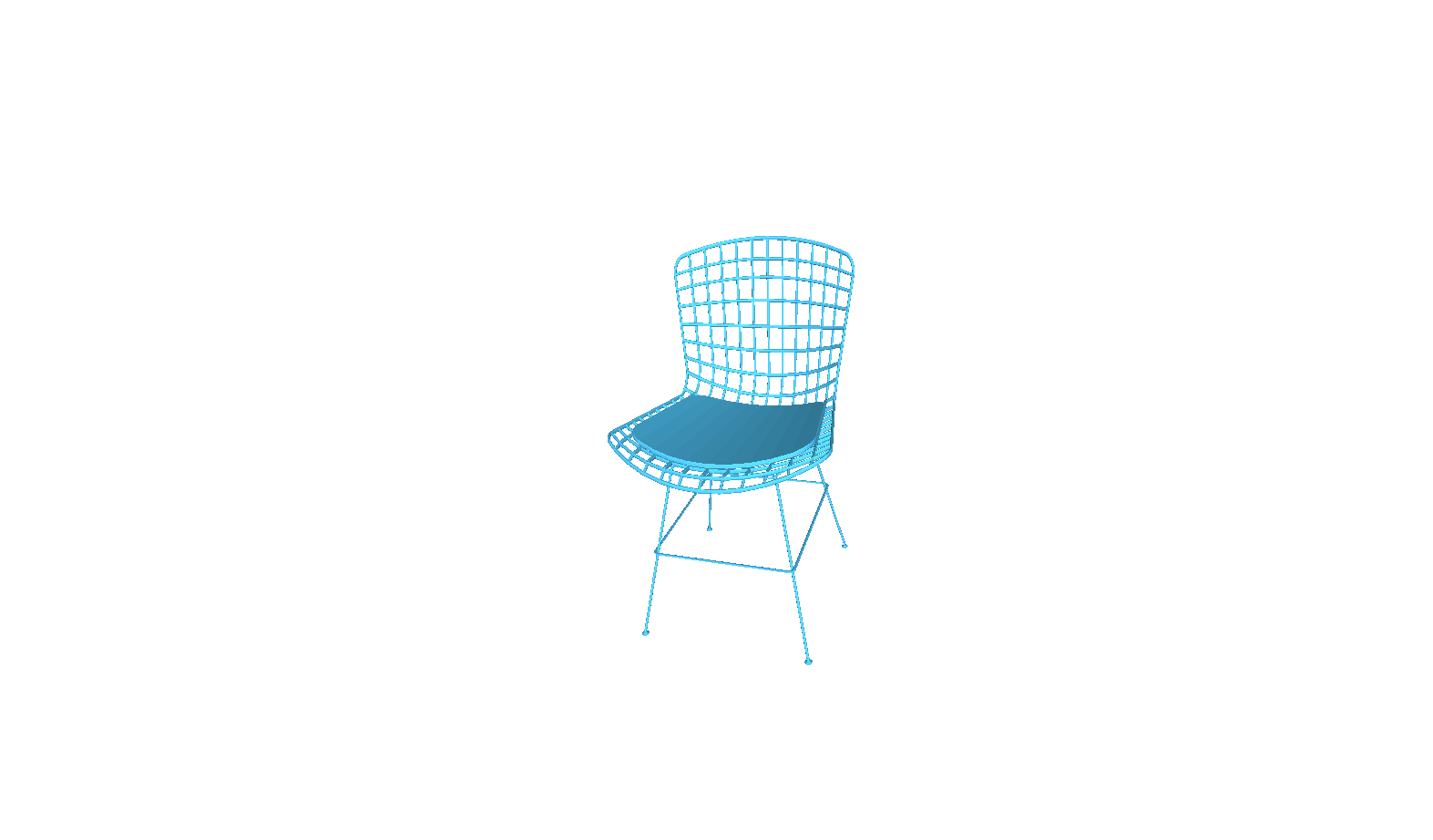} &

    \textcolor{green}{\frame{\includegraphics[trim={15.5cm 3.5cm 15cm 5.5cm},clip,width=\widthtopfv\linewidth]{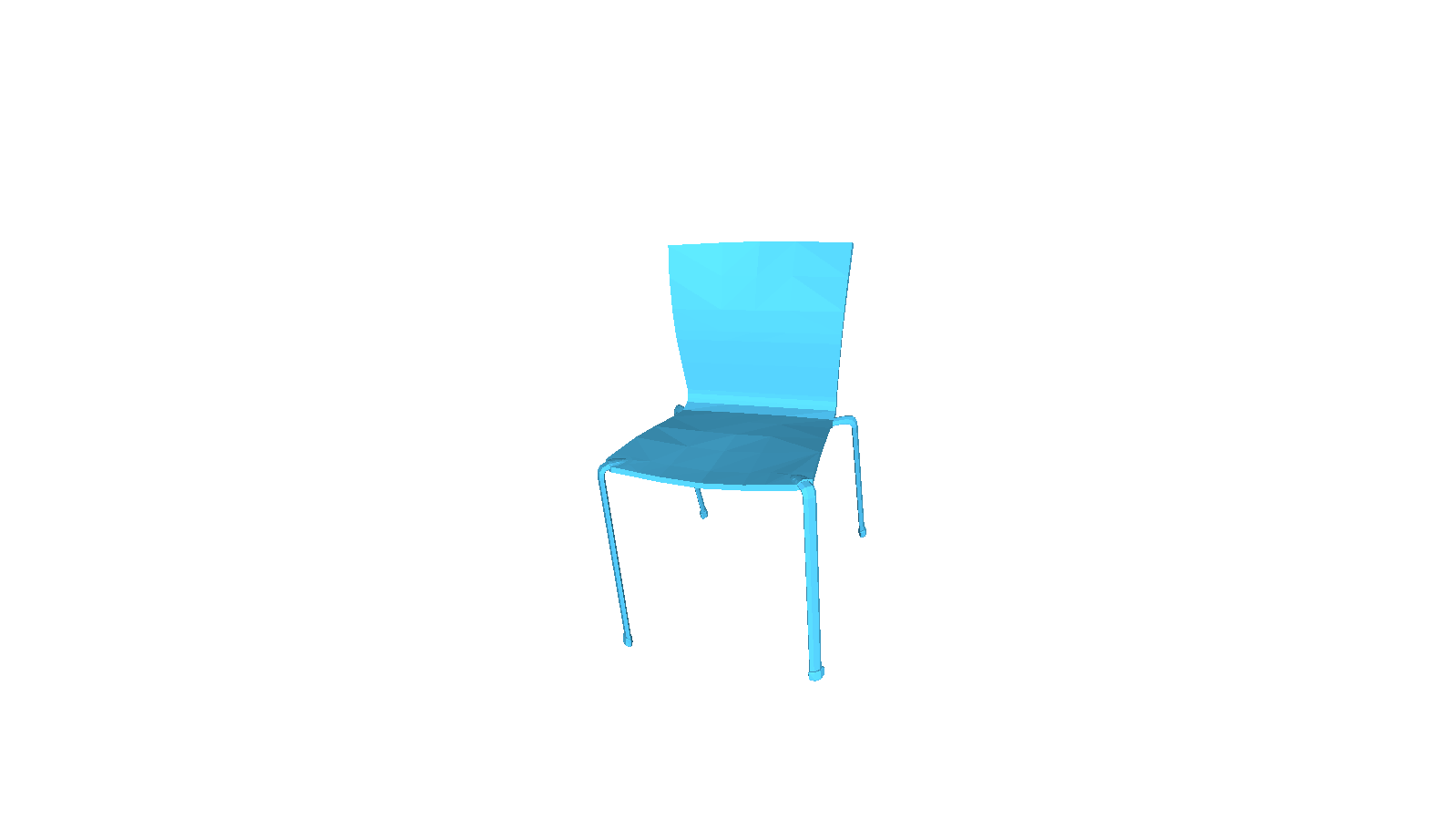}}} &
  
  \includegraphics[trim={15.5cm 3.5cm 15cm 5.5cm},clip,width=\widthtopfv\linewidth]{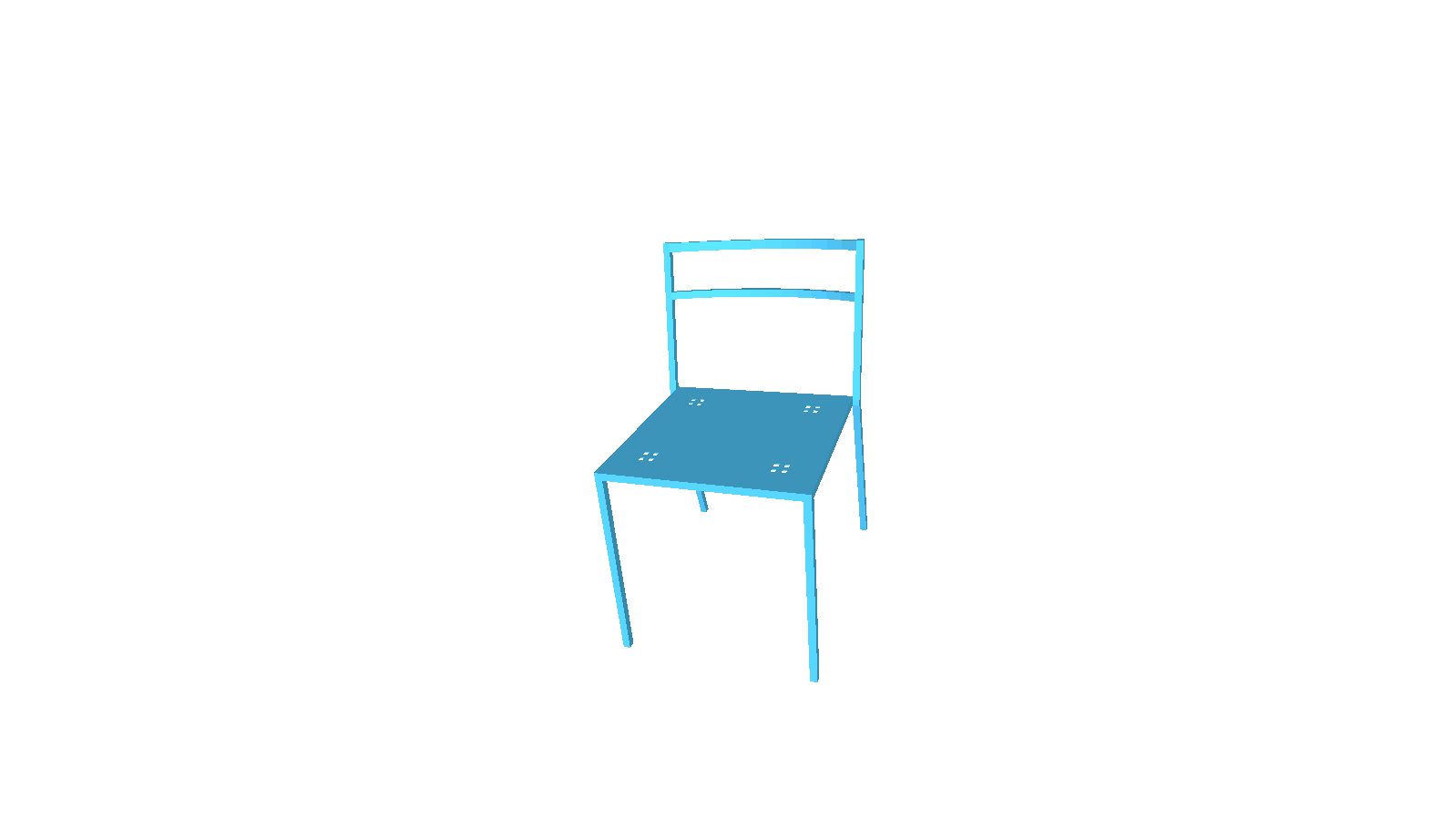} &
  \includegraphics[trim={15.5cm 3.5cm 15cm 5.5cm},clip,width=\widthtopfv\linewidth]{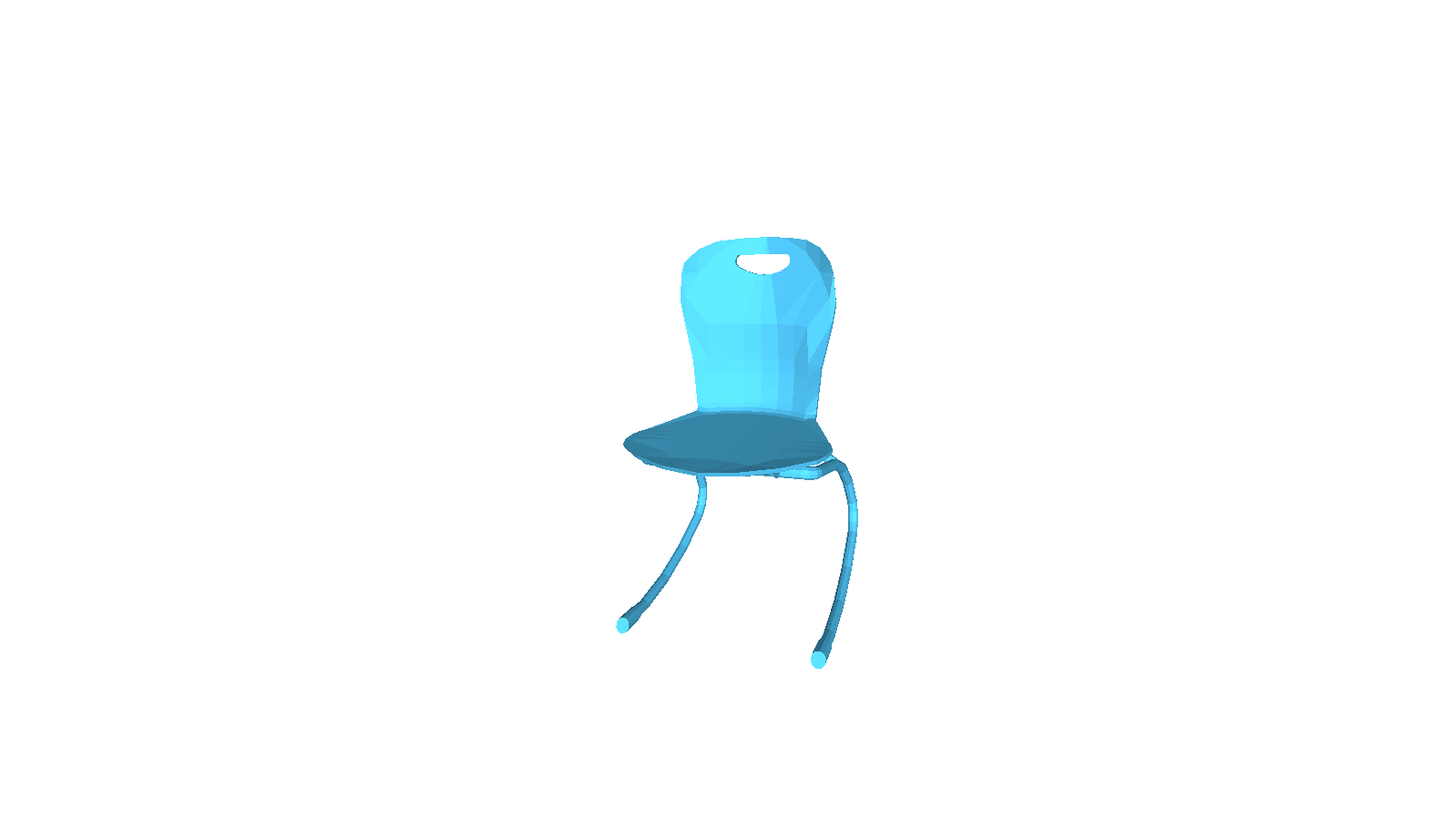} &

    \textcolor{green}{\frame{\includegraphics[trim={15.5cm 3.5cm 15cm 5.5cm},clip,width=\widthtopfv\linewidth]{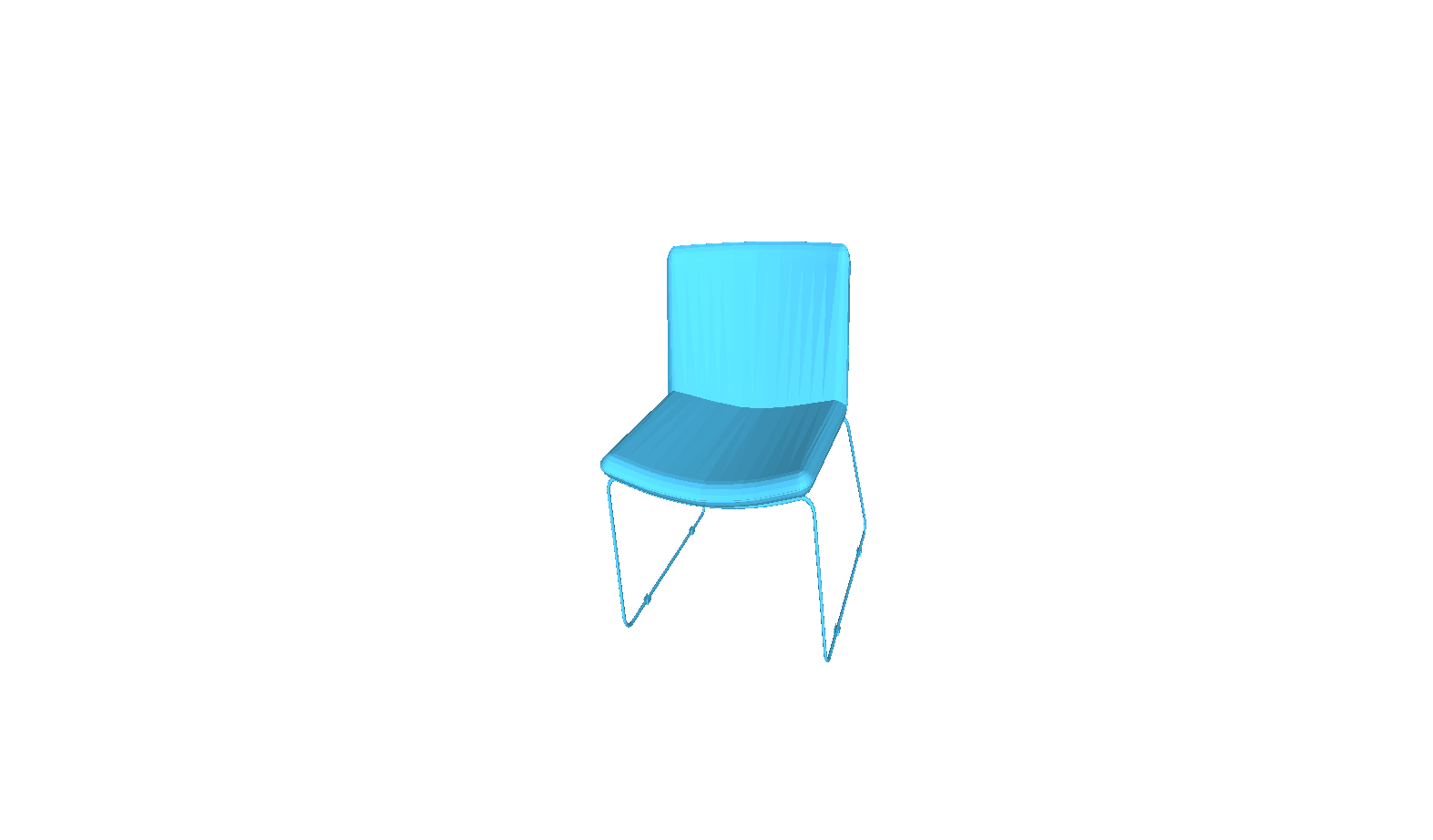}}}
  
  \\
\midrule
    \includegraphics[trim={13.5cm 3.5cm 15cm 4.5cm},clip,width=\widthtopfv\linewidth]{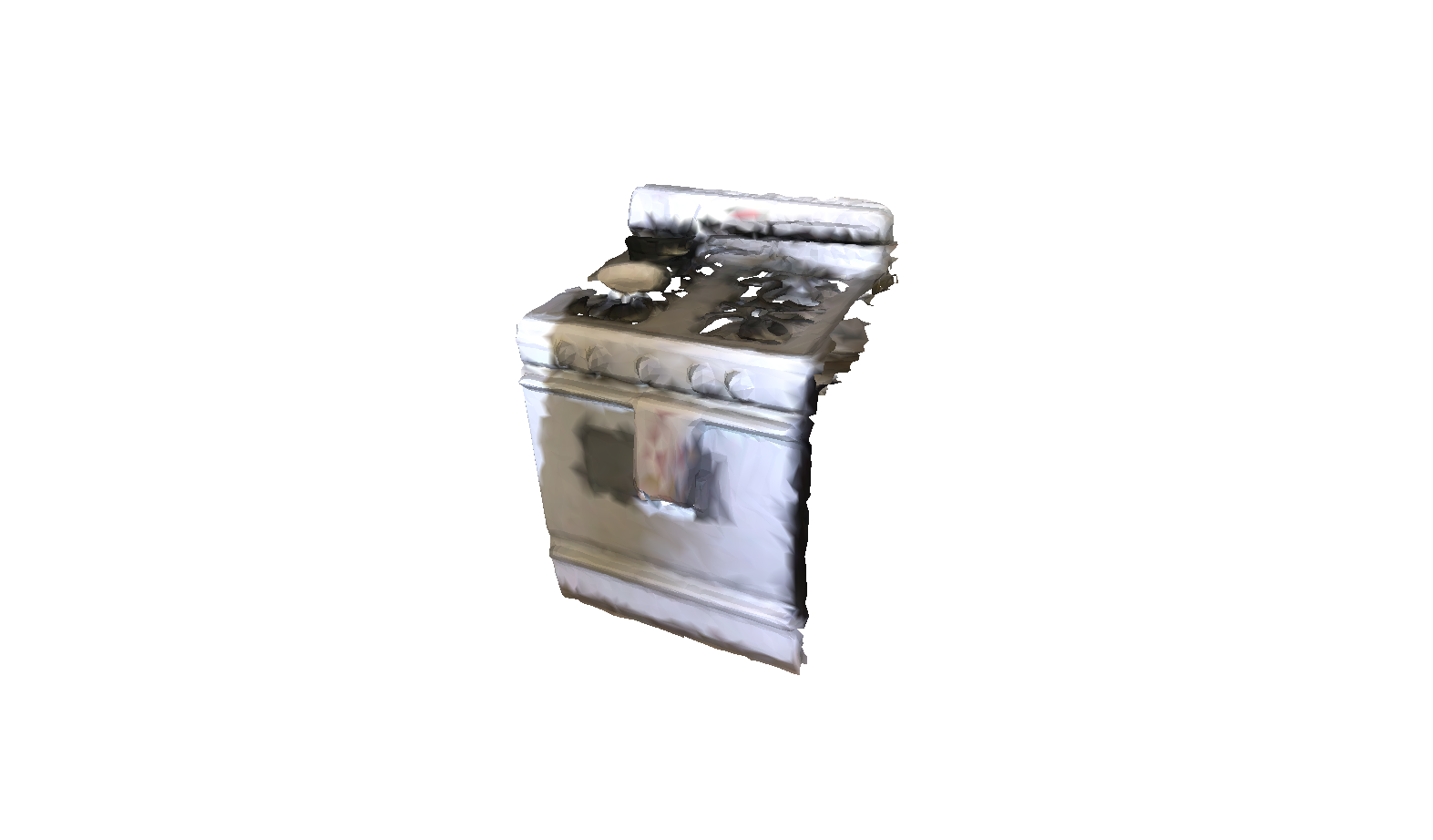} &
  \includegraphics[trim={13.5cm 3.5cm 15cm 4.5cm},clip,width=\widthtopfv\linewidth]{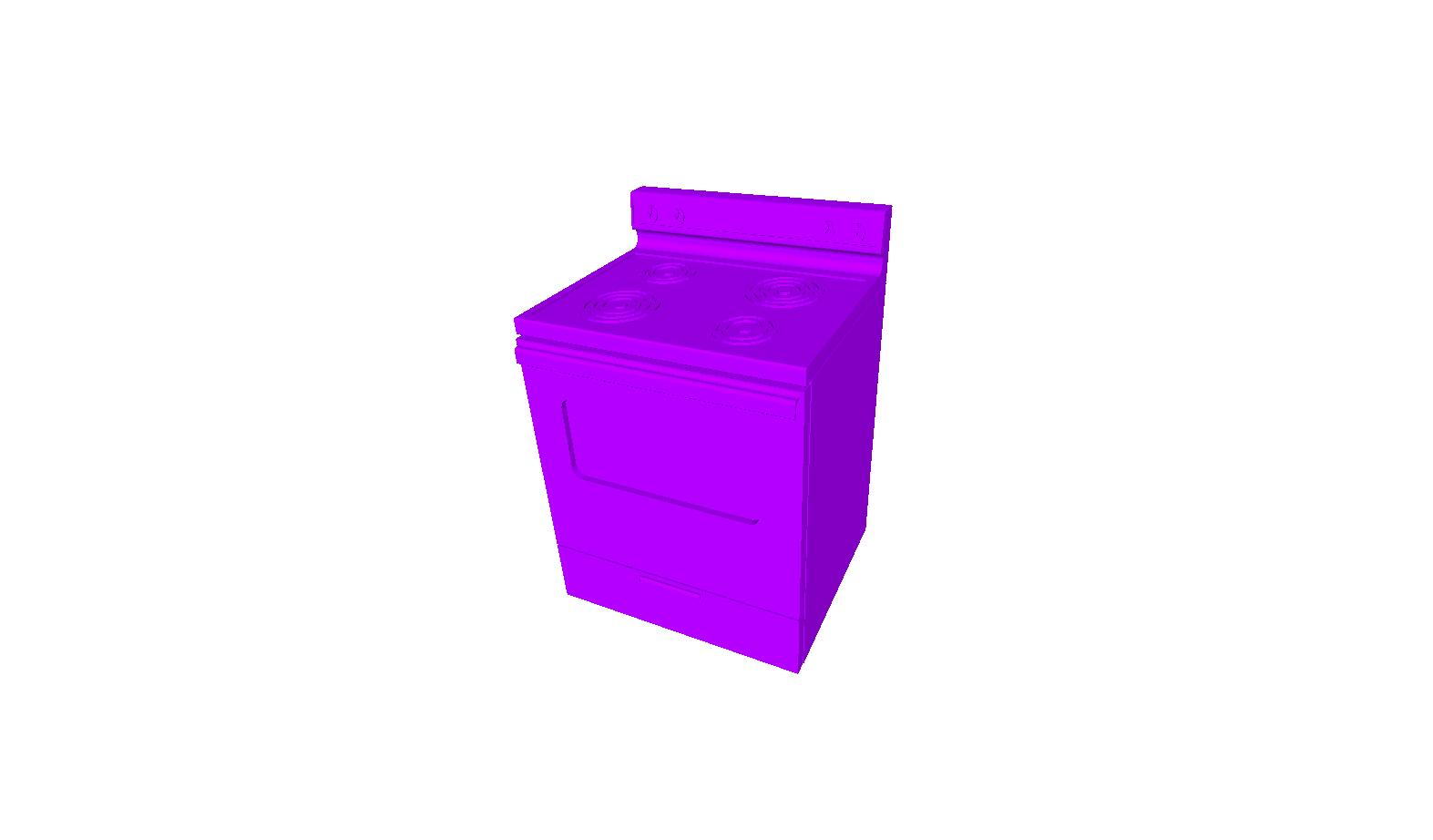} &  
  \includegraphics[trim={13.5cm 3.5cm 15cm 4.5cm},clip,width=\widthtopfv\linewidth]{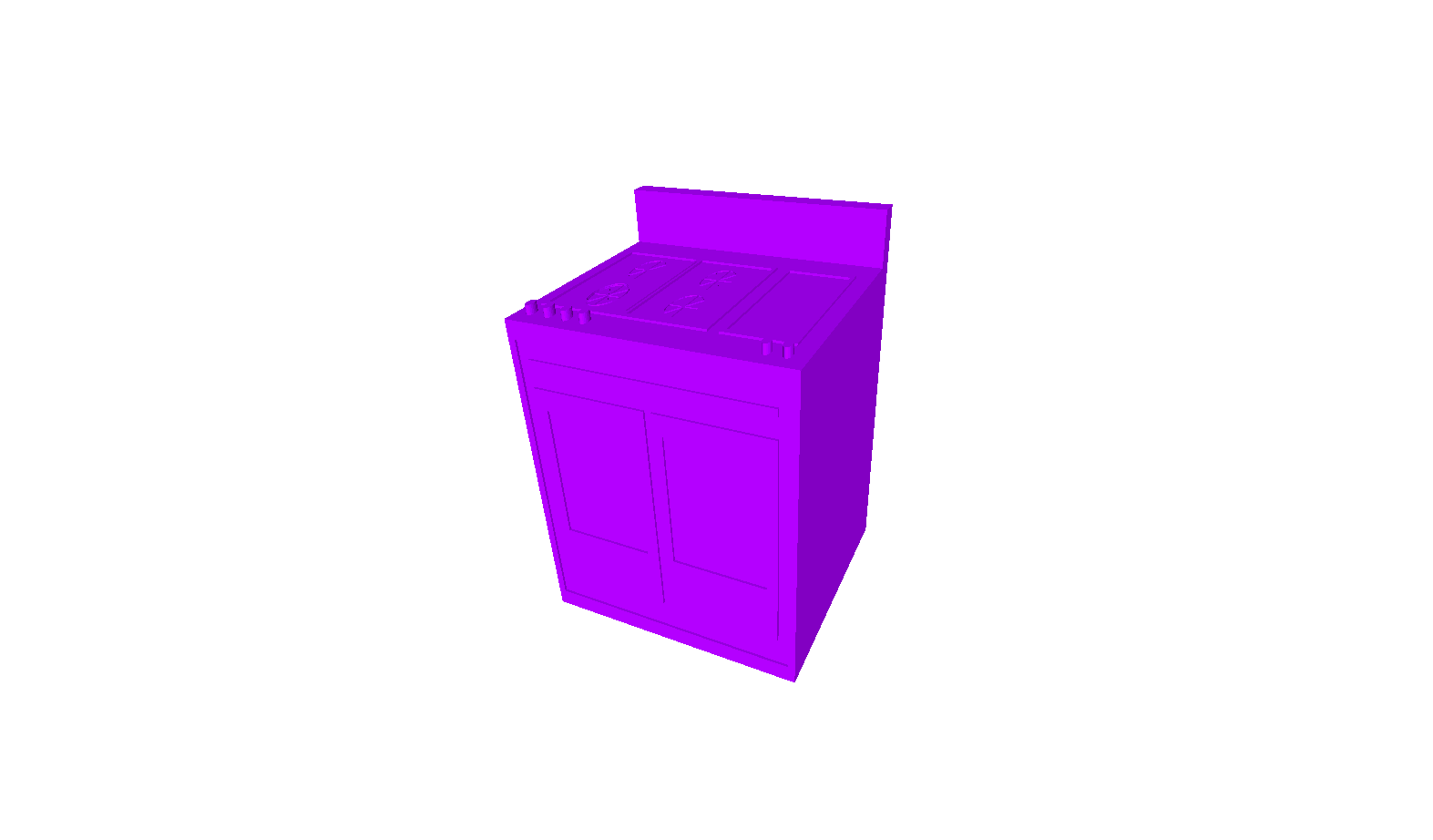} &
  \includegraphics[trim={13.5cm 3.5cm 15cm 4.5cm},clip,width=\widthtopfv\linewidth]{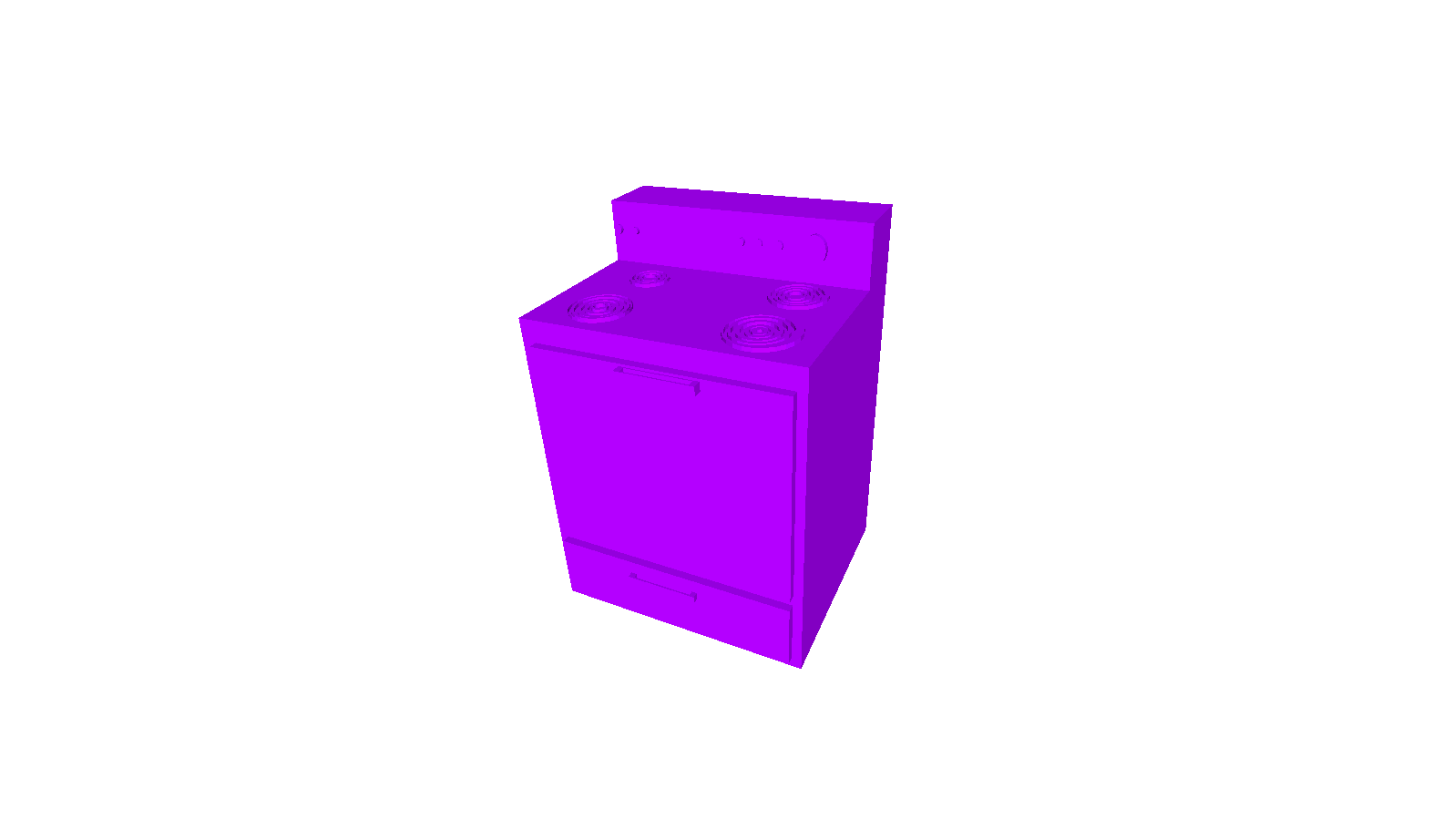} &
  \includegraphics[trim={13.5cm 3.5cm 15cm 4.5cm},clip,width=\widthtopfv\linewidth]{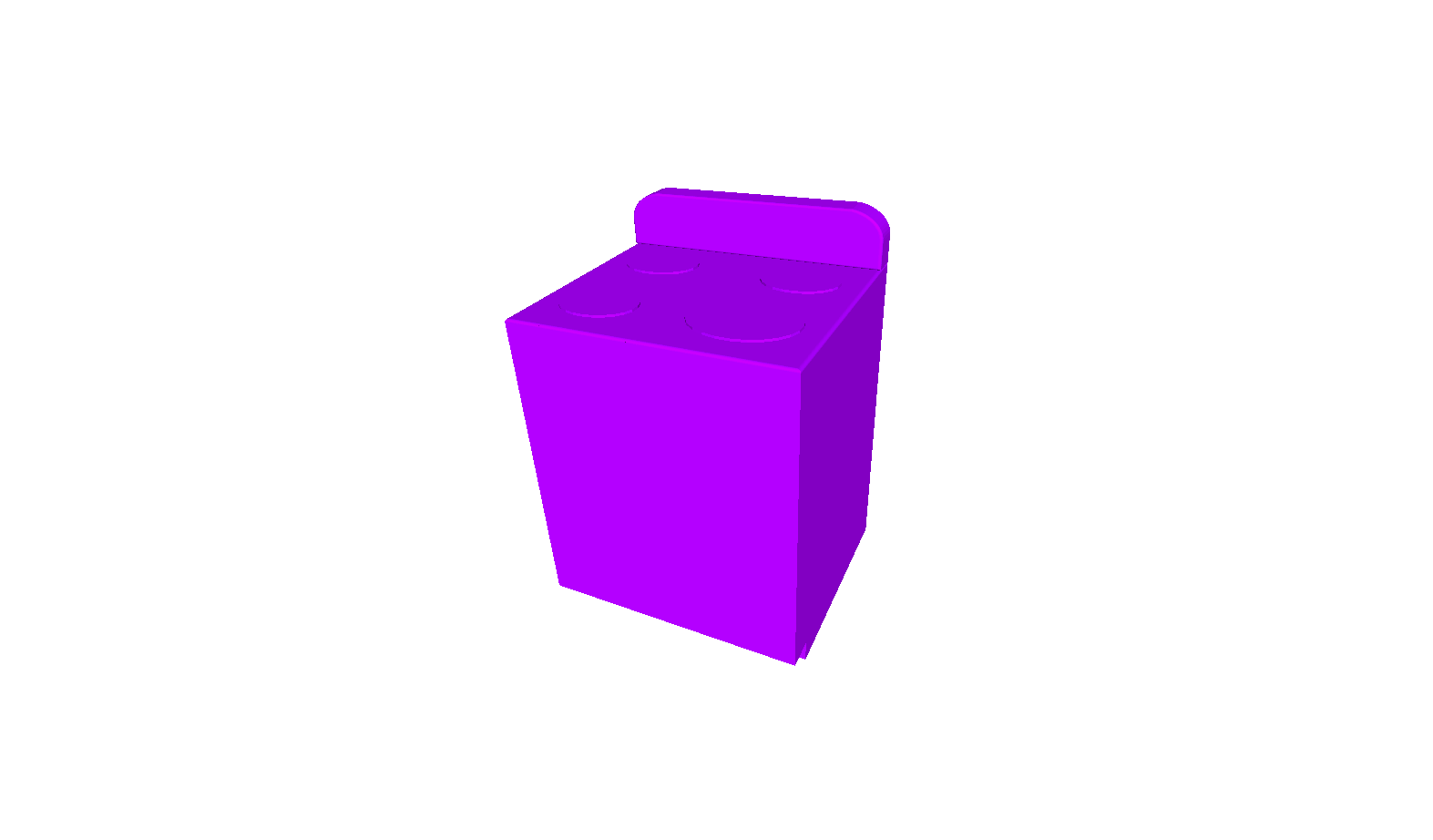} &
  \includegraphics[trim={13.5cm 3.5cm 15cm 4.5cm},clip,width=\widthtopfv\linewidth]{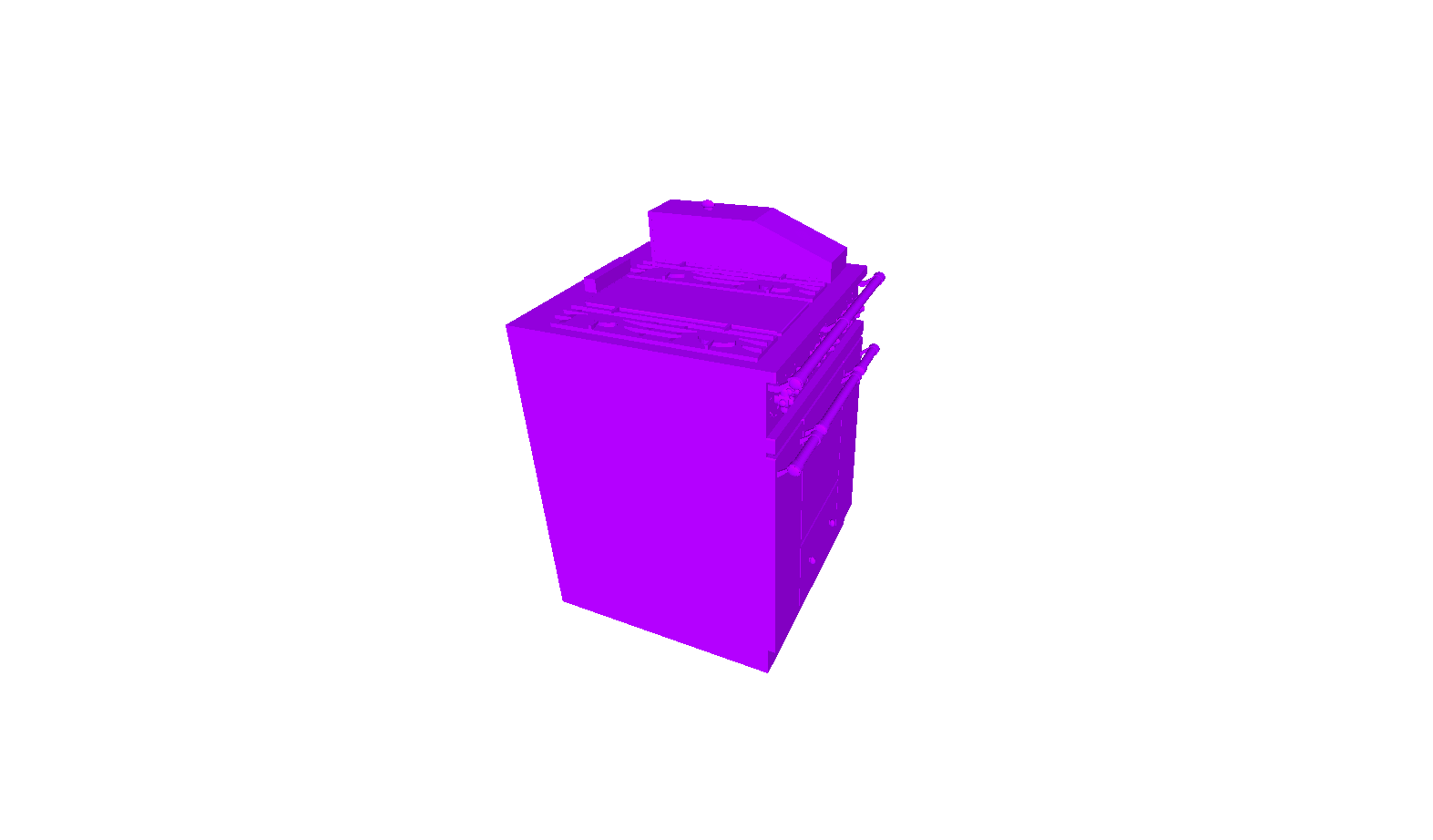} &
  \includegraphics[trim={13.5cm 3.5cm 15cm 4.5cm},clip,width=\widthtopfv\linewidth]{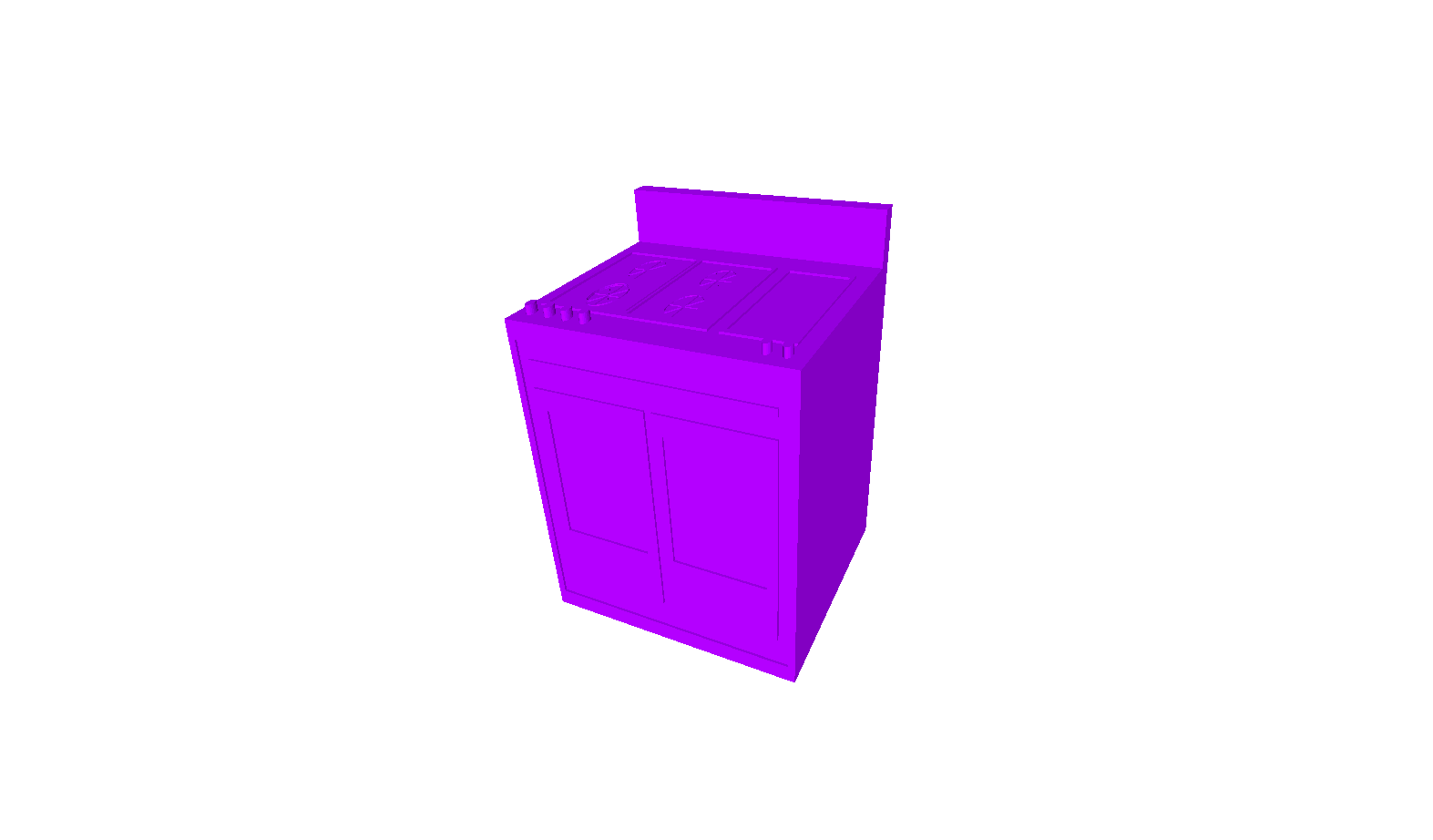} &
  \includegraphics[trim={13.5cm 3.5cm 15cm 4.5cm},clip,width=\widthtopfv\linewidth]{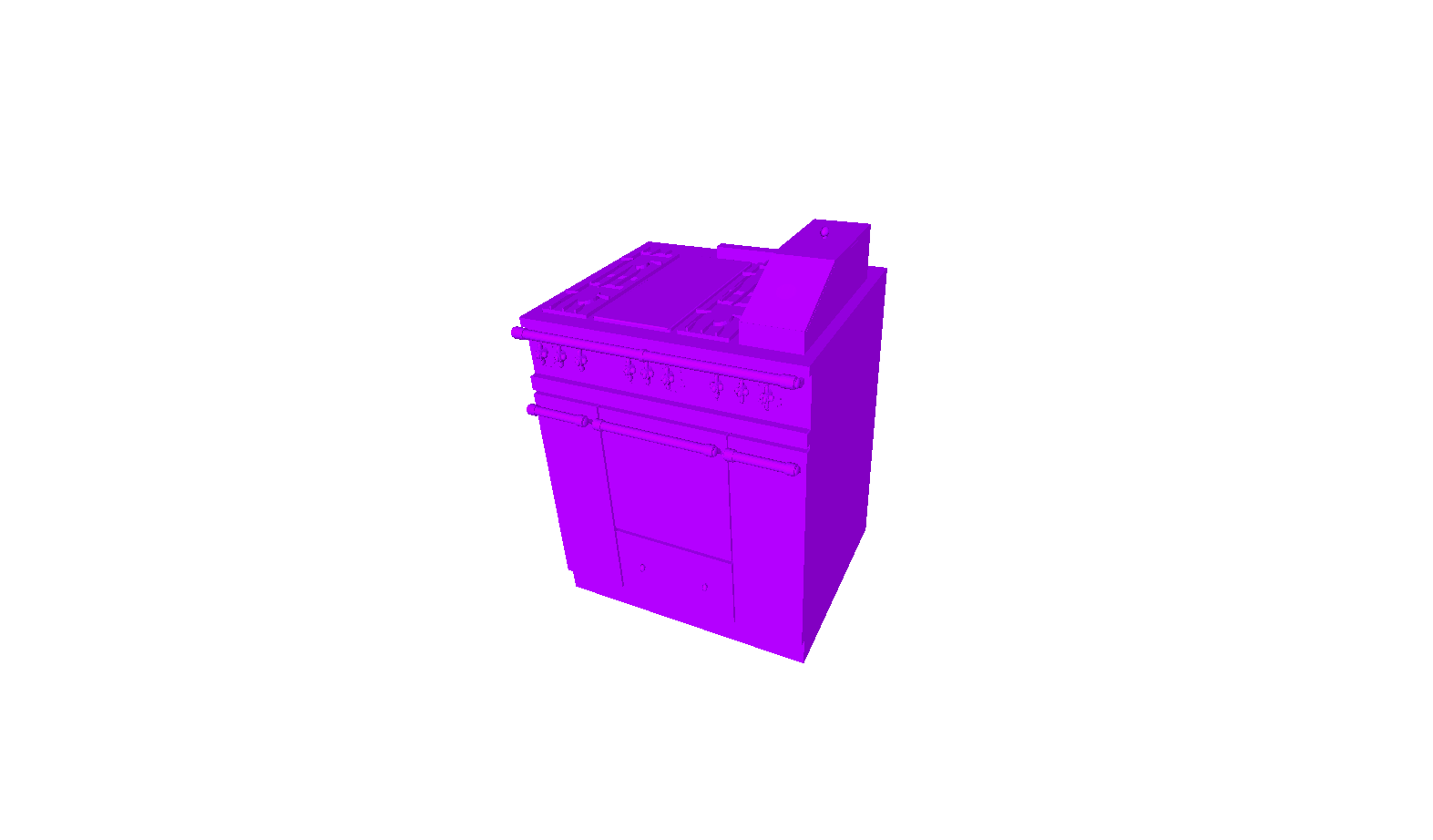} &
  \includegraphics[trim={13.5cm 3.5cm 15cm 4.5cm},clip,width=\widthtopfv\linewidth]{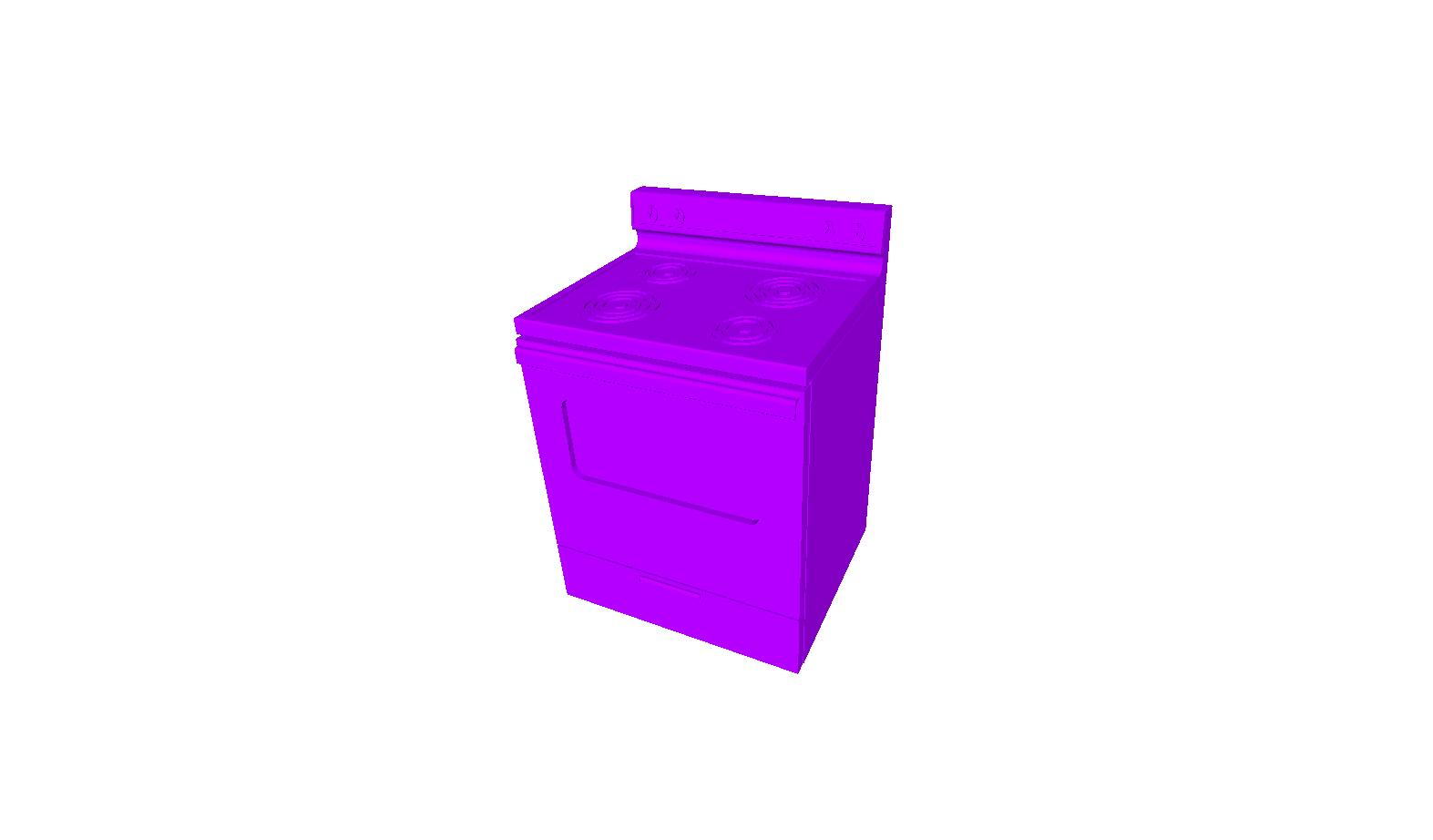} &
  \includegraphics[trim={13.5cm 3.5cm 15cm 4.5cm},clip,width=\widthtopfv\linewidth]{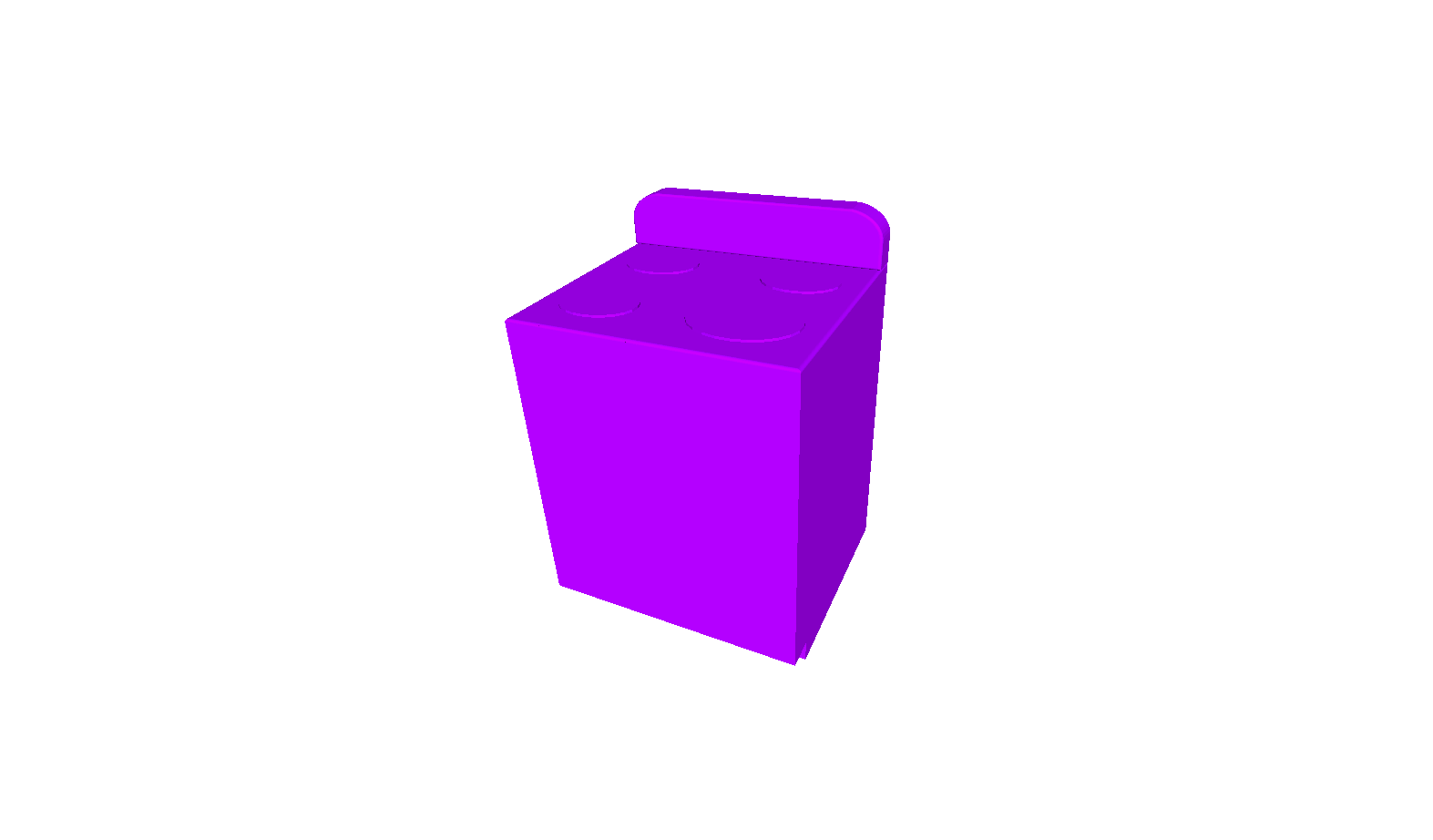} &

  %[trim={left bottom right top}

\textcolor{red}{\frame{\includegraphics[trim={13.5cm 3.5cm 15cm 4.5cm},clip,width=\widthtopfv\linewidth]{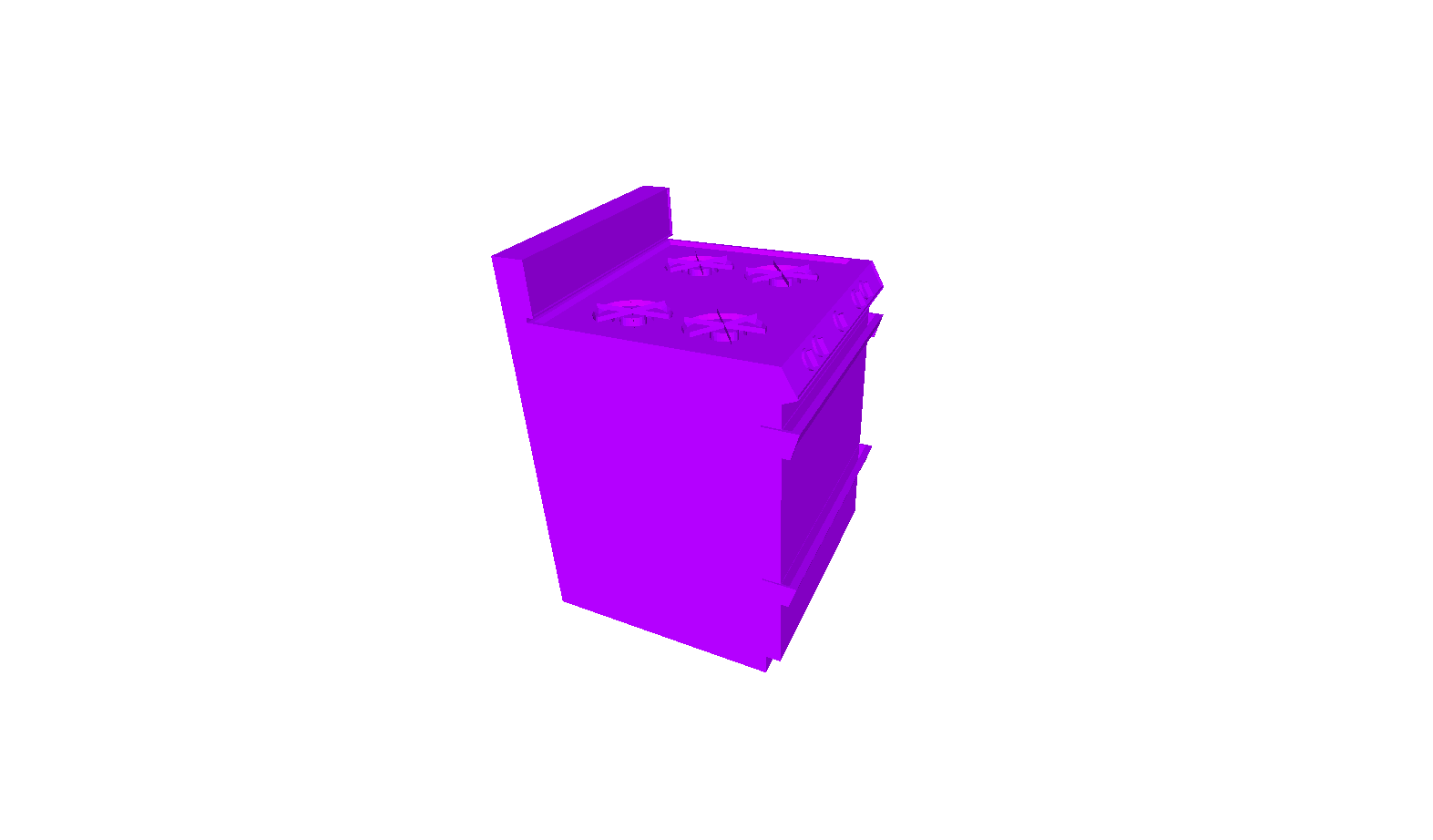}}}
  \\

    \includegraphics[trim={13.5cm 3.5cm 12cm 3.5cm},clip,width=\widthtopfv\linewidth]{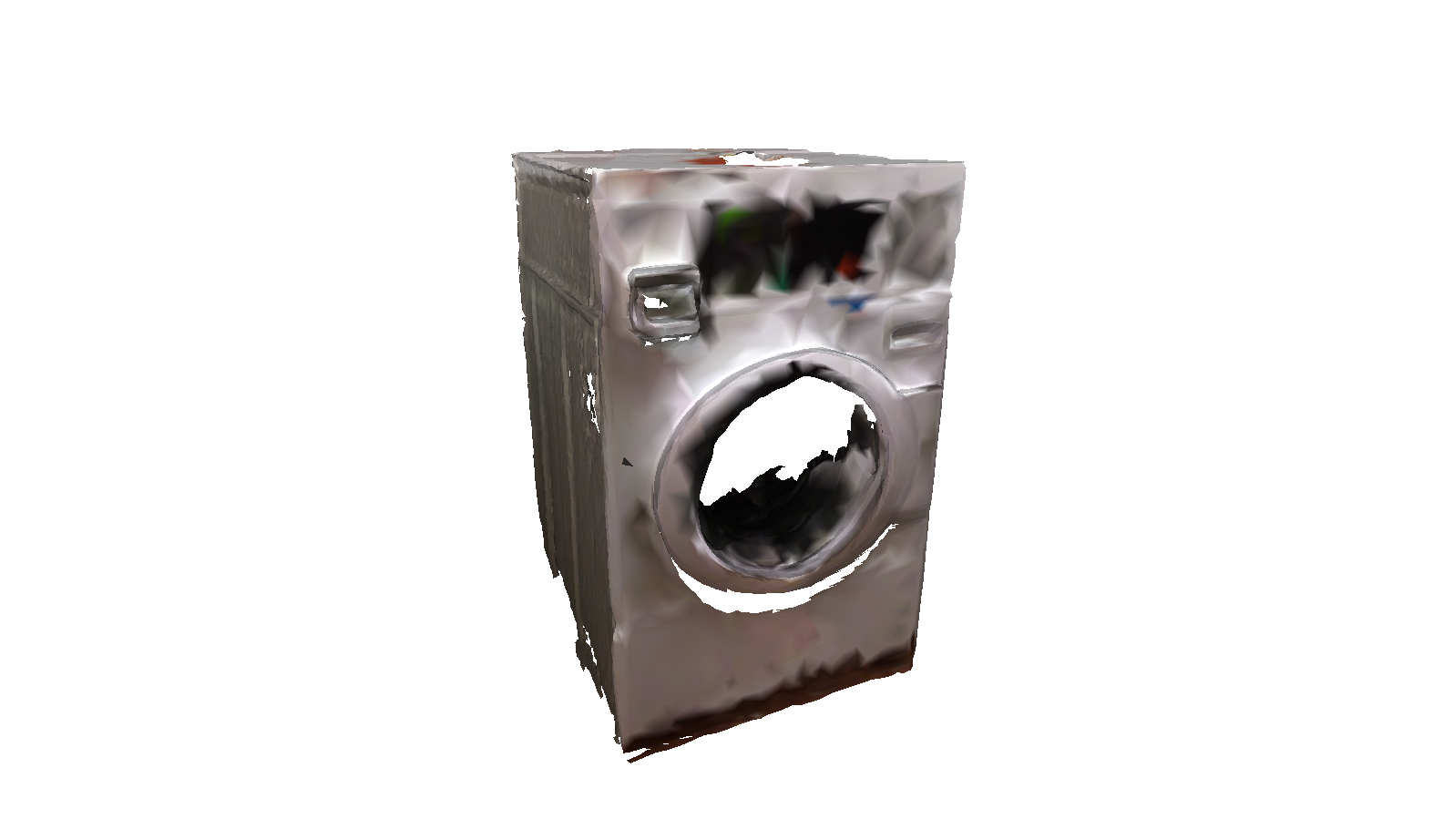} &
  \includegraphics[trim={13.5cm 3.5cm 12cm 3.5cm},clip,width=\widthtopfv\linewidth]{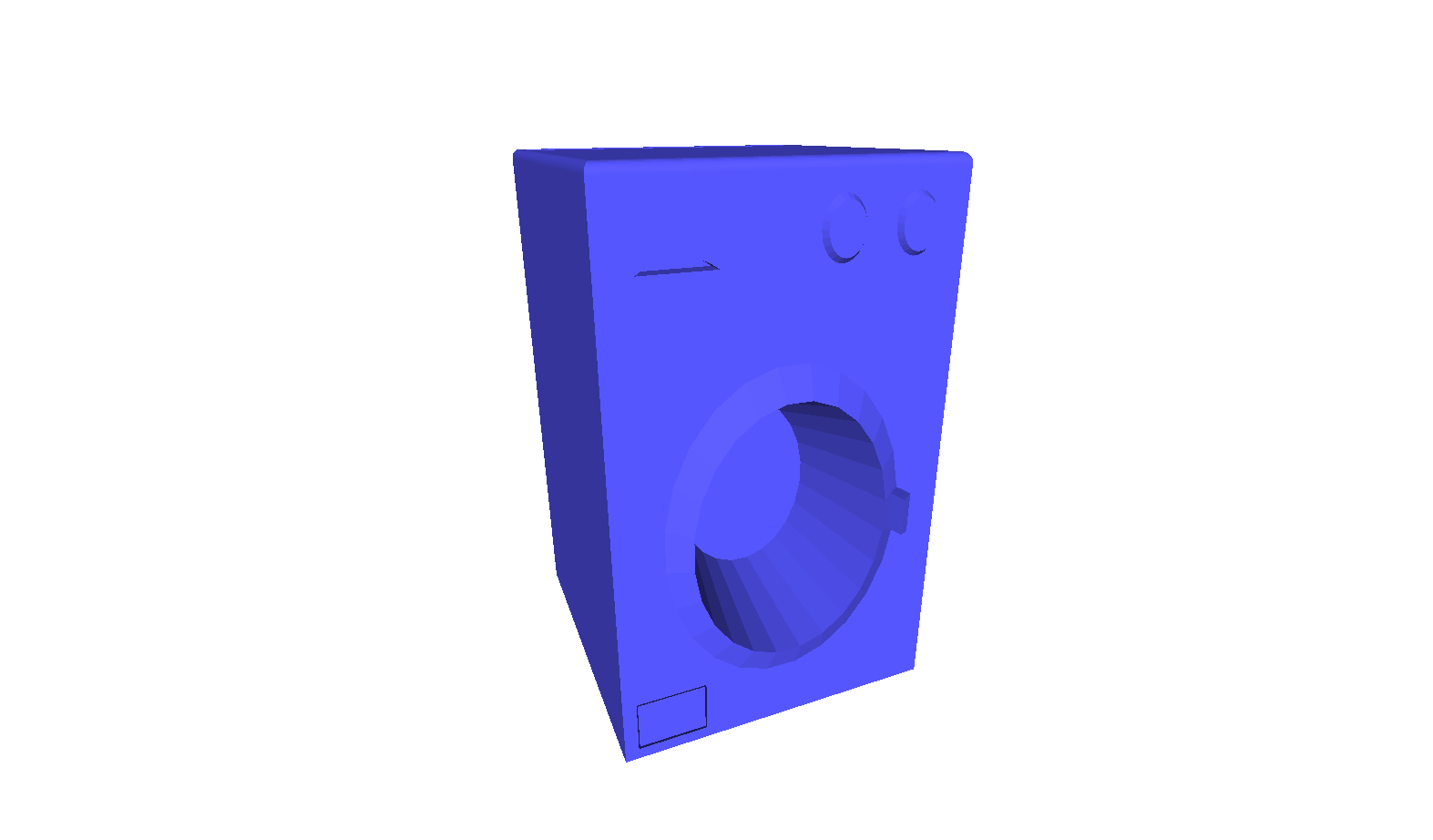} &  
  \includegraphics[trim={13.5cm 3.5cm 12cm 3.5cm},clip,width=\widthtopfv\linewidth]{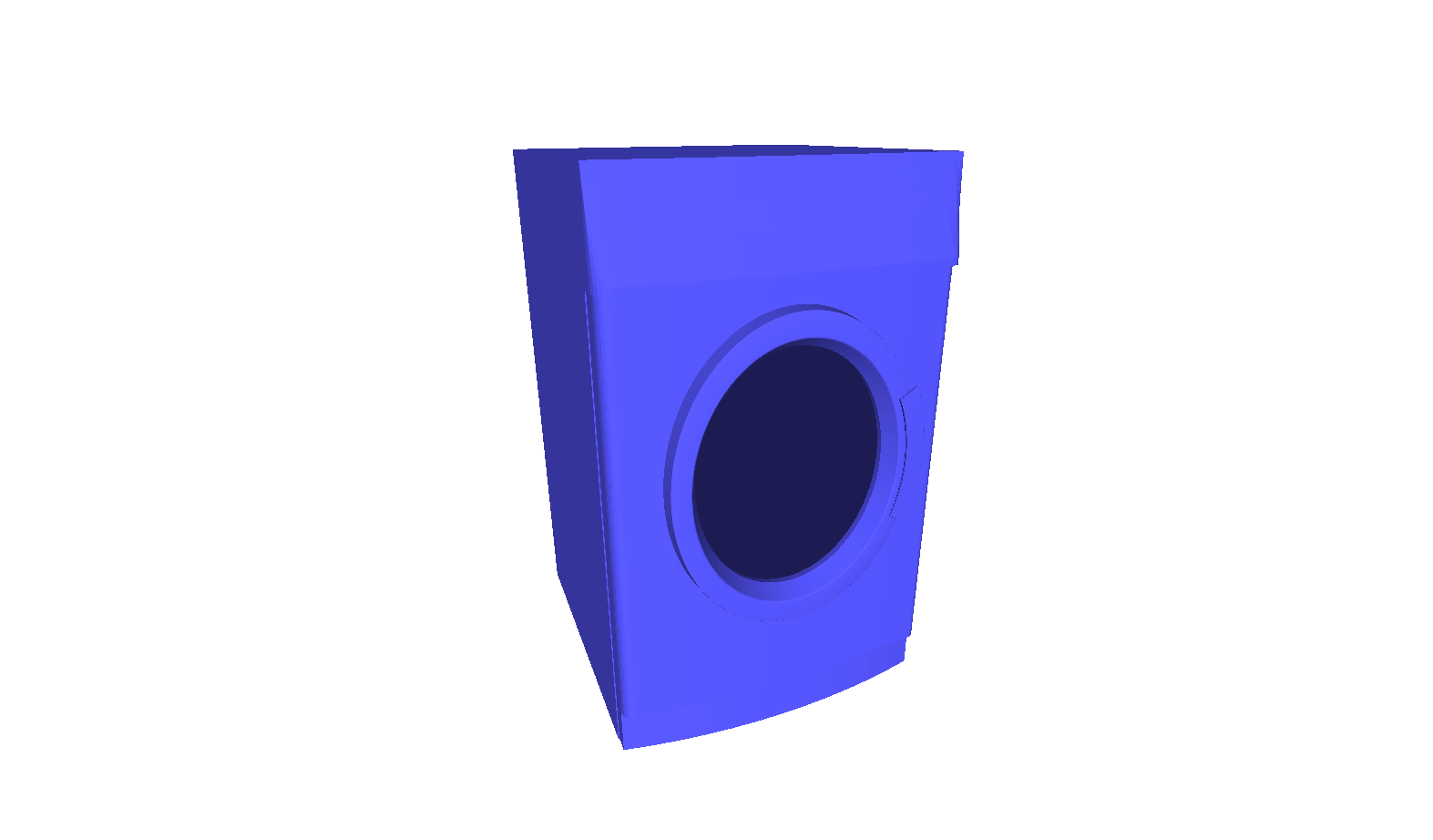} &
  \includegraphics[trim={13.5cm 3.5cm 12cm 3.5cm},clip,width=\widthtopfv\linewidth]{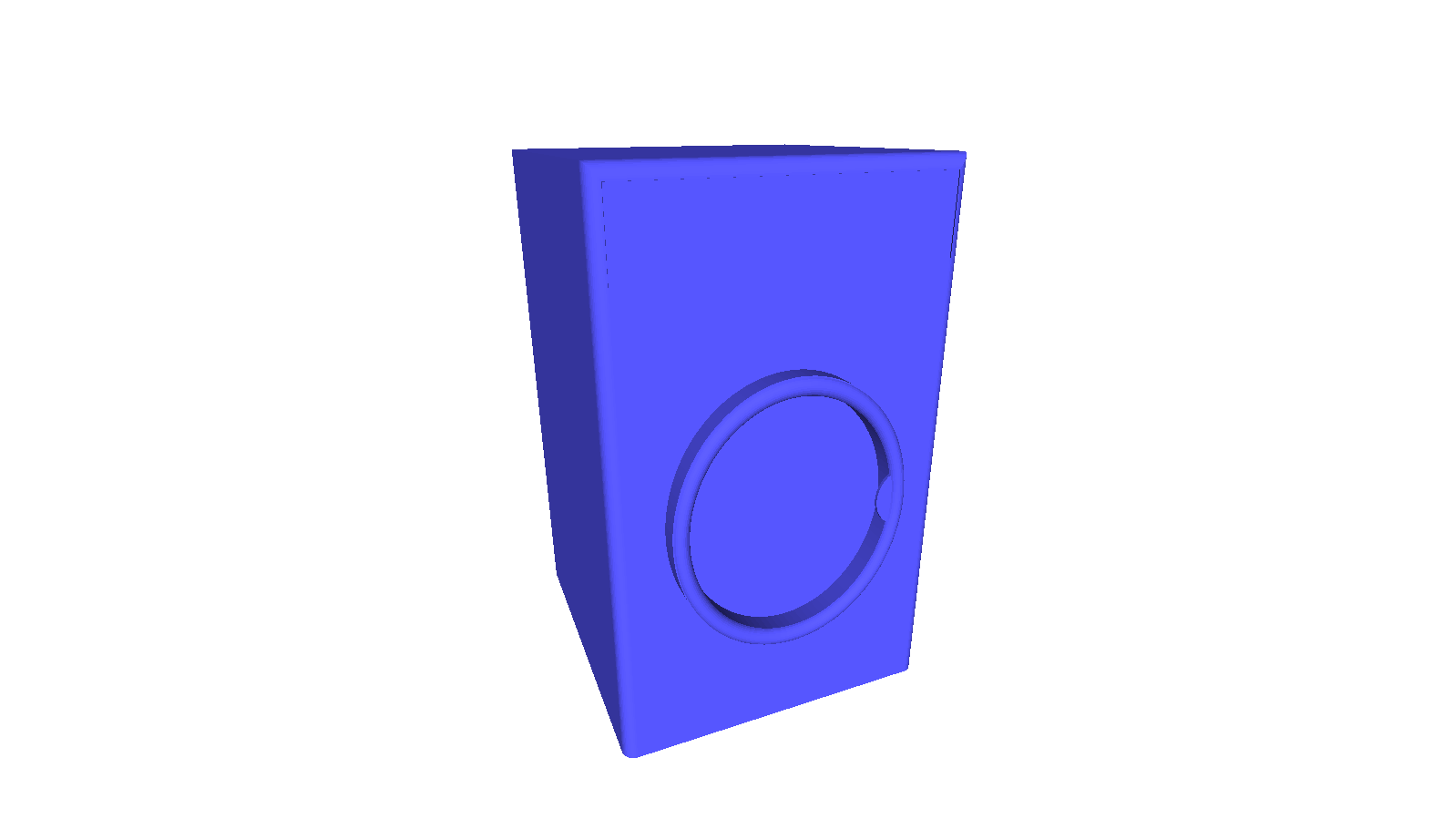} &
\textcolor{red}{\frame{\includegraphics[trim={13.5cm 3.5cm 12cm 3.5cm},clip,width=\widthtopfv\linewidth]{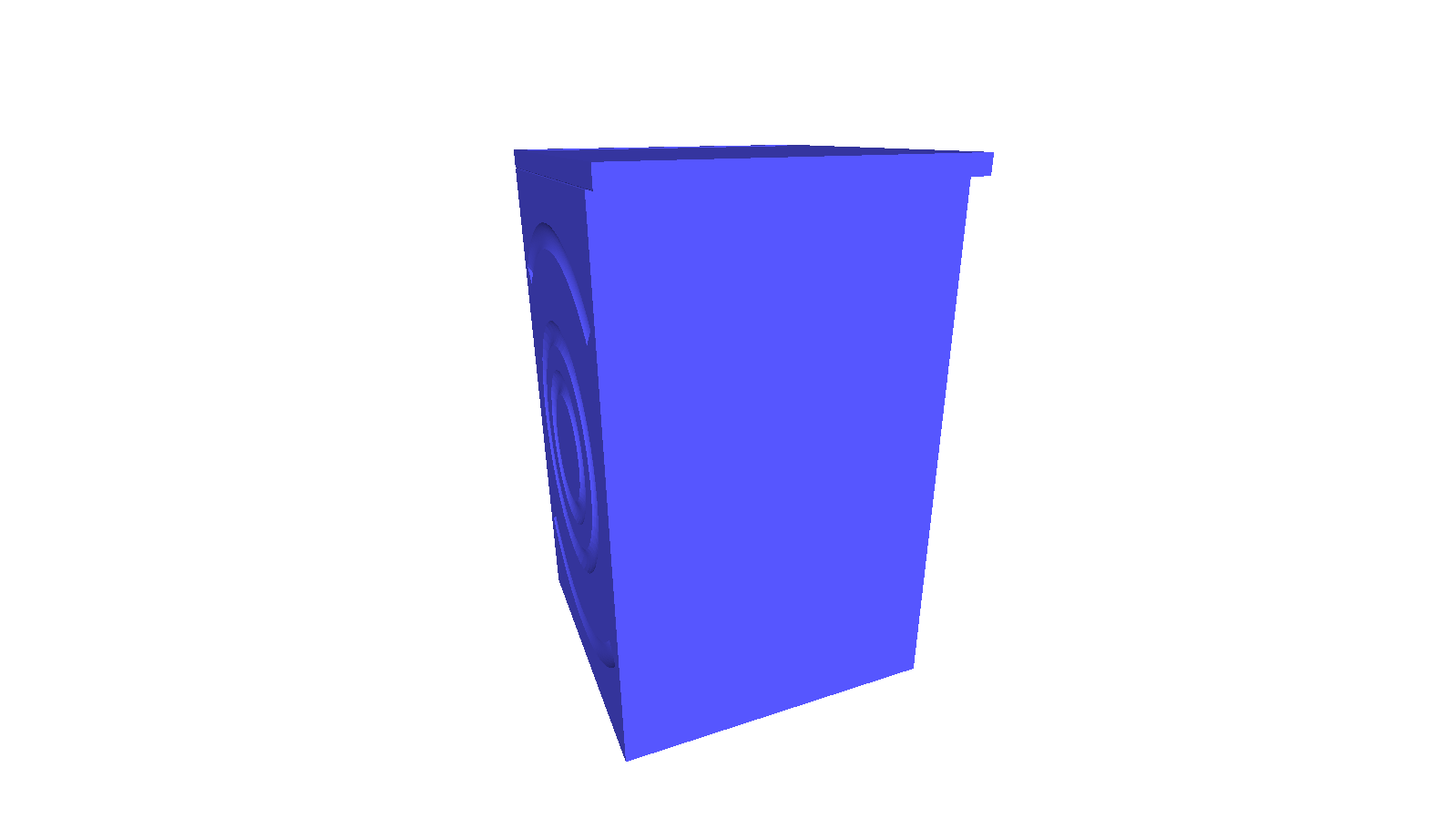}}} &
  \includegraphics[trim={13.5cm 3.5cm 12cm 3.5cm},clip,width=\widthtopfv\linewidth]{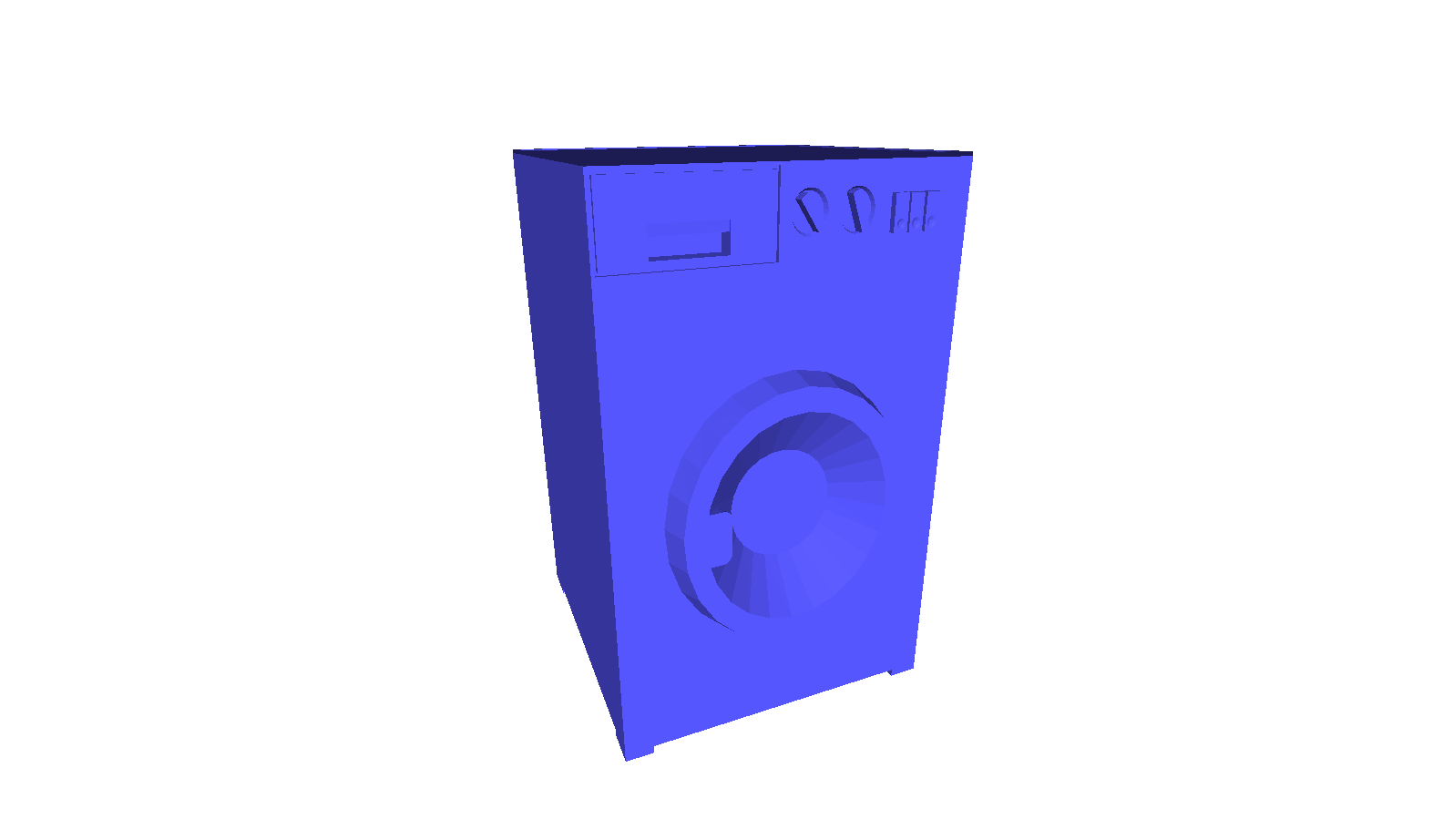} &
  \includegraphics[trim={13.5cm 3.5cm 12cm 3.5cm},clip,width=\widthtopfv\linewidth]{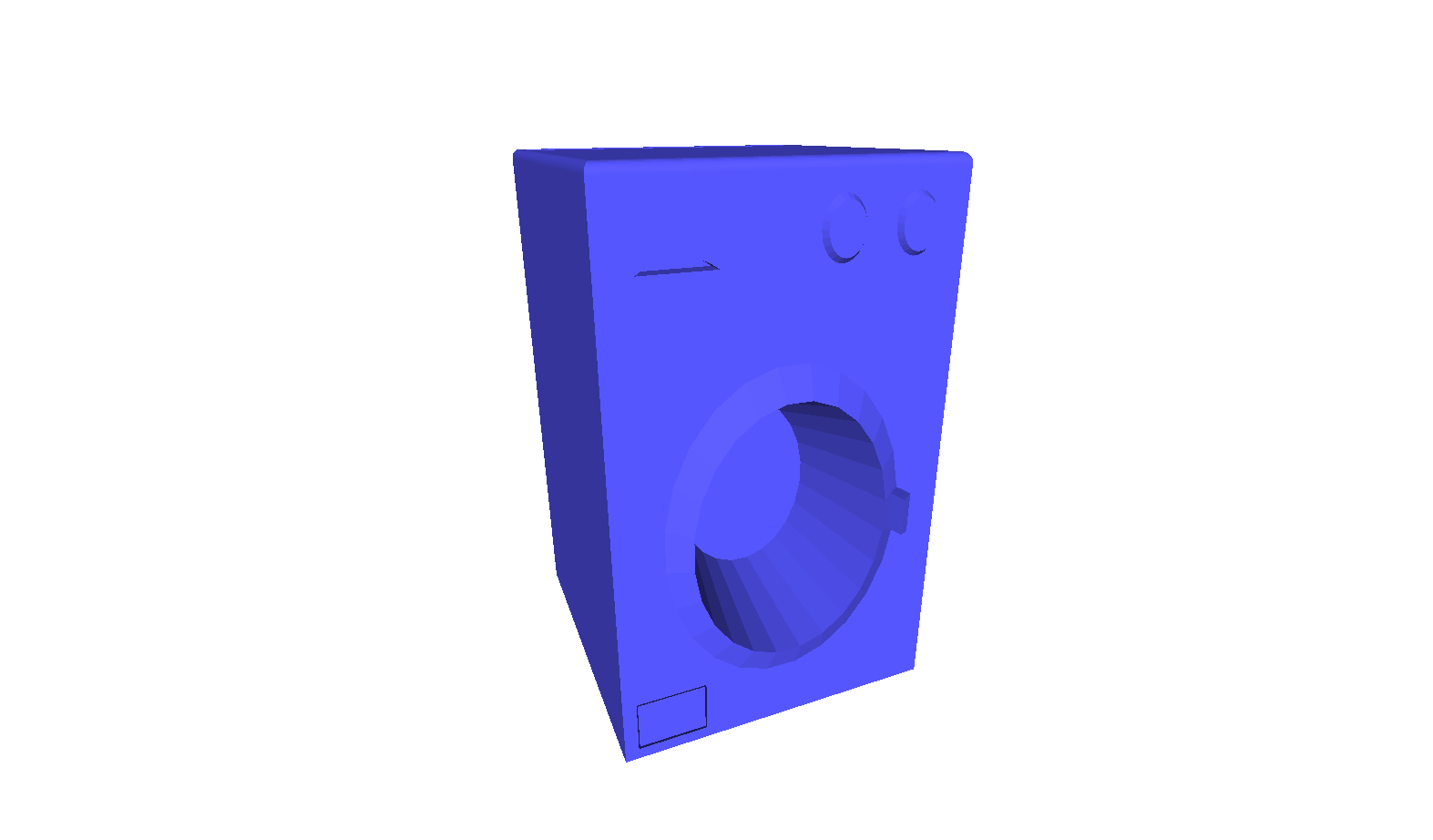} &
  \includegraphics[trim={13.5cm 3.5cm 12cm 3.5cm},clip,width=\widthtopfv\linewidth]{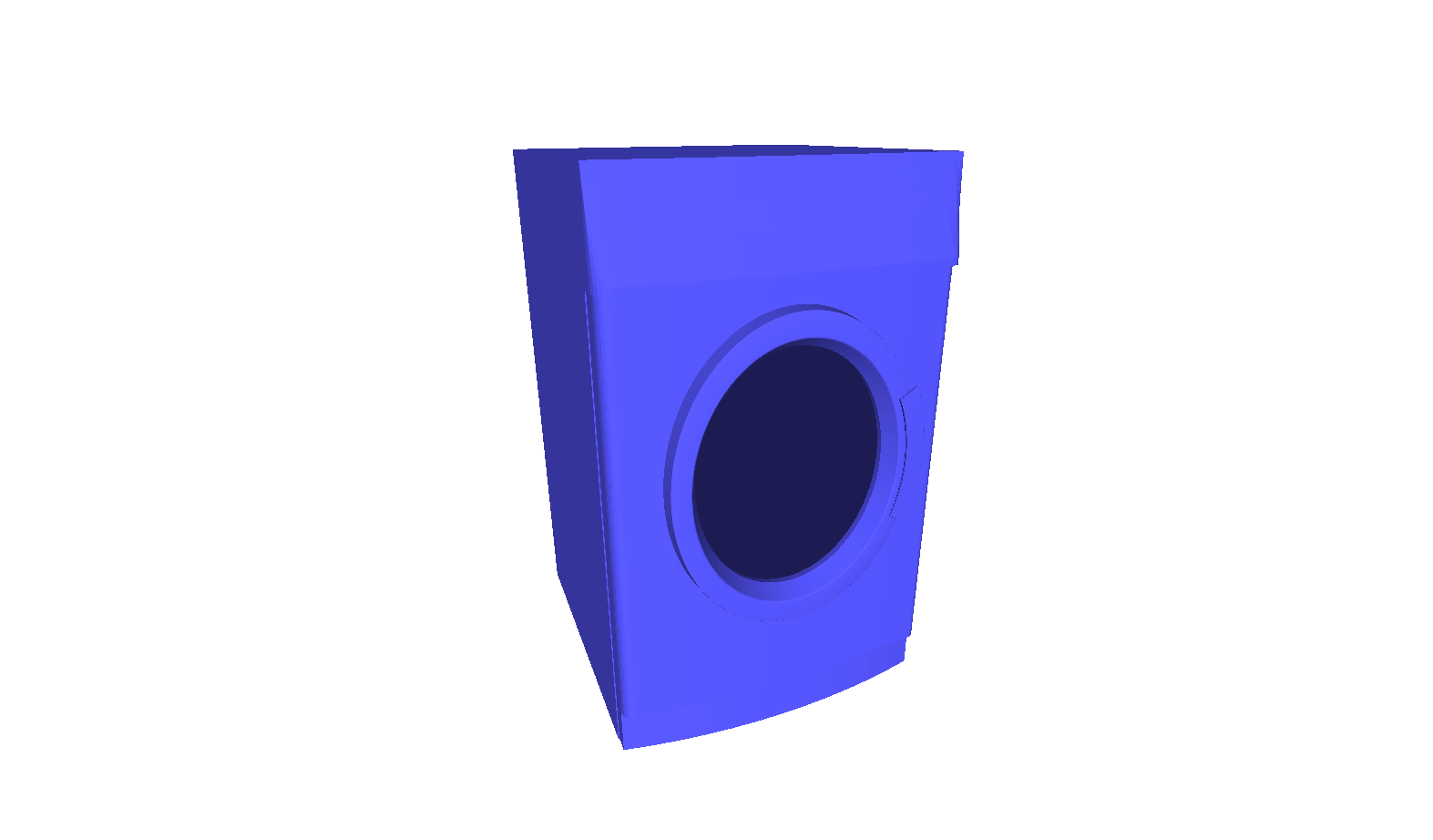} &
  \includegraphics[trim={13.5cm 3.5cm 12cm 3.5cm},clip,width=\widthtopfv\linewidth]{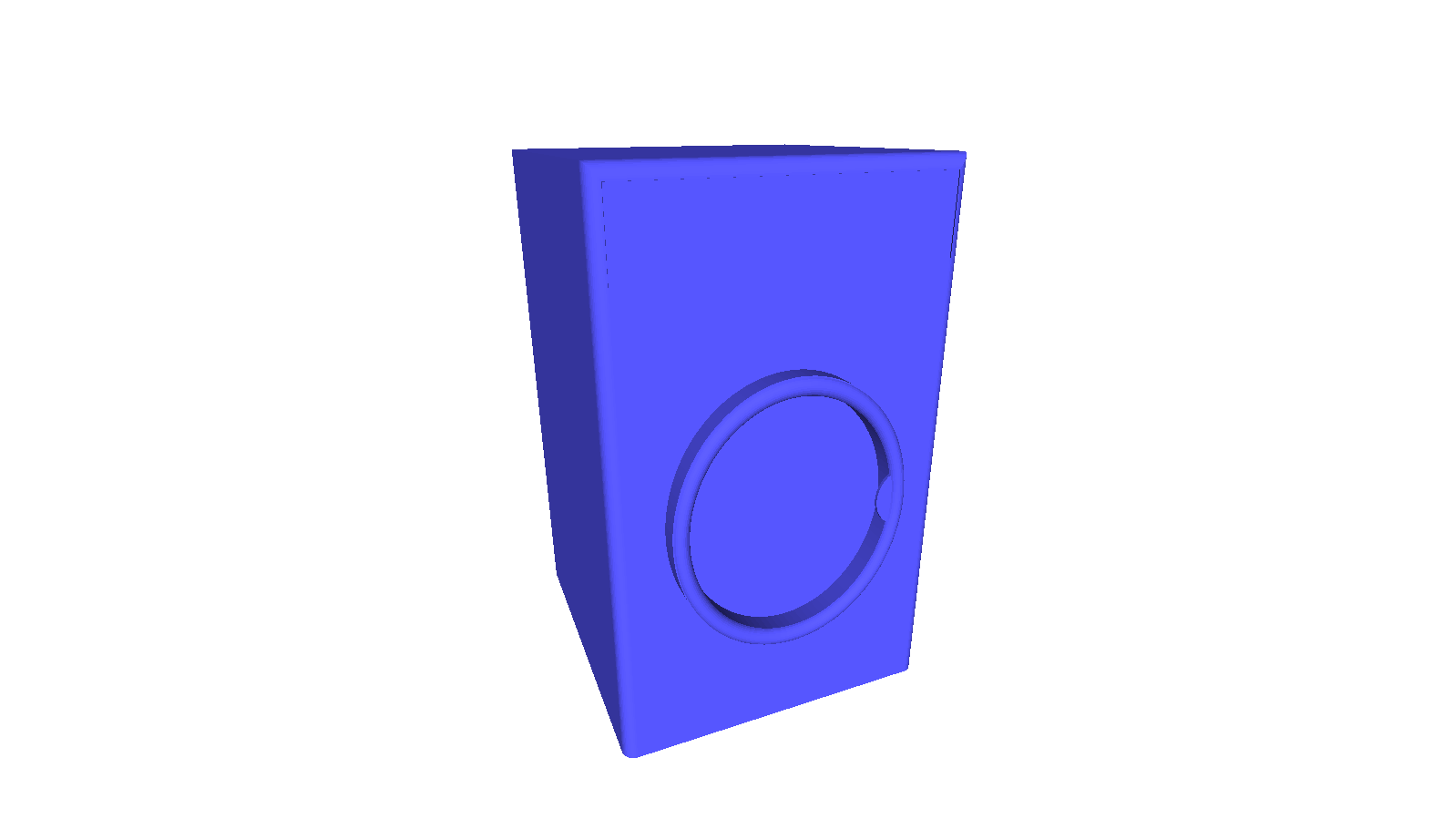} &
\textcolor{red}{\frame{\includegraphics[trim={13.5cm 3.5cm 12cm 3.5cm},clip,width=\widthtopfv\linewidth]{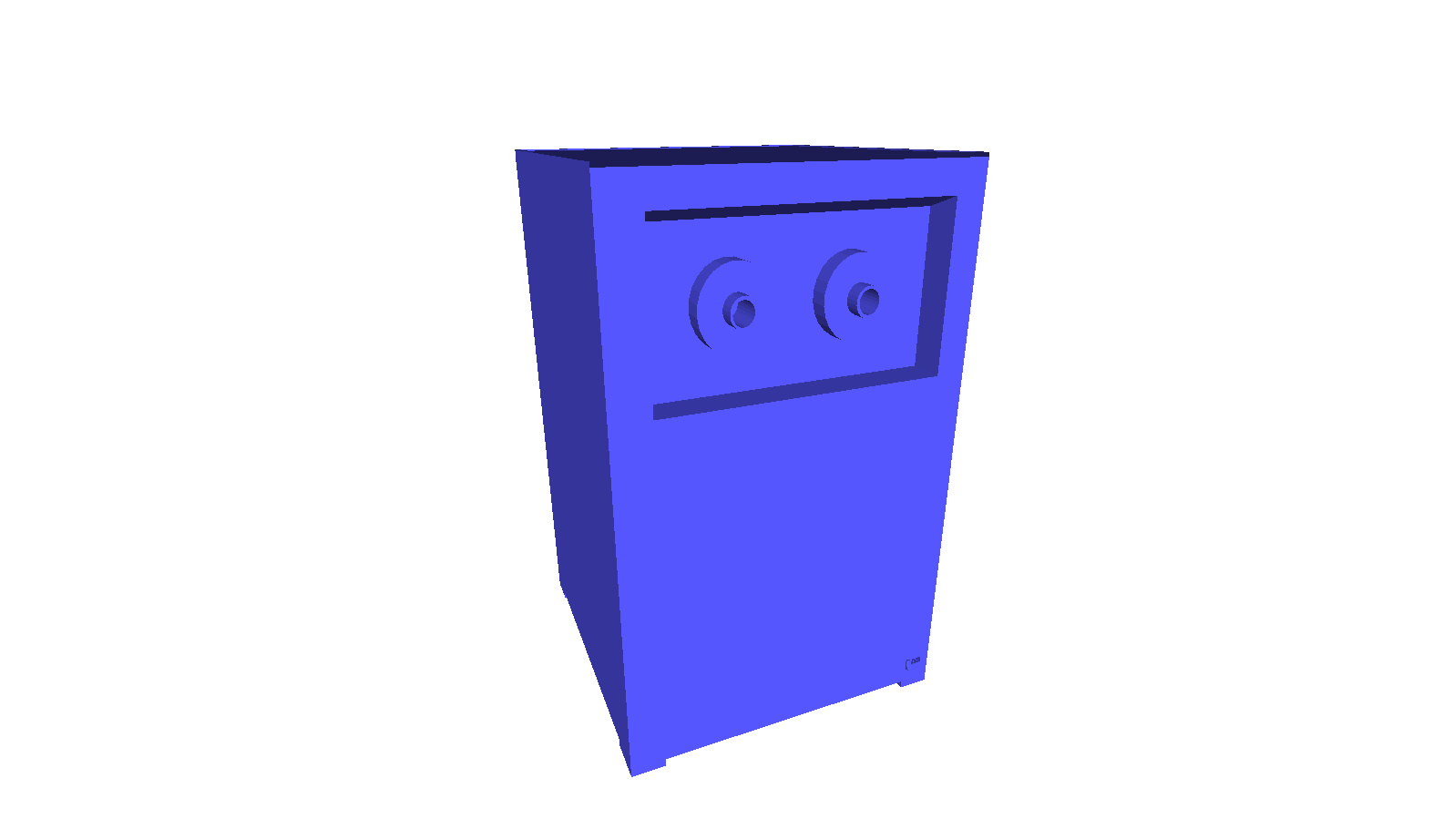}}} &
\includegraphics[trim={13.5cm 3.5cm 12cm 3.5cm},clip,width=\widthtopfv\linewidth]{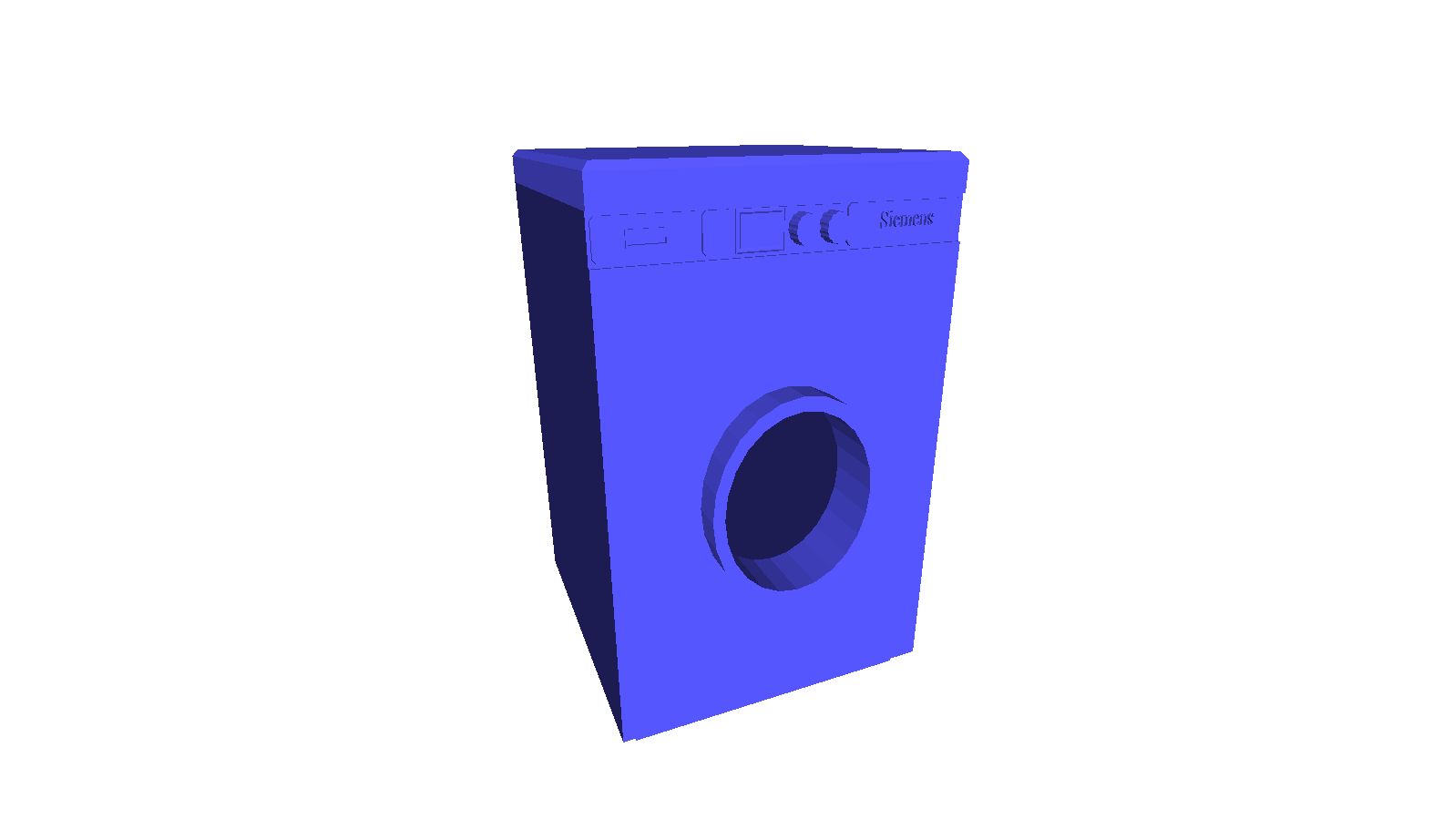}
  \\

  Target object & 
  \multicolumn{5}{c|}{Top-5 candidates, exhaustive search} &
  \multicolumn{5}{c}{Top-5 candidates, HOC-Search (ours)} \\  
  % object & &&&& & &&&&\\
\end{tabular}
}
% \vspace{-0.2cm}
\caption{Additional visualizations of top-5 candidates (in descending order) from exhaustive search compared to HOC-Search for 800 iterations. \textbf{Top two rows:} The render-and-compare objective function suffers from highly incomplete scans. Retrieved CAD models that fit the target object are highlighted in green. \textbf{Bottom two rows:} The prediction of the correct orientation for objects with a high degree of symmetry is difficult (objects with wrong orientation highlighted in red). This failure occurs for classes where most objects have box-like shapes, for example Stove, Washer or Cabinet.}
    \label{fig:supp_top5_fail}
\end{figure*}

\section{Additional Qualitative Results for Experiment: Automatic CAD Model and Pose Retrieval}
\label{sec:supp_softroup_exp}

\begin{figure*}
\centering
\scalebox{0.9}{
\begin{tabular}{cccc}
%[trim={left bottom right top}
 \includegraphics[trim={2cm 1cm 1cm 1.75cm},clip,width=0.25\linewidth]{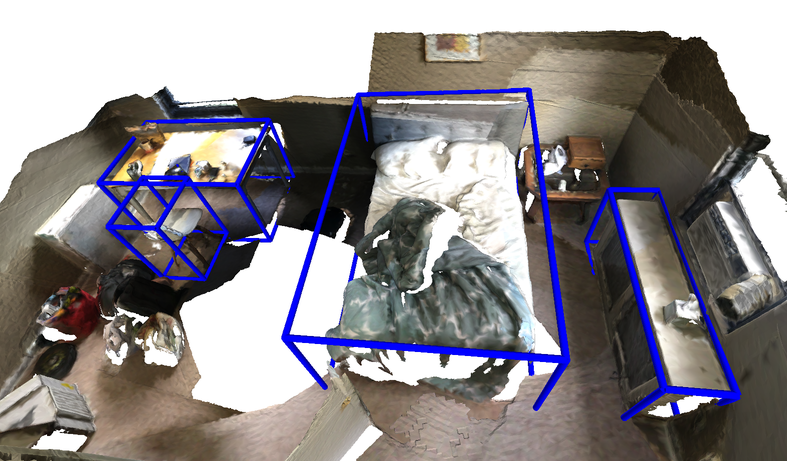} &
  \includegraphics[trim={2cm 1cm 1cm 1.75cm},clip,width=0.25\linewidth]{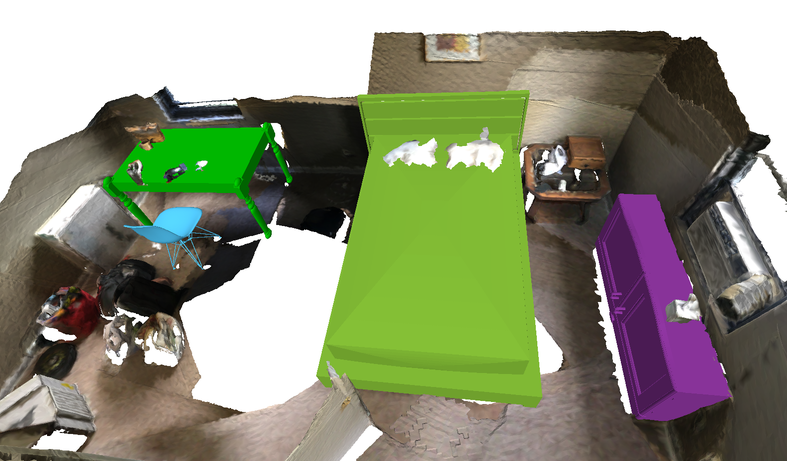} &
  \includegraphics[trim={2cm 1cm 1cm 1.75cm},clip,width=0.25\linewidth]{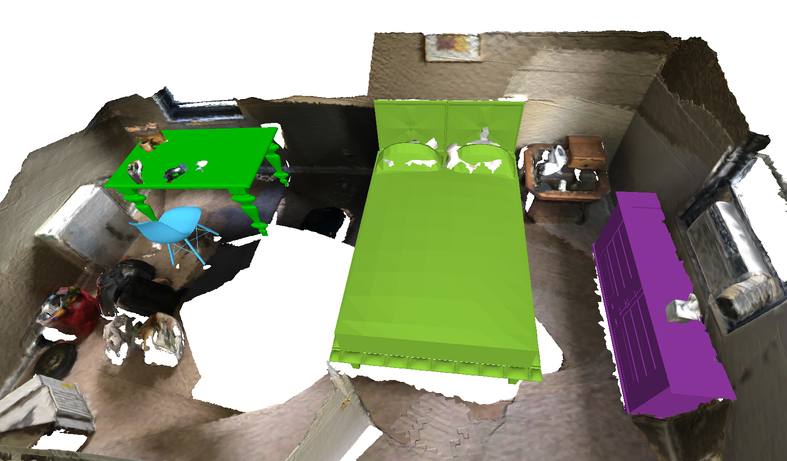} &
  \includegraphics[trim={2cm 1cm 1cm 1.75cm},clip,width=0.25\linewidth]{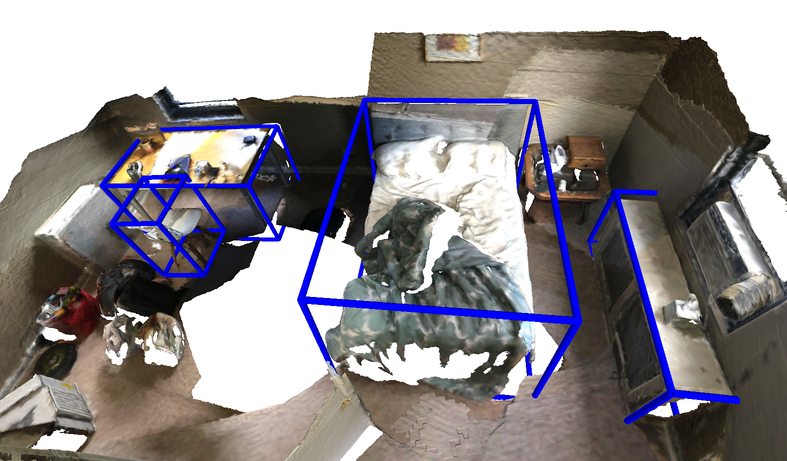} \\

     \includegraphics[trim={4.5cm .5cm 3cm 2.5cm},clip,width=0.25\linewidth]{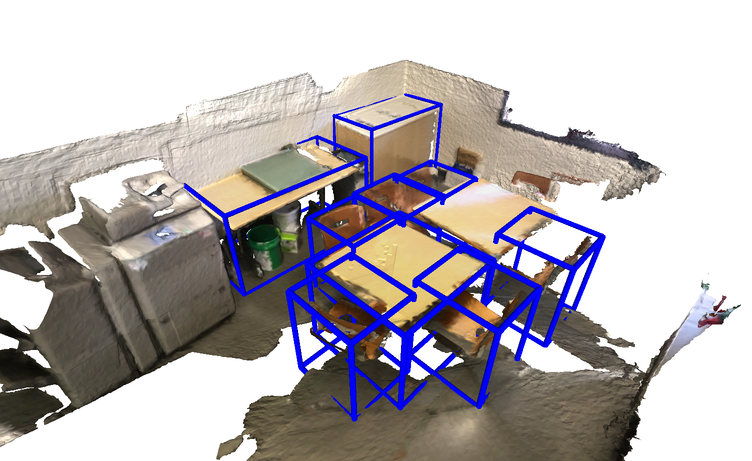} &
  \includegraphics[trim={4.5cm .5cm 3cm 2.5cm},clip,width=0.25\linewidth]{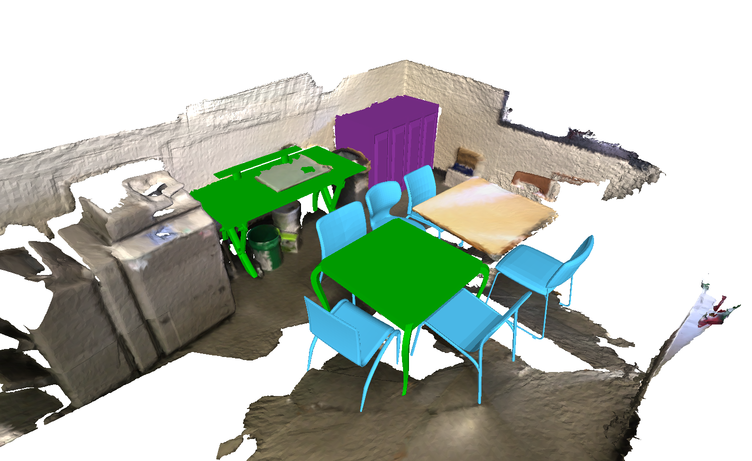} &
  \includegraphics[trim={4.5cm .5cm 3cm 2.5cm},clip,width=0.25\linewidth]{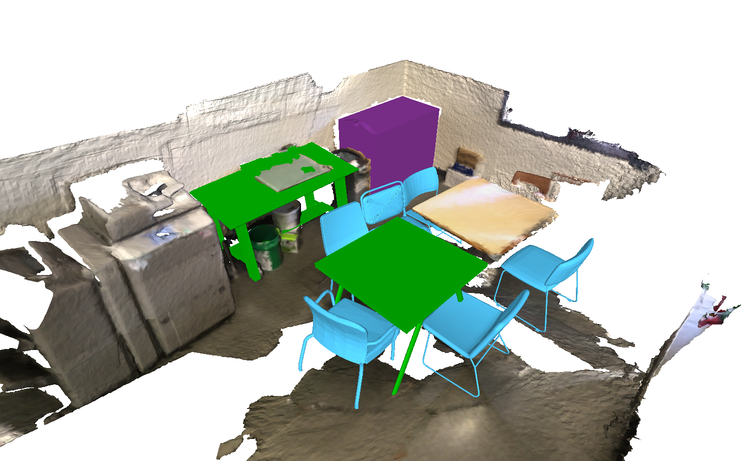} &
  \includegraphics[trim={4.5cm .5cm 3cm 2.5cm},clip,width=0.25\linewidth]{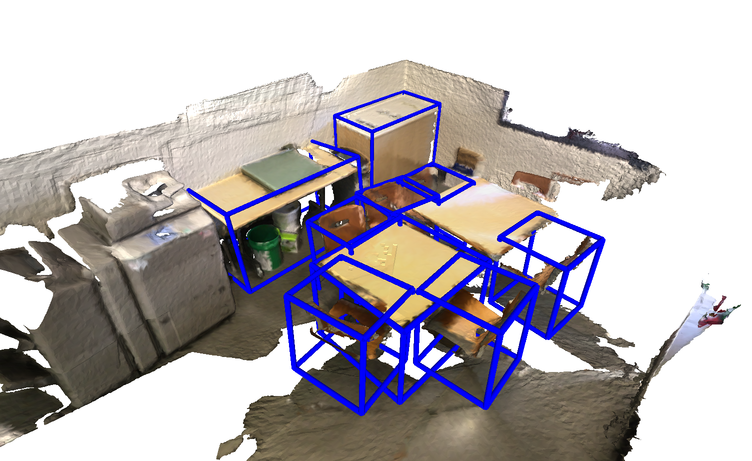} \\

     \includegraphics[trim={1cm 1cm 3cm 2.5cm},clip,width=0.25\linewidth]{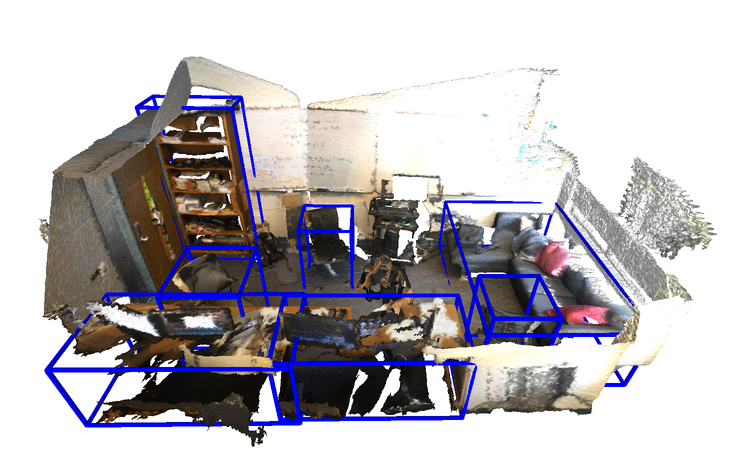} &
  \includegraphics[trim={1cm 1cm 3cm 2.5cm},clip,width=0.25\linewidth]{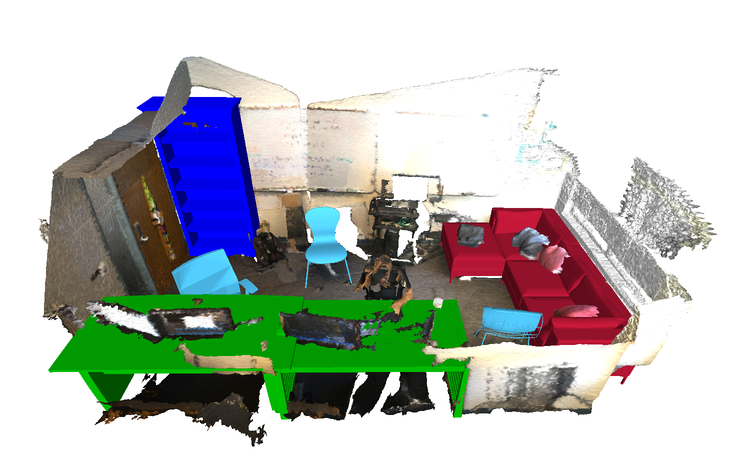} &
  \includegraphics[trim={1cm 1cm 3cm 2.5cm},clip,width=0.25\linewidth]{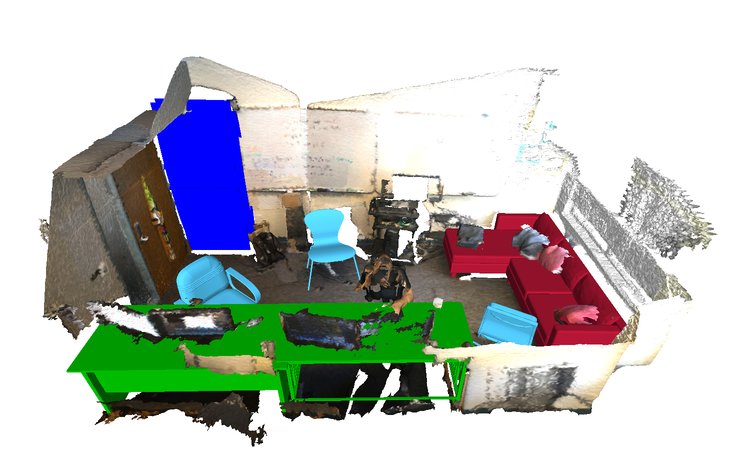} &
  \includegraphics[trim={1cm 1cm 3cm 2.5cm},clip,width=0.25\linewidth]{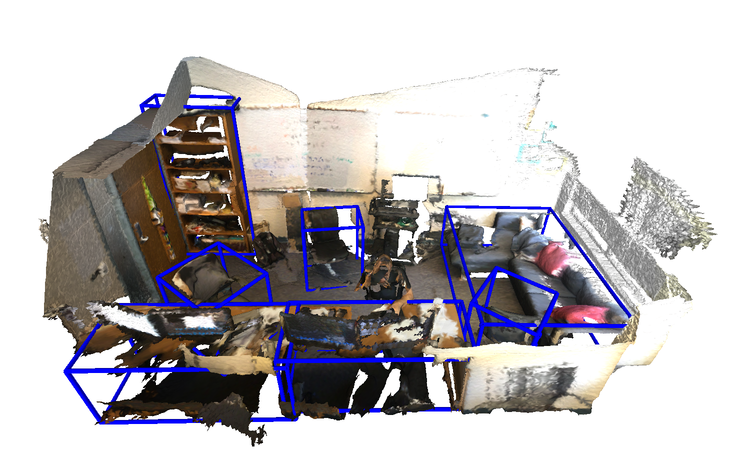} \\

       \includegraphics[trim={1cm 2.5cm 2cm 1cm},clip,width=0.25\linewidth]{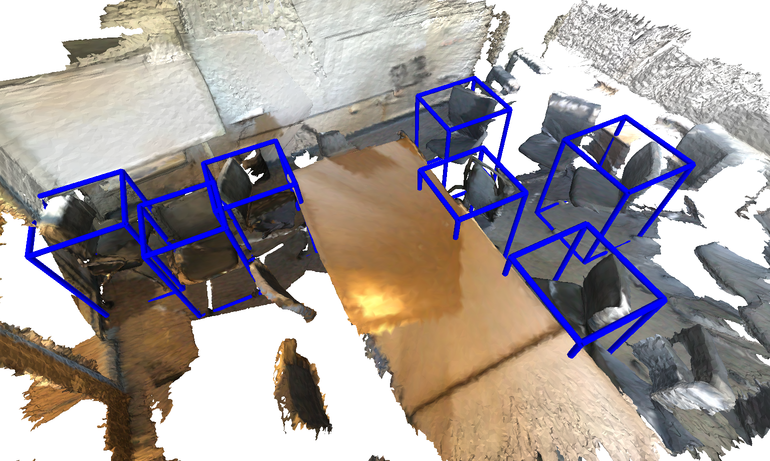} &
  \includegraphics[trim={1cm 2.5cm 2cm 1cm},clip,width=0.25\linewidth]{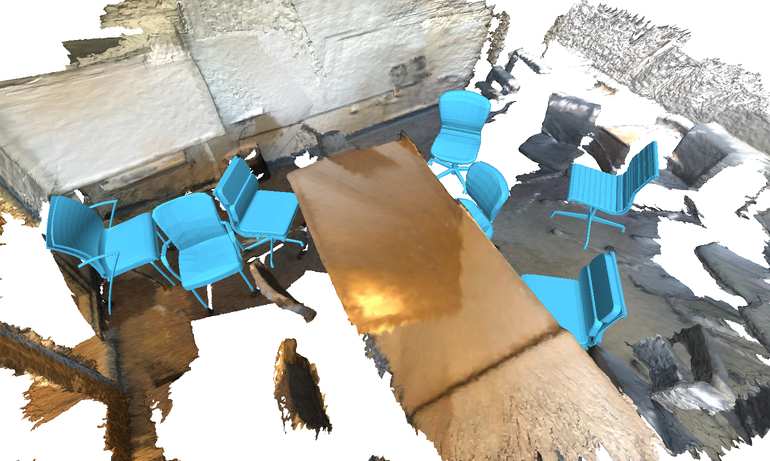} &
  \includegraphics[trim={1cm 2.5cm 2cm 1cm},clip,width=0.25\linewidth]{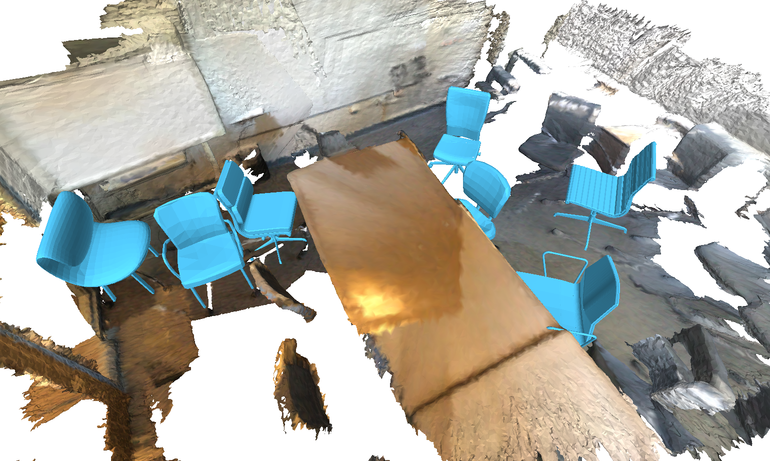} &
  \includegraphics[trim={1cm 2.5cm 2cm 1cm},clip,width=0.25\linewidth]{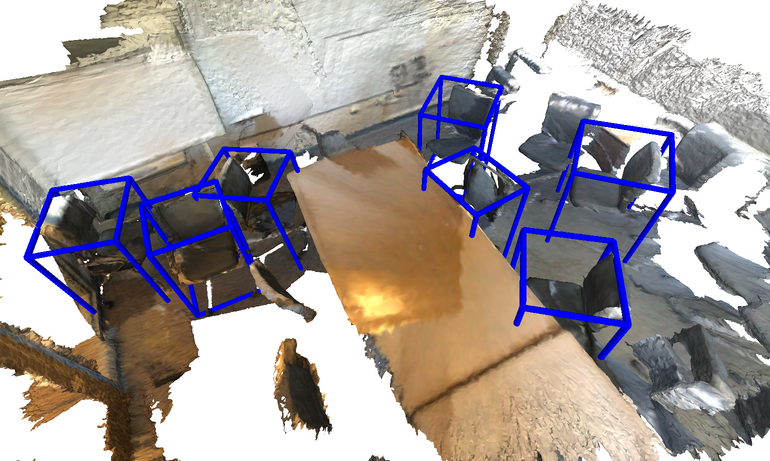} \\

       \includegraphics[trim={2.5cm 1cm 1.5cm 1.25cm},clip,width=0.25\linewidth]{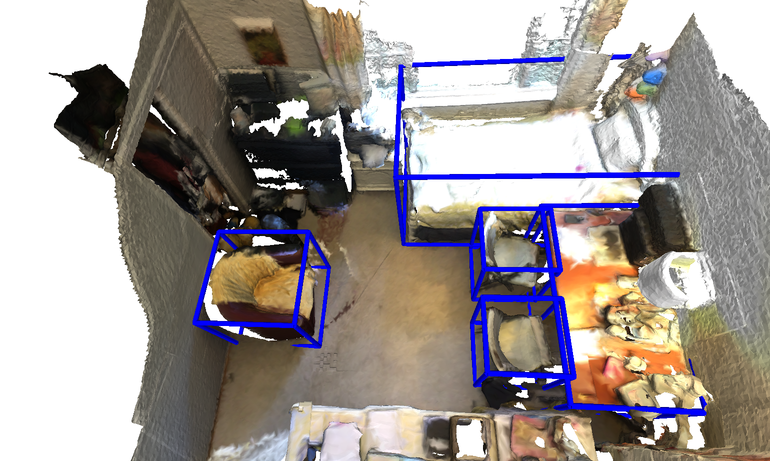} &
  \includegraphics[trim={2.5cm 1cm 1.5cm 1.25cm},clip,width=0.25\linewidth]{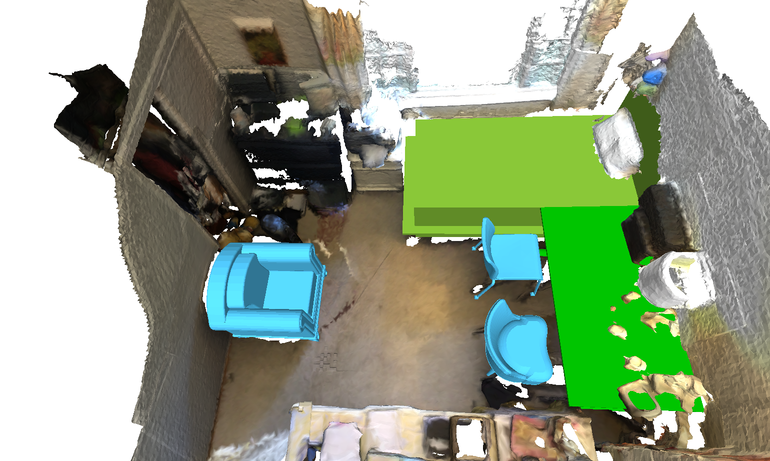} &
  \includegraphics[trim={2.5cm 1cm 1.5cm 1.25cm},clip,width=0.25\linewidth]{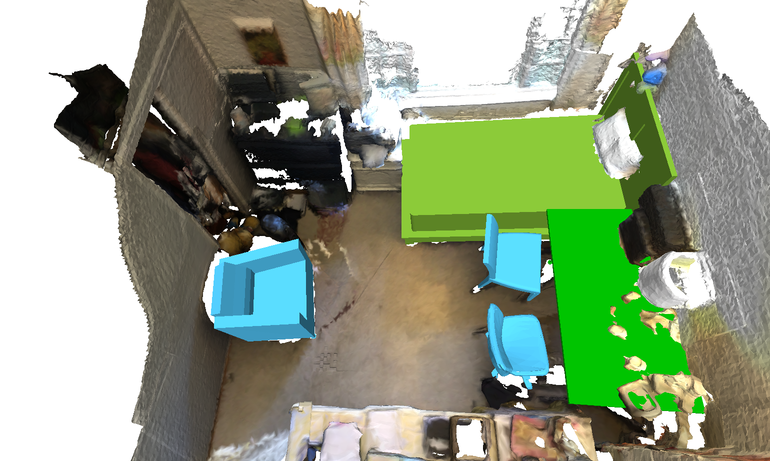} &
  \includegraphics[trim={2.5cm 1cm 1.5cm 1.25cm},clip,width=0.25\linewidth]{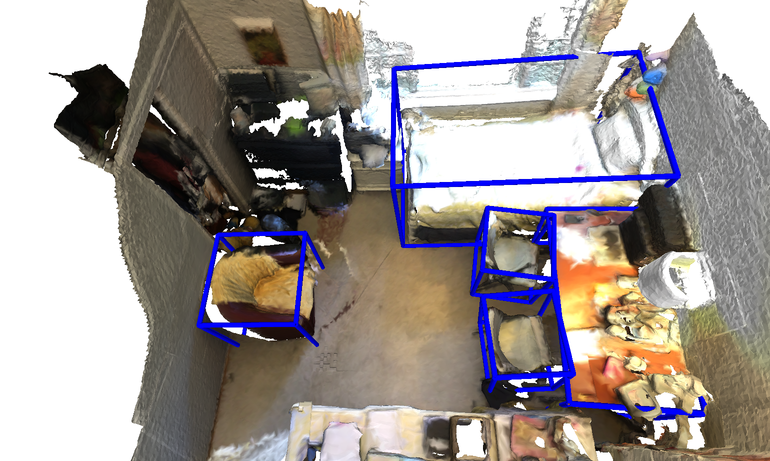} \\
  
  RGB-D Scan with &  CAD models retrieved & CAD models retrieved  &RGB-D scan with \\
   initial 3D boxes         &   with exh. search   &      with HOC-Search    & refined 3D boxes\\
      &     &      + refinement   & \\
\end{tabular}
}
\caption{Additional qualitative results of exhaustive search and HOC-Search with refinement for CAD model retrieval using SoftGroup~\cite{Vu_2022_CVPR} predictions. HOC-Search with refinement retrieves accurate CAD models, where the refined pose of the objects is often more accurate compared to the initial pose thanks to our simultaneous retrieval and pose refinement.}
\label{fig:supp_softgroup_results}
\end{figure*}

\begin{figure*}
\centering
\scalebox{0.95}{
\begin{tabular}{ccc}
%[trim={left bottom right top}

\includegraphics[trim={3cm 0.5cm 3cm 4.5cm},clip,width=0.33\linewidth]{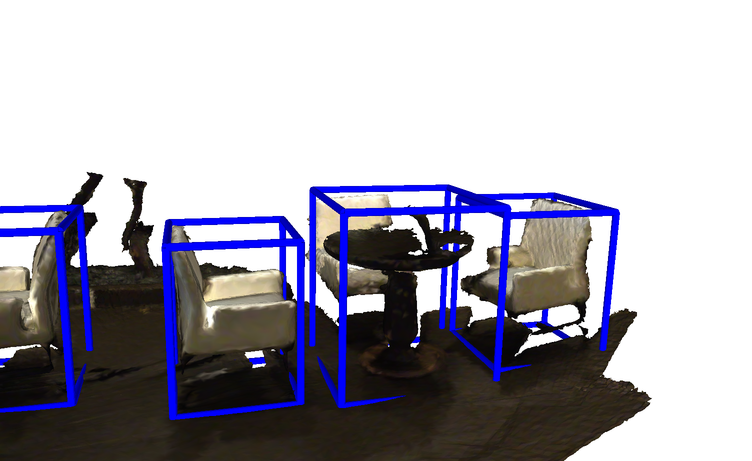} &
 \includegraphics[trim={3cm 0.5cm 3cm 4.5cm},clip,width=0.33\linewidth]{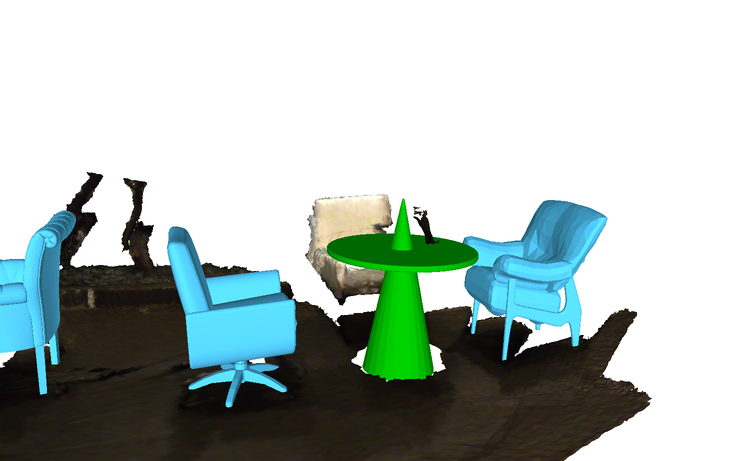} &
 \textcolor{red}{\frame{\includegraphics[trim={3cm 0.5cm 3cm 4.5cm},clip,width=0.33\linewidth]{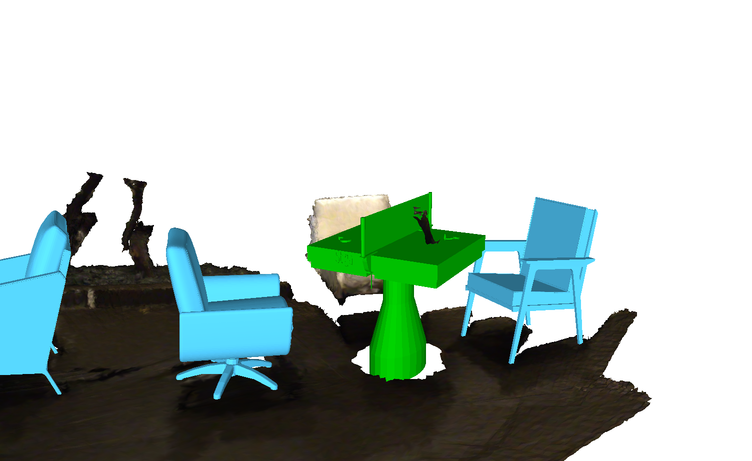}}}  \\

%[trim={left bottom right top}
     \includegraphics[trim={4.5cm 0cm 2.25cm 2.5cm},clip,width=0.33\linewidth]{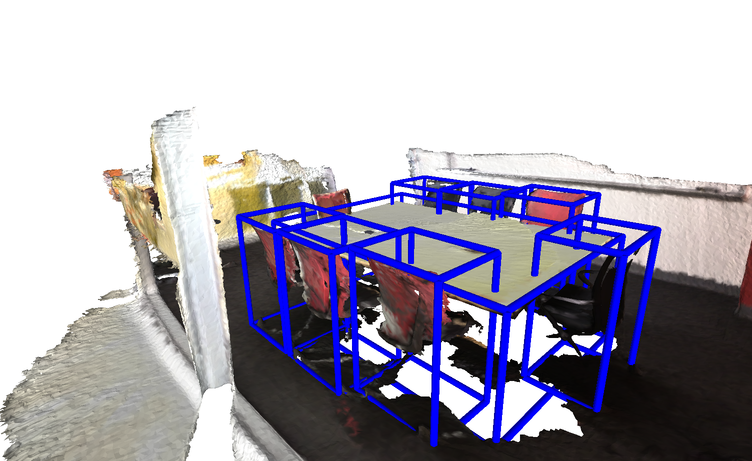} &
  \includegraphics[trim={4.5cm 0cm 2.25cm 2.5cm},clip,width=0.33\linewidth]{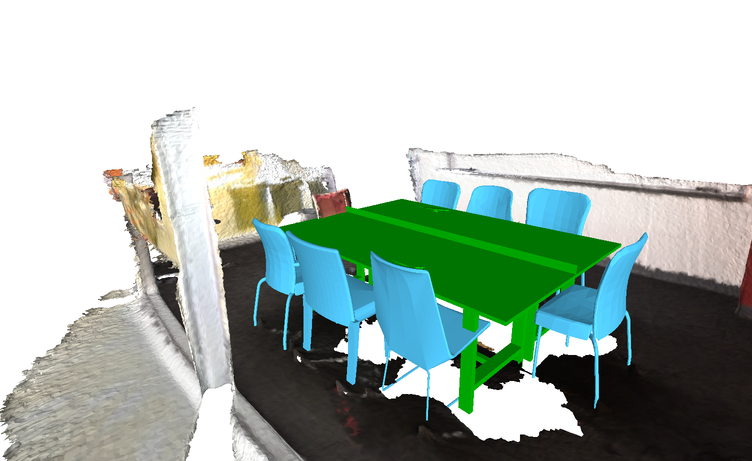} &
  \includegraphics[trim={4.5cm 0cm 2.25cm 2.5cm},clip,width=0.33\linewidth]{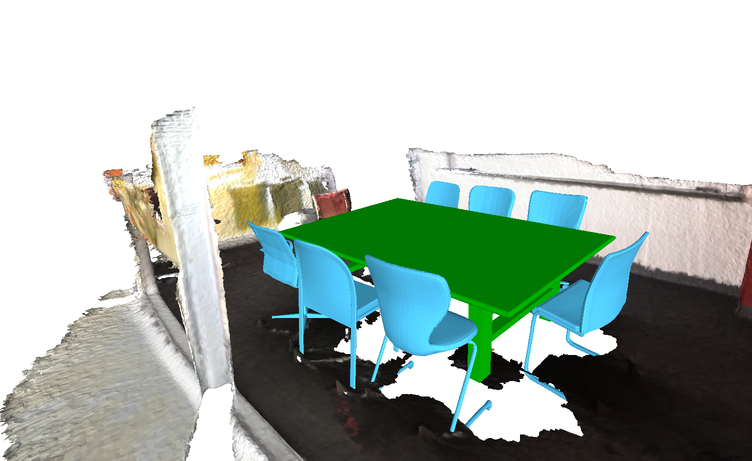} \\

       \includegraphics[trim={1.5cm 0cm 3.5cm 2.5cm},clip,width=0.33\linewidth]{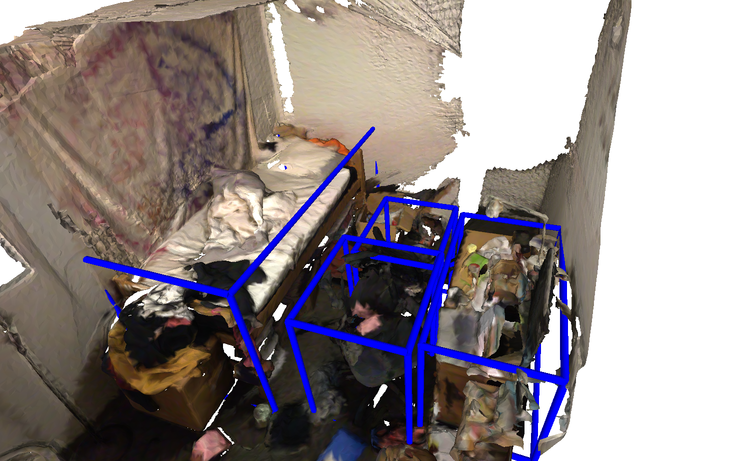} &
  \includegraphics[trim={1.5cm 0cm 3.5cm 2.5cm},clip,width=0.33\linewidth]{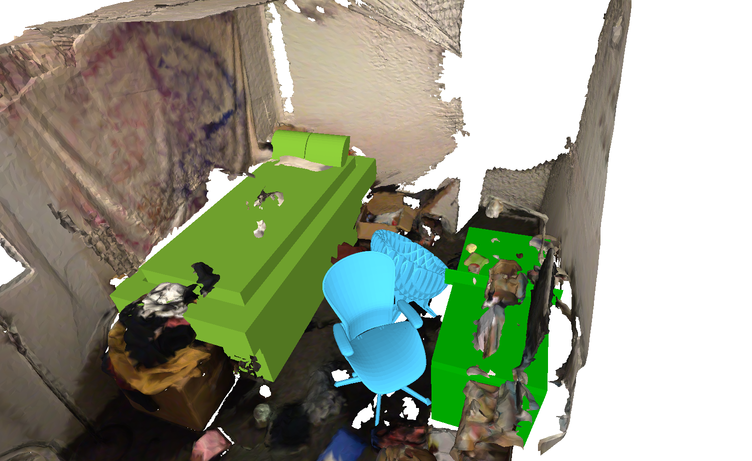} &
  \includegraphics[trim={1.5cm 0cm 3.5cm 2.5cm},clip,width=0.33\linewidth]{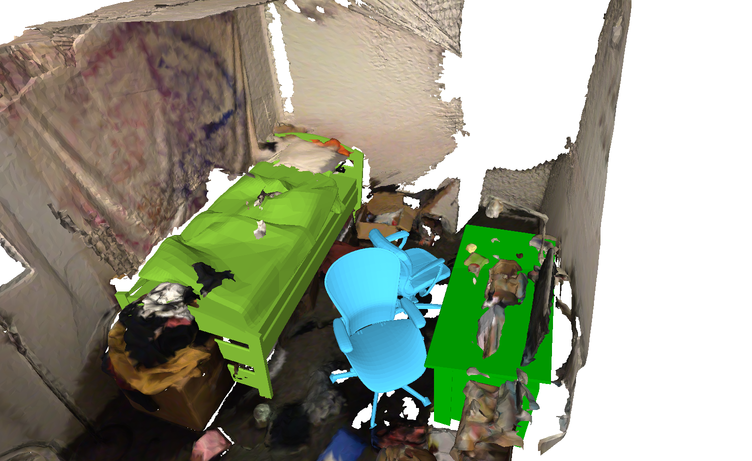} \\

  RGB-D Scan with &  CAD models retrieved & CAD models retrieved \\
   initial 3D boxes         &   with exh. search   &      with HOC-Search   \\
      &     &      + refinement   \\
\end{tabular}
}
\caption{Failure cases for CAD model retrieval using SoftGroup predictions. Inaccurate initial 3D boxes significantly reduce the quality of CAD model retrieval, see the table objects in these examples. Top row, highlighted in red: If, by chance, HOC-Search retrieves a non-suitable CAD model that has a high score for an inaccurate 3D box at an early iteration, it is not able to recover the correct pose, as none of the upcoming CAD models lead to a better score for this inaccurate box. Note that in the last two rows, HOC-Search with refinement successfully retrieves suitable CAD models for the table objects despite inaccurate initial poses. Exhaustive search fails to retrieve suitable CAD models for all table objects in these three examples.}
\label{fig:supp_softgroup_results_fail}
\end{figure*}

Figure~\ref{fig:supp_softgroup_results} shows additional visualizations for CAD model retrieval with exhaustive search and HOC-Search with refinement using SoftGroup~\cite{Vu_2022_CVPR} predictions. In general, HOC-Search with refinement retrieves CAD models that fit the target objects well, and is often able to improve the initial pose of the object. 

Figure~\ref{fig:supp_softgroup_results_fail} shows one common failure case. If a specific CAD model, by chance, fits a target object with an inaccurate pose very well, HOC-Search with refinement is not able to recover the correct pose of the target object, and consequently fails to retrieve the correct CAD model.

% \stefan{
% As described in the main paper, we are able use SoftGroup~\cite{Vu_2022_CVPR} to obtain 3D semantic
% instance segmentation from RGB-D scans for indoor scenes, which we can use to extract axis-aligned
% 3D bounding boxes for the target objects. This setup makes our approach completely automated, and we show the generalization capabilities by using it for two custom scans which were captured by the authors. Note that in this experiment, SoftGroup predictions can include small outlier predictions. To remove these outliers, we ignore all predicted instances that contain less than $1500$ points. 
% Results for the two custom scans are shown in Figure~\ref{fig:supp_softgroup_results_custom}. The retrieved CAD models fit the captured scan very well. These results show that our method generalizes well for common indoor scenes, hence it can easily be used for CAD model and pose retrieval using custom data.
% }

\section{Limitations}

\vincent{
As shown in Figure~\ref{fig:supp_top5_fail}, retrieval sometimes fails, in particular when the 3D scan is incomplete. 
Another limitation, shown in Figure~\ref{fig:supp_softgroup_results_fail}, is that HOC-Search is not always able to retrieve a correct CAD model and pose for highly inaccurate initial poses. 
Introducing more image cues to our objective function could help solving both problems.

% \section{Future Directions}

% We see a strong potential in the concept of property nodes. We showed two use-cases in our experiment by using category nodes and pose nodes. We believe that there are many more possible variations, e.g., using property nodes for different scales for different object categories, for symmetry levels of objects, or for different initial 3D bounding boxes.
}

\end{document}